\documentclass[preprint,12pt,authoryear]{elsarticle}

\usepackage{url}
\usepackage{latexsym}
\usepackage{graphicx}
\usepackage{caption}
\usepackage{subcaption}
\usepackage{booktabs}
\usepackage{tablefootnote}
\usepackage{natbib}
\usepackage{amsmath, amssymb, amsthm}
\usepackage{verbatim}
\usepackage{xcolor}

\usepackage[symbol]{footmisc}

\journal{}

\begin{document}

\begin{frontmatter}



\title{Reading Task Classification Using EEG and Eye-Tracking Data}


\author[inst1]{Nora Hollenstein\footnote[1]{Corresponding author: \textit{nora.hollenstein@hum.ku.dk}}}

\author[inst2]{Marius Tröndle}
\author[inst2]{Martyna Plomecka}
            
\author[inst3]{Samuel Kiegeland}
\author[inst3]{Yilmazcan Özyurt}

\author[inst4,inst5]{Lena A. Jäger}
\author[inst2]{Nicolas Langer}

\affiliation[inst1]{organization={Center for Language Technology},%Department and Organization
            addressline={University of Copenhagen}}
            
\affiliation[inst2]{organization={Department of Psychology},%Department and Organization
            addressline={University of Zurich}}

\affiliation[inst3]{organization={Department of Computer Science},%Department and Organization
            addressline={ETH Zurich}}
            
\affiliation[inst4]{organization={Department of Computational Linguistics},%Department and Organization
            addressline={University of Zurich}}

\affiliation[inst5]{organization={Department of Computer Sciene},%Department and Organization
            addressline={University of Potsdam}}

\begin{abstract}
The Zurich Cognitive Language Processing Corpus (ZuCo) provides eye-tracking and EEG signals from two reading paradigms, normal reading and task-specific reading. 
We analyze whether machine learning methods are able to classify these two tasks using eye-tracking and EEG features. We implement models with aggregated sentence-level features as well as fine-grained word-level features. We test the models in within-subject and cross-subject evaluation scenarios. All models are tested on the ZuCo 1.0 and ZuCo 2.0 data subsets, which are characterized by differing recording procedures and thus allow for different levels of generalizability. Finally, we provide a series of control experiments to analyze the results in more detail.
\end{abstract}

\end{frontmatter}


\section{Introduction}\label{sec:intro}

\subsection{Motivation \& Background}

\noindent Electroencephalographic (EEG) and eye tracking are considered gold-\linebreak standard physiological and behavioral measures of cognitive processes involved in reading \citep{rayner1998eye,dimigen2011coregistration}. Reading task classification, i.e., decoding mental states and detecting specific cognitive processes occurring in the brain during different reading tasks, is an important challenge in cognitive neuroscience as well as in natural language processing.

Reading is a complex cognitive process that requires the simultaneous processing of complex visual input, as well as syntactic and semantic integration. Identifying specific reading patterns can improve models of human reading and provide insights into human language understanding and how we perform linguistic tasks. This knowledge can then be applied to machine learning algorithms for natural language processing. Accurate reading task classification can improve the manual labelling process for a variety of NLP tasks, as these processes are closely related to identifying reading intents. Recognizing reading patterns for estimating reading effort has additional applications such as as the diagnosis of reading impairments such as dyslexia \citep{rello2015detecting,Raatikainen2021} and attention deficit disorder \citep{tor2021automated}. 

One of the main bottlenecks in training supervised natural language processing (NLP) and machine learning (ML) applications is that large labeled datasets are required. Generating these labels is often still an expensive and time-consuming manual process. The Zurich Cognitive Language Processing Corpus (ZuCo; \citealt{hollenstein2018zuco,hollenstein2020zuco}) addresses these challenges. ZuCo is a dataset combining electroencephalography (EEG) and eye-tracking recordings from subjects reading natural English sentences. The EEG and eye-tracking signals lend themselves to train improved ML models for various tasks, in particular for information extraction tasks. One of the advantages of the ZuCo dataset is that it provides ground truth labels for additional machine learning tasks. The availability of labelled data plays a crucial role in all supervised machine learning applications. Physiological data can be used to understand and improve the labelling process (e.g., \citealt{tokunaga2017eye}), and, for instance, to build cost models for active learning scenarios \citep{tomanek2010cognitive}. Is it possible to replace this expensive manual work with models trained on physiological activity data recorded from humans while reading? That is to say, can we find and extract the relevant aspects of text understanding and annotation directly from the source, i.e., eye-tracking and brain activity signals during reading? Using cognitive signals of language processing could be used directly to (pre-)annotate samples to generate training data for ML models.  Moreover, identifying reading intents can help to improve the labelling processes by detecting tiredness from brain activity data and eye-tracking data, and subsequently to suggest breaks or task switching.

We leverage the ZuCo dataset with EEG and eye-tracking recordings from reading English sentences for this work. This dataset contains two different reading paradigms: Normal reading (with the only task of reading naturally for reading comprehension) and task-specific reading (with the purpose of finding specific information in the text). We train machine learning models on eye-tracking and EEG features to solve a binary classification to identify the two reading tasks as accurately as possible. We investigate how two different reading tasks affect both eye movements and brain activity.

Understanding the physiological aspects of the reading
process can advance our understanding of human language processing as well as provide benefits for natural language processing. Recent advances in machine learning are providing new methods to approach reading task classification \citep{haynes2006decoding,mathur2021dynamic}. 
Moreover, the availability of a neurolinguistic dataset with co-registered EEG and eye-tracking signals such as the ZuCo facilitates this work.
The simultaneous recording of EEG and eye-tracking allows us to investigate specific feature sets on different levels of analysis, e.g., sentence level, word level, fixation level. Additionally, the naturalistic setup of the experiments used in this work are crucial for this work and for the ecological validity of experiment in neuroscience in general \citep{nastase2020keep}.

\subsection{Previous Work}

\noindent The ZuCo dataset is freely available and has recently been used in a variety of applications including leveraging EEG and eye-tracking data to improve natural language processing tasks \citep{barrett2018sequence, mathias2020survey,mcguire2021relation}, evaluating the cognitive plausibility of computational language models \citep{hollenstein2019cognival, hollenstein2021relative}, investigating the neural dynamics of reading \citep{pfeiffer2020neural}, developing models of human reading \citep{bautista2020towards,bestgen2021last}. This shows that ZuCo can also be used for other machine learning benchmark tasks.
In a recent study, the ZuCo data has been used already for reading task identification \citep{mathur2021dynamic}. The authors propose a complex convolutional network combining eye-tracking and EEG features, which is evaluated on a fixed cross-subject scenario (trained on 12 subjects, validated on 2, and tested on 4 subjects) on the sentences from ZuCo 2.0. The authors achieve 69.79\% accuracy on this binary classification task. However, this performance measured on a fixed evaluation setting still leaves room for improvement and open research questions regarding the selection of features.

\subsection{Contributions}
\noindent We propose a machine learning approach for reading task classification based on eye-tracking and EEG data. We investigate a large range of features and implement word-level and sentence-level models. We test all models on both ZuCo datasets. Finally, we present a series of control analyses to validate the results. The code for all experiments is available online.\footnote{\url{https://github.com/norahollenstein/reading-task-classification}} 

The results show substantially higher performance for sentence-level than for word-level models.
Generally, we find that the models achieve high accuracy on the ZuCo 1.0 data, and lower accuracy -- but still higher than the baselines -- for the ZuCo 2.0 data. Additional analyses show that these differences in performance might be attributed to the session effects present in ZuCo 1.0. Moreover, while the within-subject evaluation yields good results, there is still room for improvement in the cross-subject settings in future work, which is crucial for practical machine learning applications.

\section{The Zurich Cognitive Language Processing Corpus}\label{sec:data}

\noindent In this section, we describe the compilation of the Zurich Cognitive Language Processing Corpus (ZuCo). ZuCo is a dataset combining electroencephalography (EEG) and eye-tracking recordings from subjects reading natural sentences. ZuCo includes high-density EEG and eye-tracking data of 30 healthy adult native English speakers, each reading natural English text for 3–6 hours. We recorded two separate datasets with different participants. The first dataset, ZuCo 1.0, encompasses EEG and eye-tracking data of 21,629 words in 1107 sentences for each of the 12 subjects. The second dataset, ZuCo 2.0, encompasses the same type of recordings of 15,138 words and 739 sentences for each of the 18 subjects.
The recordings of ZuCo 1.0 include three reading paradigms. In this work, we consider two paradigms only (present also in ZuCo 2.0): a normal reading experiment and a task-specific reading experiment. 

Both datasets, including the raw data and the extracted features, are freely available on the Open Science Framework\footnote{ZuCo 1.0: \url{https://osf.io/q3zws/} and ZuCo 2.0: \url{https://osf.io/2urht/}.}. Moreover, both datasets have been extensively described in previous publications (ZuCo 1.0 in \citealt{hollenstein2018zuco} and ZuCo 2.0 in \citealt{hollenstein2020zuco}). Therefore, in this article we provide a higher-level general description of the data collection focusing mainly on the reading task paradigms relevant for the benchmark task.

One of the main advantages of the ZuCo dataset is its naturalistic reading setup, defined by the following characteristics: (1) We present full sentences on the screen spanning multiple lines, as opposed to a rapid serial visual paradigm where each word is presented in isolation. (2) There are no time constraints on reading speed. The participants are able to read each sentence in their own pace. (3) The presented stimuli are naturally occurring sentences and not hand-picked or manually constructed for experimental purposes.

\subsection{Reading Materials \& Experimental Design}

\noindent The reading materials recorded for the ZuCo corpus contain sentences from movie reviews from the Stanford Sentiment Treebank \citep{socher2013recursive} and Wikipedia articles from a dataset provided by \citet{culotta2006integrating}. These resources were chosen since they provide ground truth labels for various natural language processing ML tasks. In this work, we focus on the Wikipedia sentences, which were used in the normal reading (NR) and task-specific (TSR) experiment paradigms. Descriptive statistics about the datasets used in this work are presented in Table \ref{tab:dataset}.

% experiment design
For the recording sessions, the sentences were presented one at a time at the same position on the screen. Text was presented in black with font size 20-point Arial on a light grey background resulting in a letter height of 0.8 mm or 0.674° of visual angle. The lines were triple-spaced, and the words double-spaced. A maximum of 80 letters or 13 words were presented per line in both tasks. Long sentences spanned multiple lines (max. 7 lines). 
%A maximum of 7 lines for SR, 5 lines for NR and 7 lines for Task TSR were presented simultaneously on the screen.

\begin{table}[t]
\centering
\small
\begin{tabular}{l|ll|ll}
\toprule
 & \multicolumn{2}{c|}{\textbf{ZuCo 1.0}} & \multicolumn{2}{c}{\textbf{ZuCo 2.0}} \\
 & NR & TSR & NR & TSR \\\midrule
sentences & 300 &  & 349 & 390 \\
sent. length  & 21.3 ($\pm10.6$) & 20.1. ($\pm10.1$) & 19.6 (8.8) & 21.3 (9.5) \\
total words & 6386 & 8164 & 6828 & 8310 \\
word types & 2657 & 2995 & 2412 & 2437 \\
word length & 6.7 (2.7) & 6.7 (2.6) & 4.9 (2.7) & 4.9 (2.7) \\
Flesch score & 51.33 & 51.43 & 55.38 & 50.76 \\\bottomrule
\end{tabular}
\caption{Descriptive statistics of reading materials (SD = standard deviation), including Flesch readibility scores.}
\label{tab:dataset}
\end{table}

\begin{figure}[t]
    \centering
    \includegraphics[width=1\textwidth]{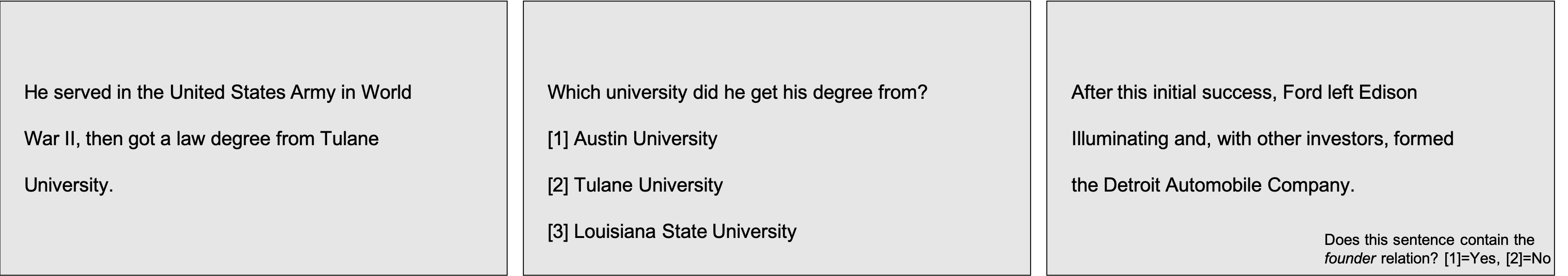} 
    \caption{Example sentences on the recording screen: (left) a normal reading sentence, (middle) a control question for a normal reading sentence, and (right) a task-specific  annotation sentence.}
    \label{fig:sentences}
\end{figure}

\subsection{Normal Reading (NR)} 

\noindent In the first task, participants were
instructed to read the sentences naturally, without any specific task other than comprehension. The participants were instructed to read one sentence at a time at their own pace and use the control pad to move to trigger the onset of the next sentence. They were informed that a portion of the sentences would be followed by a comprehension question. The task was explained to the subjects orally, followed by instructions on the screen. 
Figure \ref{fig:sentences} (left) shows an example sentence as it was depicted on the screen during recording. As shown in Figure \ref{fig:sentences} (middle), the control condition for this task consisted of single-choice questions about the content of the previous sentence. The questions are presented with three answer options, out of which only one is correct. 12\% of randomly selected sentences were followed by such a comprehension question with three answer options on a separate screen. The task was preceded by a practice round.

\subsection{Task-specific Reading (TSR)}
\noindent In the second task, the subjects were presented with similar sentences as in the normal reading task, but with specific instructions to search for a specific relation in each sentence they read. The following relation types were contained in the sentences: \textit{award}, \textit{education}, \textit{employer}, \textit{founder}, \textit{job\_title}, \textit{nationality}, \textit{political\_affiliation}, \textit{visited} and \textit{wife/husband}. This allows us to compare the EEG and eye-tracking signals during normal reading to task-specific reading while searching for a specific relation type.

Instead of comprehension questions, the participants had to decide for each sentence whether it contained the relation or not, i.e. they were actively annotating each sentence. Figure \ref{fig:sentences} (right) shows an example screen for this task. 17\% of the sentences did not include the relation type and were used as control conditions. All sentences within one recording block involved the same relation type. The blocks started with a practice round, which described the relation and was followed by three sample sentences, so that the participants would be familiar with the respective relation type.

Purposefully, there are some duplicate sentences that appear in both the normal reading and the task-specific reading tasks (48 sentences in ZuCo 1.0 and 63 sentences in ZuCo 2.0.). The intention of these duplicate sentences is to provide a set of sentences read twice by all participants with a different task in mind. Hence, this enables the comparison of eye-tracking and brain activity data when reading normally and when annotating specific relations (see examples in Section \ref{sec:prel-data-analysis}).

\subsection{Recording Procedure}
\noindent The main difference and reason for recording ZuCo 2.0 consisted in the experiment procedure, namely, the number of sessions and the order of the reading tasks. 
For ZuCo 1.0, the normal reading and task-specific reading paradigms were recorded in different sessions on different days. The order of the sessions and sentences within the sessions was identical for all subjects.

Therefore, the recorded data is not fully appropriate as a means of comparison between natural reading and annotation, since the differences in the brain activity data might result mostly from the different sessions due to the sensitivity of EEG and session-specific effects in the eye-tracking signal. This, and extending the dataset with more sentences and more subjects, were the main factors for recording the ZuCo 2.0 dataset. 

For ZuCo 2.0, we recorded 14 blocks of approx. 50 sentences for each subject, alternating between tasks: 50 sentences of normal reading, followed by 50 sentences of task-specific reading. The order of blocks and sentences within blocks was identical for all subjects. Each sentence block was preceded by a practice round of three sentences and followed by a short break to ensure a clear separation between the reading tasks.

The differing recording procedures between the two datasets allow us to investigate the impact of possible session biases in the data. As we show in Section \ref{sec:results}, these aspects affect the results and are discussed through various control analyses in Section \ref{sec:control}.

\subsection{Participants}

\noindent For ZuCo 1.0, data were recorded from 12 healthy adults (between 22 and 54 years, all right-handed; 5 female participants). For ZuCo 2.0, data were recorded from 18 healthy adults (between 23 and 52 years old; 2 left-handed; 10 female participants). The native language of all participants is English. See \ref{app:data-collection} for more details on participant demographics and linguistic assessment.

\subsection{Technical Set-up \& Preprocessing}

\paragraph{Eye-Tracking Acquisition} Eye movements and pupil size were recorded with an infrared video-based eye tracker (EyeLink 1000 Plus, SR Research) at a sampling rate of 500 Hz. The participant was seated at a distance of 68cm from a 24-inch monitor (ASUS ROG, Swift PG248Q, display dimensions 531x299 mm, resolution 800x600 pixels resulting in a display: 400x298.9 mm, a vertical refresh rate of 100 Hz). The eye tracker was calibrated with a 9-point grid at the beginning of the session and re-validated before each block of sentences. 
The eye-tracker computed eye position data and identified events such as saccades, fixations, and blinks. Saccade onsets were detected using the eye-tracking software default settings: acceleration larger than 8000°/s2, a velocity above 30°/s, and a deflection above 0.1°. 

\paragraph{Eye-Tracking Feature Extraction} The datasets provide the data as provided by the eye-tracker, consisting of $(x,y)$ gaze location entries for all individual fixations. Coordinates were given in pixels with respect to the monitor coordinates (the upper left corner of the screen was coded as $(0,0)$ and down/right was positive). Additionally, we provide various engineered reading time features. We extend the features provided in the dataset with a range of fixation and saccade based metrics, both on word level and aggregated on the sentence level (see Table \ref{tab:sent-et-feats}).

\paragraph{EEG Acquisition}

High-density EEG data were recorded simultaneously at a sampling rate of 500 Hz with a bandpass of 0.1 to 100 Hz, using a 128-channel EEG Geodesic Hydrocel system (Electrical Geodesics). To ensure good contact, the impedance of each electrode was checked before recording and was kept below 40 $k\Omega$. In addition, electrode impedance levels were checked approx.\, every 30 mins and reduced if necessary. The EEG data shared in this project are available as raw data but also preprocessed with Automagic \citet{pedroni2019automagic} toolbox, a tool for automatic EEG data cleaning and validation. 
Before the EEG preprocessing with the Automagic toolbox, data from all 14 blocks (7 NR and 7 TSR) were first merged to avoid high predictive power based on the differences resulting from the preprocessing itself. Afterwards, all subfiles whose average standard deviation exceeded $100 \mu V$ were excluded.

\paragraph{EEG Preprocessing}
First, bad channels were detected by the algorithms implemented in the EEGlab plugin \texttt{clean\_rawdata}\footnote{\url{http://sccn.ucsd.edu/wiki/Plugin\_list\_process}}. A channel was defined as a bad electrode when recorded data from that electrode was correlated at less than 0.85 to an estimate based on other channels. Furthermore, a channel was defined as bad if it had more line noise relative to its signal than all other channels (4 standard deviations). Finally, if a channel had a longer flat-line than 5 seconds, it was considered bad. These bad channels were automatically removed and later interpolated using a spherical spline interpolation (EEGLAB function \texttt{eeg\_interp.m}). The interpolation was performed as a final step before the automatic quality assessment of the EEG files. Next, data were filtered using a 2 Hz high-pass filter and line noise artifacts were removed by applying Zapline \cite{de2020zapline}, removing seven power line components. Subsequently, independent component analysis (ICA) was performed. Components reflecting artifactual activity were classified by the pre-trained classifier ICLabel \cite{pion2019iclabel}. Components that were classified as any class of artifacts (line noise, channel noise, muscle activity, eye activity, and heart artifacts) with a probability higher than 0.8 were removed from the data. Subsequently, residual bad channels were excluded if their standard deviation exceeded a threshold of $25 \mu V$. Very high transient artifacts ($> \pm100 \mu V$) were excluded from calculating the standard deviation of each channel. However, if this resulted in a significant loss of channel data ($>$ 50\%), the channel was removed from the data. After this, the pipeline automatically assessed the quality of the resulting EEG files based on four criteria: First, a data file was marked as bad-quality EEG and not included in the analysis if the proportion of high-amplitude data points in the signals ($>30 \mu V$) was larger than 0.20. Second, more than 20\% of time points showed a variance larger than $15 \mu V$ across channels. Third, 30\% of the channels showed high variance ($>15 \mu V$). Fourth, the ratio of bad channels was higher than 0.3.

After Automagic preprocessing, 13 of the 128 electrodes in the outermost circumference (chin and neck) were excluded from further processing as they capture little brain activity and mainly record muscular activity. Additionally, 10 EOG electrodes were separated from the data and not used for further analysis, yielding a total number of 105 EEG electrodes. Subsequently, the EEG and eye-tracking data were synchronized using the “EYE EEG” extension \cite{dimigen2011coregistration} to enable EEG analyses time-locked to the onsets of fixations and saccades, and subsequently segment the EEG data based on the eye-tracking measures.
The synchronization was performed in two steps. 
First, the eye-tracking data were upsampled by linear interpolation to match the number of EEG sampling points. Subsequently, the algorithm identified the  “shared” events. Next, a linear function was fitted to the shared event latencies to refine the start- and end-event latency estimation in the eye tracker recording. Finally, synchronization quality was ensured by comparing the trigger latencies recorded in the EEG and eye-tracker data. All synchronization errors did not exceed 2 ms (one sample).
The remaining eye artifacts in data were modelled and removed with Unfold toolbox \cite{ehinger2019unfold}
Finally, the data was referenced to the common average reference.
 
\paragraph{EEG Feature Extraction}
To compute oscillatory power measures, we band-pass filtered the continuous EEG signals across an entire reading task for four different frequency bands, resulting in a time-series for each frequency band. The distinct frequency bands were determined as follows: \textit{theta} (4-8 Hz), \textit{alpha} (8.5-13 Hz), \textit{beta} (13.5-30 Hz), and \textit{gamma} (30.5-49.5 Hz). Afterwards, we applied a Hilbert transformation to each of these time-series resulting in a complex time series. The Hilbert phase and amplitude estimation method yields results equivalent to sliding window FFT and wavelet approaches \cite{bruns2004fourier}. We chose specifically the Hilbert transformation to maintain temporal information for the  amplitude of the frequency bands to enable  the power computation of the different frequencies for time segments defined through fixations from the eye tracking. Finally, for each word in each sentence, the EEG features consist of a vector of 105 dimensions (one value for each EEG channel).

%More detailed information about the data acquisition and preprocessing can be found in the Appendix. 
%\ref{app:data-collection}.

\section{Method}

\noindent In this section, we first present a preliminary data analysis to investigate the differences between normal reading and task-specific reading reflected in eye-tracking and EEG data.
Following, we describe the machine learning models we developed for the reading task classification. In the first approach (Section \ref{sec:word-level}) the models receive word-level features as input, while in the second approach (Section \ref{sec:sentence-level}) the models learn from aggregated sentence-level features.

\subsection{Preliminary Data Analysis}\label{sec:prel-data-analysis}

\begin{figure}[t]
    \centering
     \begin{subfigure}[A]{0.99\textwidth}
     \centering
    \includegraphics[width=0.31\textwidth]{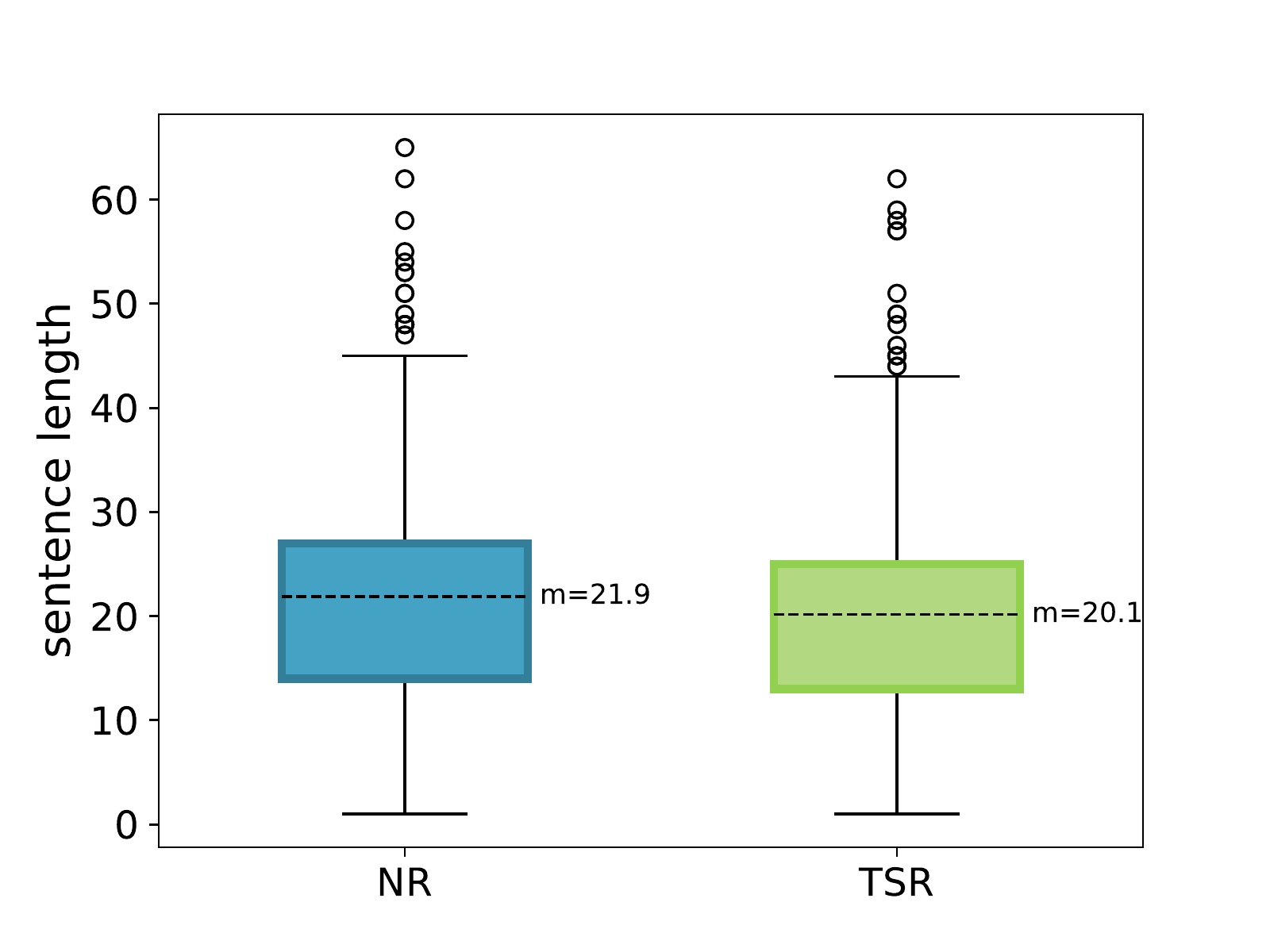} 
    \includegraphics[width=0.31\textwidth]{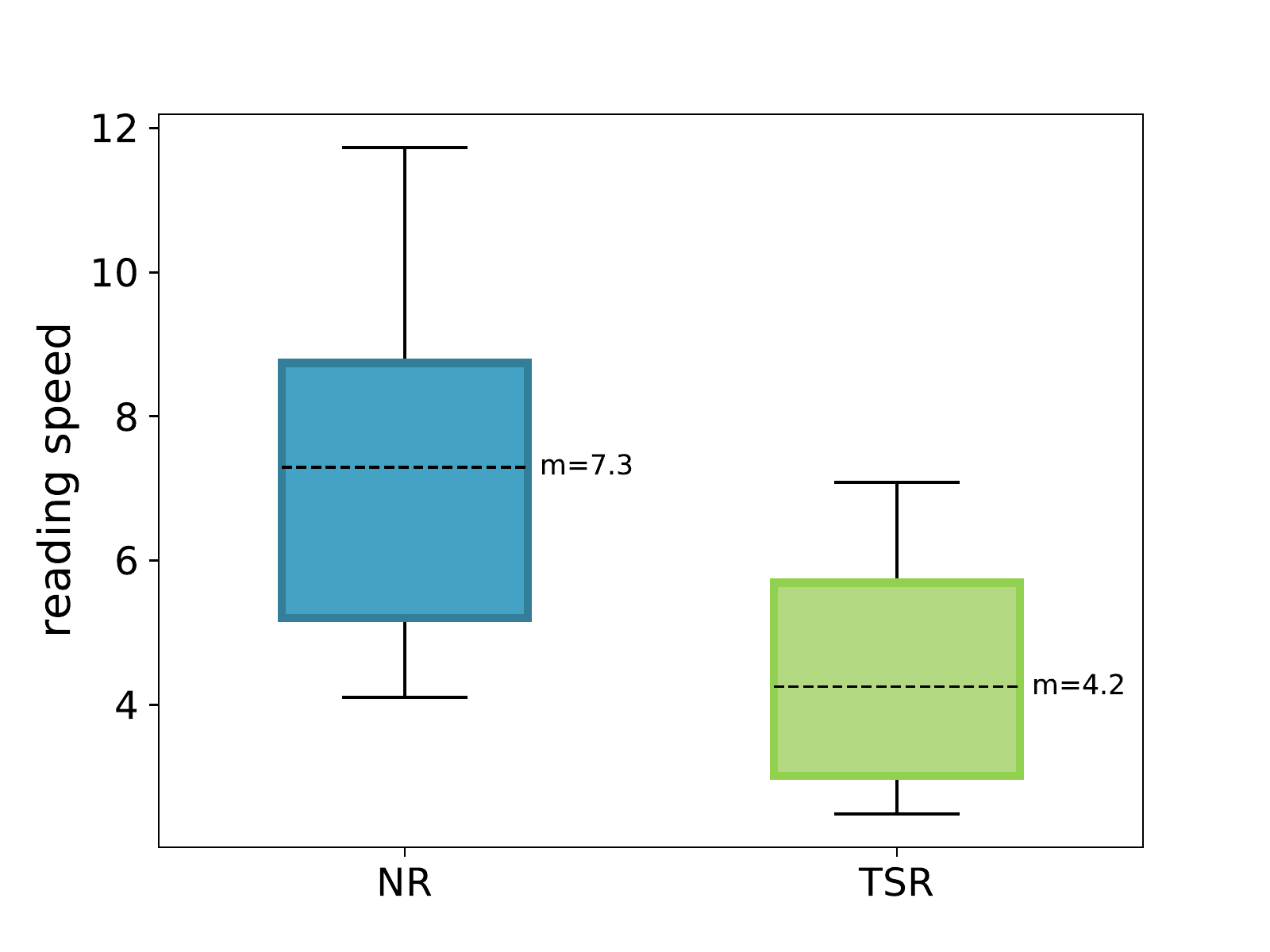}
    \includegraphics[width=0.31\textwidth]{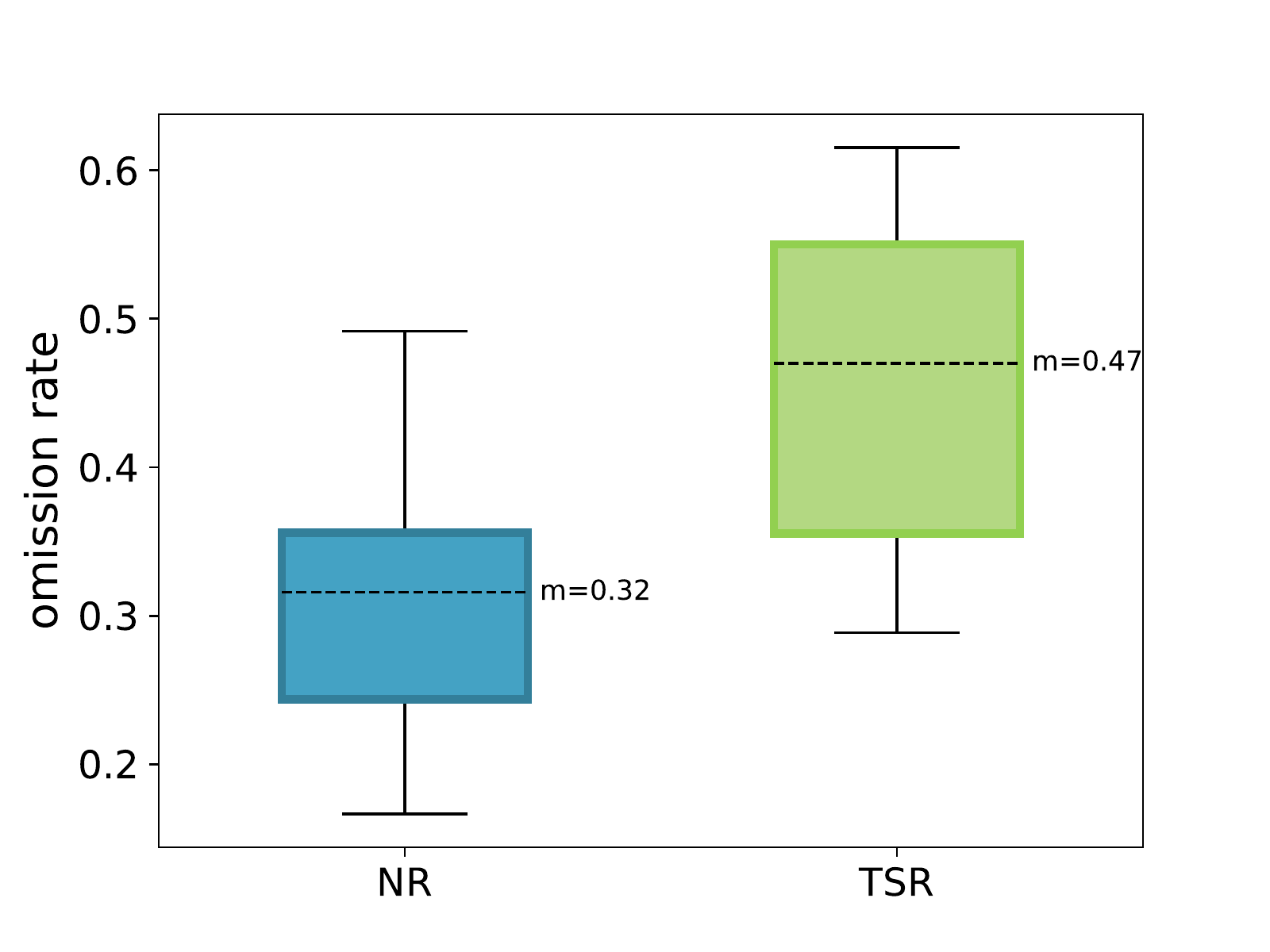}\caption{ZuCo 1.0}
    \end{subfigure}
    \centering
     \begin{subfigure}[A]{0.99\textwidth}
     \centering
    \includegraphics[width=0.31\textwidth]{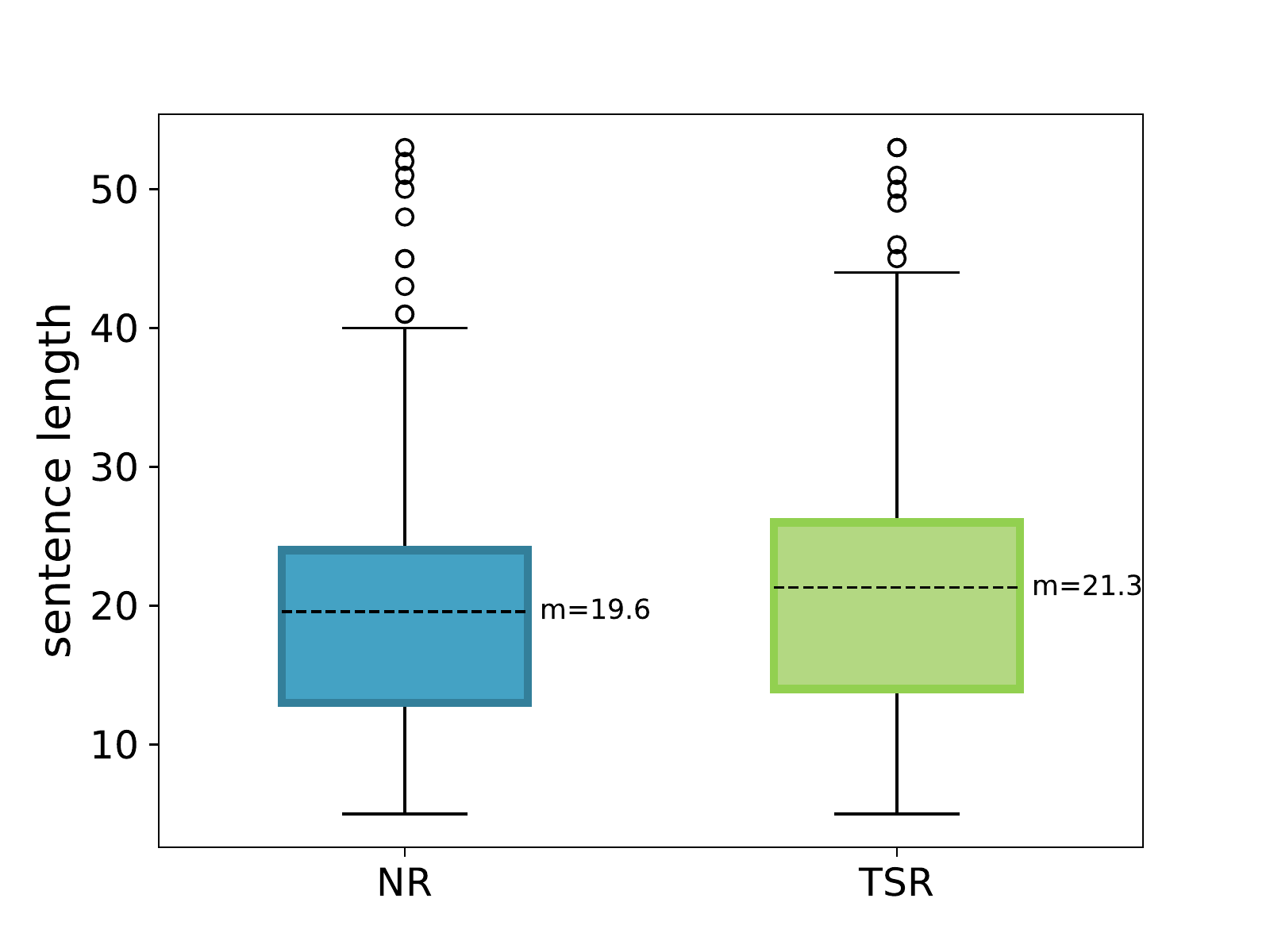} 
    \includegraphics[width=0.31\textwidth]{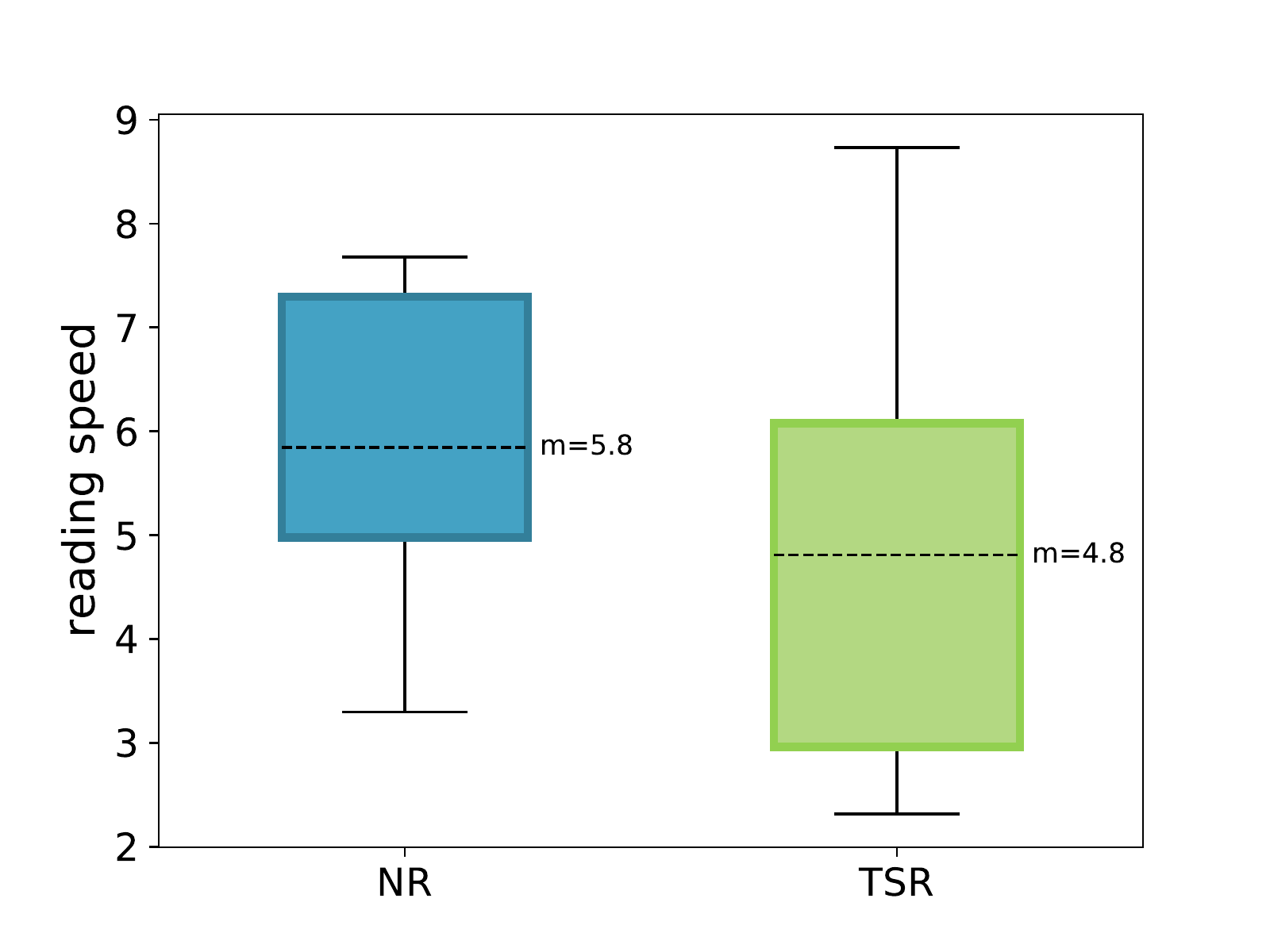}
    \includegraphics[width=0.31\textwidth]{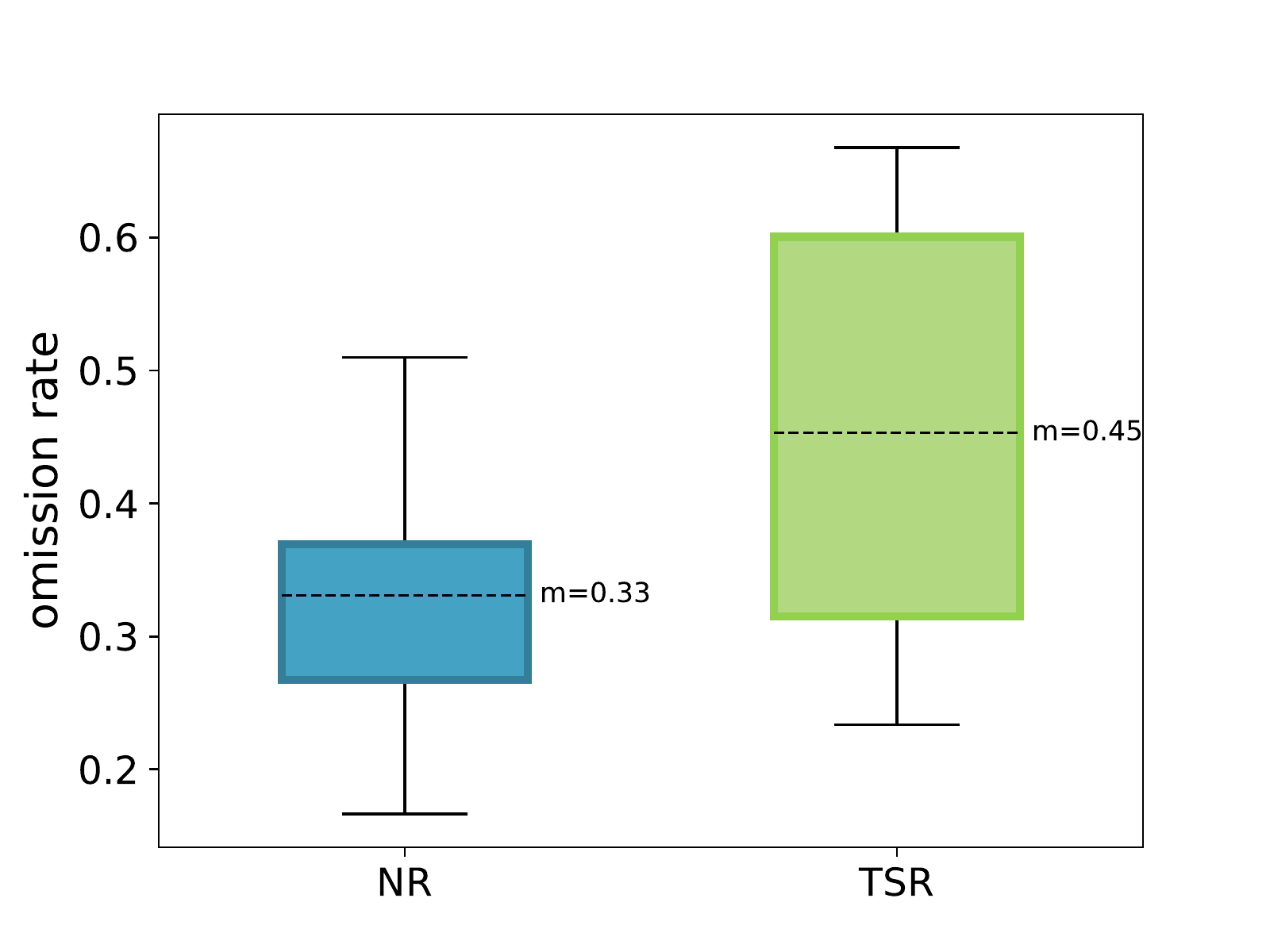}\caption{ZuCo 2.0}
    \end{subfigure}
    \caption{Sentence length (words per sentence), reading speed (seconds per sentence) and omission rate (percentage of words not fixated) comparison between normal reading (NR) and task-specific reading (TSR) in ZuCo 1.0 (a) and ZuCo 2.0 (b). The differences in omission rate and reading speed between the two tasks are significant in both datasets.}
    \label{fig:data-analysis}
\end{figure}
  
\begin{figure}[t]
    \centering
    \includegraphics[width=1\textwidth]{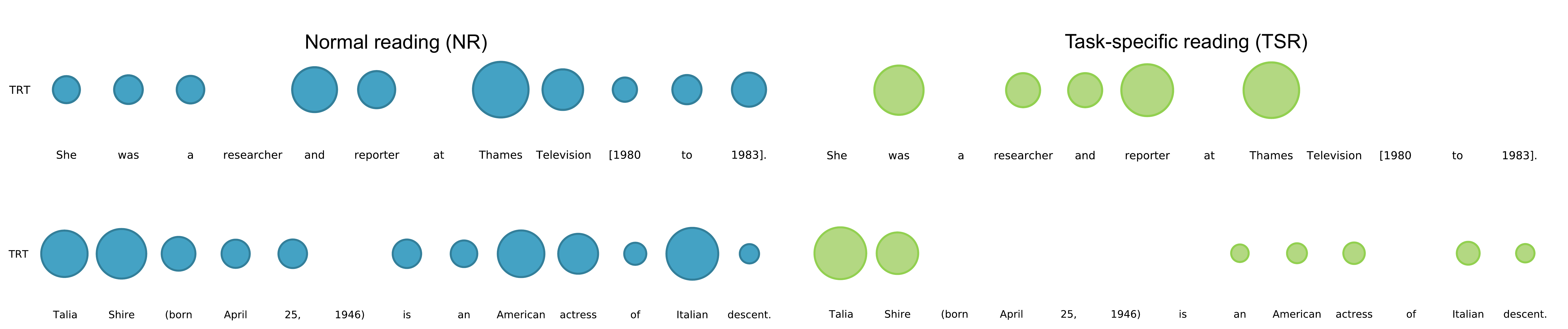}
    \caption{Example sentences read in the normal reading paradigm (left) as well as the task-specific experiment paradigm (right) for a single subject. The dots denote the total reading time (TRT) of each word; a larger dot means longer reading time.}
    \label{fig:nr-tsr-examples}
\end{figure}

\noindent We analyze the properties of the dataset and compare the two reading tasks. First, we compare the sentence length, reading speed, and omission rate based on the collected eye-tracking data. The sentence length (i.e., the number of words per sentence) was controlled in the selection of reading materials, so that it would not differ significantly between the two tasks (ZuCo 1.0: NR mean 21.9, std 11.1; TSR mean 20.1, std 9.9. ZuCo 2.0: NR mean 19.6, std 8.8; TSR mean 21.3, std 9.5). This is shown in Figure \ref{fig:data-analysis}, left. 
Reading speed is defined as the number of seconds spent on a given sentence and the omission rate is the percentage of words that were skipped (i.e., not fixated) per sentence.
As expected, in both datasets, the omission rate is higher during task-specific reading (ZuCo 1.0: NR mean 0.32, std 0.09; TSR mean 0.47, std 0.11, $p<0.002$. ZuCo 2.0: NR mean 0.33, std 0.09; TSR mean 0.45, std 0.14, $p<0.008$), where words are more often skipped, which in turn reduces the reading speed (ZuCo 1.0: NR mean 7.3, std 2.5; TSR mean 4.2, std 1.5, $p<0.0005$. ZuCo 2.0: NR mean 5.8, std 1.4; TSR mean 4.8, std 2.0, $p<0.03$). This behavior is visualized in Figure \ref{fig:data-analysis}, middle and right. Additionally, Figure \ref{fig:nr-tsr-examples} exemplifies the fixation times of two overlapping sentences between the two reading tasks.

\begin{figure}[t]
\centering
 \includegraphics[width=\textwidth]{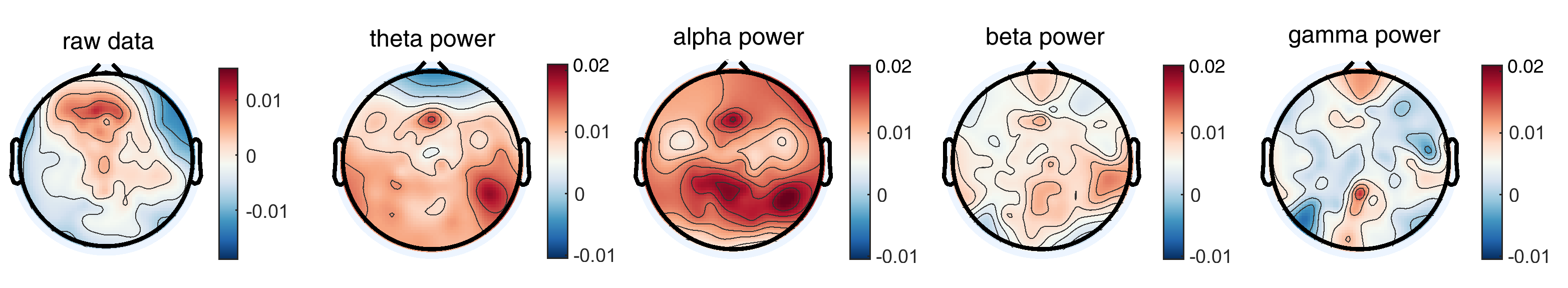}
\caption{Topography plots showing the EEG activity of the difference between the tasks (NR minus TSR), averaged over all sentences across all subjects from ZuCo 2.0, and averaged across all subjects (scalp viewed from above, nose at the top).}\label{fig:topoplots}
\end{figure}

Second, we compare the EEG activity between normal reading and task-specific reading (Figure \ref{fig:topoplots}.) These topography plots show higher frontal raw EEG activity in the NR task compared to the TSR task. However, when calculating EEG power in different frequency bands, the lower frequencies (theta and alpha) show widespread increased activity in the NR condition. Contrary, in the high frequency gamma band -- typically associated with higher cognitive functions -- there is less power in the NR condition compared to TSR.

\clearpage

\subsection{Word-level Models}\label{sec:word-level}

\noindent First, we describe the reading task classification models leveraging word-level eye-tracking or EEG features. Every sample is assigned a feature vector containing values for each word in the sentence.

\paragraph{Eye-tracking Features} Table \ref{tab:word-et-feats} shows the word-level eye-tracking features we extracted for the reading task classification models. We include standard fixation-based psycholinguistic measures such as the number of fixations and the total reading time. Additionally, we leverage saccade-based features to capture additional reading patterns.
For these models, the input for each word is a vector of 5 or 17 feature values, depending on whether saccade features are included or not.

\begin{table}[t]
\small
\centering
\begin{tabular}{llc}
\toprule
\textbf{Name} & \textbf{Definition} & \textbf{Values} \\\midrule
\textsc{Fixation features} & & \\\midrule
nFix & number of fixations & 1 \\\hline
FFD & first fixation duration & 1 \\\hline
TRT & total reading time & 1 \\\hline
GD & gaze duration & 1 \\\hline
GPT & go-past time & 1 \\\hline
\textsc{Saccade features} & & \\\midrule
inSacc\_velocity\_mean & mean incoming saccade velocity & 1 \\\hline
inSacc\_duration\_mean & mean incoming saccade duration & 1 \\\hline
inSacc\_amplitude\_mean & mean incoming saccade amplitude & 1 \\\hline
outSacc\_velocity\_mean & mean outgoing saccade velocity & 1 \\\hline
outSacc\_duration\_mean & mean outgoing saccade duration & 1 \\\hline
outSacc\_amplitude\_mean & mean outgoing saccade amplitude & 1 \\\hline
inSacc\_velocity\_max & max incoming saccade velocity & 1 \\\hline
inSacc\_duration\_max & max incoming saccade duration & 1 \\\hline
inSacc\_amplitude\_max & max incoming saccade amplitude & 1 \\\hline
outSacc\_velocity\_max & max outgoing saccade velocity & 1 \\\hline
outSacc\_duration\_max & max outgoing saccade duration & 1 \\\hline
outSacc\_amplitude\_max & max outgoing saccade amplitude & 1 \\
\bottomrule
\end{tabular}
\caption{Eye-tracking word-level features. \textit{Name} denotes the variable names as used in the dataset.}
\label{tab:word-et-feats}
\end{table}

\begin{table}[t]
\small
\centering
\begin{tabular}{llc}
\toprule
\textbf{Name} & \textbf{Definition} & \textbf{Values} \\\midrule
theta & theta frequency band (4-8 Hz) & 105 \\\hline
alpha & alpha frequency band (8.5-13 Hz) & 105  \\\hline
beta & beta frequency band (13.5-30 Hz) & 105 \\\hline
gamma & gamma frequency band (30.5-49.5 Hz) & 105 \\\hline
raw EEG & mean broadband EEG (0-50 Hz) & 105 \\\bottomrule
\end{tabular}
\caption{EEG word-level features. \textit{Name} denotes the variable names as used in the dataset.}
\label{tab:word-eeg-feats}
\end{table}

\paragraph{EEG Features} \citet{dimigen2011coregistration} demonstrated that EEG indices of semantic processing can be obtained in natural reading and compared to eye movement behavior. The eye tracking data provides millisecond-accurate fixation times for each word. Therefore, the co-registration of both modalities allows the mapping of EEG signals to the visual processing of each word. Hence, by extracting fixation-related potentials (FRPs), we extract the EEG brain activity during the fixation duration of individual words.

Table \ref{tab:word-eeg-feats} shows the EEG feature values used for the word-level models. The EEG data for each of these features was extracted for the duration of the total reading time (TRT) of a word. We include a feature composed of the full broadband EEG signals as well as features filtered by frequency bands (theta, alpha, beta, and gamma waves).
The input for each word is a vector with 105 electrode values for each frequency band. To capture the total reading time of a word, the EEG signals were averaged over all fixations of a word.

\paragraph{Model} Long-Short Term Memory Networks (LSTMs) have been widely \linebreak adopted for sequence classification tasks \citep{yu2019review}. We implement a bidirectional LSTM to train a reading task classifier based on the word-level features. We train and test the models over five runs with different random seeds, using early stopping to end the training when the validation accuracy is not improving further. Table \ref{tab:params-lstm} presents the values of all hyper-parameters.
For the within-subject evaluation, we perform 3-fold cross-validation in each run and average the results across folds and random seeds. We use 10\% of the training data used for validation.
For the cross-subject evaluation, we train on $n-1$ subjects and test on the data of the left-out subject.

\subsection{Sentence-level Models}\label{sec:sentence-level}

\begin{figure}[t]
    \centering
    \includegraphics[width=0.32\textwidth]{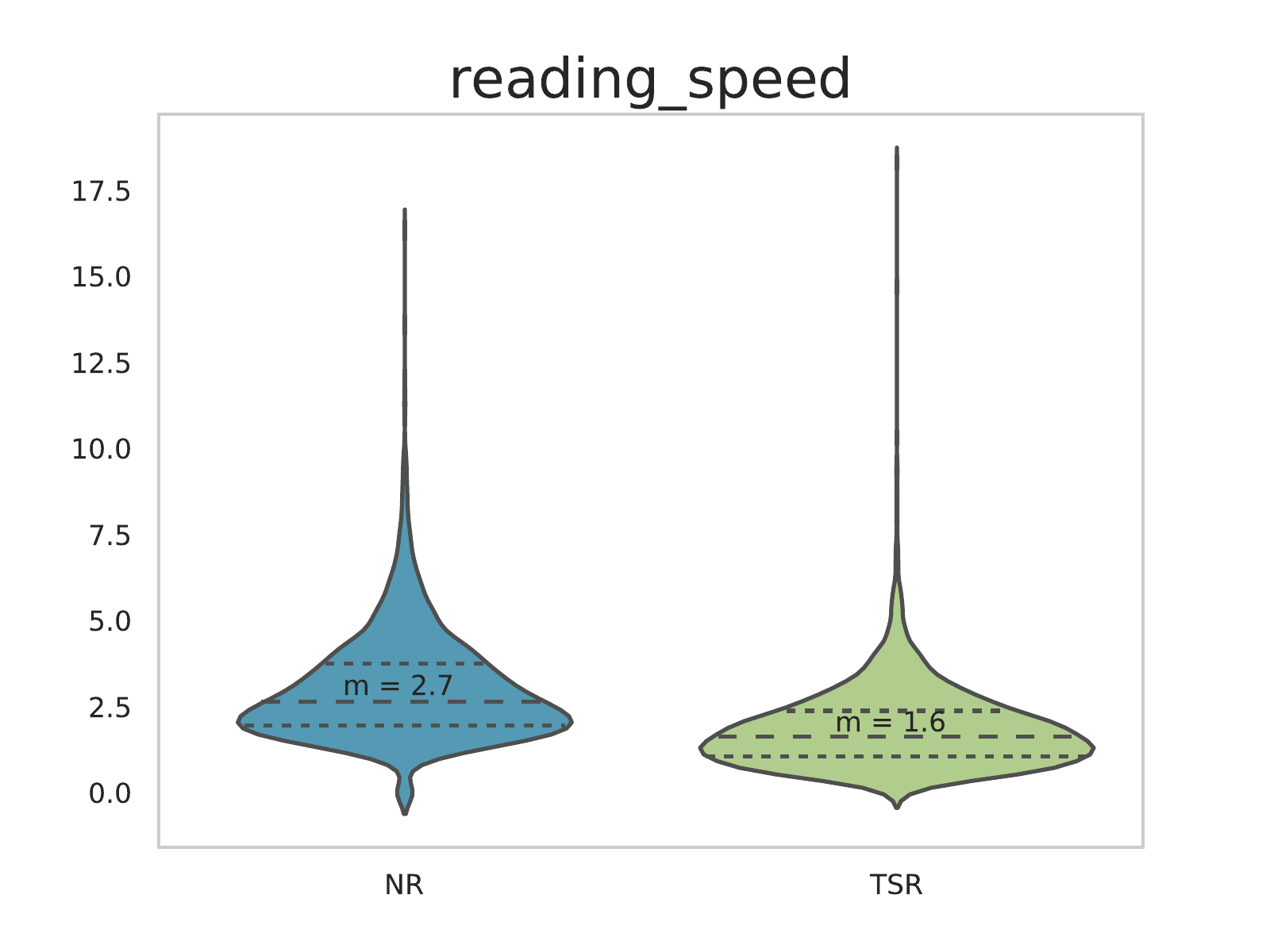}
    \includegraphics[width=0.32\textwidth]{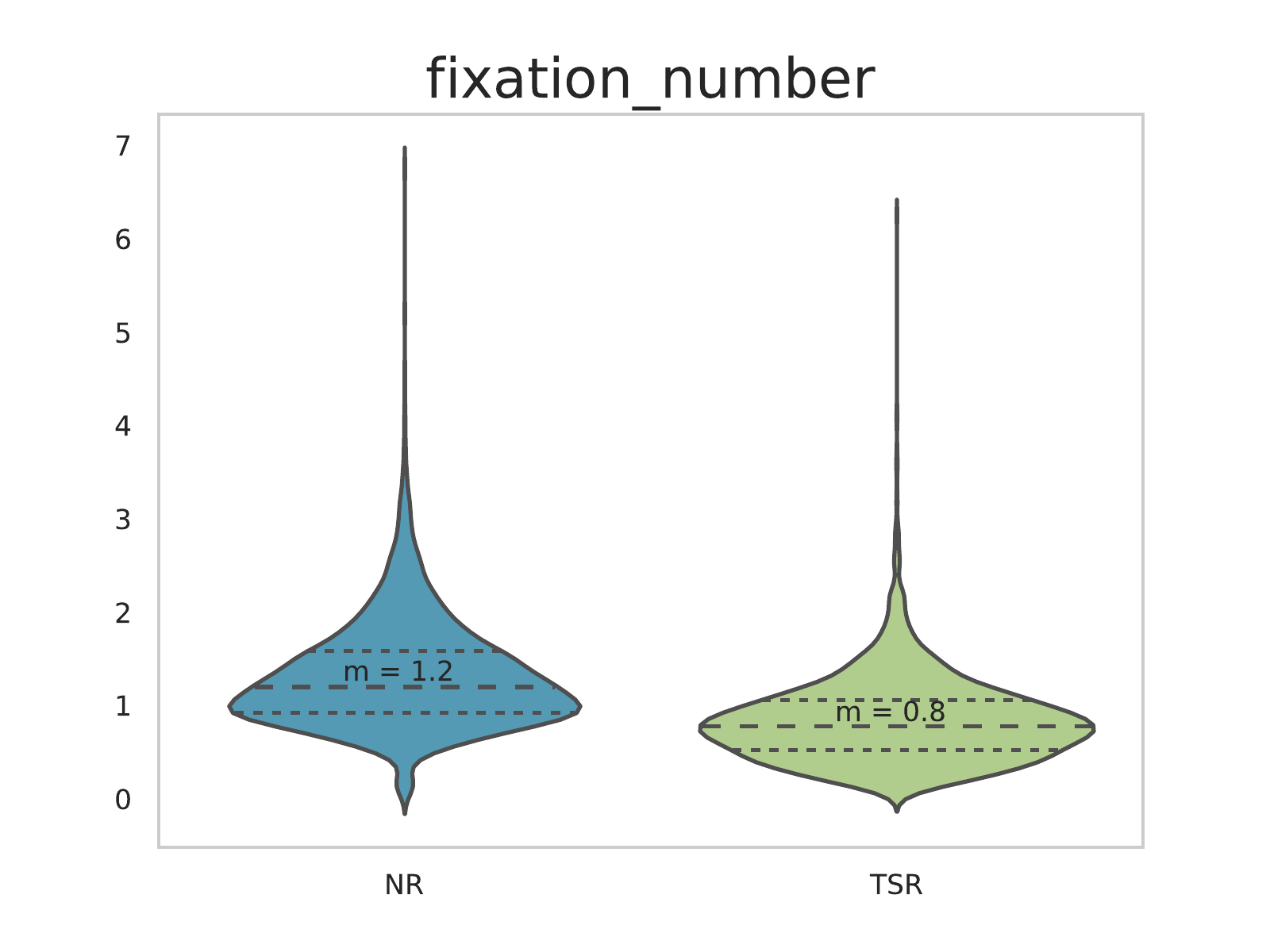}
    \includegraphics[width=0.32\textwidth]{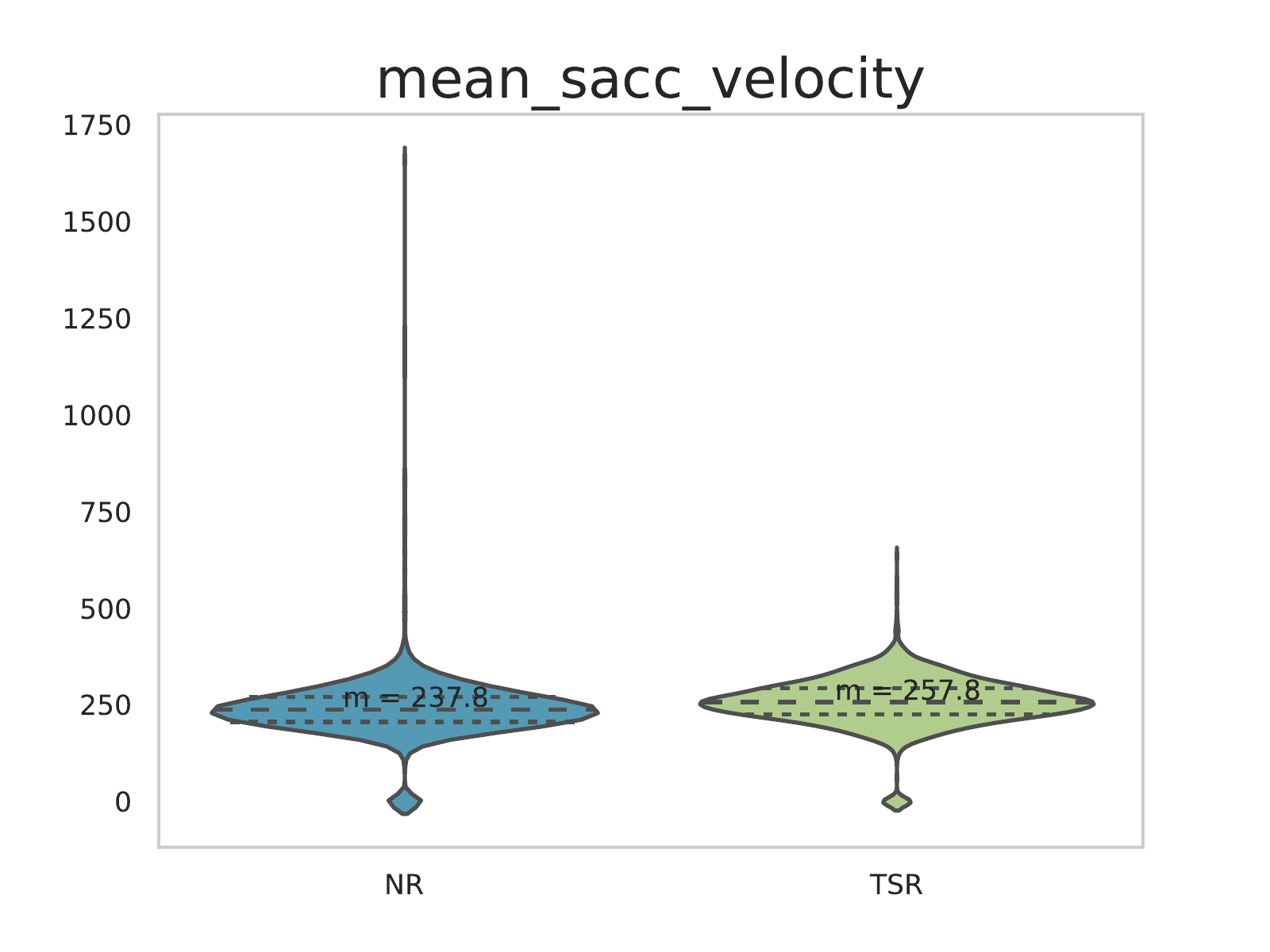}
    \includegraphics[width=0.32\textwidth]{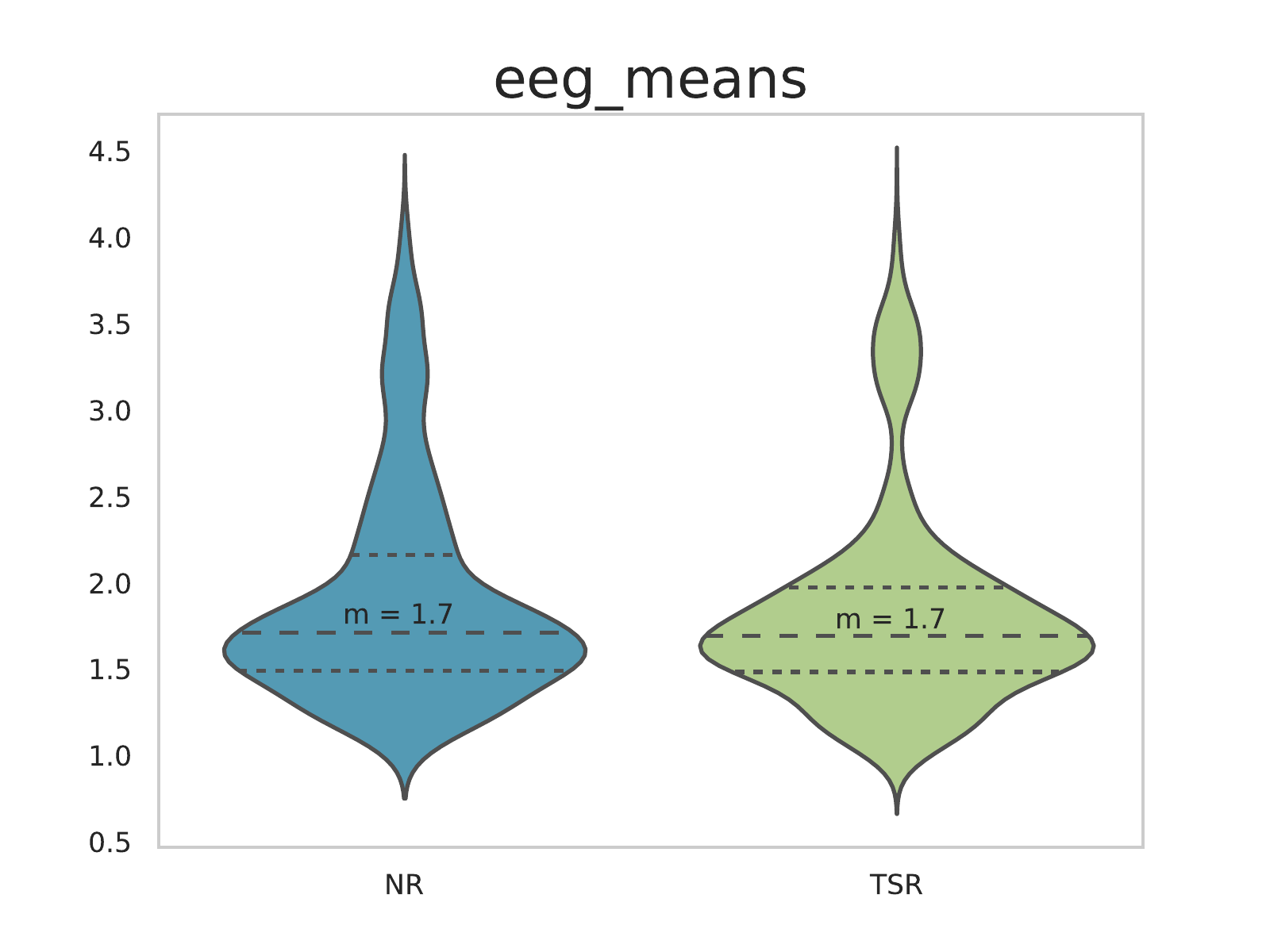}
    \includegraphics[width=0.32\textwidth]{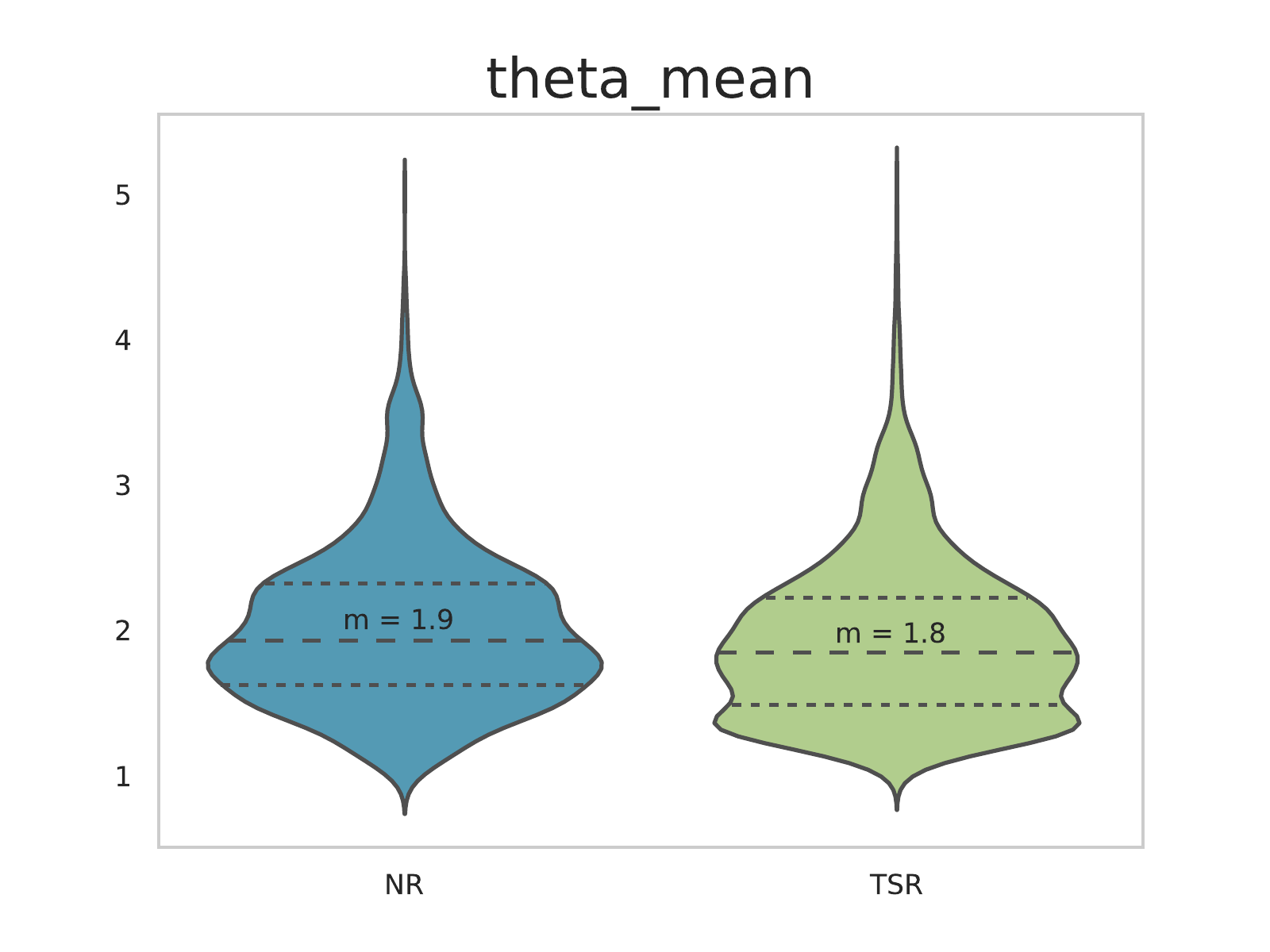} 
    \includegraphics[width=0.32\textwidth]{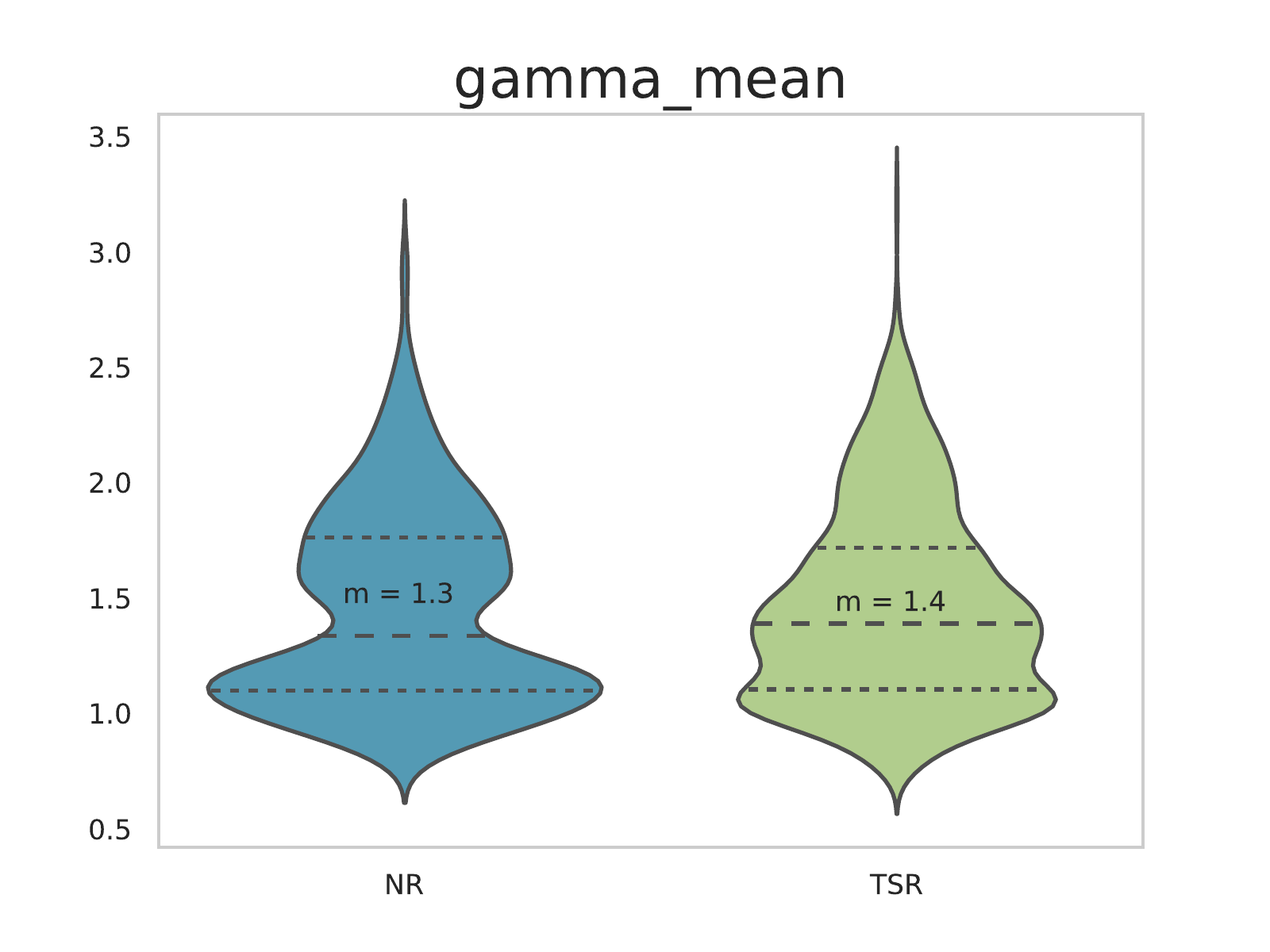} 
    \caption{Examples of feature distributions across all subjects for ZuCo 1.0.}
    \label{fig:z1-features}
\end{figure}

\begin{figure}[t]
    \centering
    \includegraphics[width=0.32\textwidth]{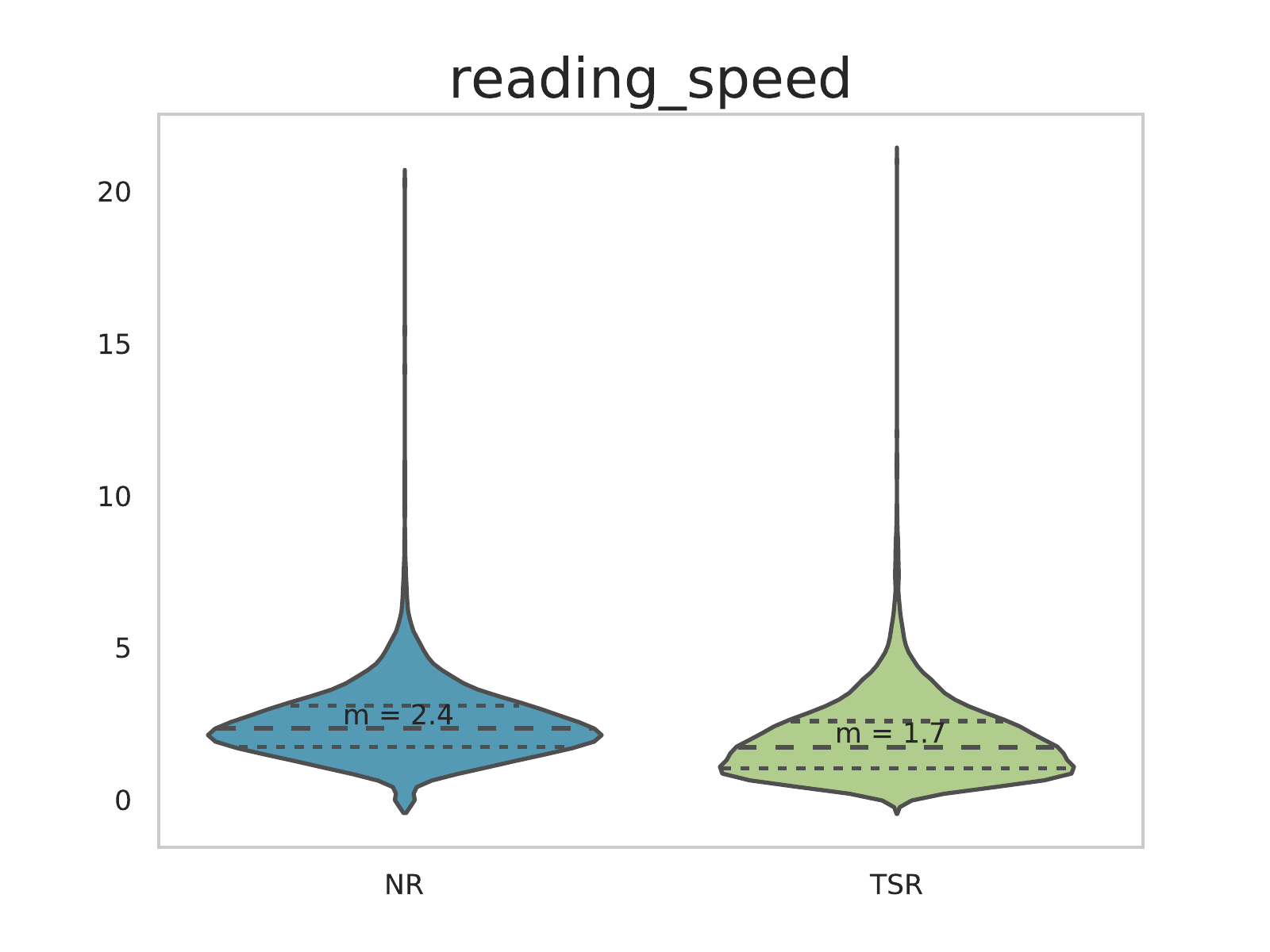}
    \includegraphics[width=0.32\textwidth]{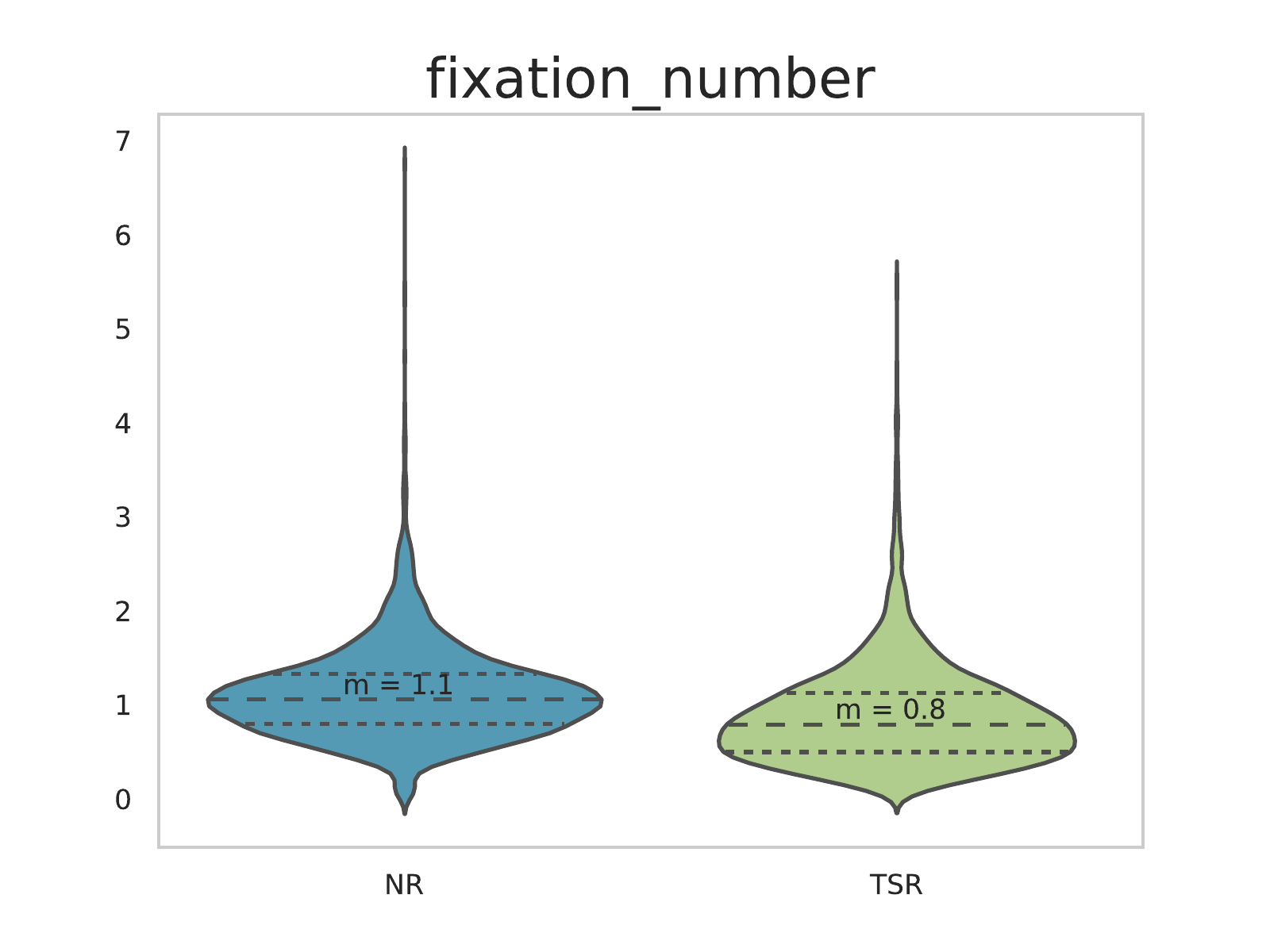}
    \includegraphics[width=0.32\textwidth]{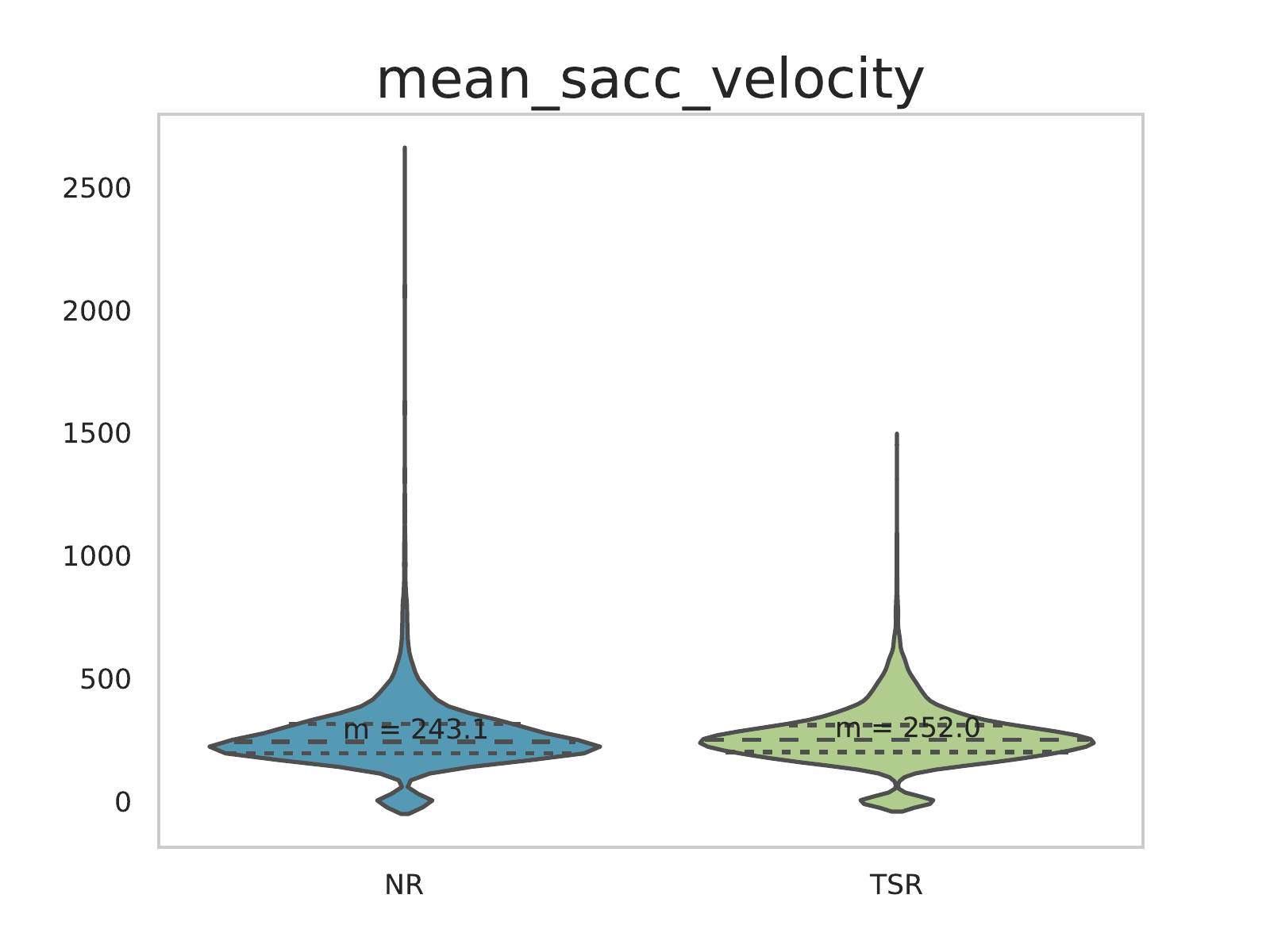}
    \includegraphics[width=0.32\textwidth]{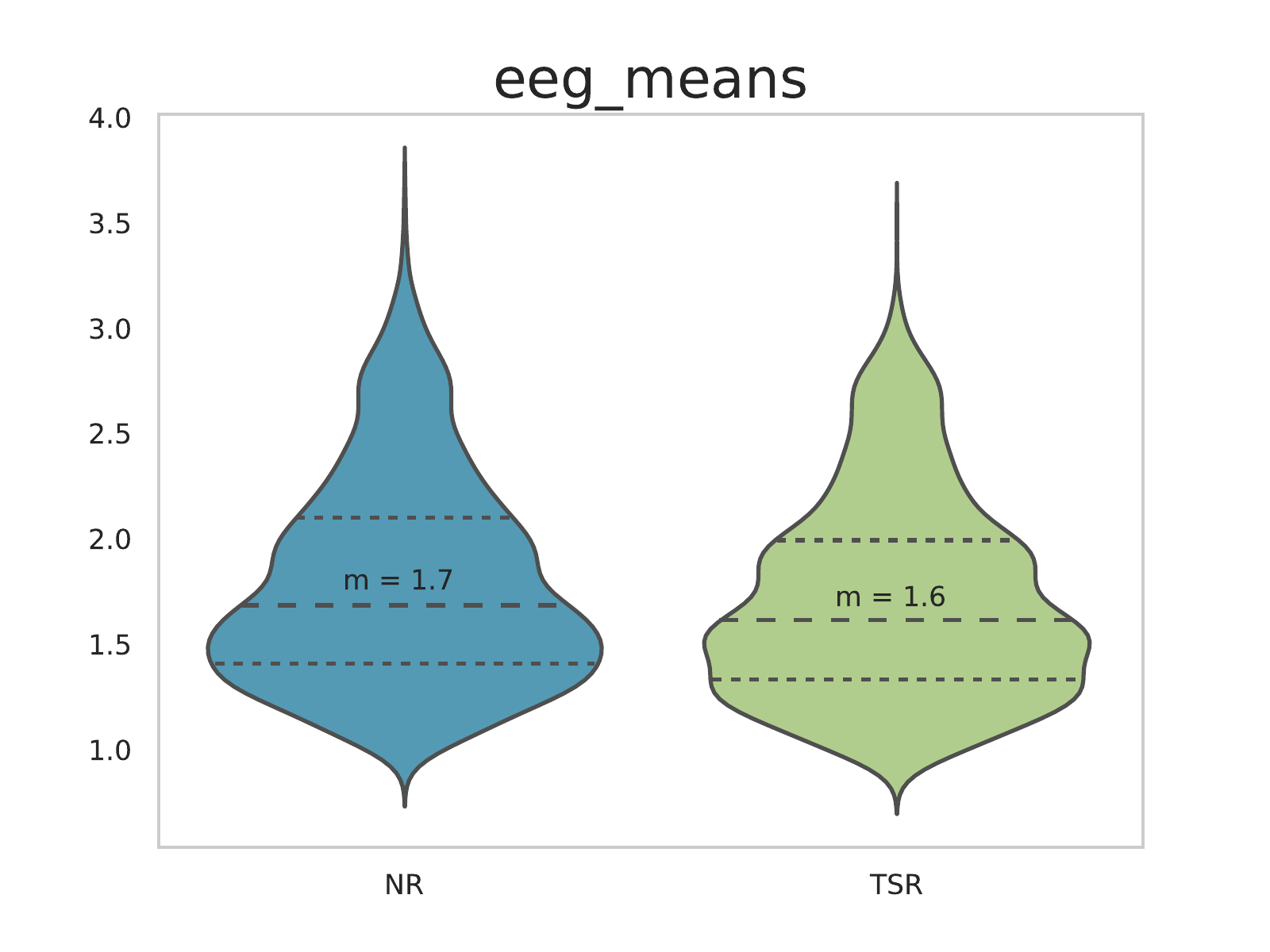}
    \includegraphics[width=0.32\textwidth]{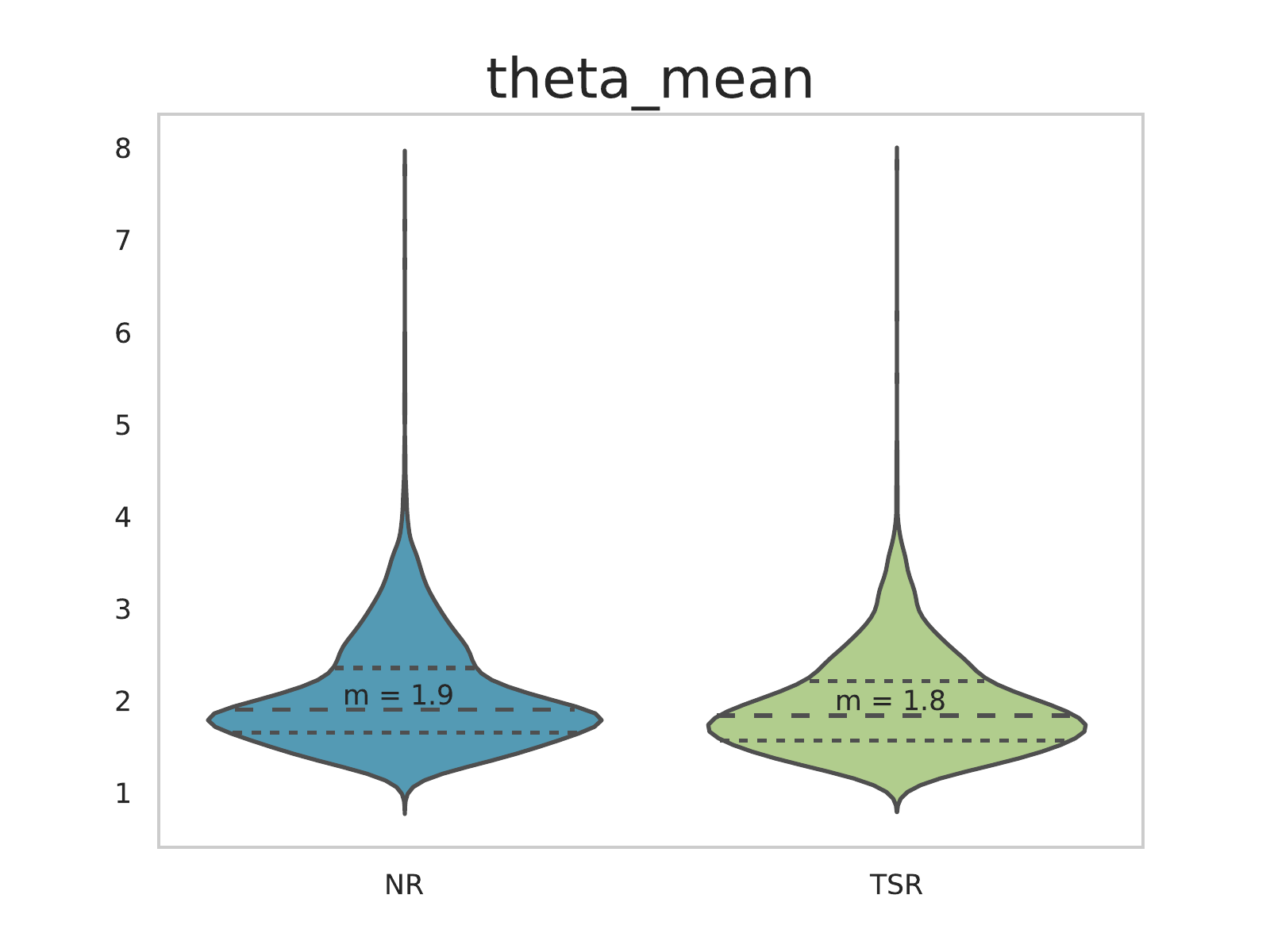} 
    \includegraphics[width=0.32\textwidth]{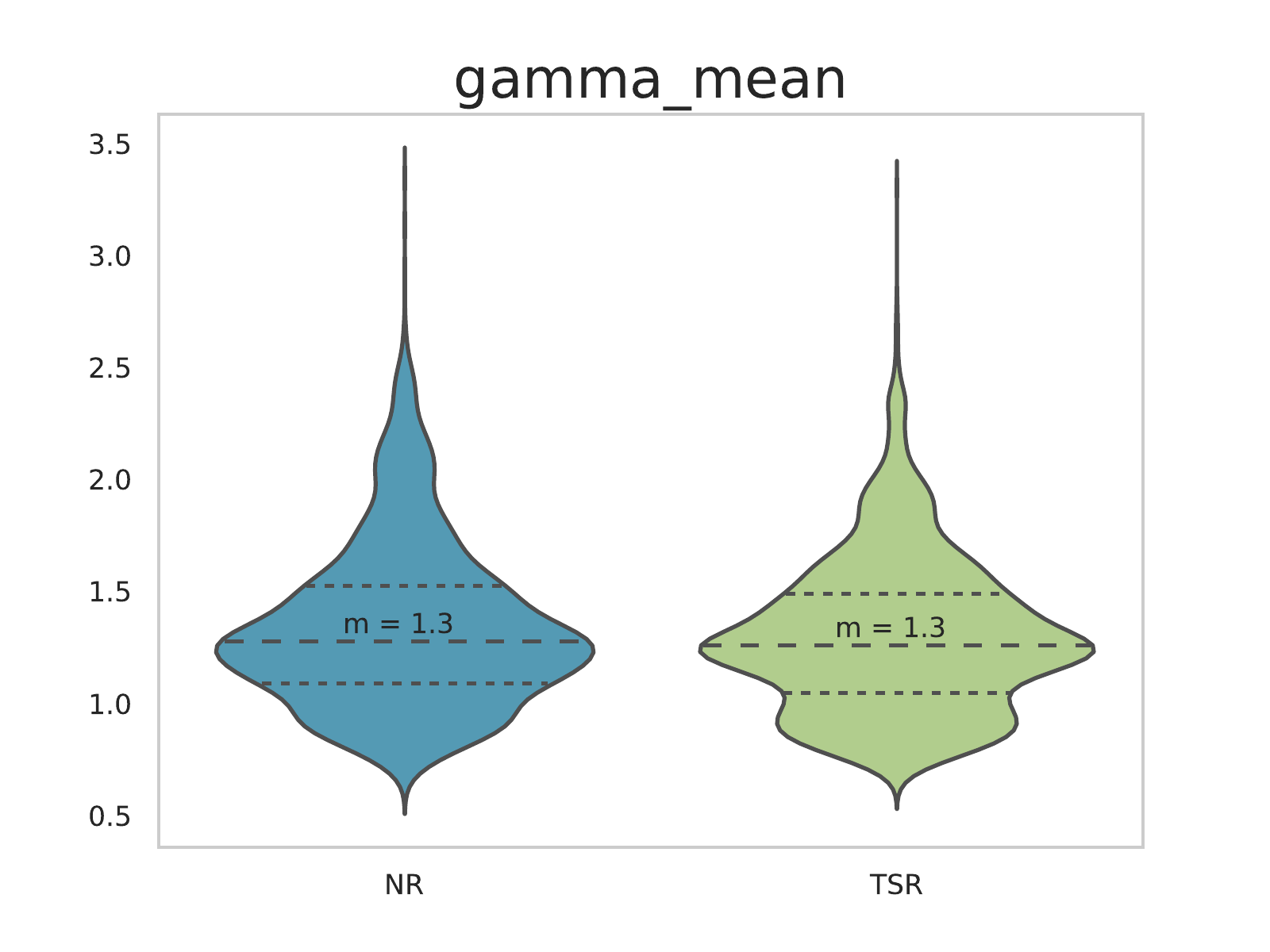} 
    \caption{Examples of feature distributions across all subjects for ZuCo 2.0.}
    \label{fig:z2-features}
\end{figure}

\noindent In addition to word-level features, we investigate the potential of using \linebreak sentence-level eye tracking and EEG features for the reading task classification. The advantages of sentence-level features consist of the possibility of using simpler machine learning models and reduced training times. Our goal is to investigate if sentence-level features provide the same accuracy as word-level features.
Sentence-level features are defined as metrics aggregated over all words in a given sentence. Therefore, every sentence is assigned a single feature value.

\paragraph{Eye-tracking Features} We include two types of sentence-level eye-tracking features. The features are summarized in Table \ref{tab:sent-et-feats}. First, the fixation-based features - omission rate, number of fixations and reading speed - are aggregated metrics normalized by sentence length, i.e., the number of words in a sentence. Analogous to the word-level models, we also include saccade-based features. These include the mean and maximum duration, velocity and amplitude across all saccades that occurred within the reading time of a give sentence. We test these features individually and combined to investigate the performance increase achieved by adding more features.
Examples of these features across all subjects, split by class (normal reading vs. task-specific reading) are shown in Figures \label{fig:z1-features} and \ref{fig:z2-features}, for ZuCo 1.0 and ZuCo 2.0, respectively.

\begin{table}[t]
\centering
\small
\begin{tabular}{llc}
\toprule
\textbf{Name} & \textbf{Definition} & \textbf{Values} \\\midrule
\textsc{Fixation Features} & & \\\midrule
omission\_rate & Percentage of words that is \textit{not} fixated in a sentence  & 1 \\\hline
fixation\_number & \begin{tabular}[c]{@{}l@{}}Number of fixations in the sentence \\ divided by the number of words   \end{tabular} & 1 \\\hline
reading\_speed & \begin{tabular}[c]{@{}l@{}}Sum of the duration of all fixations in the sentence \\ divided by the number of words in the sentence  \\ \end{tabular} & 1 \\\hline
sent\_gaze & \begin{tabular}[c]{@{}l@{}}Concatenation of the three features described \\ above\end{tabular} & 3 \\\midrule
\textsc{Saccade Features} & & \\\midrule
mean\_sacc\_dur & \begin{tabular}[c]{@{}l@{}}Sum of the duration of all saccades in the sentence \\ divided by the number of words  \end{tabular} & 1 \\\hline
max\_sacc\_dur & \begin{tabular}[c]{@{}l@{}}Maximum saccade duration per sentence  \end{tabular} & 1 \\\hline
mean\_sacc\_velocity & \begin{tabular}[c]{@{}l@{}}Sum of the velocity of all saccades in the sentence \\ divided by the number of saccades \end{tabular} & 1 \\\hline
max\_sacc\_velocity & \begin{tabular}[c]{@{}l@{}}Maximum saccade velocity per sentence \end{tabular} & 1 \\\hline
mean\_sacc\_amplitude & \begin{tabular}[c]{@{}l@{}}Sum of the amplitude of all saccades in the sentence \\ divided by the number of saccades \end{tabular} & 1 \\\hline
max\_sacc\_amplitude & \begin{tabular}[c]{@{}l@{}}Maximum saccade amplitude per sentence \end{tabular} & 1 \\\hline
sent\_saccade & \begin{tabular}[c]{@{}l@{}}Concatenation of the six features described \\ above\end{tabular} & 6 \\\midrule
\textsc{Combined Features} & & \\\midrule
sent\_gaze\_sacc & Concatenation of sent\_gaze and sent\_sacc & 9 \\
\bottomrule
\end{tabular}
\caption{Sentence-level eye-tracking features. \textit{Name} denotes the variable names as used in the dataset.}
\label{tab:sent-et-feats}
\end{table}

\begin{table}[t]
\centering
\small
\begin{tabular}{llc}
\toprule
\textbf{Name} & \textbf{Definition} & \textbf{Values} \\\midrule
\textsc{Mean Features} & & \\\midrule
theta\_mean & \begin{tabular}[c]{@{}l@{}}Mean theta band power averaged \\ over all electrodes\end{tabular} & 1 \\\hline
alpha\_mean & \begin{tabular}[c]{@{}l@{}}Mean alpha band power averaged \\ over all electrodes \end{tabular} & 1 \\\hline
beta\_mean & \begin{tabular}[c]{@{}l@{}}Mean beta band power averaged \\ over all electrodes\end{tabular} & 1 \\\hline
gamma\_mean & \begin{tabular}[c]{@{}l@{}}Mean gamma band power averaged \\ over all electrodes\end{tabular} & 1 \\\hline
eeg\_means & \begin{tabular}[c]{@{}l@{}}Mean frequency band features averaged \\ over all electrodes, resulting in 1 feature \\ value for each of the 8 frequency bands  \end{tabular} & 8 \\\midrule
\textsc{Electrode Features} & & \\\midrule
electrode\_features\_theta & \begin{tabular}[c]{@{}l@{}}Mean theta1 and theta2 values of \\ all 104 electrodes \end{tabular} & 104 \\\hline
electrode\_features\_alpha & \begin{tabular}[c]{@{}l@{}}Mean alpha1 and alpha2 values of \\ all 104 electrodes \end{tabular} & 104 \\\hline
electrode\_features\_beta & mean(sent.mean\_b1, sent-mean\_b2) & 105 \\\hline
electrode\_features\_gamma & mean(sent.mean\_g1, sent-mean\_g2) & 105 \\\hline
electrode\_features\_all & \begin{tabular}[c]{@{}l@{}}Concatenation of the four features above \end{tabular} & 420 \\
\bottomrule
\end{tabular}
\caption{Sentence-level EEG features. \textit{Name} denotes the variable names as used in the dataset.}
\label{tab:sent-eeg-feats}
\end{table}

\paragraph{EEG Features} The sentence-level EEG features take into account the EEG activity over the whole sentence duration (even when no words were fixated). We aggregate over the pre-processed EEG signals of the full reading duration of a sentence. The sentence-level EEG features are described in Table \ref{tab:sent-eeg-feats}.

\paragraph{Model} The input to the sentence-level model is a single vector representing each sentence. We scale the feature values to a range between $\{-1,1\}$.
We train a support vector machine for classification with a linear kernel. We use the \texttt{scikit-learn} SVC implementation\footnote{\url{https://scikit-learn.org/stable/modules/generated/sklearn.svm.SVC.html}}. 
For the within-subject evaluation, we average the results over 50 runs with different random seeds, hence the training and test data are shuffled and split into different sets at every run. The test data is always unseen during training. The models are trained on 90\% and tested on 10\% of all samples of a single subject during each run.
For the cross-subject evaluation, the models are trained on all samples from $n-1$ subjects and tested on the samples from the left-out subject. Hence, only 1 run is necessary since the samples in the test set remain the same.

\subsection{Baseline Methods}

\noindent We compare all models against three baselines: (i) a random baseline, i.e., chance level of 50\% for binary classification, (ii) a word embedding baseline, and (iii) a text difficulty baseline.

\paragraph{Word Embedding Baseline} We compare our models to a textual baseline as a sanity check to ensure the sentences in the data are not easily separable merely by their linguistic characteristics. For this purpose, we use pre-trained textual representations, namely the state-of-the-art contextualized BERT word embeddings \citep{devlin2019bert}. We concatenate the embeddings of all words in a sentence feed the into the LSTM model. This word embedding baseline yield a classification accuracy of 58\% and 65\% for ZuCo 1.0 and ZuCo 2.0, respectively.

\begin{figure}[t]
    \centering
    \includegraphics[width=0.45\textwidth]{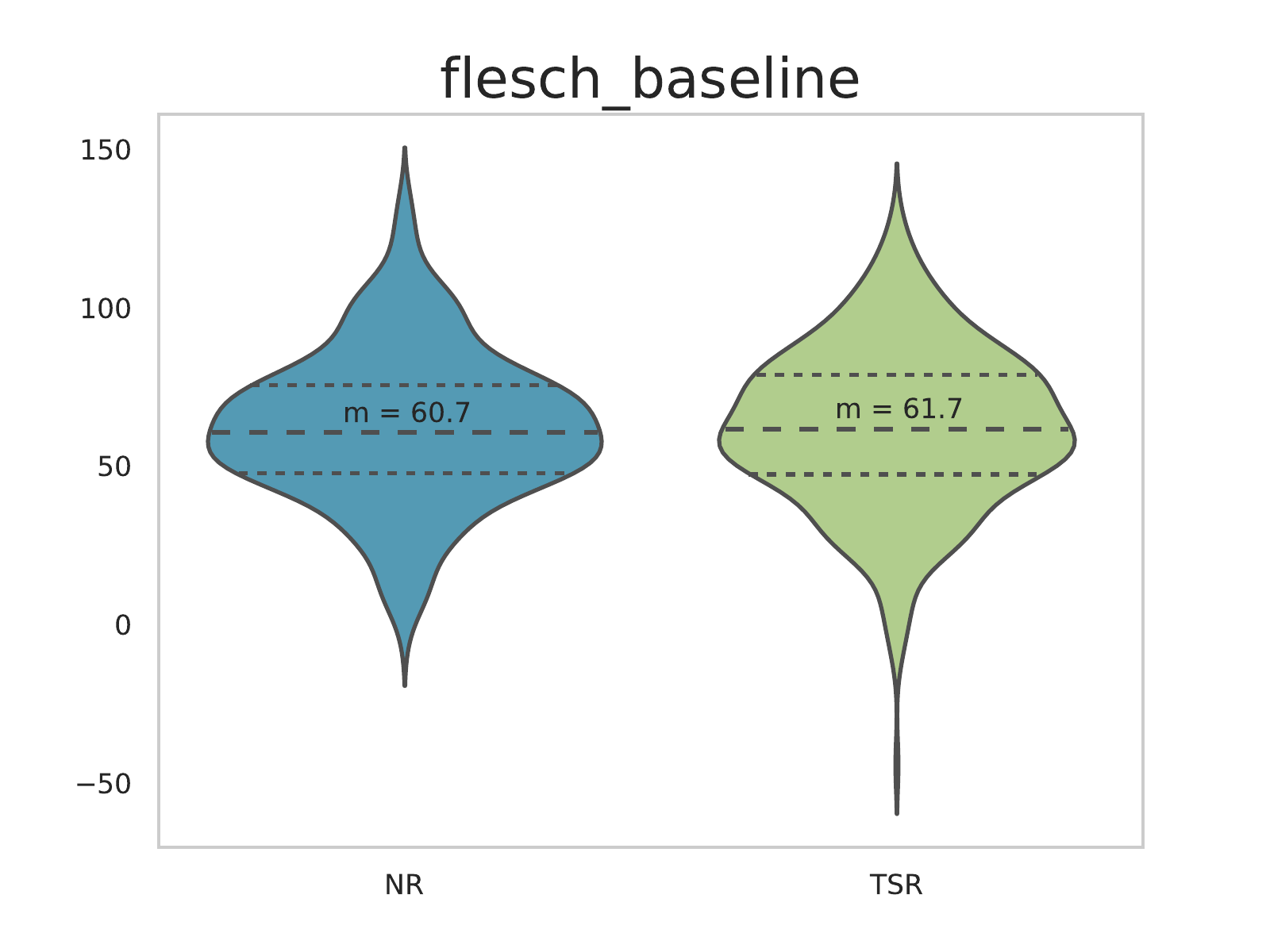} 
    \includegraphics[width=0.45\textwidth]{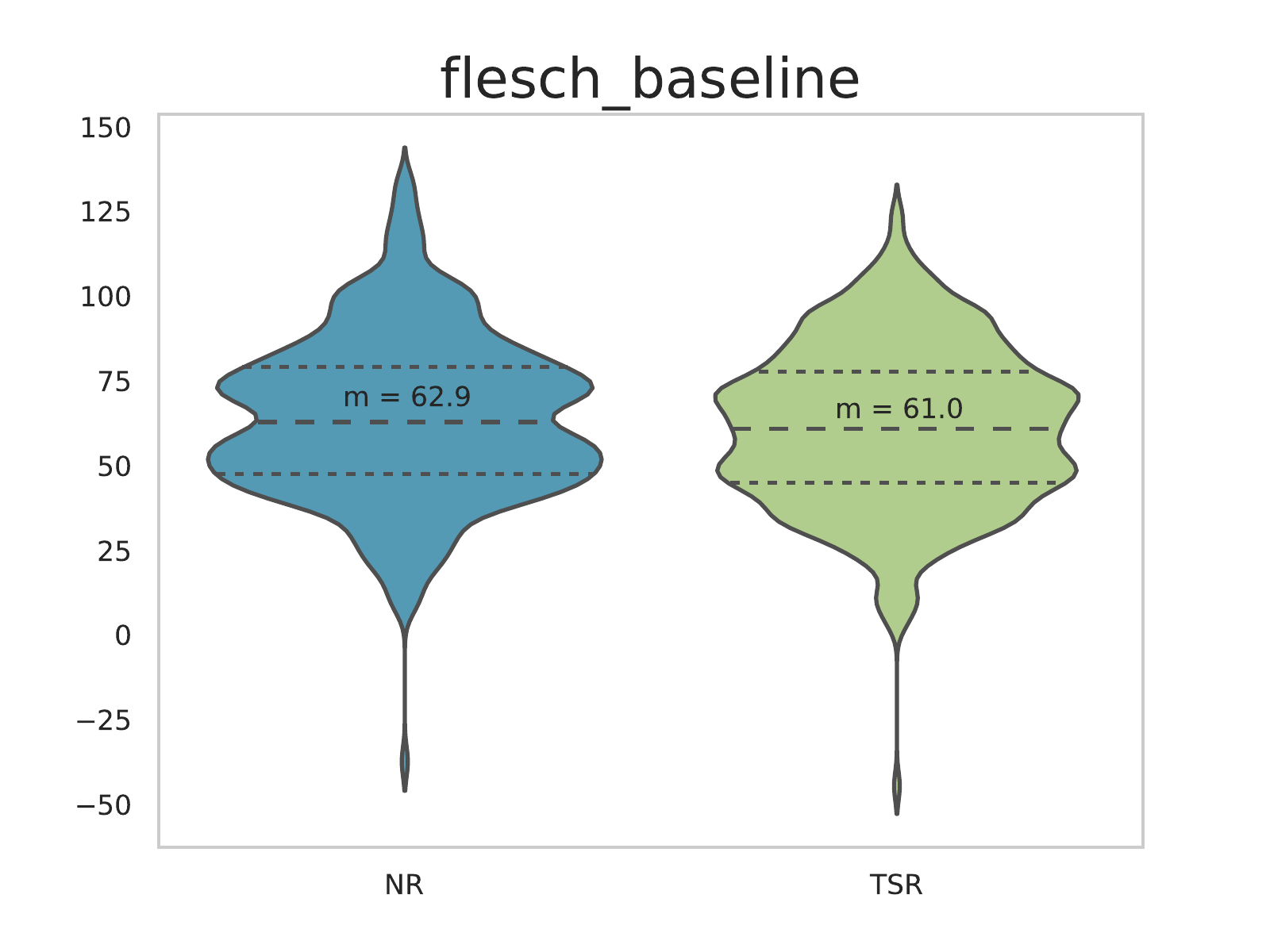} 
    \caption{Flesch reading ease (FRE) scores for ZuCo 1.0. (left) and ZuCo 2.0 (right).}
    \label{fig:flesch-features}
\end{figure}

\paragraph{Text Difficulty Baseline} We also provide a baseline based on text readability. Although the sentences for both reading tasks were chosen to be of similar length and from the same text genre, we want to ensure that both tasks are not separable merely by the difficulty of the sentences. Therefore, we implement a text difficulty baseline, which classifies the sentences into NR and TSR based on their Flesch reading ease score (FRE; \citealt{flesch1948new}). This score indicates how difficult an English text passage is to understand based on the average number of words in a sentence and the average number of syllables in a word:

\begin{align}
    FRE = x - y  \left( \frac{words}{sentences} \right) - z \left( \frac{ syllables}{words} \right)
\end{align}

\noindent where $x$, $y$ and $z$ are language-specific weighting factors (for English $x=206.835$, $y=1.015$, $z=84.6$). We compute FRE scores for each of the English sentences in the ZuCo data. Figure \ref{fig:flesch-features} shows the distribution of the FRE across the sentences of ZuCo 1.0 and ZuCo 2.0. This baseline is also above random performance with a classification accuracy of 58\% for ZuCo 1.0 and 53\% for ZuCo 2.0. Therefore, we also report it in our results.

\clearpage
\section{Results}\label{sec:results}

\noindent In the following, we present the results of both word-level and sentence-level classification models. We present the results on all EEG and eye-tracking feature sets in two settings: within-subject and cross-subjects evaluations. We report all results for both ZuCo datasets. For ZuCo 1.0, we used the data from 11 subjects, and for ZuCo 2.0 from 16 subjects. 

\subsection{Word-level Results}

\subsubsection{Within-subject Evaluation}

\begin{table}[t]
\centering
\small
\begin{tabular}{lcccc}
\toprule
 & \multicolumn{2}{c}{ZuCo 1.0} & \multicolumn{2}{c}{ZuCo 2.0} \\
\textbf{feature set} & \textbf{median} & \textbf{MAD} & \textbf{median} & \textbf{MAD} \\\midrule
text difficulty baseline & 0.58 & - & 0.53 & - \\
word embedding baseline & 0.58 & - & 0.65 & -\\\midrule
eye\_tracking & 0.75 & 0.07 & \textbf{0.65} & 0.08 \\
eye\_tracking (w/sacc) & \textbf{0.75} & 0.05 & \textbf{0.65} & 0.04 \\\midrule
eeg\_raw & 0.88 & 0.06 & 0.64 & 0.05\\
eeg\_theta & 0.98 & 0.01 & 0.62 & 0.07 \\
eeg\_alpha & 0.98 & 0.01 & 0.62 & 0.07 \\
eeg\_beta & \textbf{0.99} & 0.01 & 0.63 & 0.08 \\
eeg\_gamma & \textbf{0.99} & 0.00 & \textbf{0.67} & 0.10 \\\bottomrule
\end{tabular}
\caption{Summary of all word-level within-subject results. The best results per category are marked in bold.}
\label{tab:word-level-res-summary}
\end{table}

\noindent Table \ref{tab:word-level-res-summary} summarizes all word-level within-subject results for both eye-tracking and EEG features. We report the median and median absolute deviation (MAD) of accuracies across all subjects.

\begin{figure}[t!]
    \centering
    \includegraphics[width=0.48\textwidth]{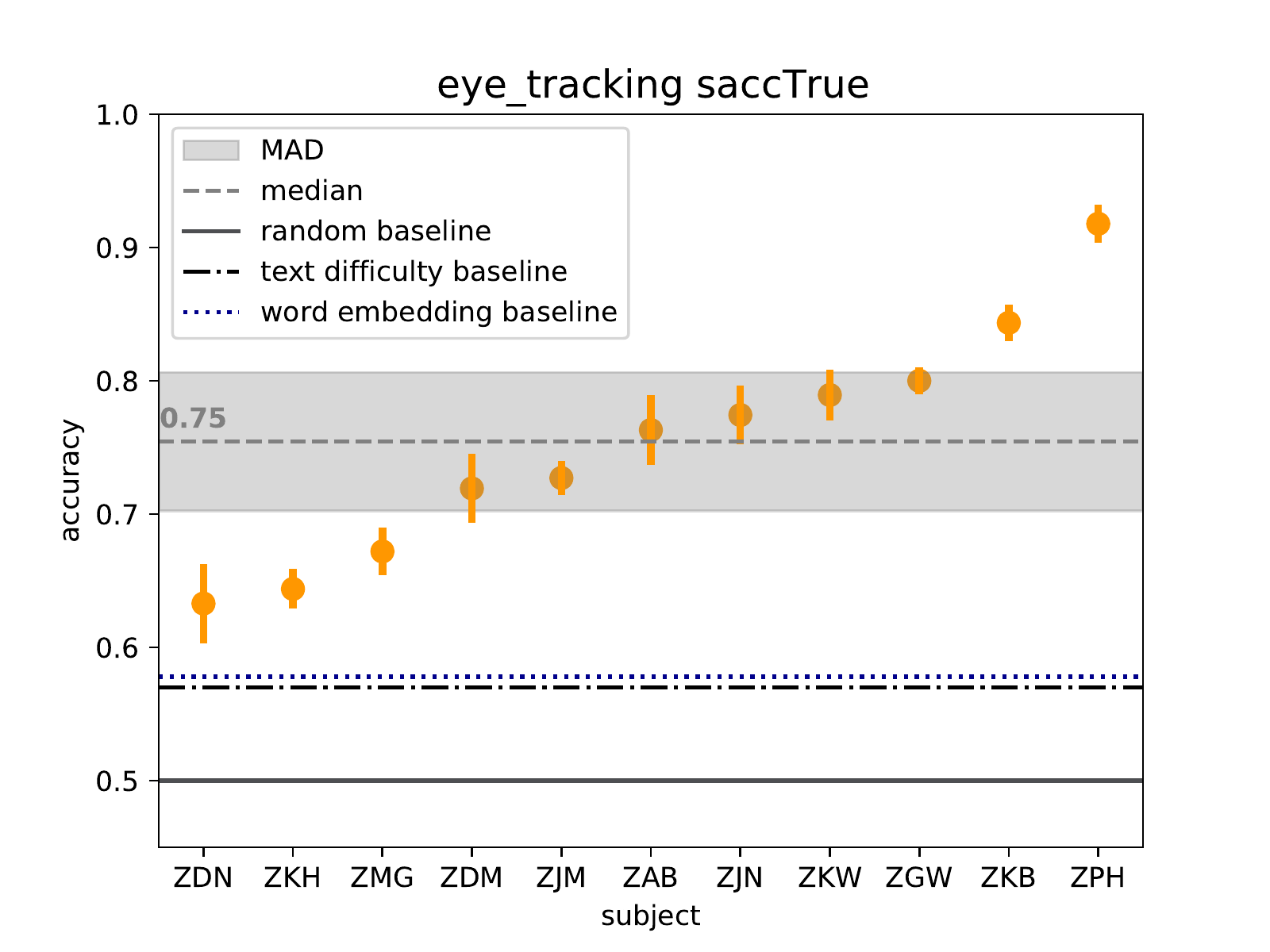} 
    \includegraphics[width=0.49\textwidth]{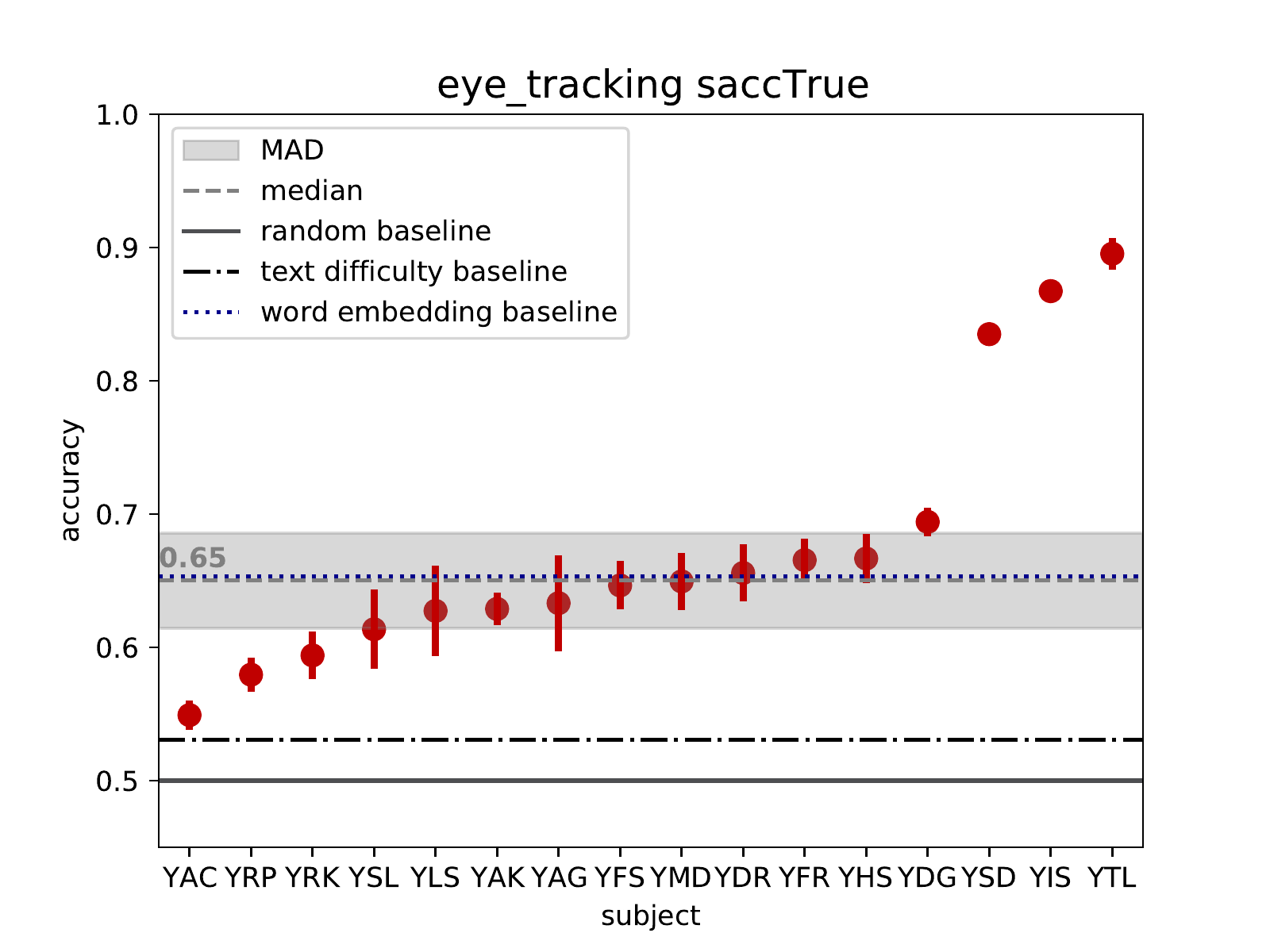} 
    \caption{Classification accuracy for each subject on word-level eye-tracking features on ZuCo 1.0 (left) and ZuCo 2.0 (right). }
    \label{fig:word-res-z1-z2-et}
\end{figure}

Figure \ref{fig:word-res-z1-z2-et} show the results of the LSTM models using word-level eye-tracking features including saccade features for ZuCo 1.0 and 2.0. The results without saccade features per subject were almost identical. All within-subject eye-tracking models achieve a performance higher than chance. The median accuracy without saccade features for ZuCo 1.0 is 75\% and for ZuCo 2.0 is 65\%. For both datasets, the results show that the saccade features do not increase the accuracy, but the variance between multiple runs of individual subjects as well as across all subjects is reduced by including the saccade features.

\begin{figure}[t!]
    \centering
    \includegraphics[width=0.49\textwidth]{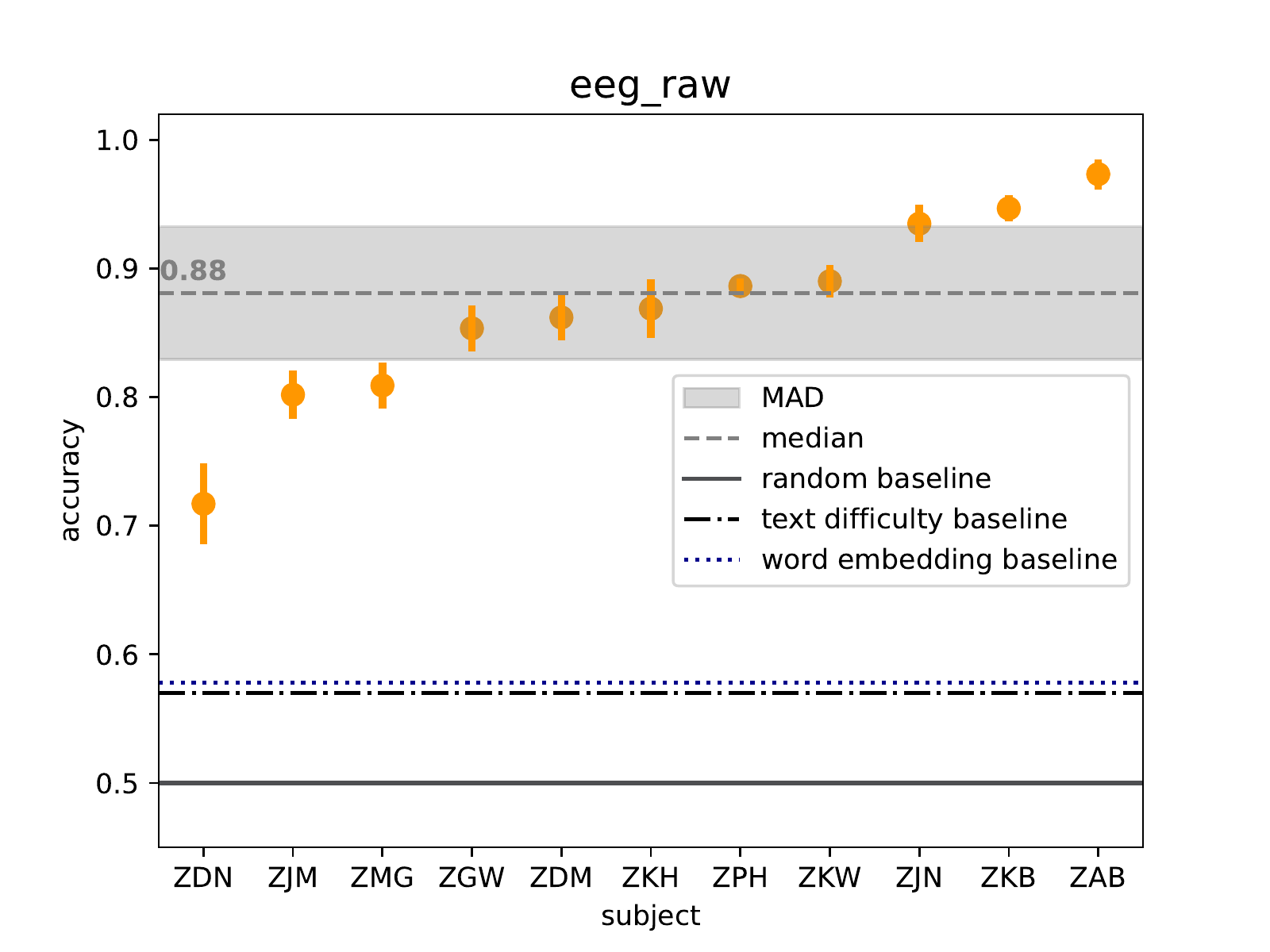} 
    \includegraphics[width=0.49\textwidth]{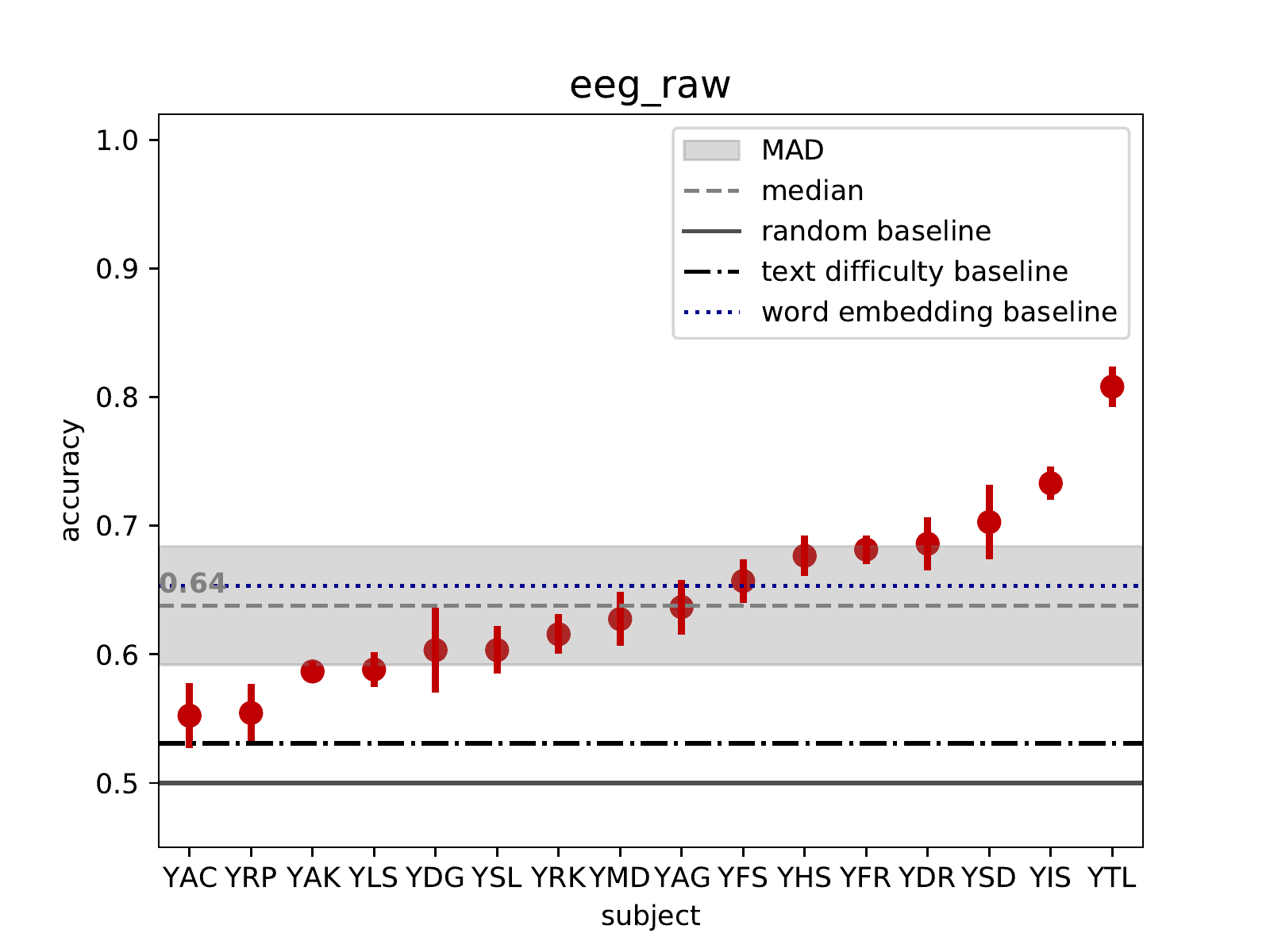} 
    \caption{Classification accuracy for each subject on word-level full broadband EEG features on ZuCo 1.0 (left) and ZuCo 2.0 (right).}
    \label{fig:word-res-eegraw}
\end{figure}

Figure \ref{fig:word-res-eegraw} presents the results achieved by training word-level models on the full broadband EEG signals (\textit{raw\_EEG}). The models yield substantially higher classification accuracy when trained and tested on data from ZuCo 1.0 subjects.

\begin{figure}[h]
    \centering
    \includegraphics[width=0.49\textwidth]{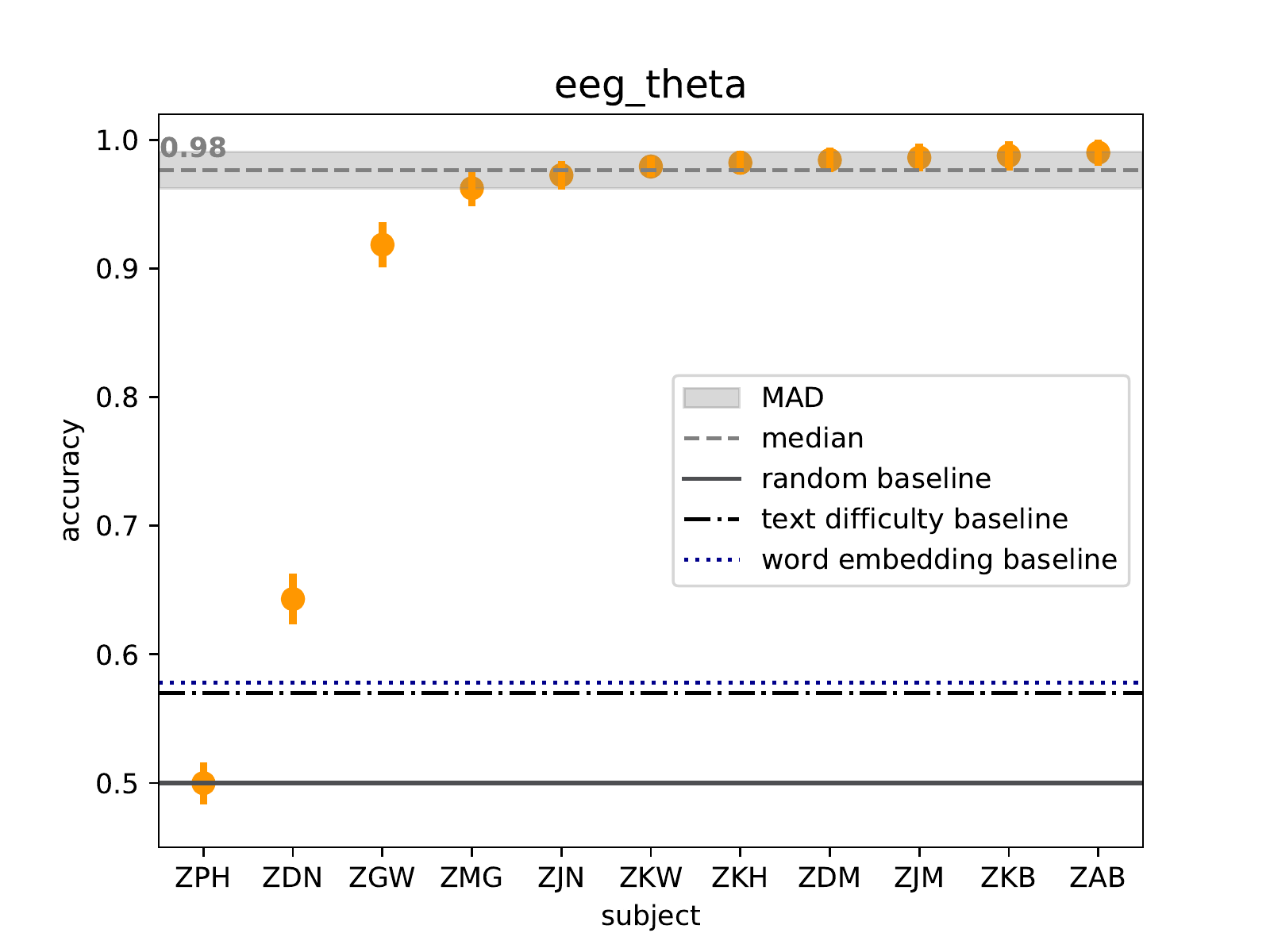} 
    \includegraphics[width=0.49\textwidth]{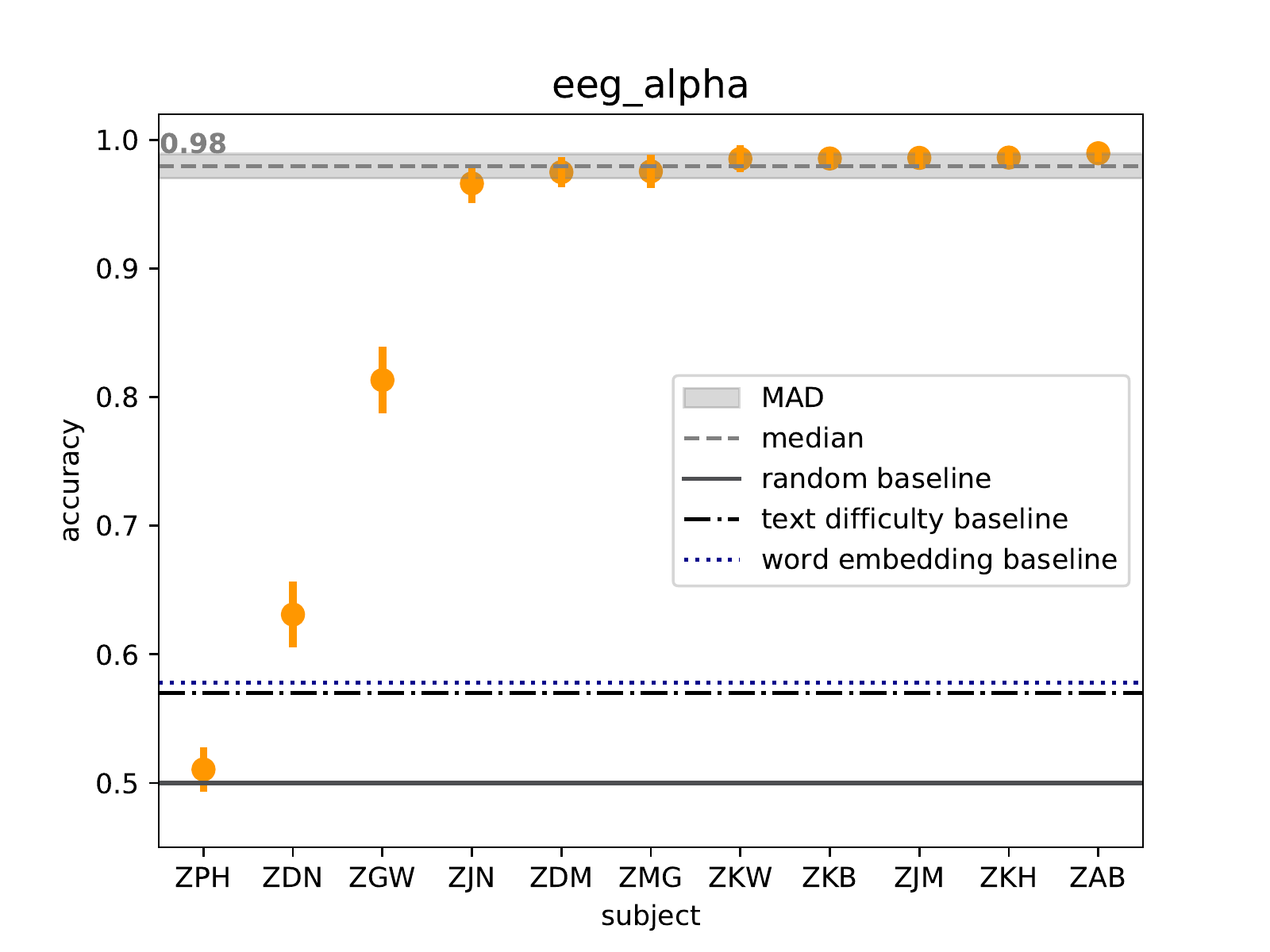}
    \includegraphics[width=0.49\textwidth]{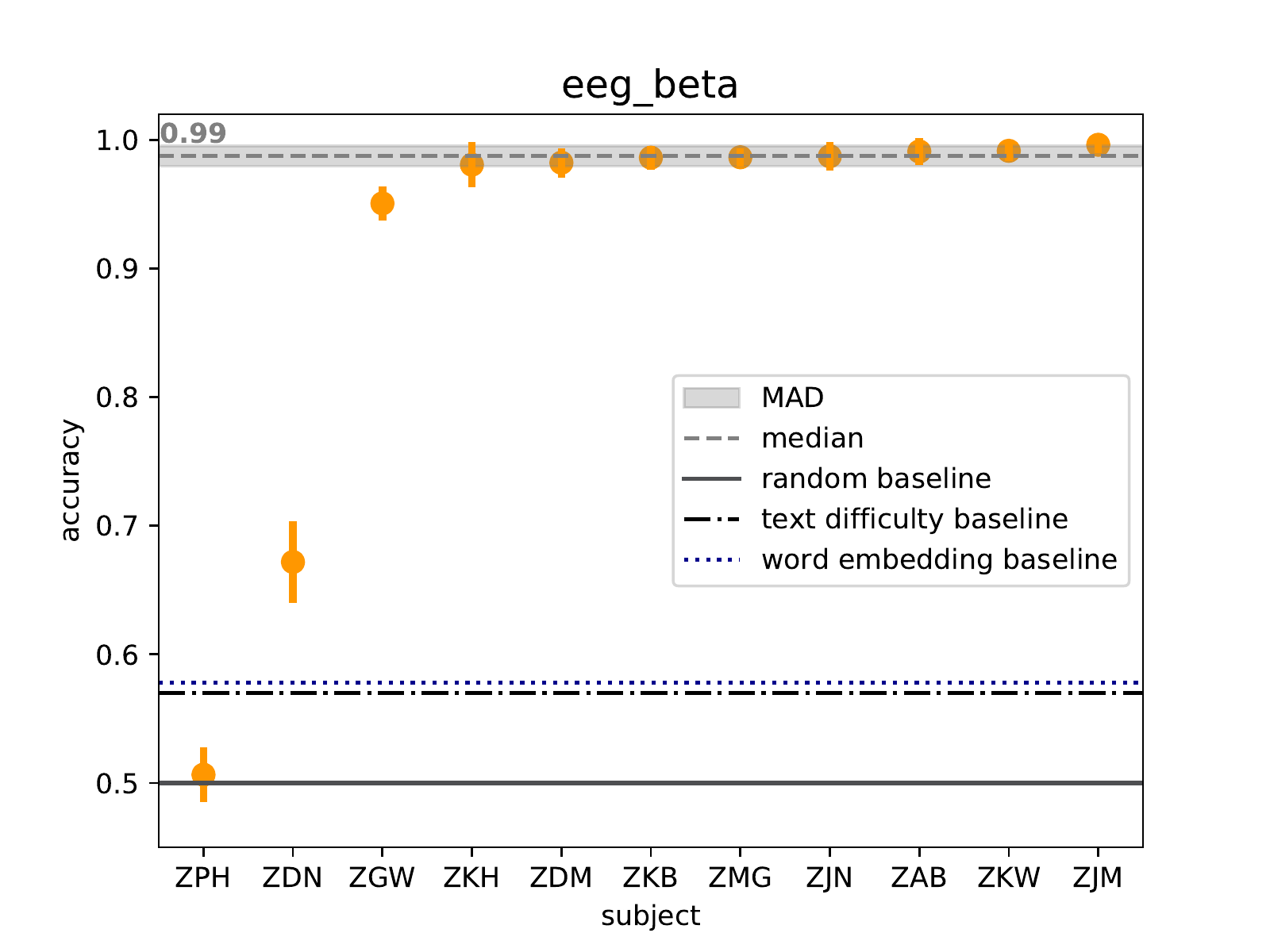}
    \includegraphics[width=0.49\textwidth]{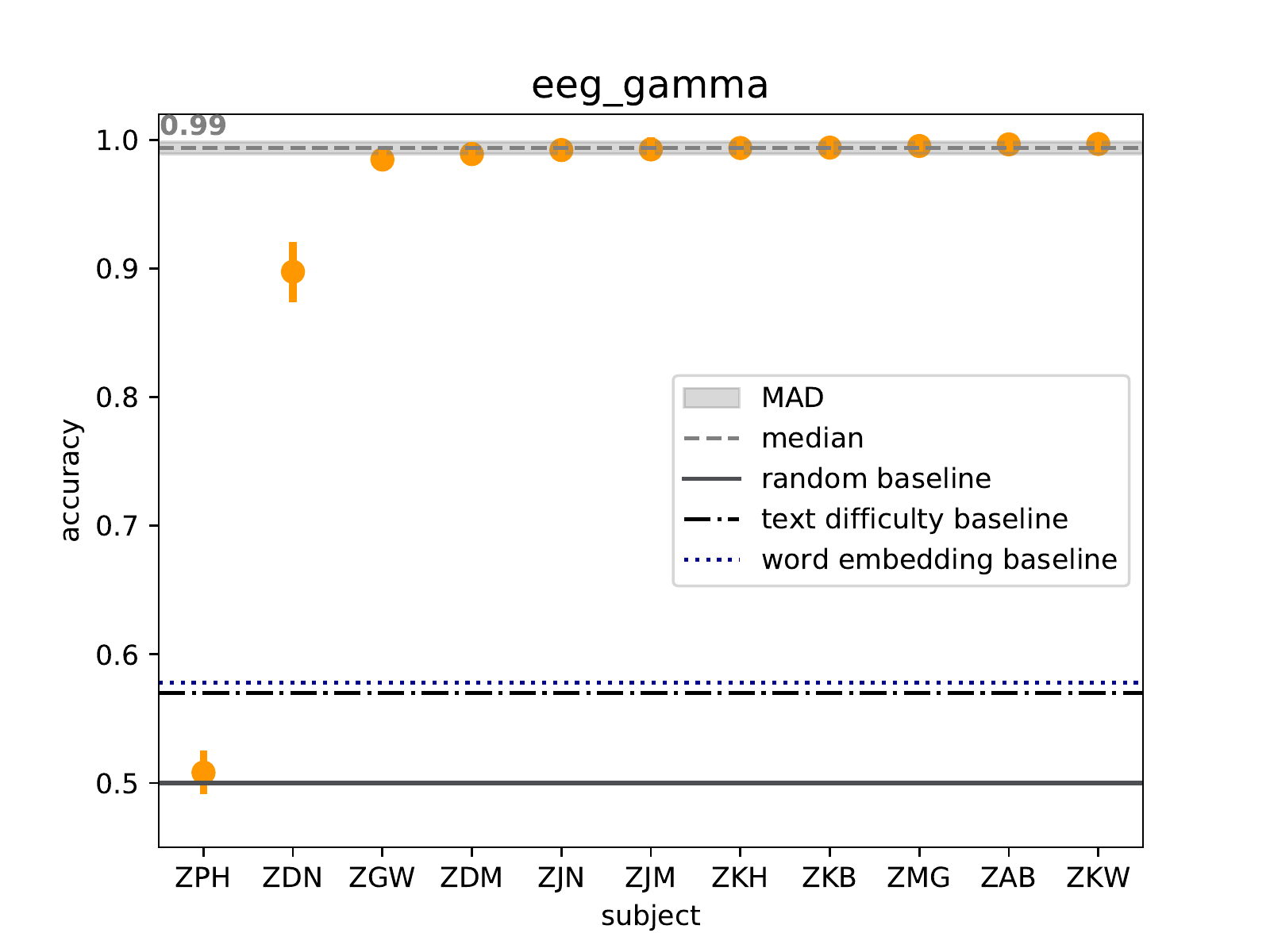}
    \caption{Classification accuracy for within-subject models on word-level EEG frequency band features on ZuCo 1.0.}
    \label{fig:word-res-z1-eeg-freq}
\end{figure}

\begin{figure}[h]
    \centering
    \includegraphics[width=0.49\textwidth]{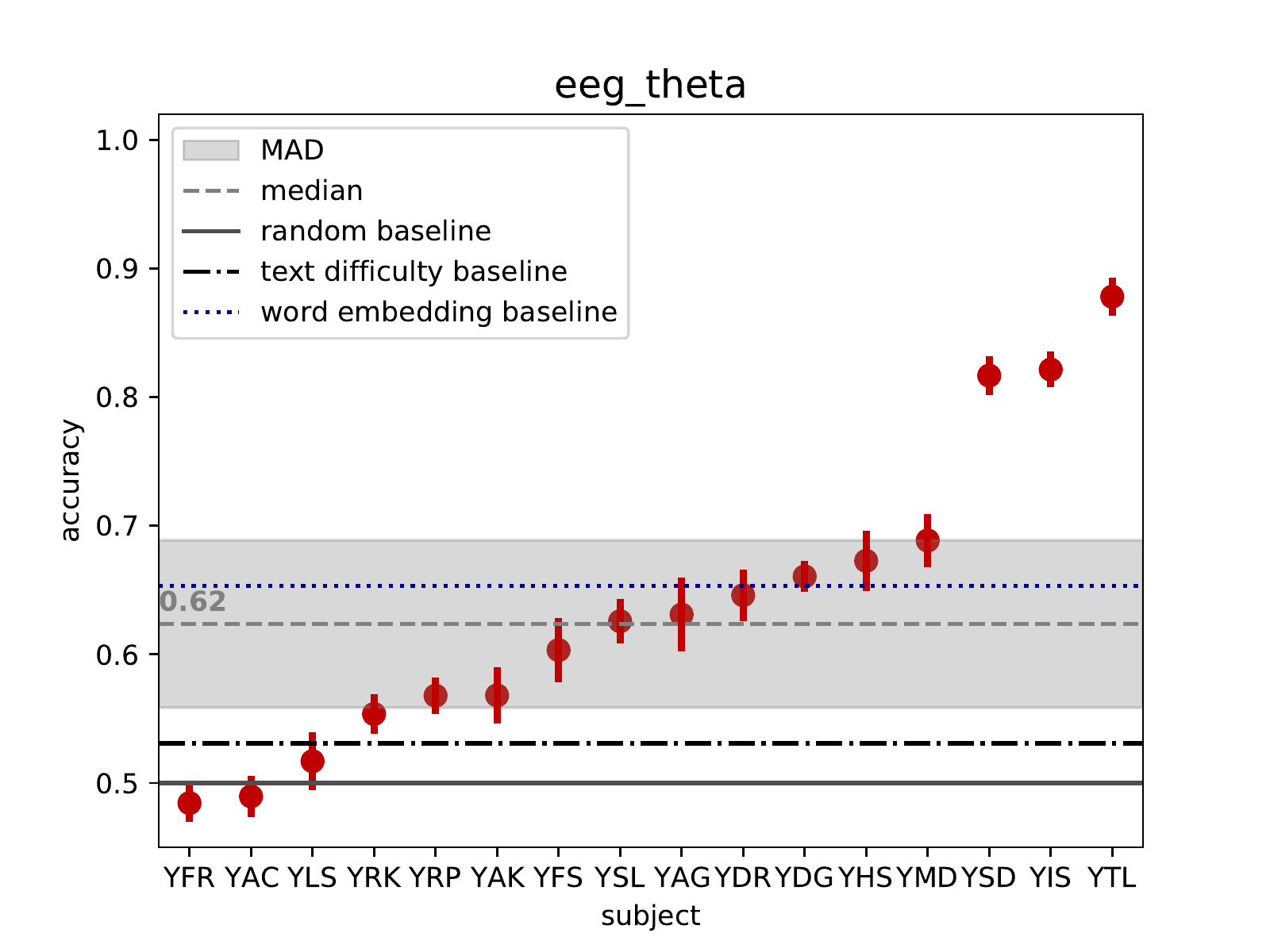} 
    \includegraphics[width=0.49\textwidth]{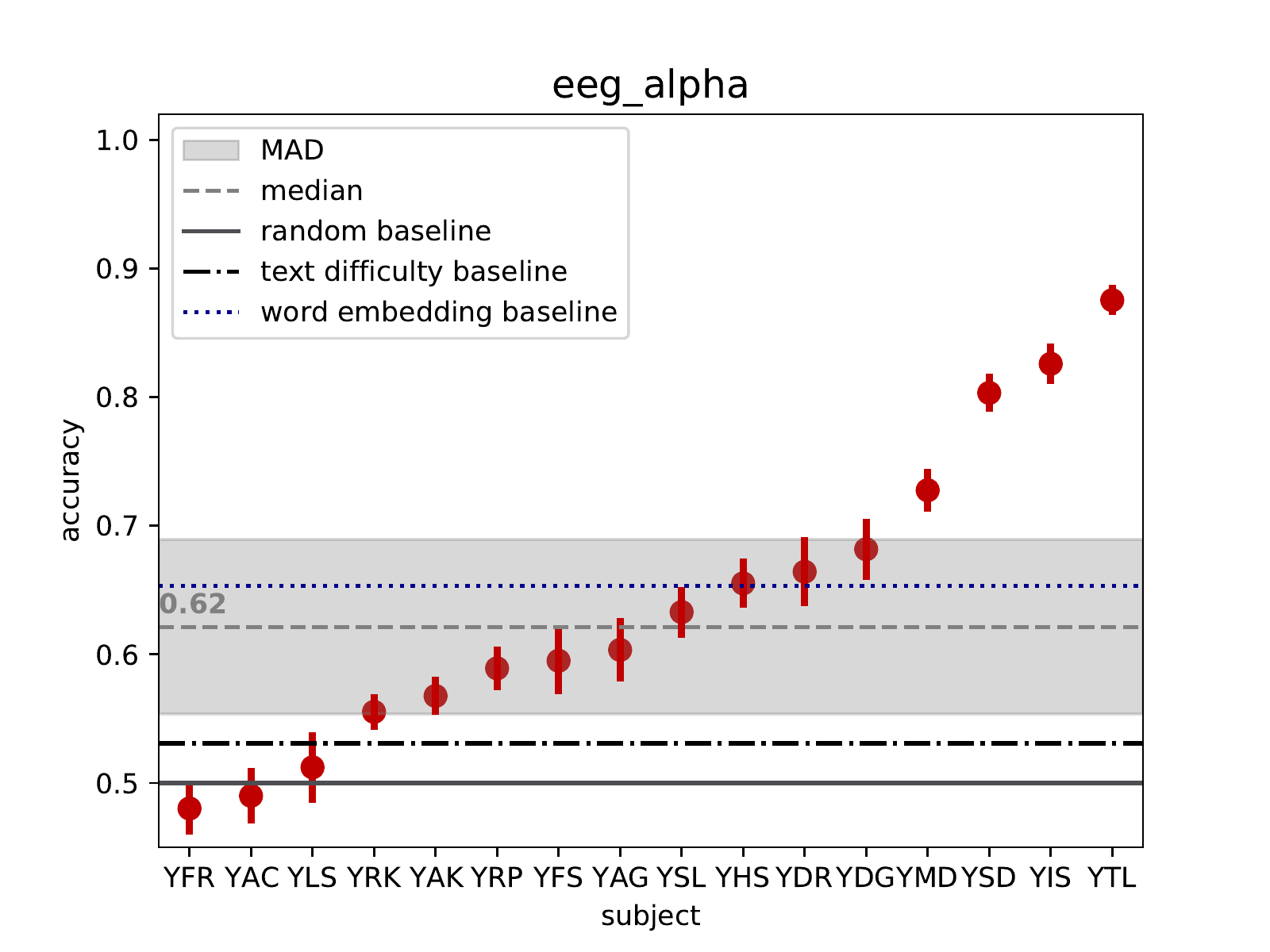}
    \includegraphics[width=0.49\textwidth]{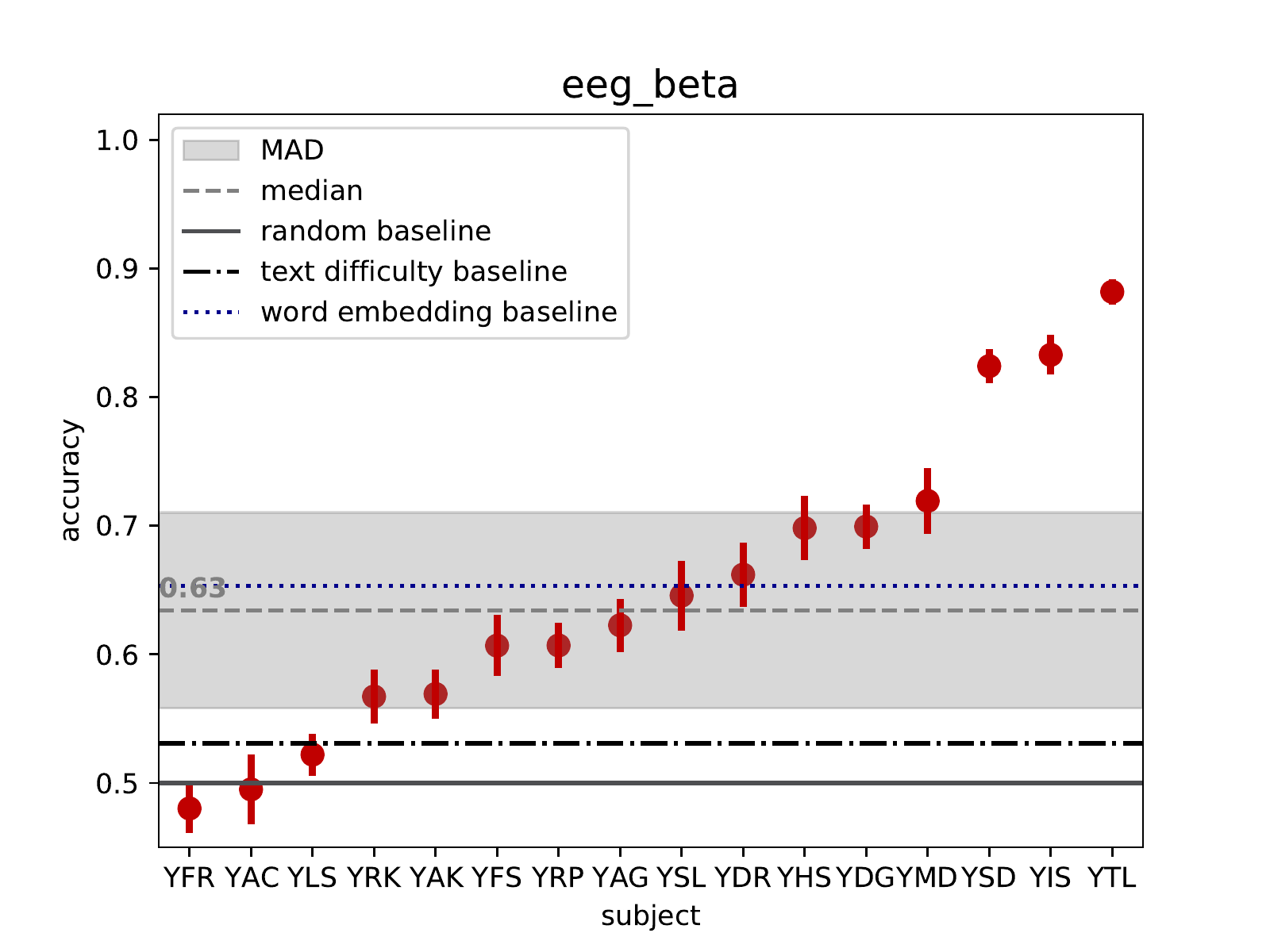}
    \includegraphics[width=0.49\textwidth]{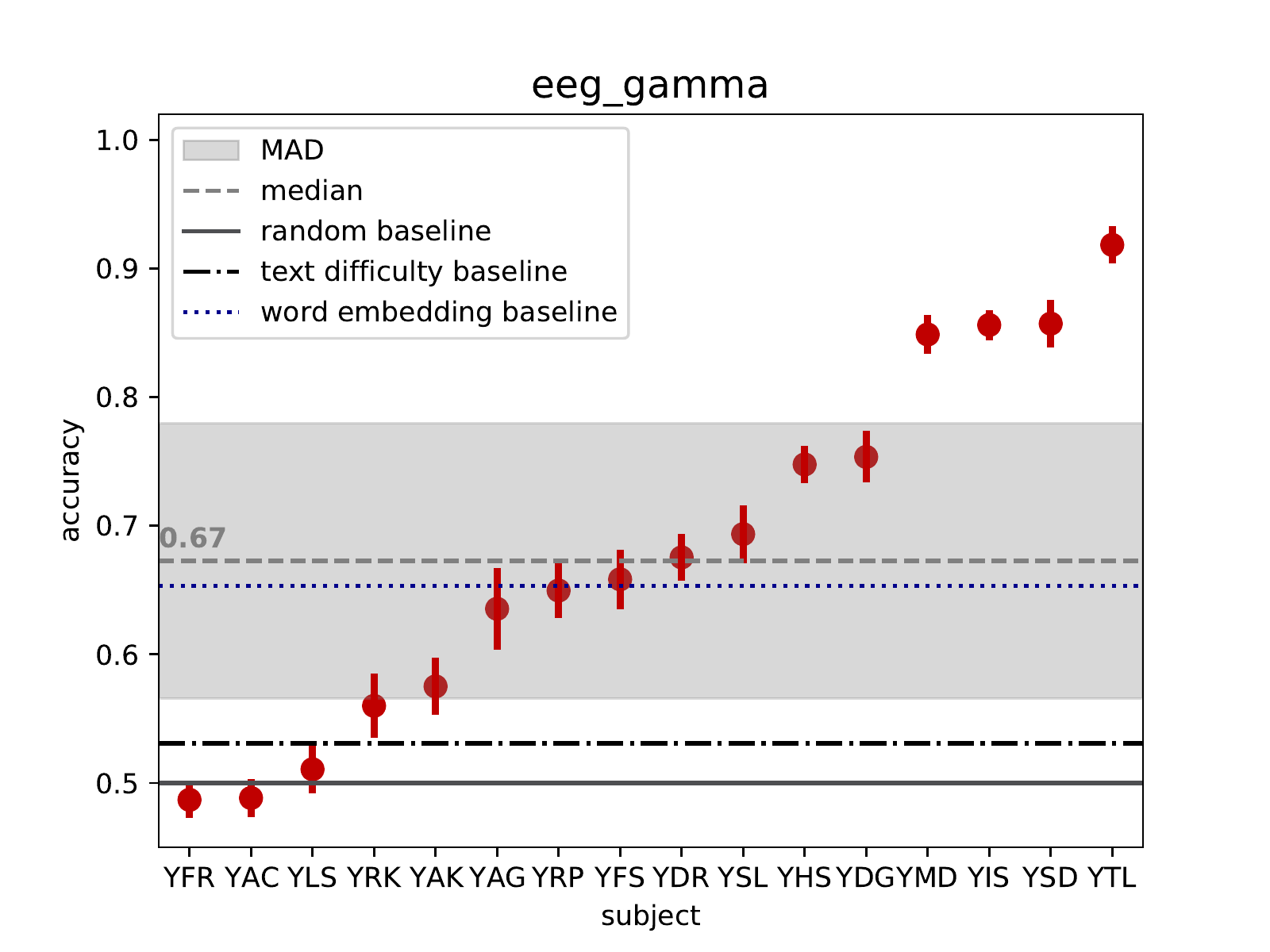}
    \caption{Classification accuracy for each within-subject model on word-level EEG frequency band features on ZuCo 2.0.}
    \label{fig:word-res-z2-eeg-freq}
\end{figure}

The detailed EEG results for all four frequency bands are presented in Figures \ref{fig:word-res-z1-eeg-freq} and \ref{fig:word-res-z2-eeg-freq}. We observe near-perfect results (98\% accuracy) on all four frequency bands for most subjects of the ZuCo 1.0 dataset, while for ZuCo 2.0 the results show much larger mean absolute deviations and merely 67\% accuracy for the best frequency band (i.e., gamma). Generally, all sentence-level feature sets perform better on ZuCo 1.0 than on ZuCo 2.0. This might be due to the session effect, which we discuss in Section \ref{sec:control}.

\subsubsection{Leave-one-out Cross-subject Evaluation}

\begin{figure}[ht]
    \centering
    \includegraphics[width=0.49\textwidth]{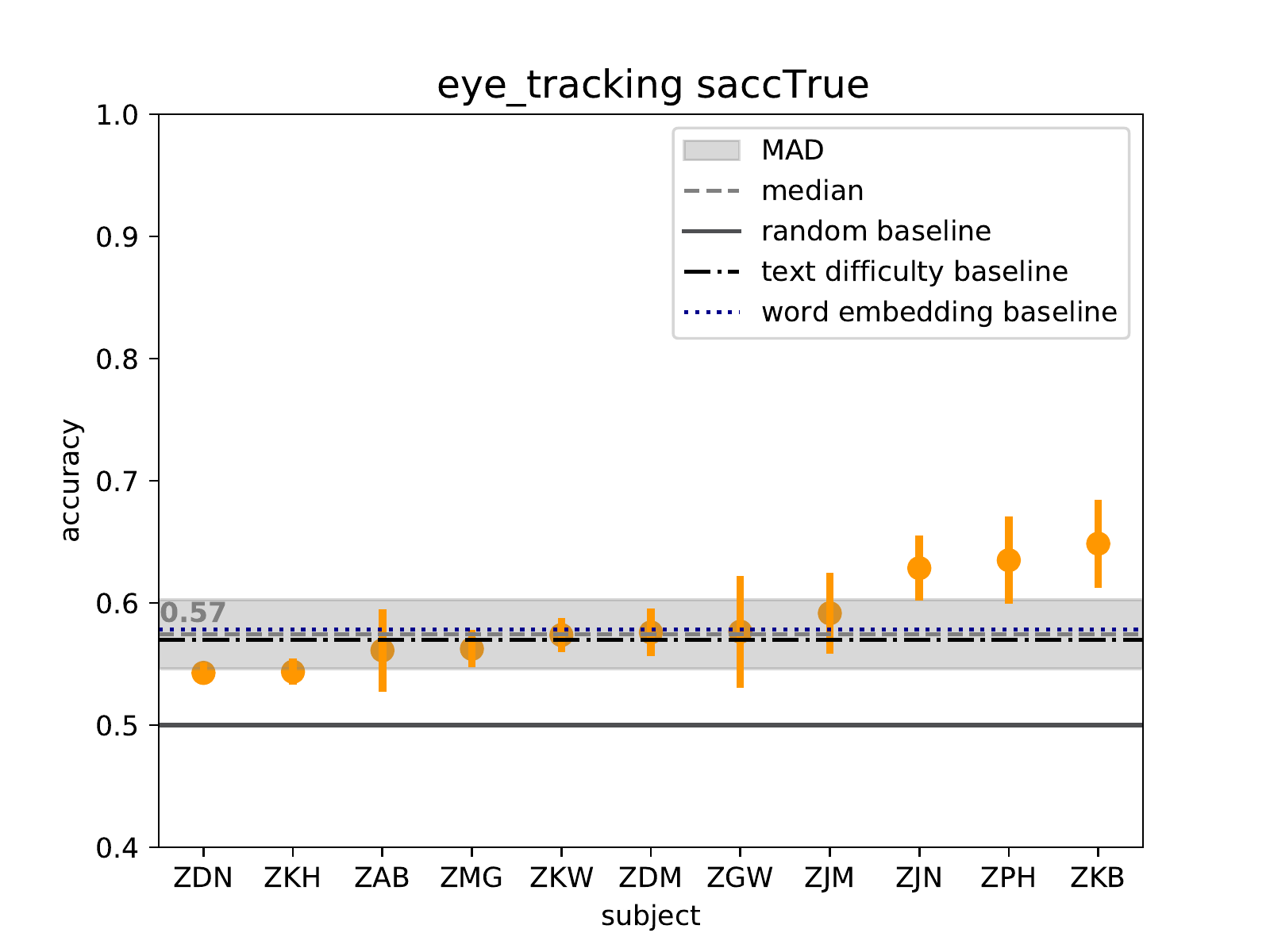} 
    \includegraphics[width=0.49\textwidth]{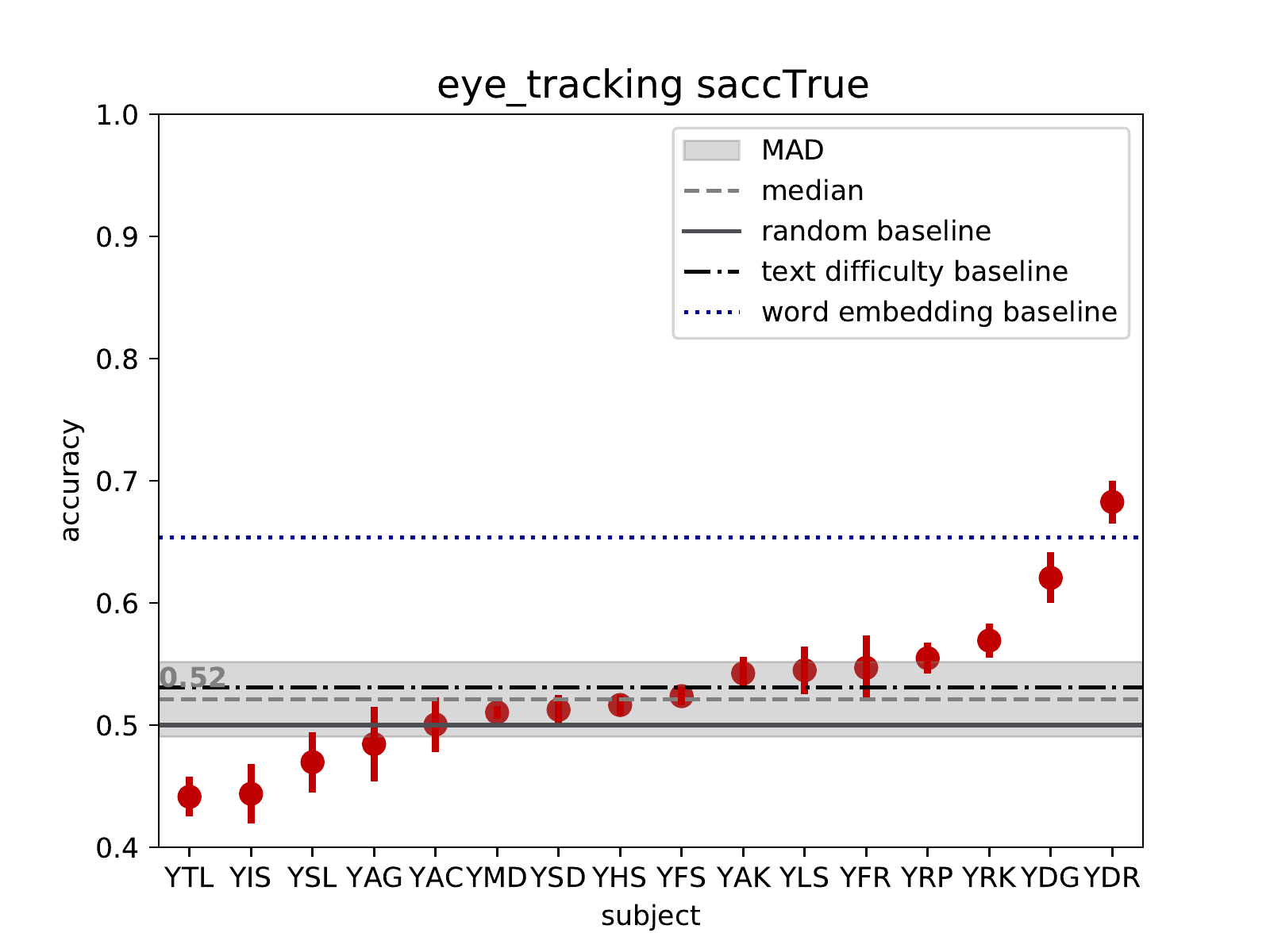} 
    \caption{Accuracy results of the cross-subject eye-tracking word-level classification including saccade features on ZuCo 1.0 (left) and ZuCo 2.0 (right).}
    \label{fig:word-res-et-cross}
\end{figure}

\begin{figure}[ht]
    \centering
    \includegraphics[width=0.49\textwidth]{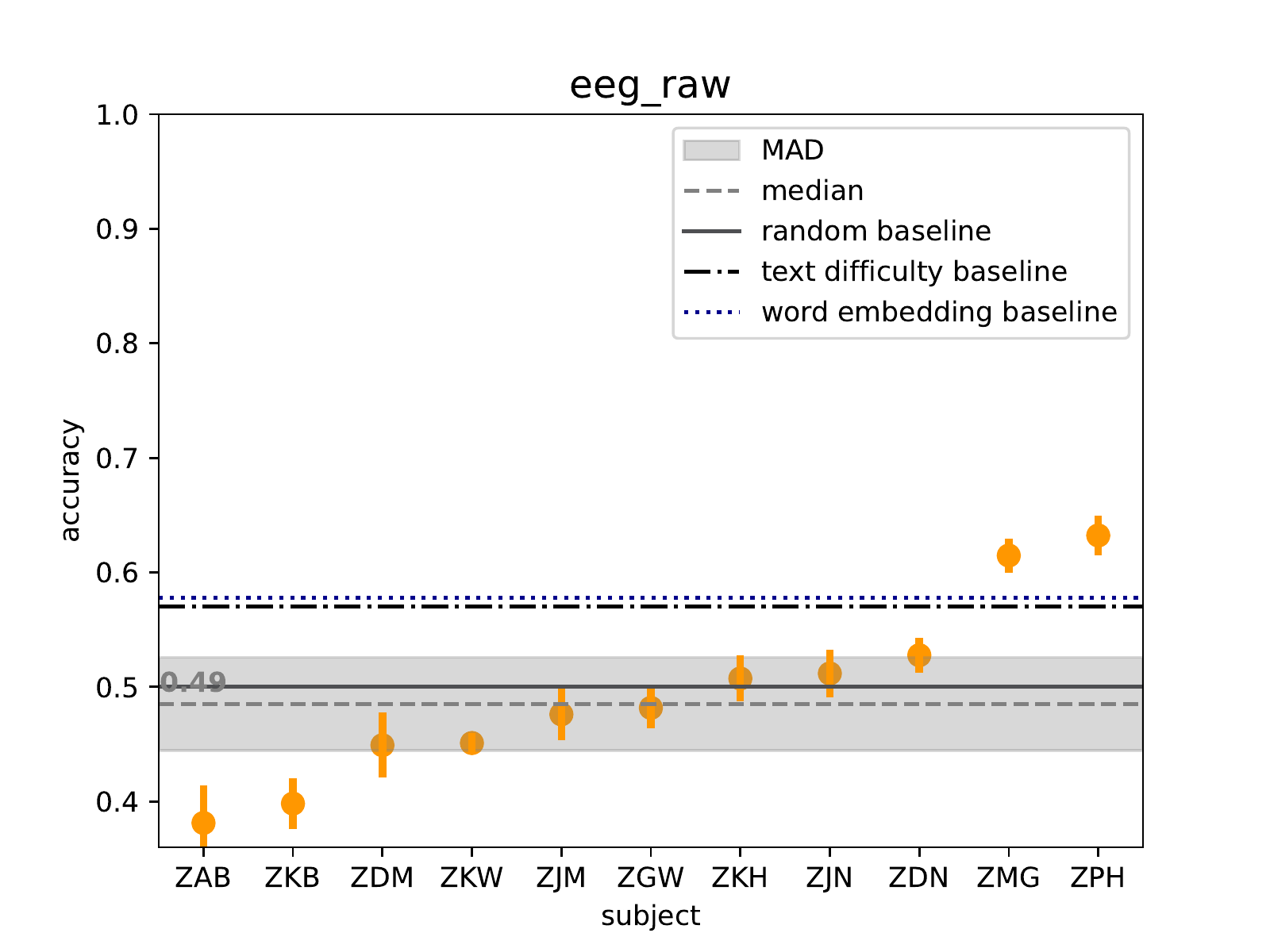} 
    \includegraphics[width=0.49\textwidth]{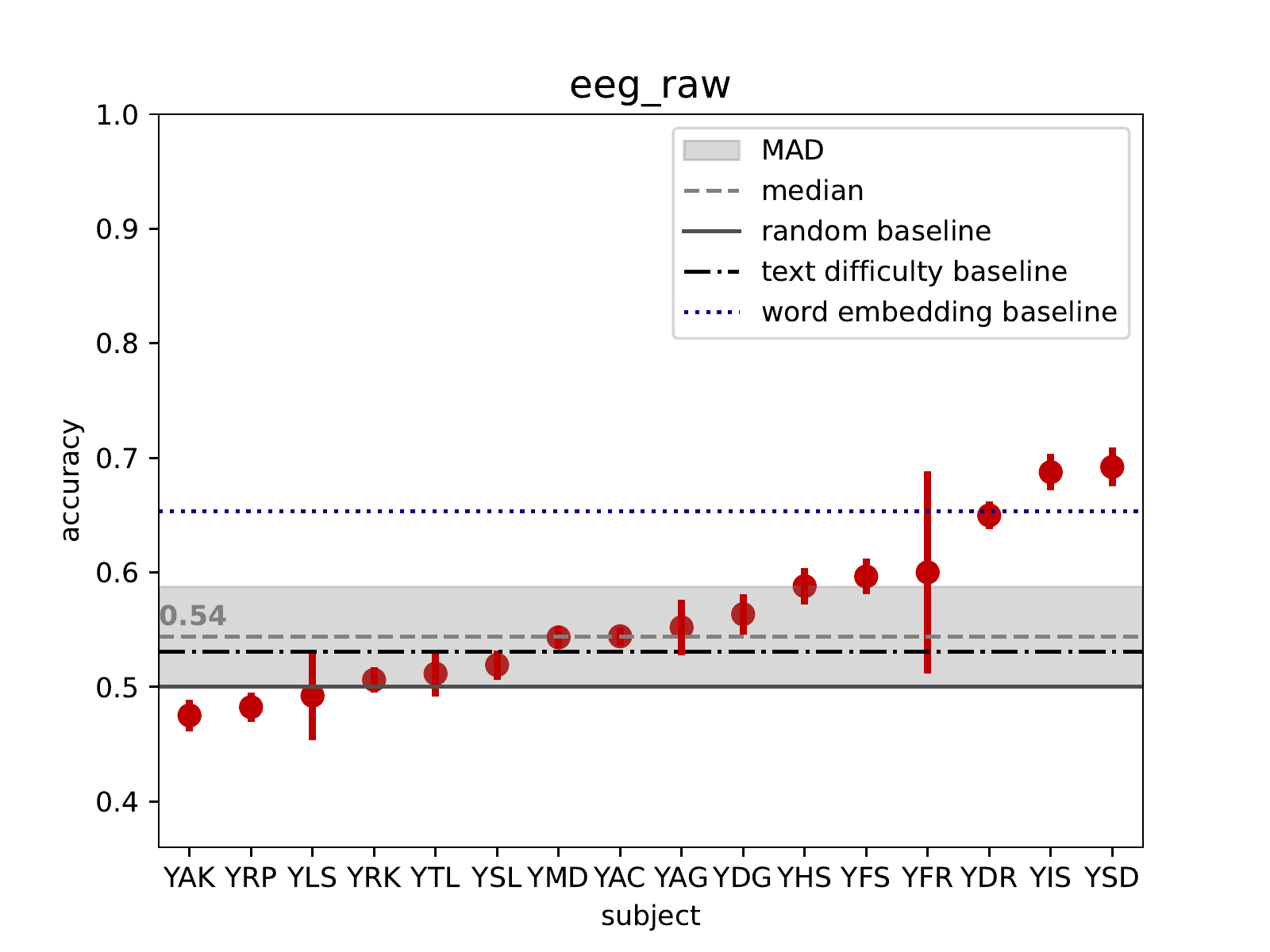} 
    \caption{Accuracy results of the cross-subject EEG raw word-level classification including saccade features on ZuCo 1.0 (left) and ZuCo 2.0 (right).}
    \label{fig:word-res-eeg-raw-cross}
\end{figure}

\begin{figure}[ht]
    \centering
    \includegraphics[width=0.23\textwidth]{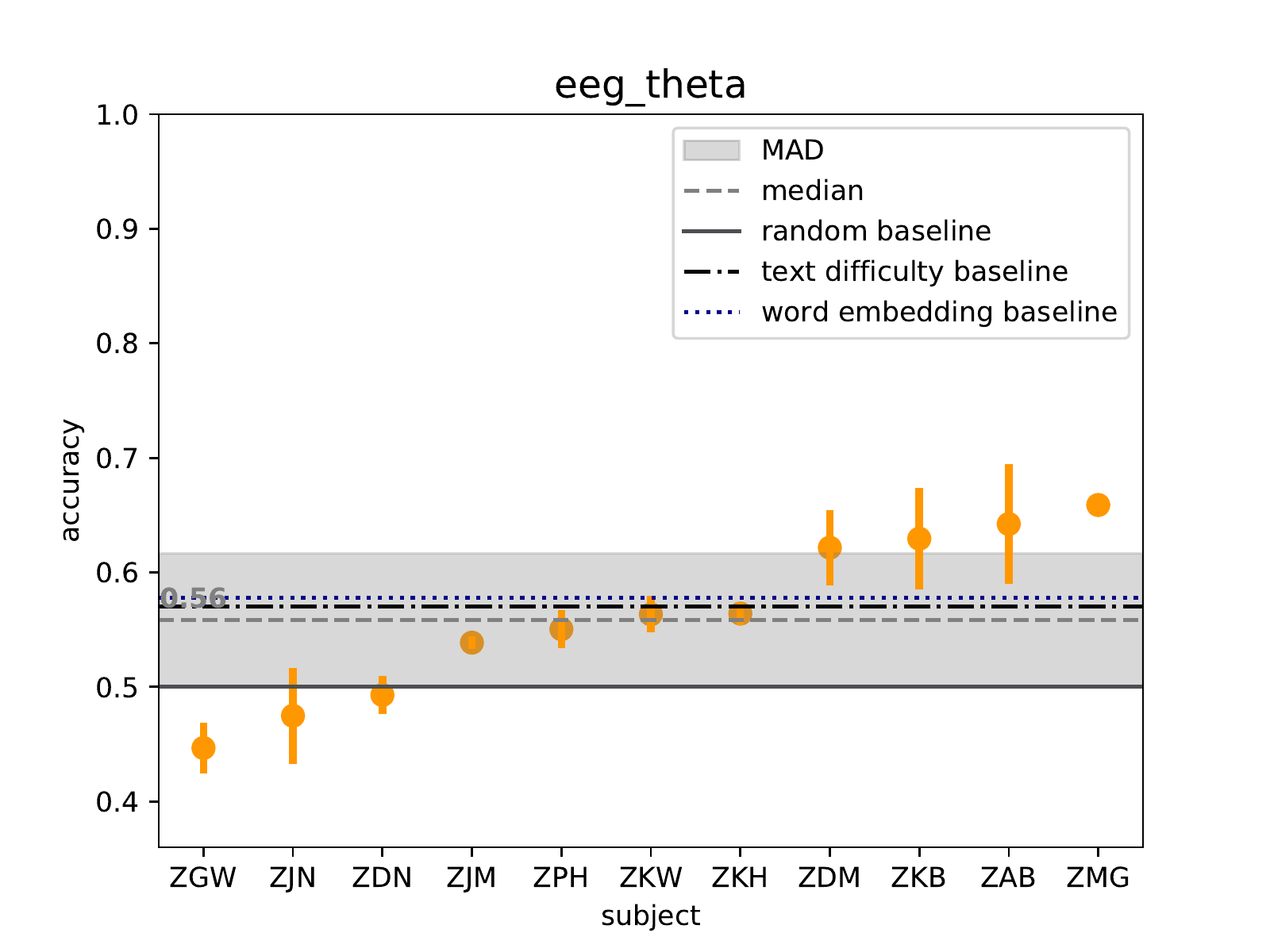}
    \includegraphics[width=0.23\textwidth]{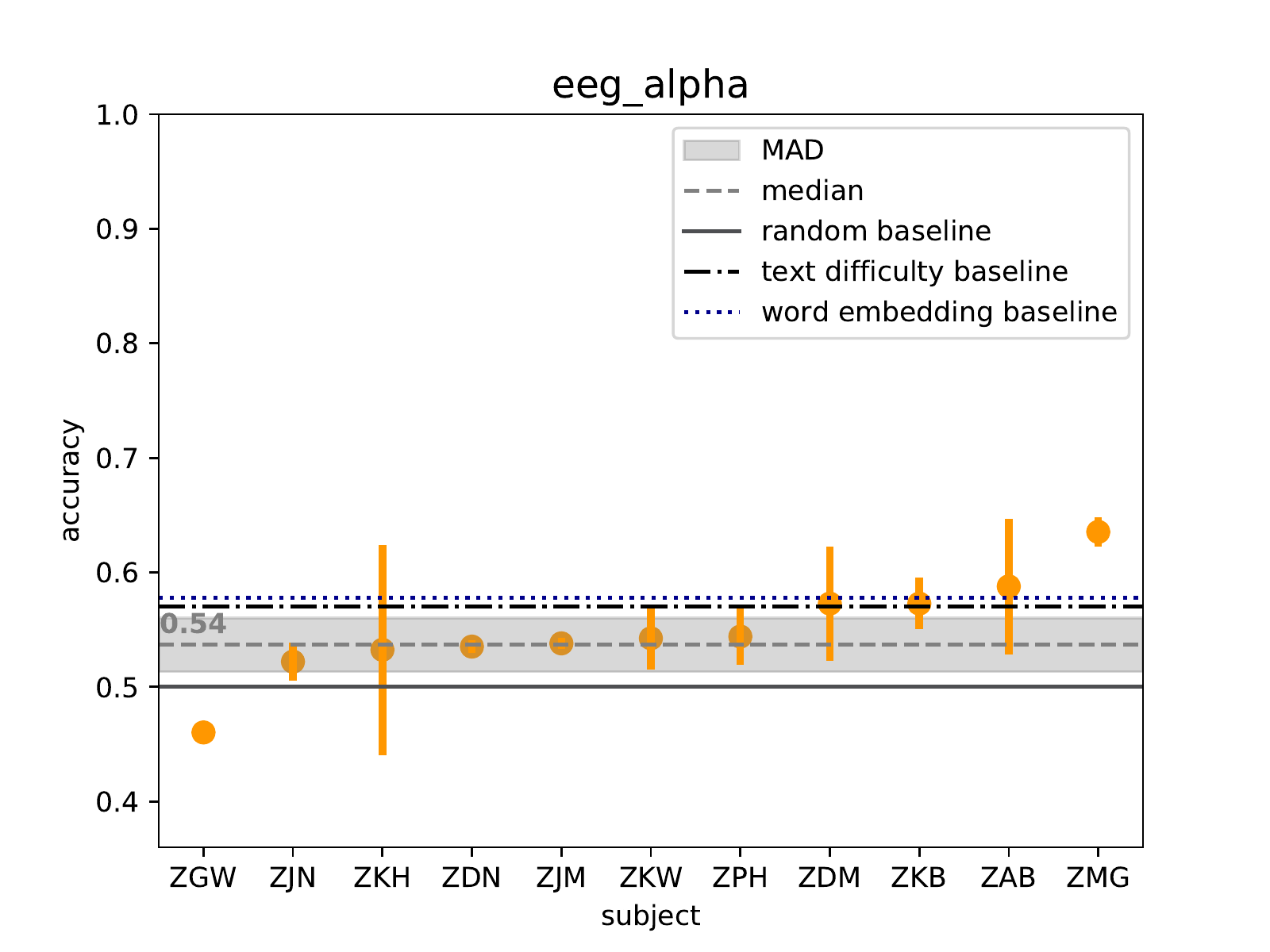}
    \includegraphics[width=0.23\textwidth]{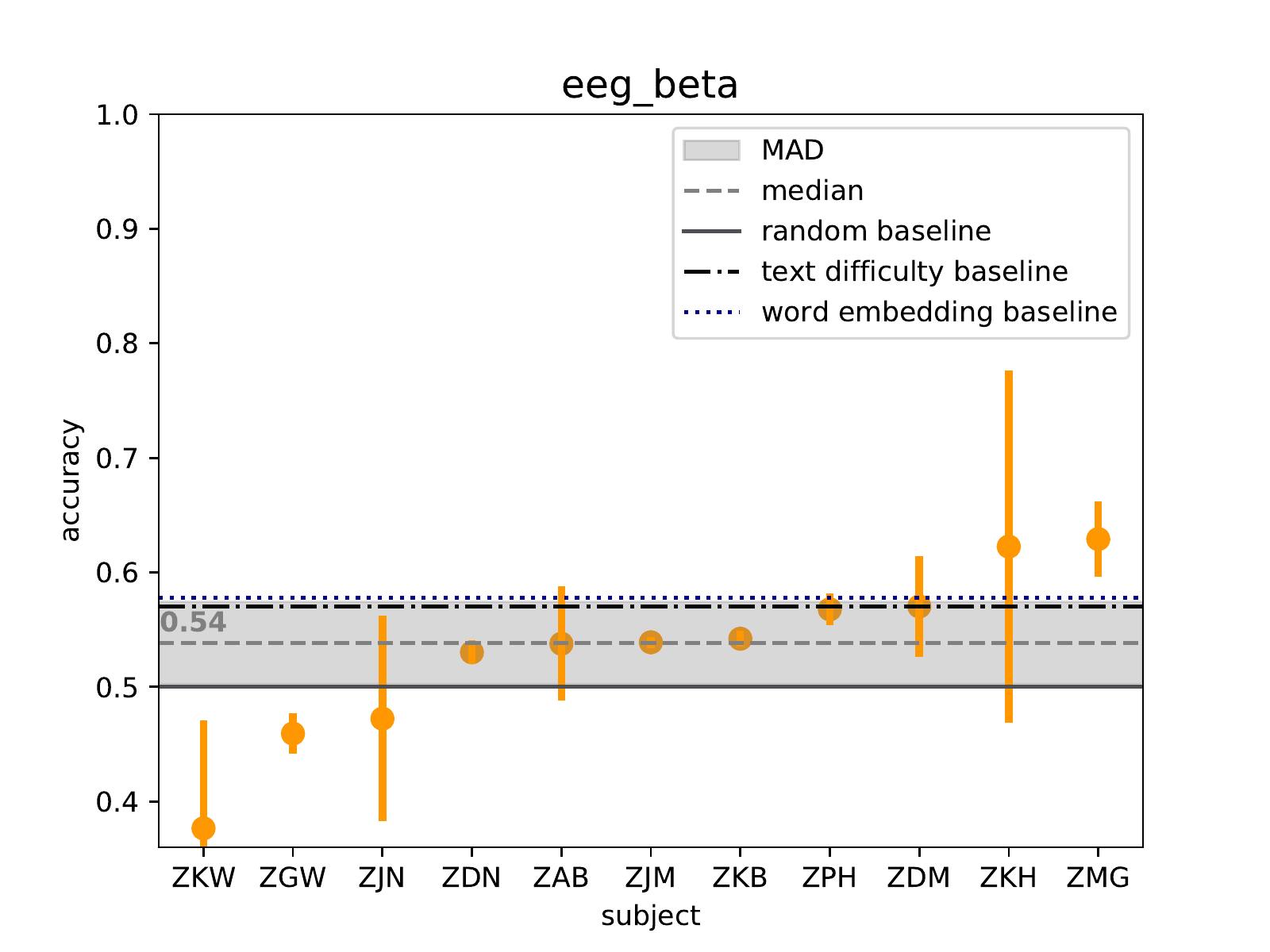} 
    \includegraphics[width=0.23\textwidth]{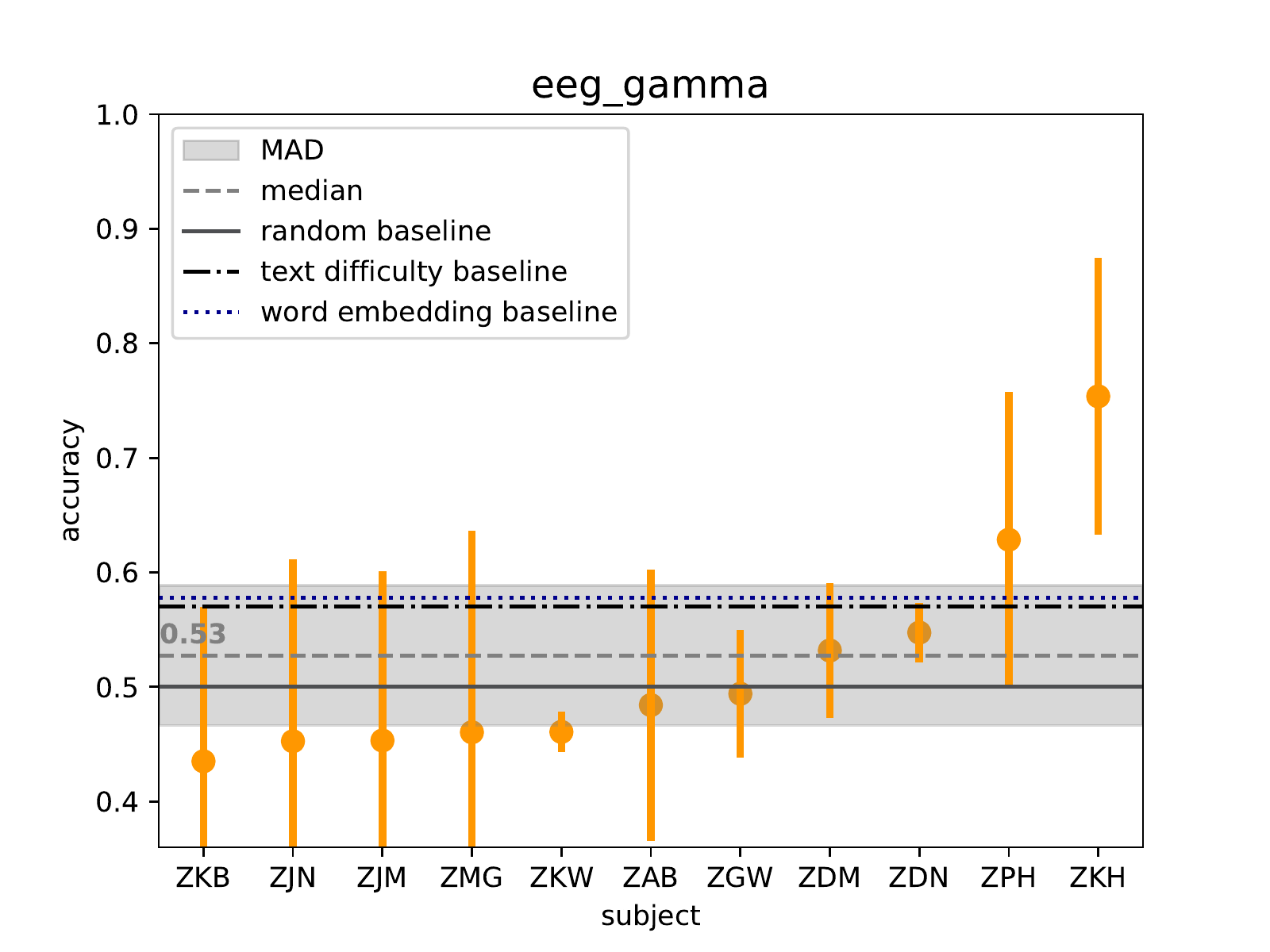}
    \includegraphics[width=0.23\textwidth]{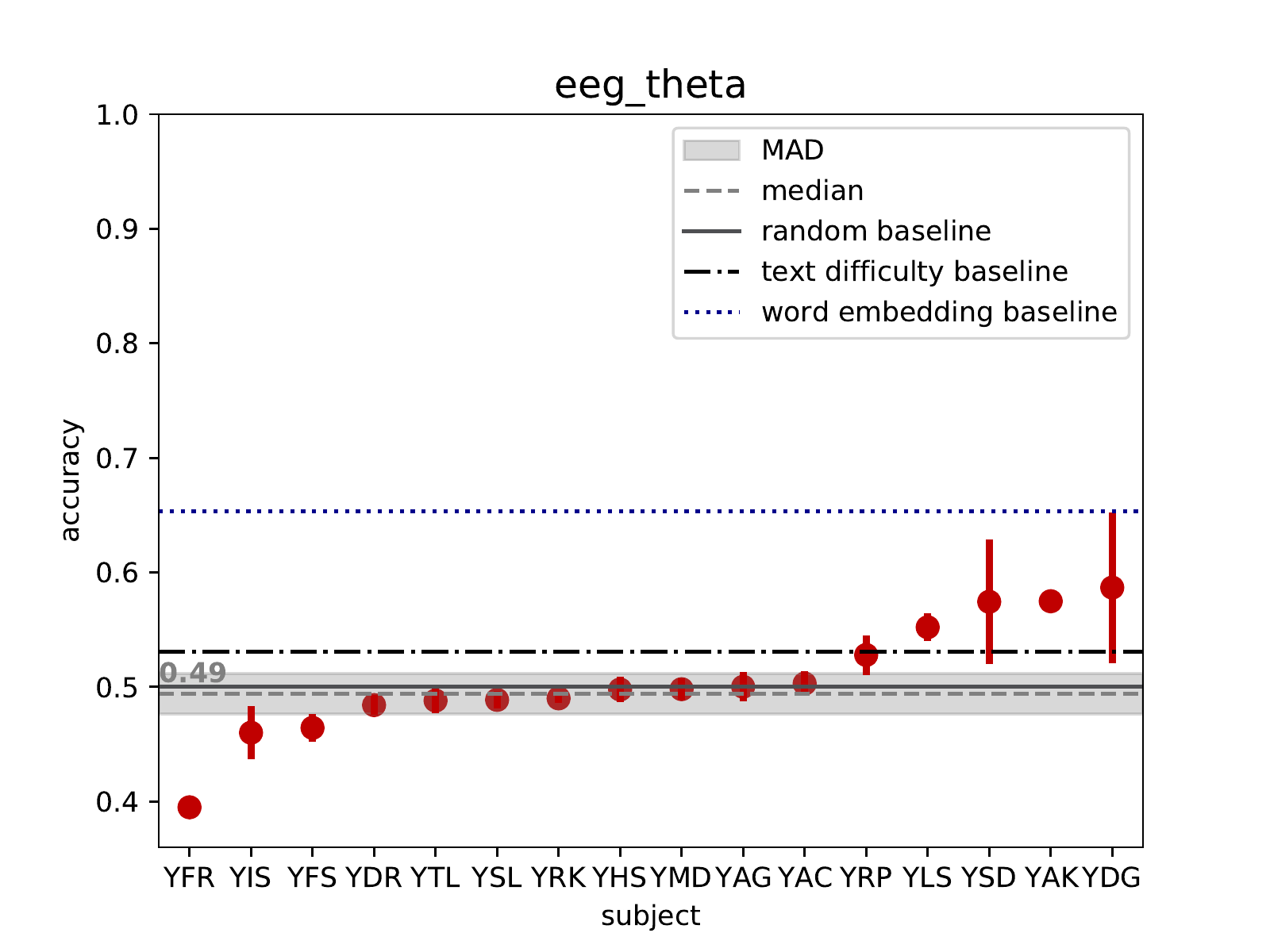}
    \includegraphics[width=0.23\textwidth]{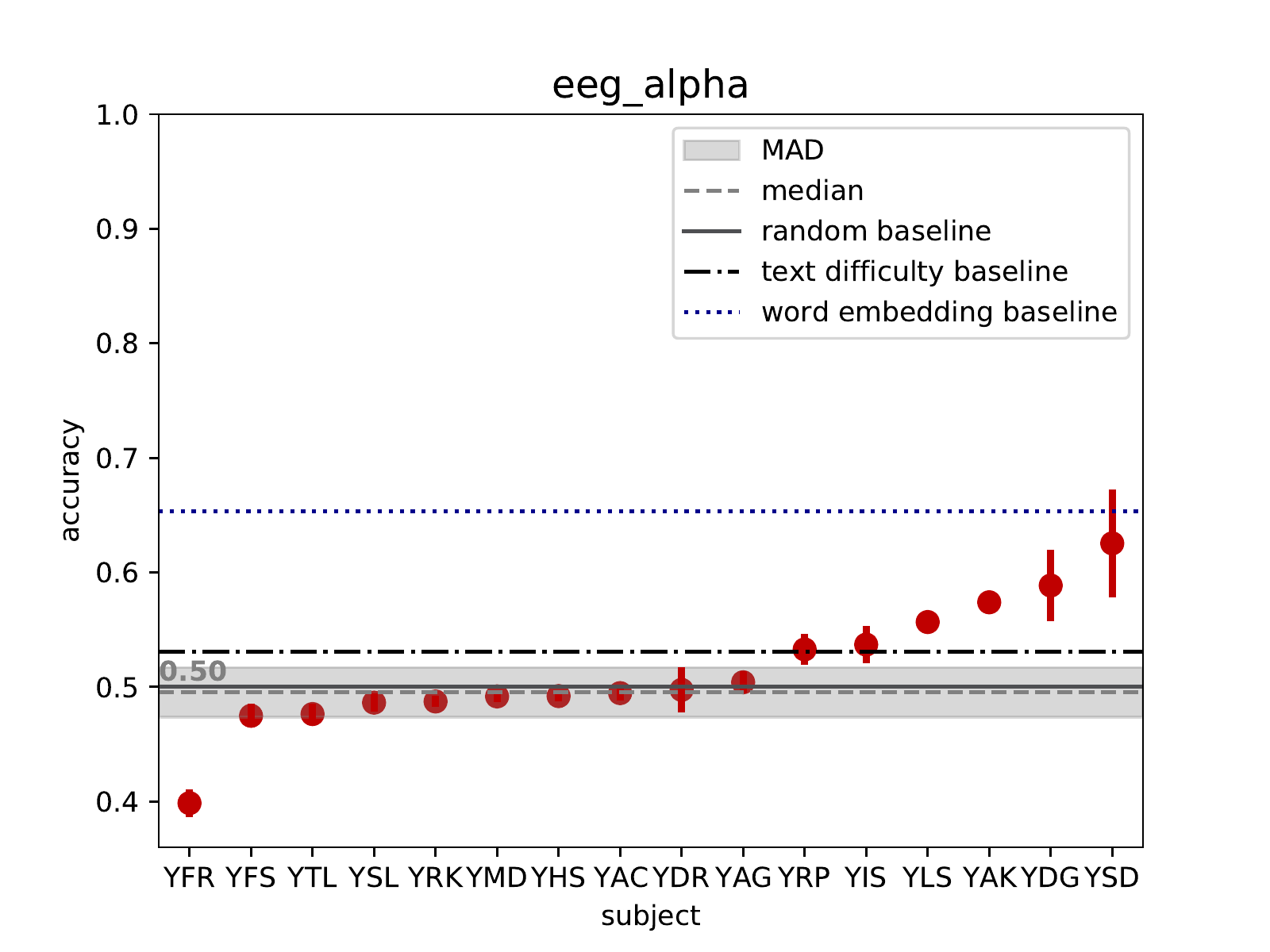}
    \includegraphics[width=0.23\textwidth]{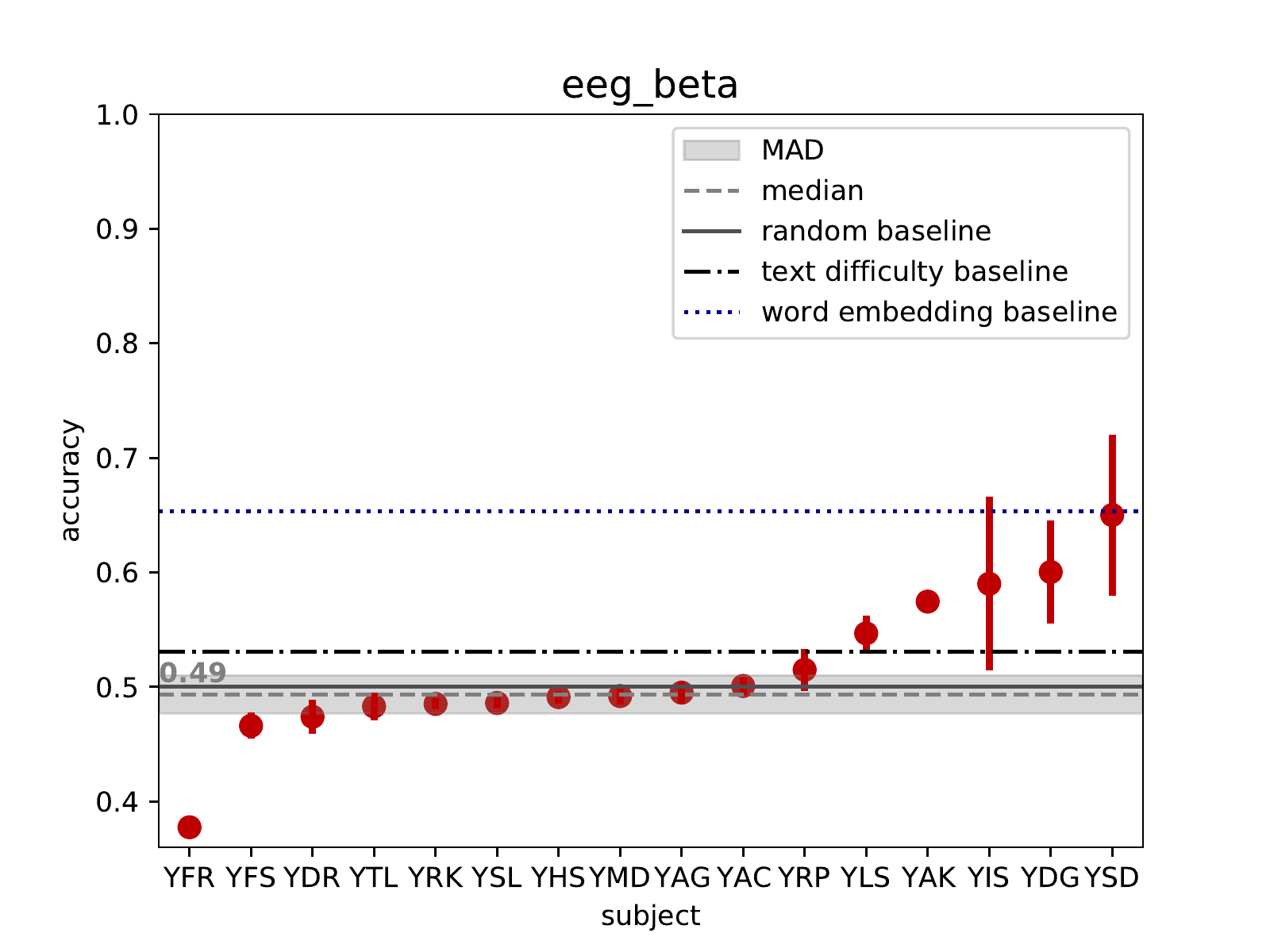} 
    \includegraphics[width=0.23\textwidth]{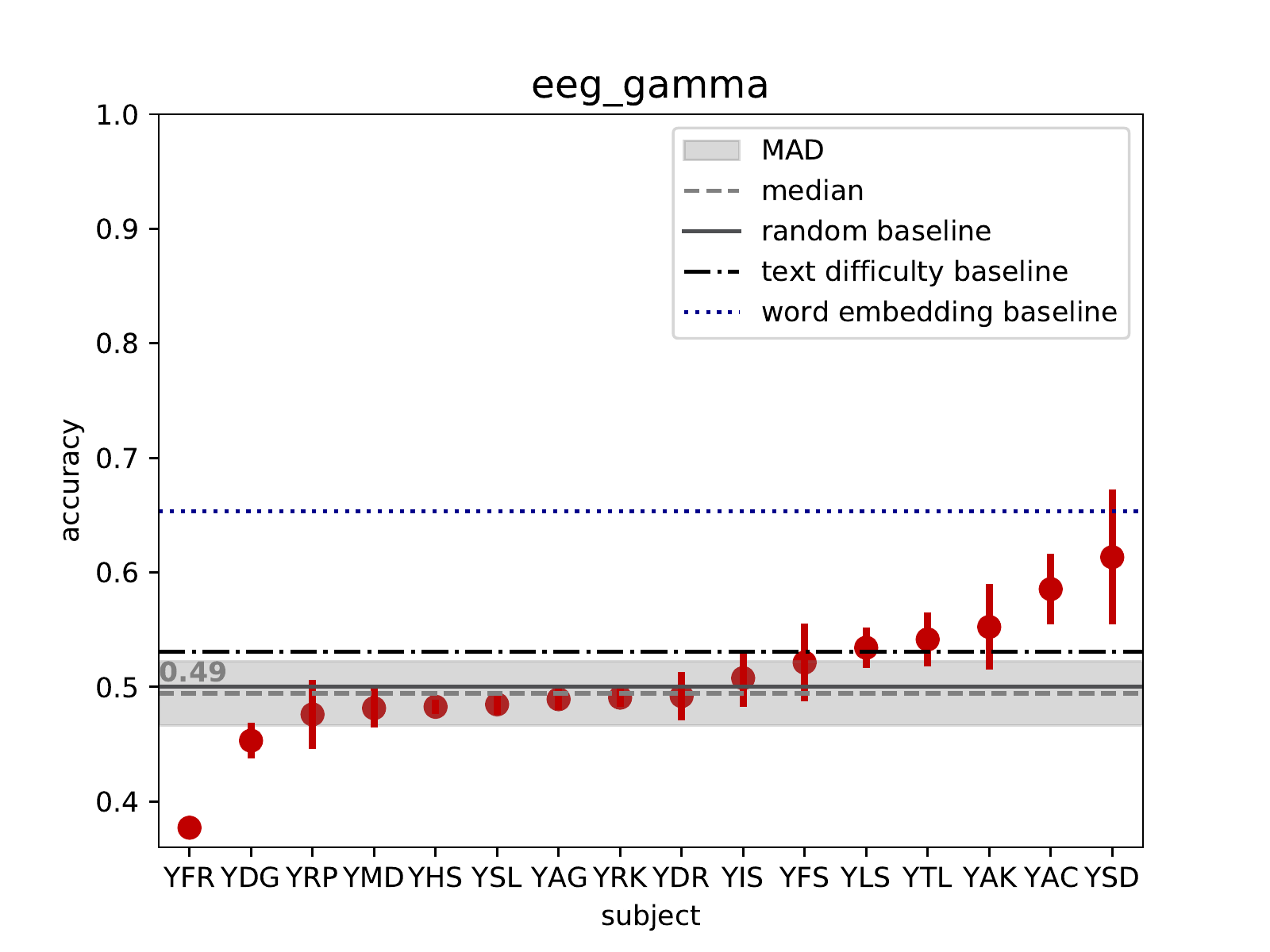}
    \caption{Cross-subject EEG frequency band word-level classification accuracy on ZuCo 1.0 (top) and ZuCo 2.0 (bottom).}
    \label{fig:word-res-eeg-cross}
\end{figure}

\noindent We also explore the generalization capabilities of the word-level features across subjects in a leave-one-out scenario. The results for the eye-tracking features are shown in Figure \ref{fig:word-res-et-cross} and for the EEG features in Figures \ref{fig:word-res-eeg-raw-cross} and \ref{fig:word-res-eeg-cross}. The results show that the accuracy of cross-subject models is much lower than that of within-subject models for both datasets. Using the EEG features from ZuCo 2.0, the models of all frequency bands merely reach chance level.

\begin{table}[ht]
\centering
\begin{tabular}{lcccc}
\toprule
 & \multicolumn{2}{c}{ZuCo 1.0} & \multicolumn{2}{c}{ZuCo 2.0} \\
\textbf{feature set} & \textbf{median} & \textbf{MAD} & \textbf{median} & \textbf{MAD} \\\midrule
eye\_tracking & 0.55 & 0.02 & 0.54 & 0.03 \\
eye\_tracking sacc & 0.57 & 0.03 & 0.52 & 0.03 \\\midrule
eeg\_raw & 0.49 & 0.04 & 0.54  & 0.04 \\
eeg\_gamma & 0.53 & 0.06 & 0.49 & 0.03 \\
eeg\_beta & 0.54 & 0.04 & 0.49 & 0.02 \\
eeg\_alpha & 0.54 & 0.02 & 0.50 & 0.02 \\
eeg\_theta & 0.56 & 0.06 & 0.49 & 0.02 \\
\bottomrule
\end{tabular}
\caption{Word-level cross-subject result summary.}
\label{tab:word-cross-subj-results}
\end{table}

\clearpage

\subsection{Sentence-level Results}

\begin{table}[t]
\centering
\small
\begin{tabular}{lcccc}
\toprule
 & \multicolumn{2}{c}{ZuCo 1.0} & \multicolumn{2}{c}{ZuCo 2.0} \\
\textbf{feature set} & \textbf{median} & \textbf{MAD} & \textbf{median} & \textbf{MAD} \\\midrule
text difficulty baseline & 0.58 & - & 0.53 & - \\
word embedding baseline & 0.58 & - & 0.65 & -\\\midrule
fixation\_number & 0.70 & 0.10 & \textbf{0.66} & 0.08 \\
omission\_rate & 0.70 & 0.06 & 0.65 & 0.09 \\
reading\_speed & \textbf{0.72} & 0.10 & 0.65 & 0.09 \\\midrule
sent\_gaze & \textbf{0.76} & 0.08 & \textbf{0.68} & 0.08\\\midrule
mean\_sacc\_dur & 0.59 & 0.06 & 0.55 & 0.05 \\
max\_sacc\_velocity & 0.61 & 0.06 & 0.57 & 0.04 \\
mean\_sacc\_velocity & 0.63 & 0.07 & 0.57 & 0.05 \\
max\_sacc\_dur & 0.58 & 0.04 & 0.54 & 0.04 \\
max\_sacc\_amp & 0.59 & 0.04 & 0.57 & 0.07 \\
mean\_sacc\_amp & \textbf{0.63} & 0.06 & \textbf{0.61} & 0.07 \\\midrule
sent\_saccade & 0.77 & 0.06 & 0.68 & 0.08 \\
sent\_gaze\_sacc & \textbf{0.82} & 0.06 & \textbf{0.70} & 0.07 \\\midrule
theta\_mean & 0.63 & 0.07 & 0.56 & 0.04 \\
alpha\_mean & 0.62 & 0.07 & 0.58 & 0.05 \\
beta\_mean & 0.66 & 0.08 & \textbf{0.58} & 0.04 \\
gamma\_mean & \textbf{0.67} & 0.08 & 0.57 & 0.05 \\\midrule
eeg\_means & 0.79 & 0.09 & 0.62 & 0.05 \\
sent\_gaze\_eeg\_means & \textbf{0.88} & 0.06 & \textbf{0.72} & 0.06\\\midrule
electrode\_features\_theta & \textbf{1.00} & 0.00 & 0.70 & 0.05 \\
electrode\_features\_alpha & \textbf{1.00} & 0.00 & 0.71 & 0.05 \\
electrode\_features\_beta & \textbf{1.00} & 0.00 & 0.76 & 0.05 \\
electrode\_features\_gamma & \textbf{1.00} & 0.00 & \textbf{0.92} & 0.05 \\
electrode\_features\_all & \textbf{1.00} & 0.00 & 0.90 & 0.06 \\
\bottomrule
\end{tabular}
\caption{Summary of all within-subject sentence-level results. The best results per category are marked in bold.}
\label{tab:sent-level-res-summary}
\end{table}

\subsubsection{Within-subject Evaluation}

\begin{figure}[t]
    \centering
    \includegraphics[width=0.32\textwidth]{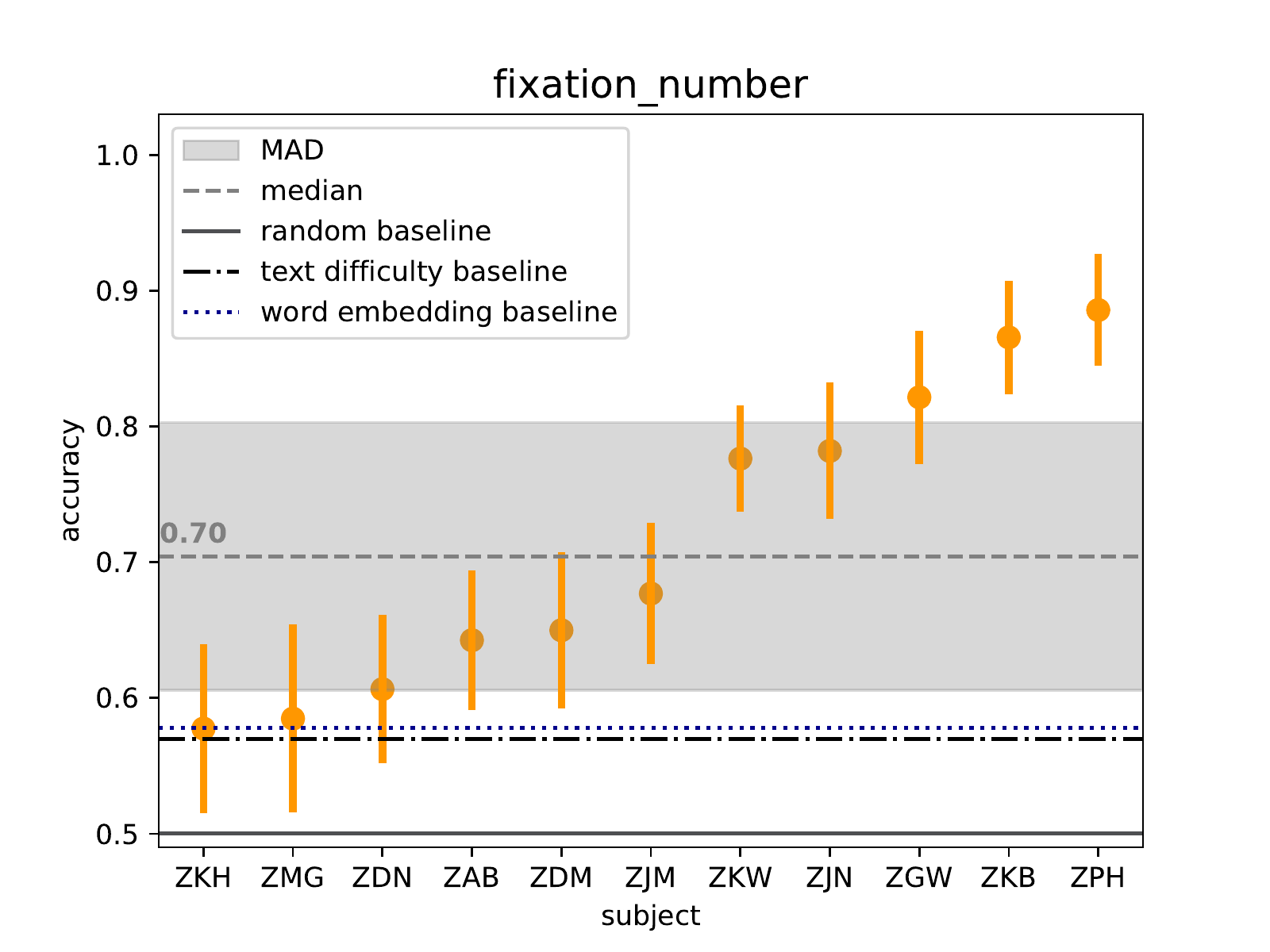} 
    \includegraphics[width=0.32\textwidth]{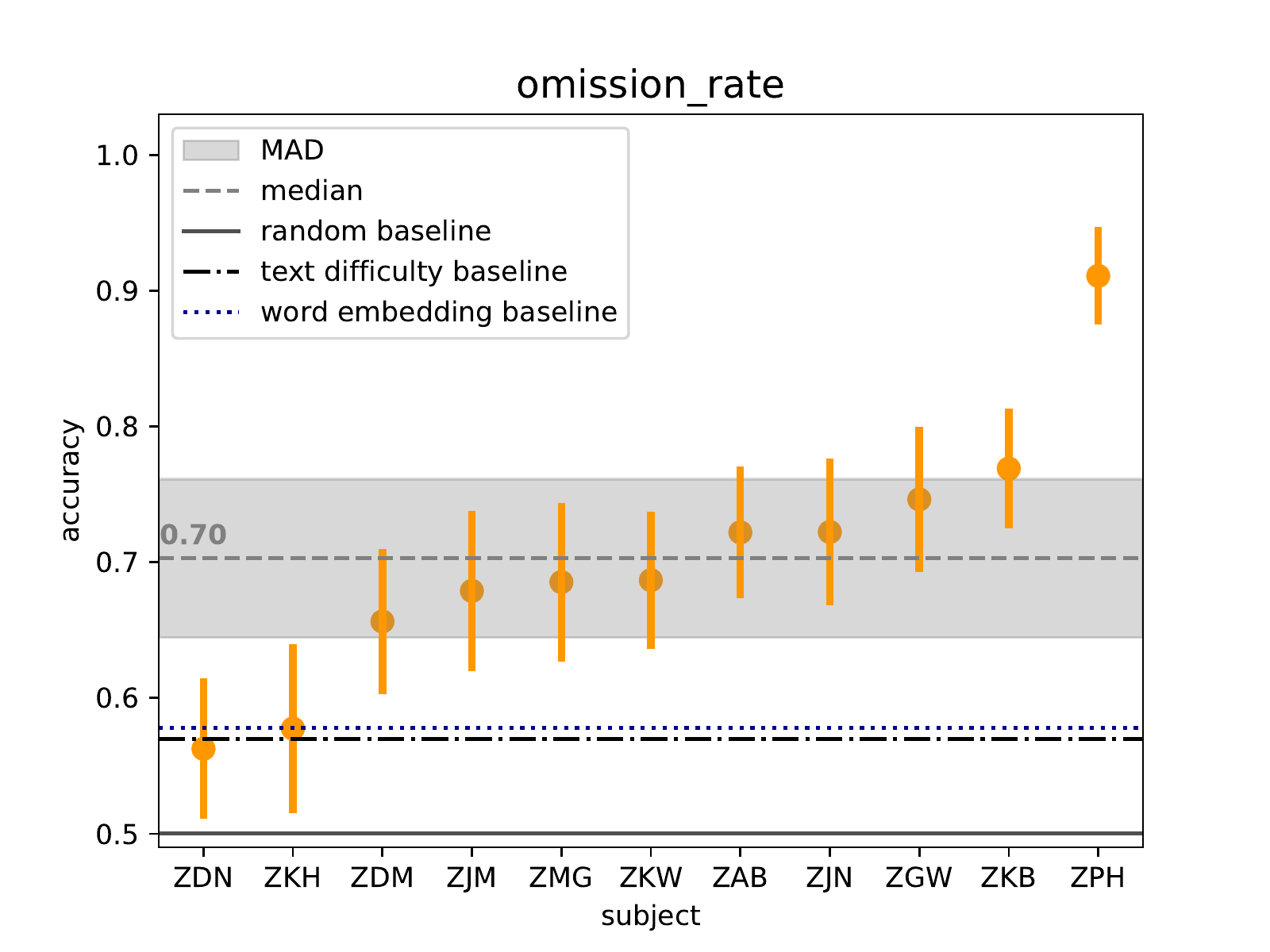} 
    \includegraphics[width=0.32\textwidth]{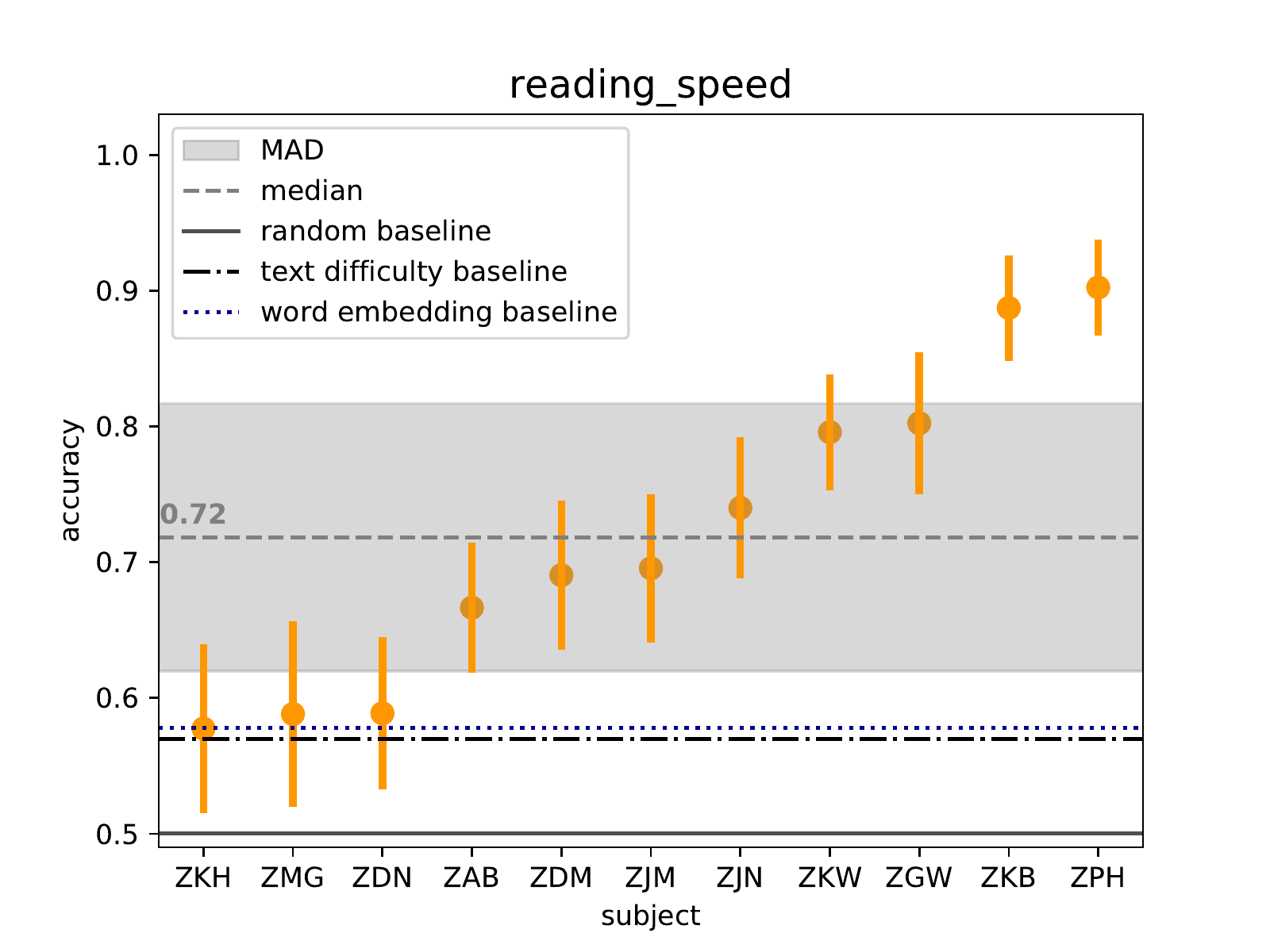} 
    \includegraphics[width=0.32\textwidth]{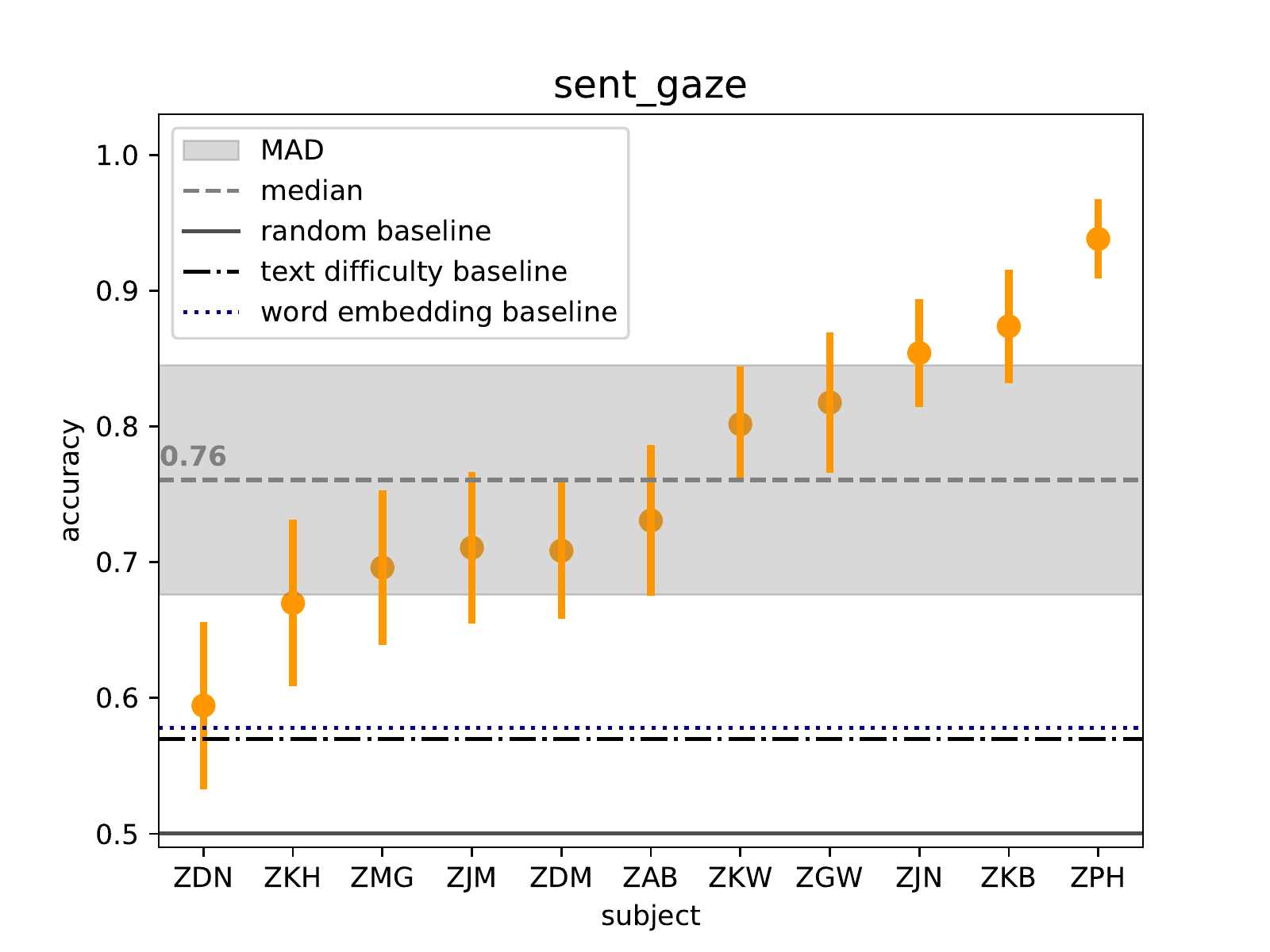} 
    \includegraphics[width=0.32\textwidth]{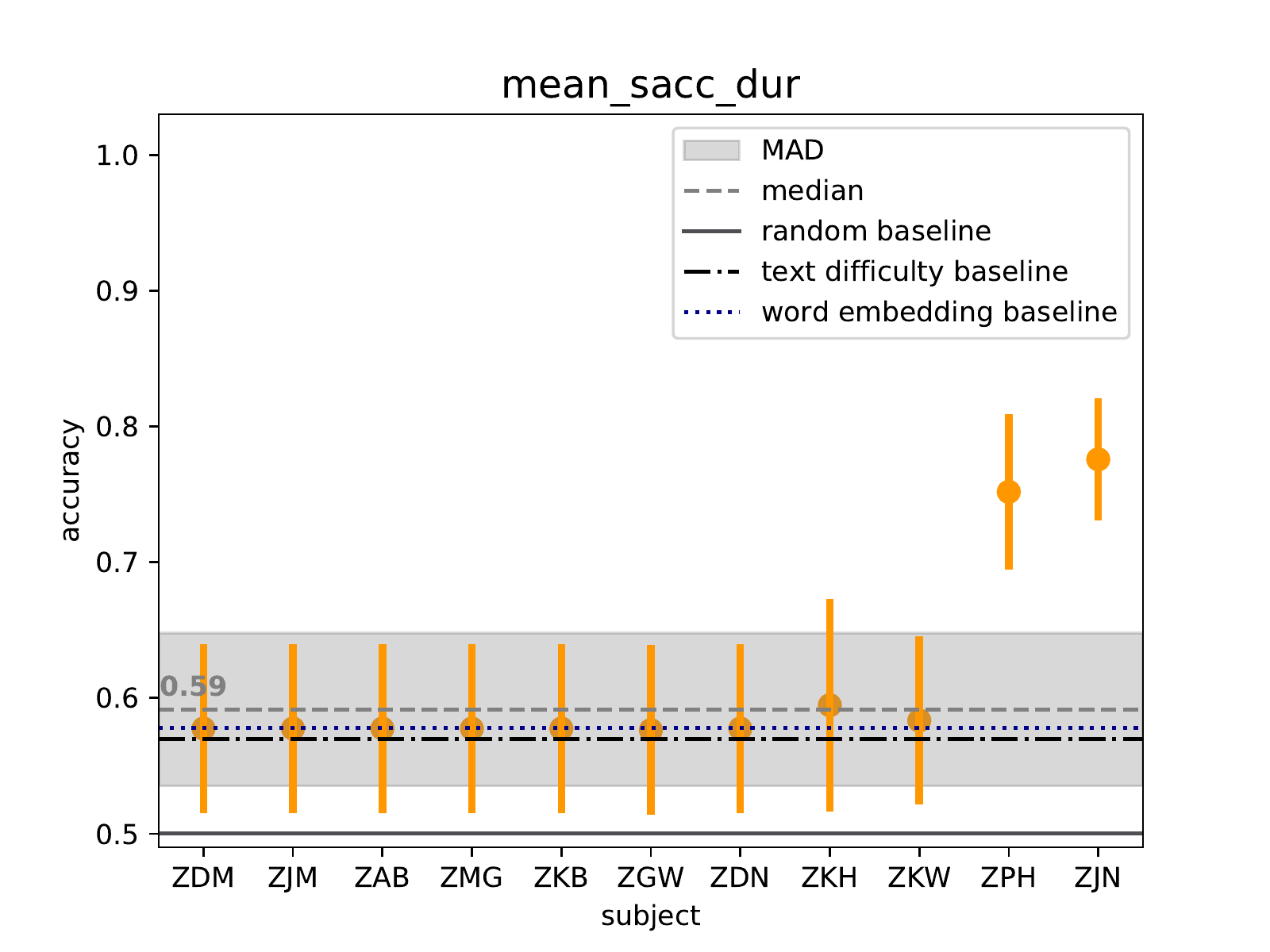} 
    \includegraphics[width=0.32\textwidth]{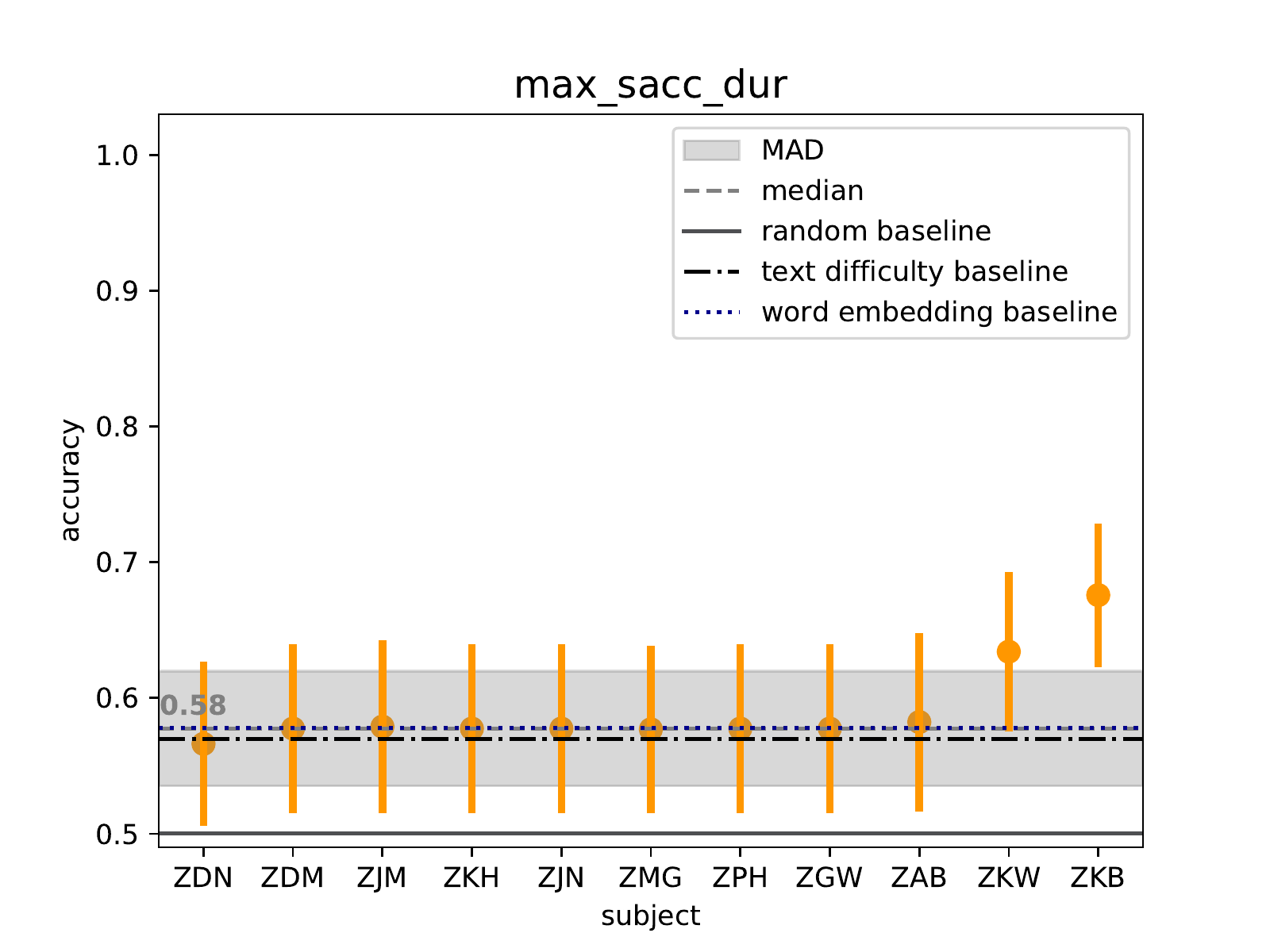} 
    \includegraphics[width=0.32\textwidth]{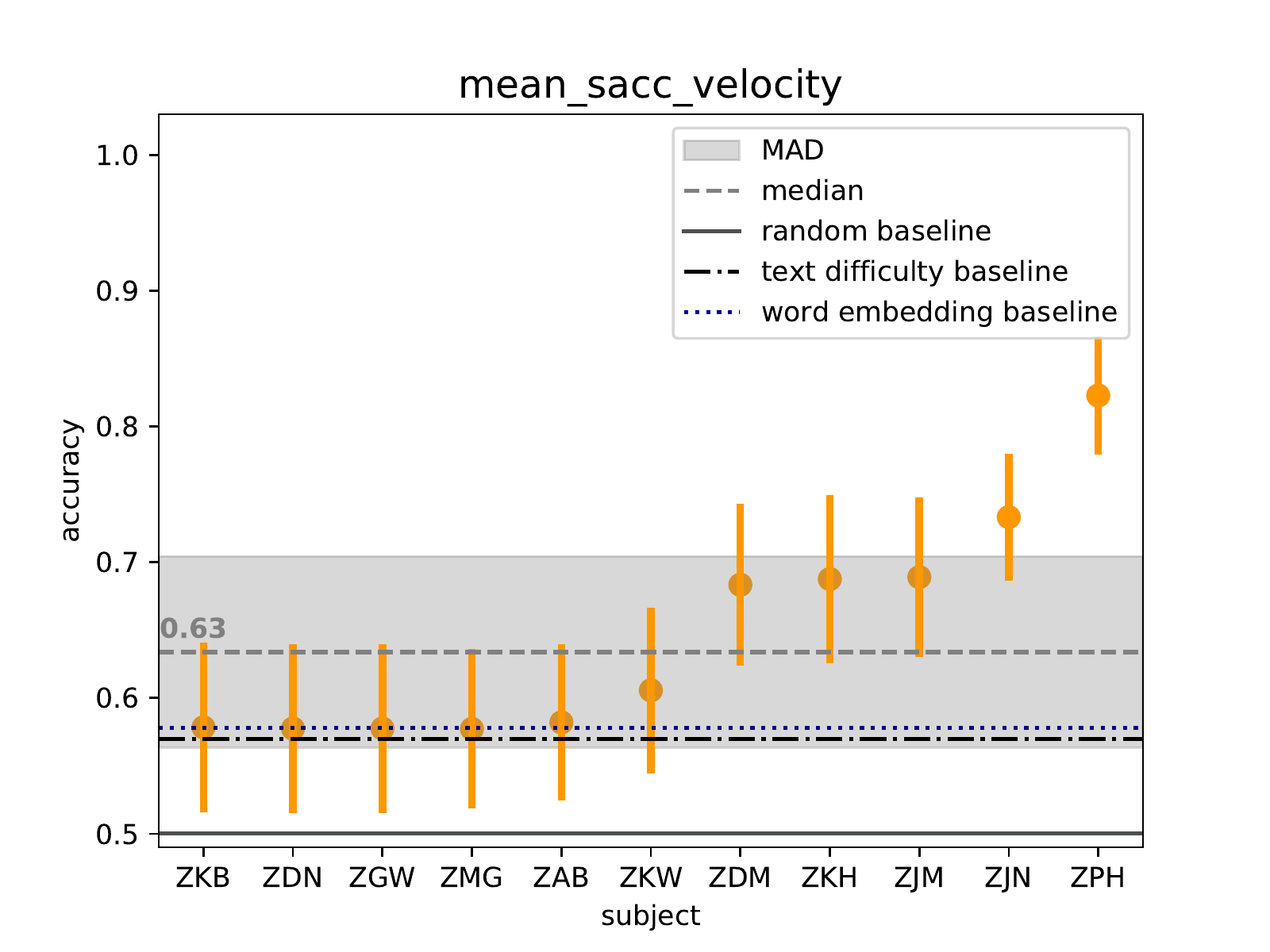} 
    \includegraphics[width=0.32\textwidth]{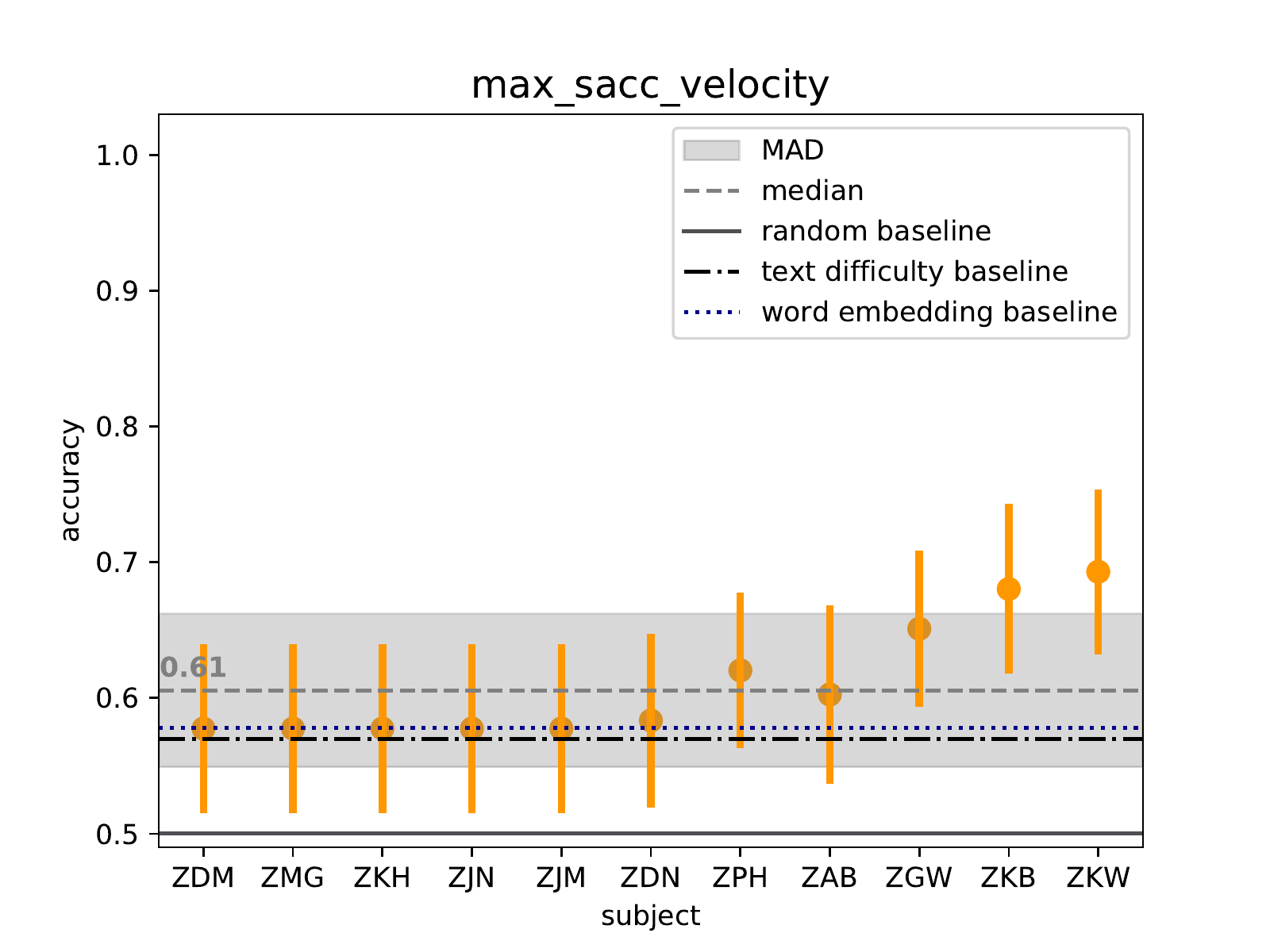} 
    \includegraphics[width=0.32\textwidth]{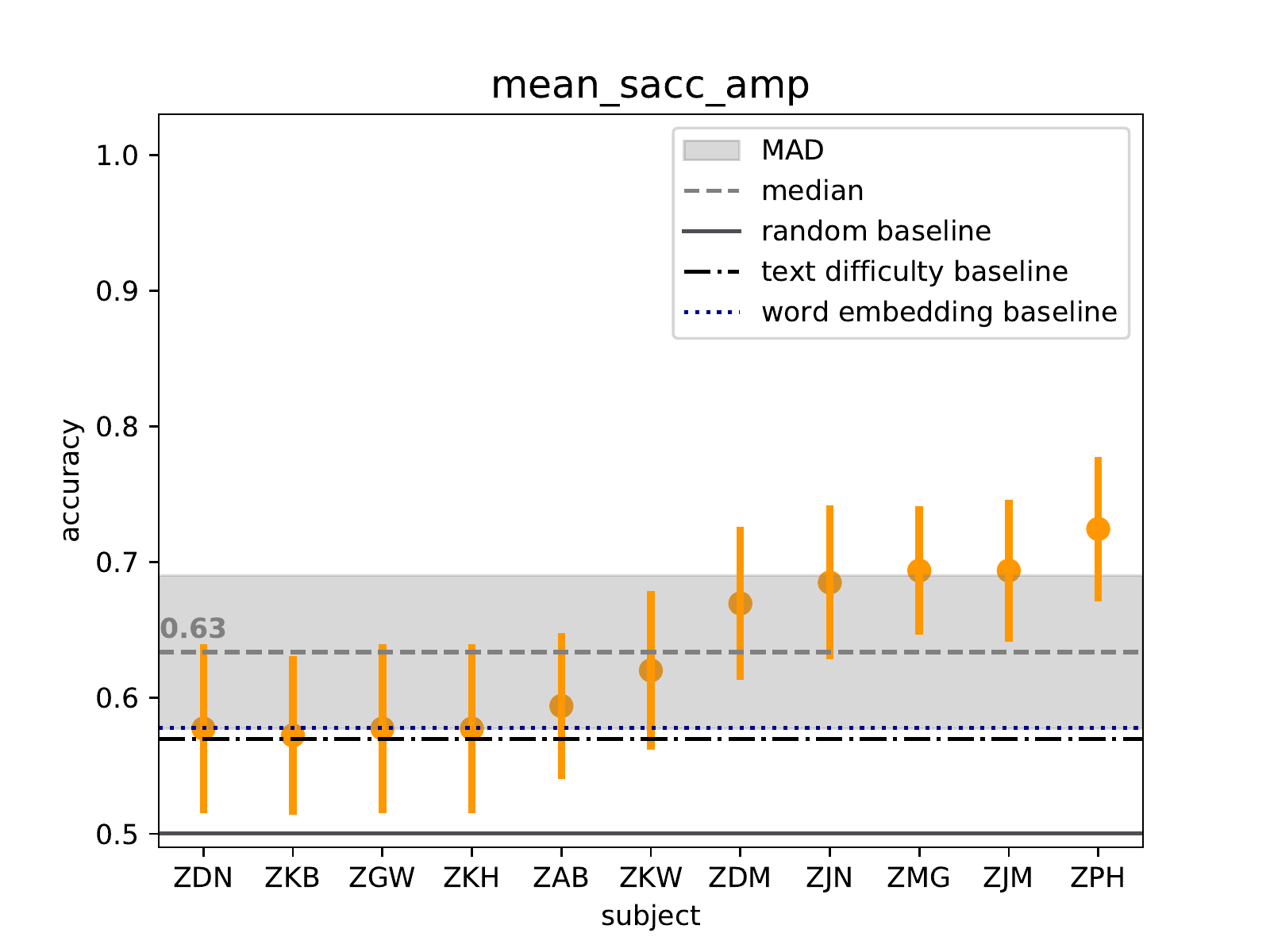} 
    \includegraphics[width=0.32\textwidth]{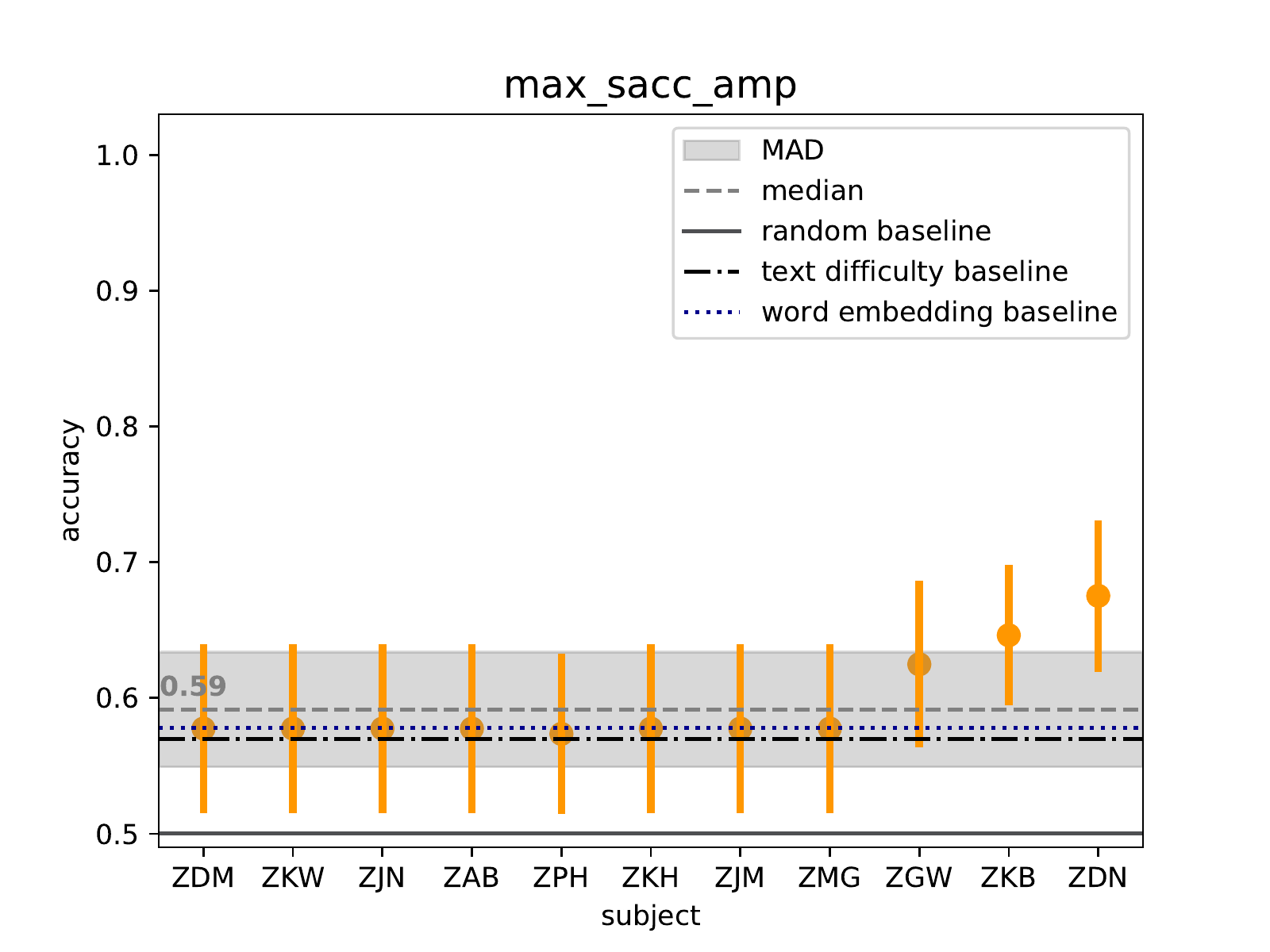} 
    \includegraphics[width=0.32\textwidth]{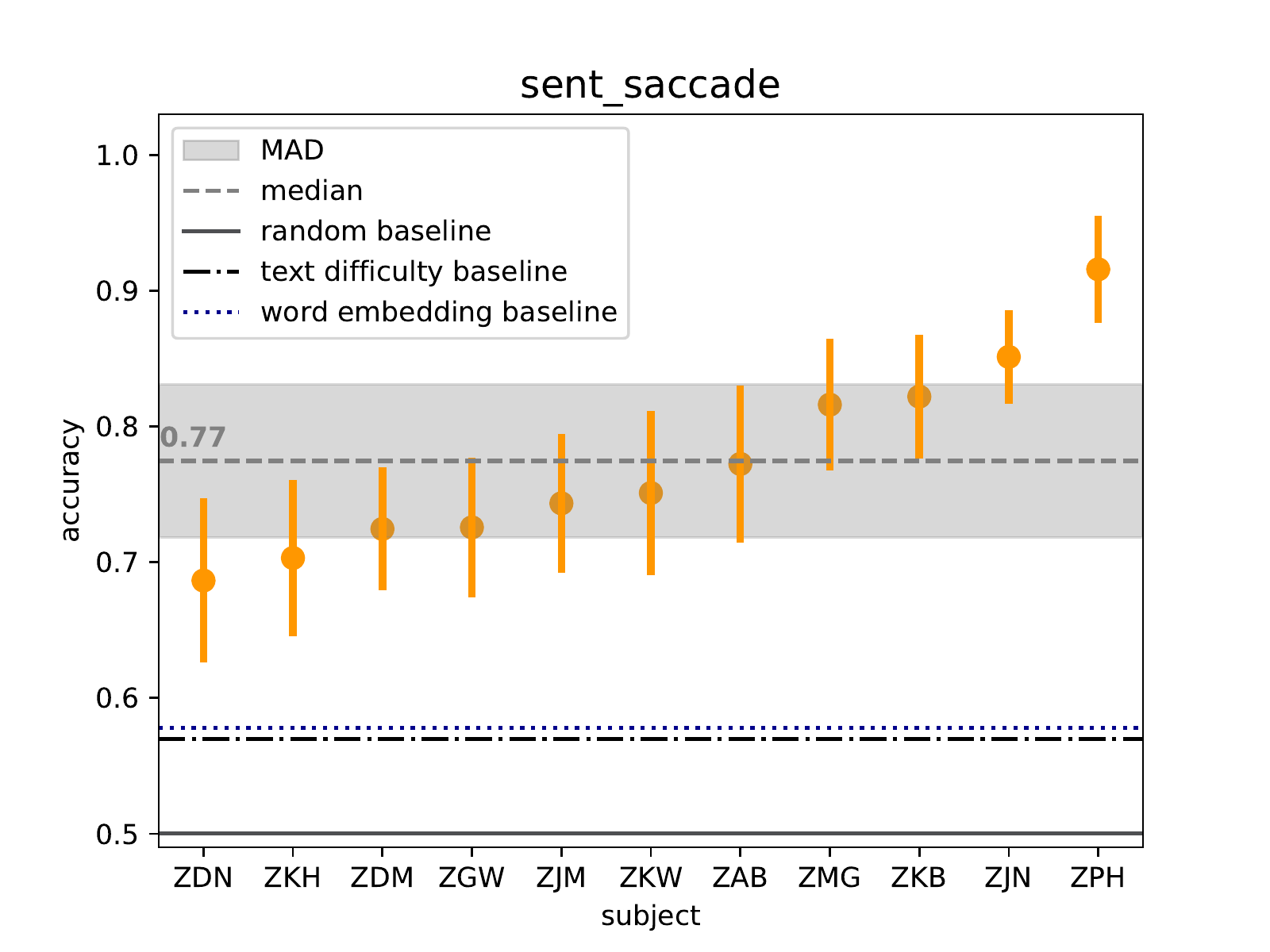} 
    \includegraphics[width=0.32\textwidth]{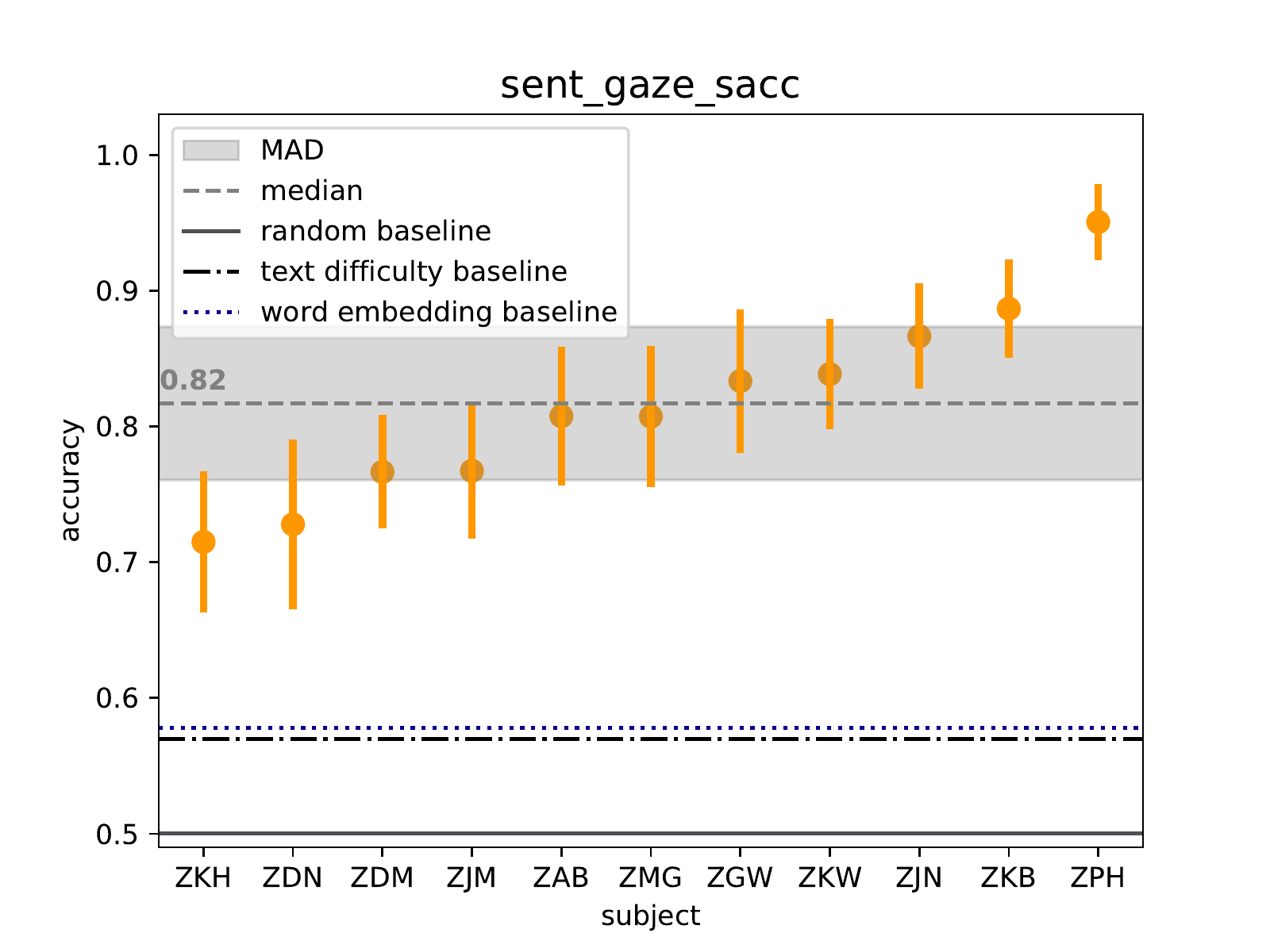} 
    \caption{Eye-tracking sentence-level classification accuracy on the ZuCo 1.0 data.}
    \label{fig:sent-res-z1-et}
\end{figure}

\begin{figure}[h]
    \centering
    \includegraphics[width=0.32\textwidth]{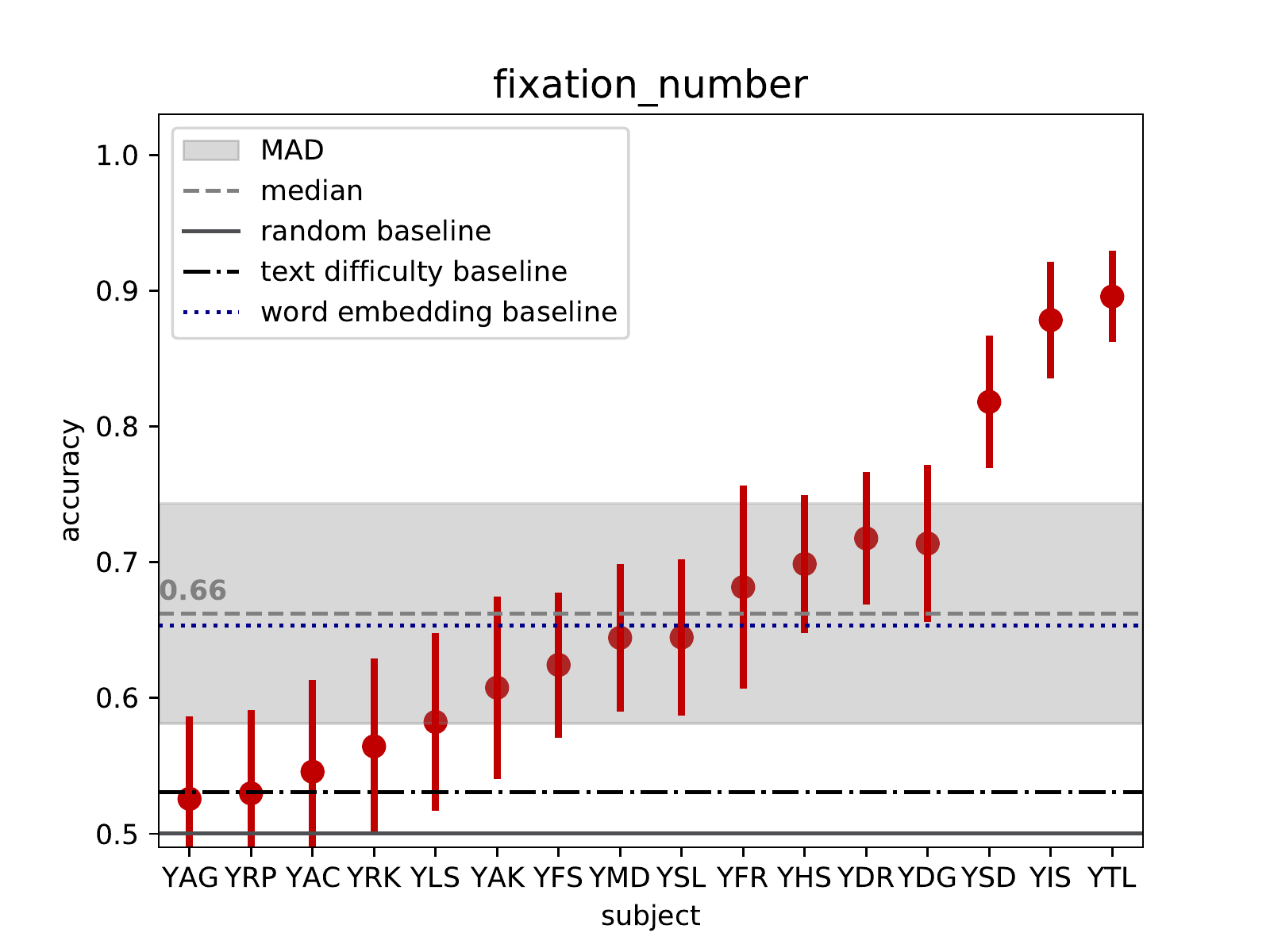} 
    \includegraphics[width=0.32\textwidth]{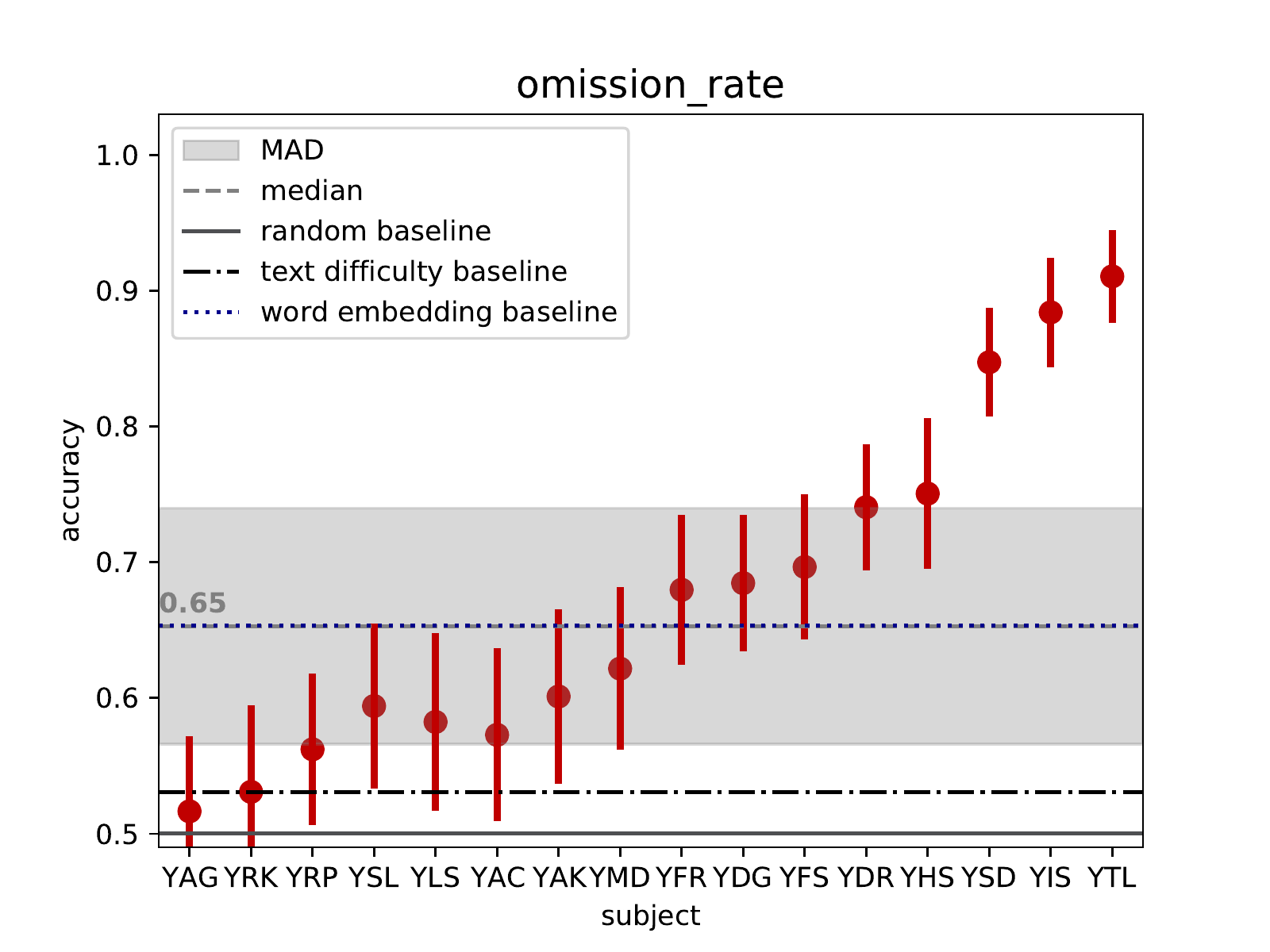} 
    \includegraphics[width=0.32\textwidth]{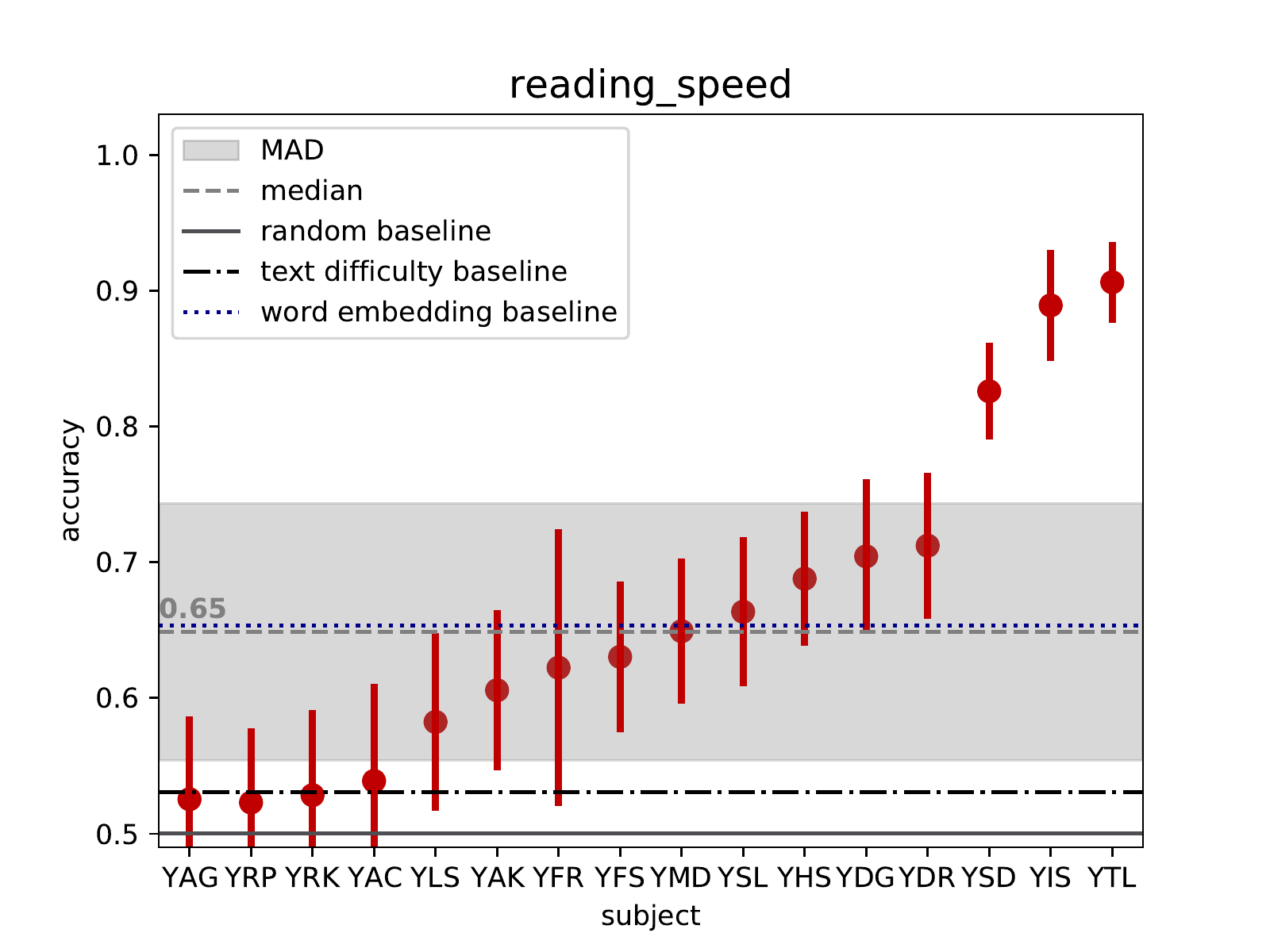} 
    \includegraphics[width=0.32\textwidth]{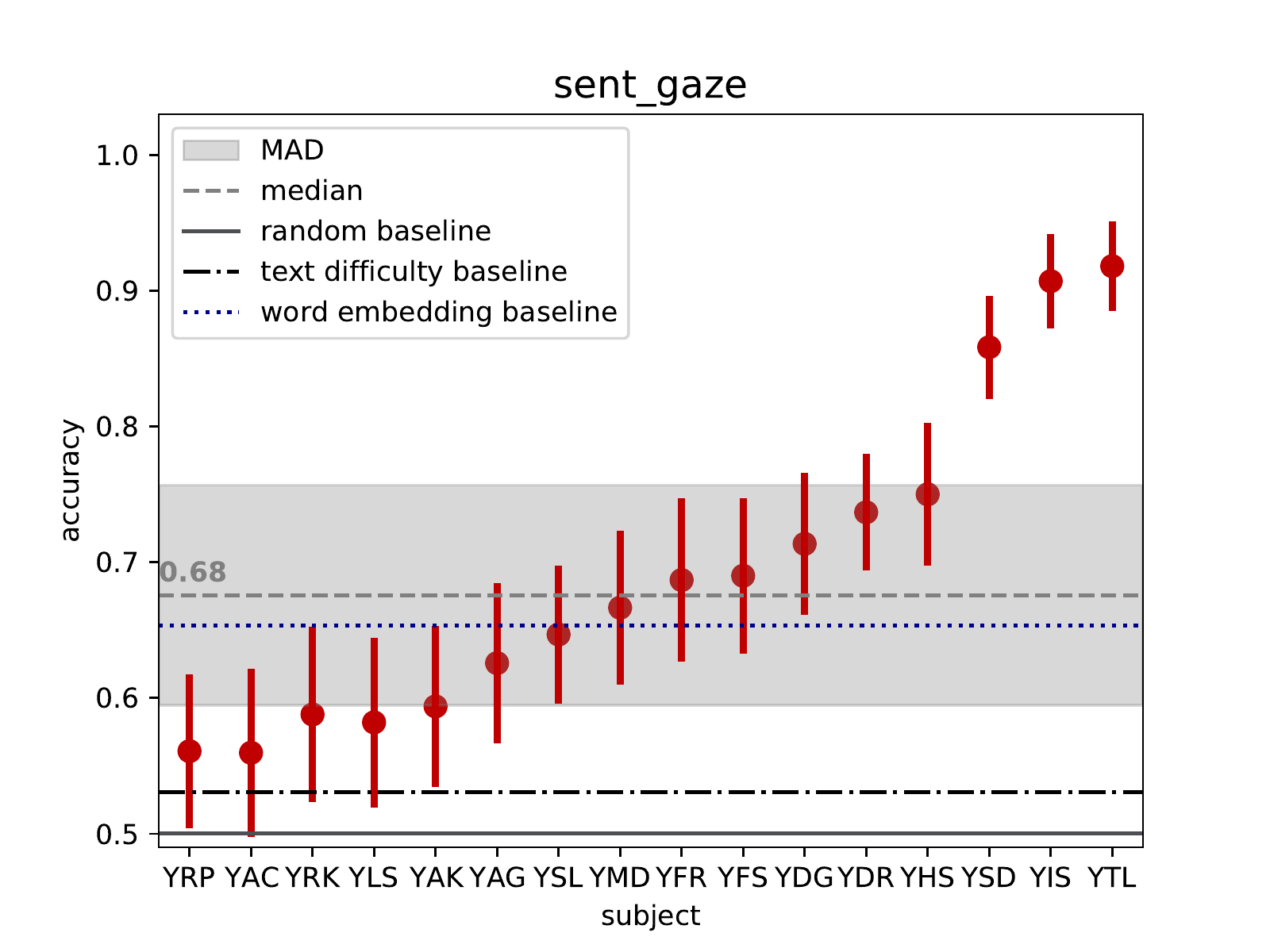} 
    \includegraphics[width=0.32\textwidth]{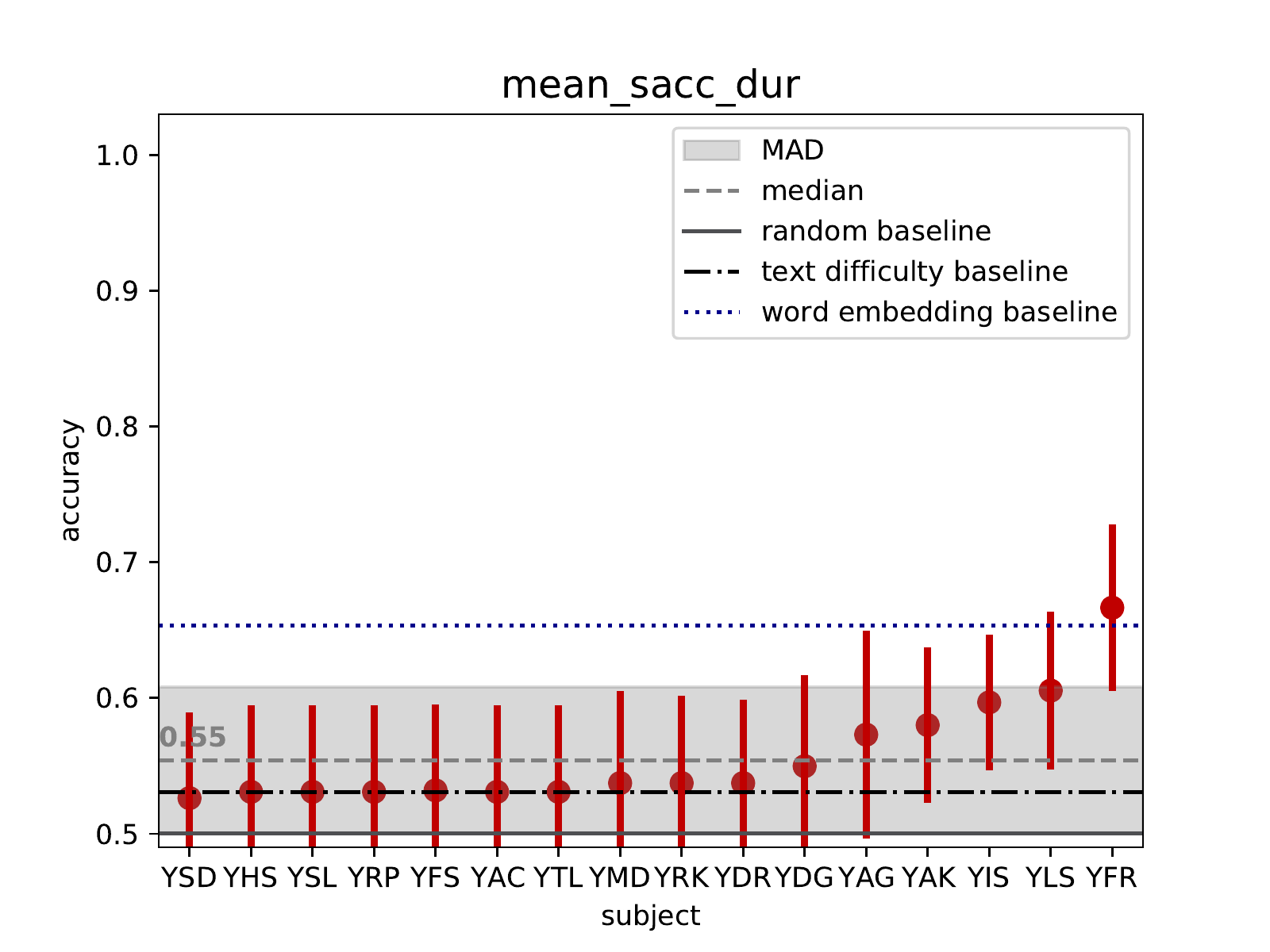} 
    \includegraphics[width=0.32\textwidth]{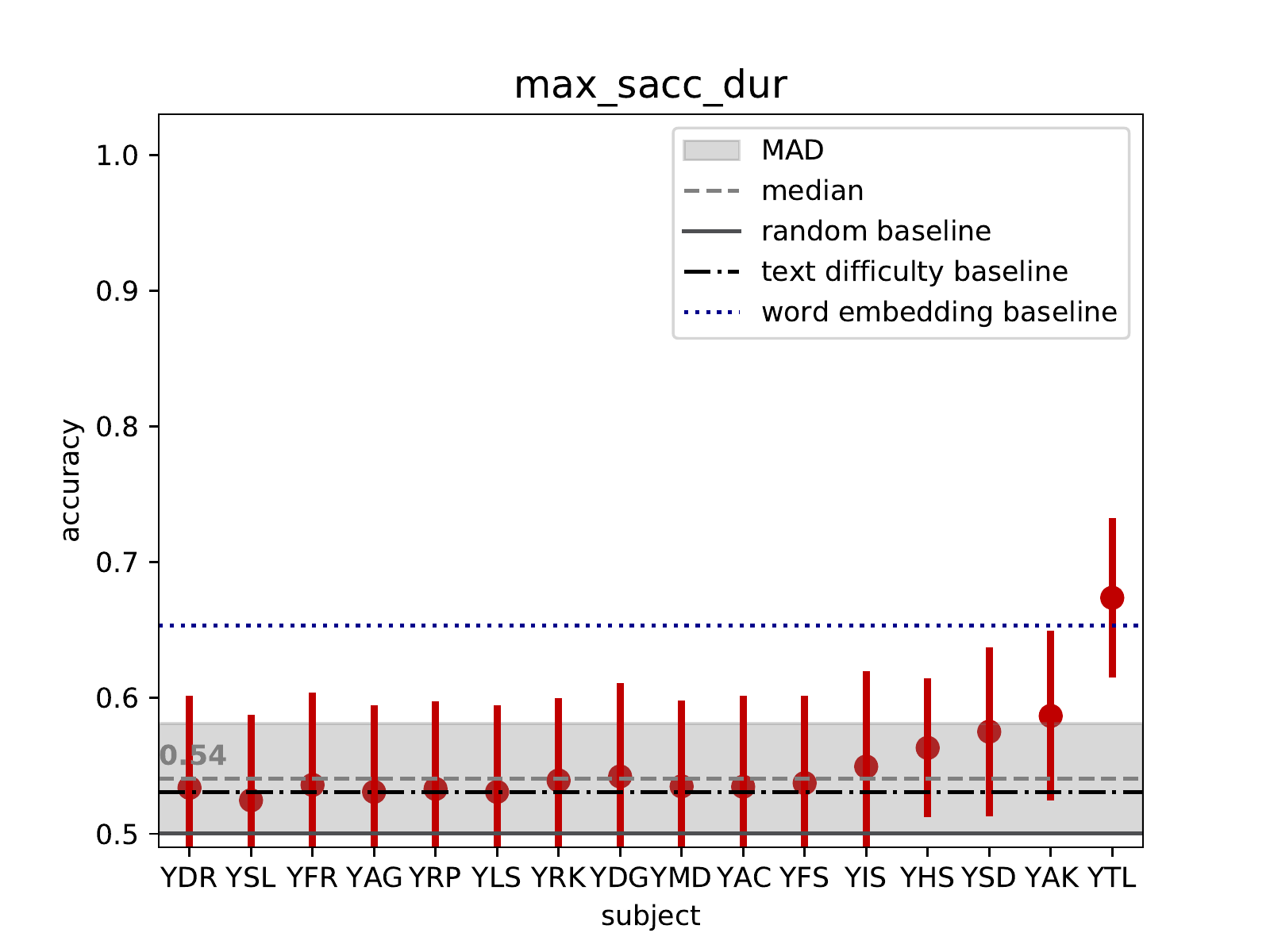} 
    \includegraphics[width=0.32\textwidth]{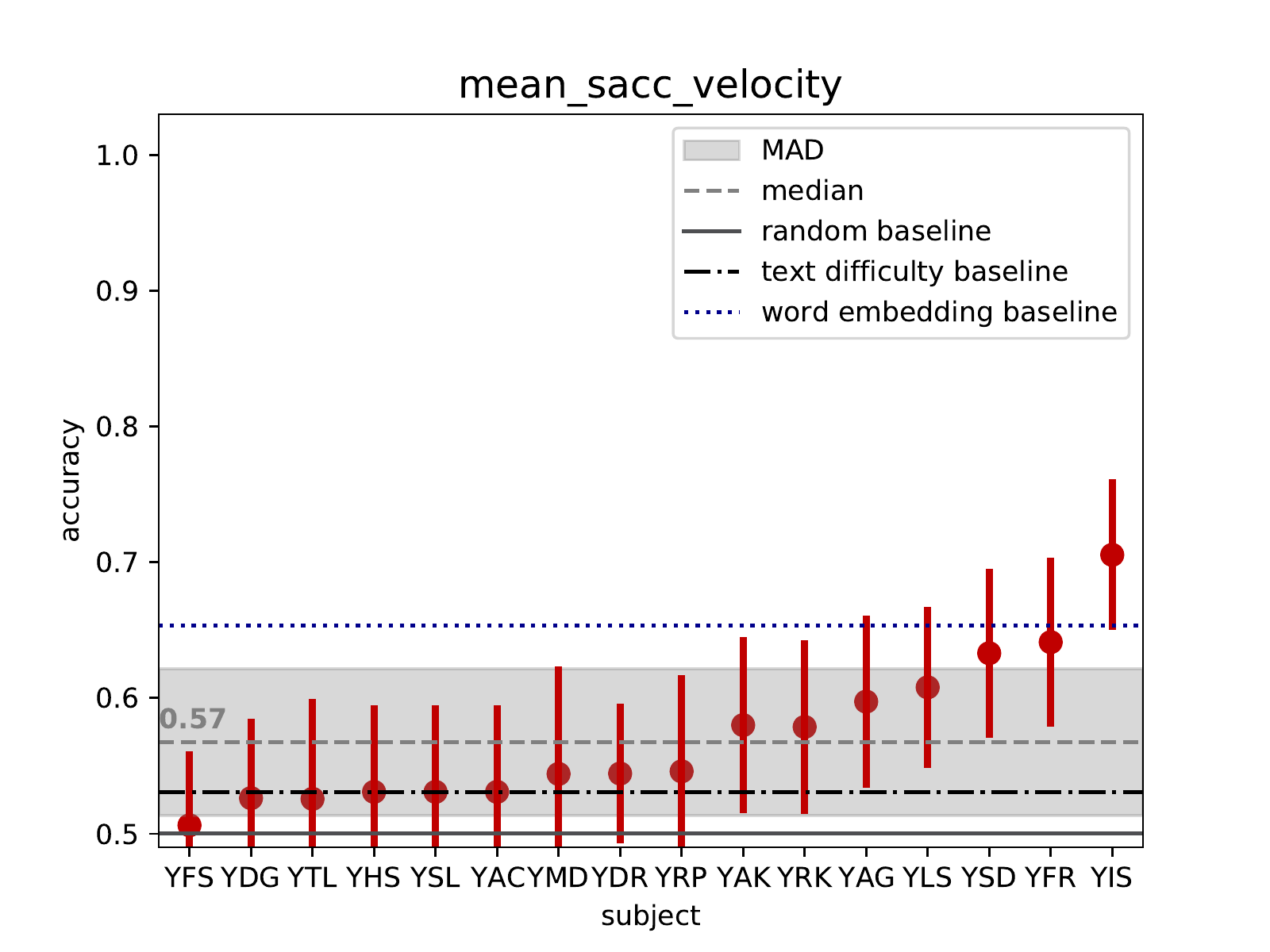} 
    \includegraphics[width=0.32\textwidth]{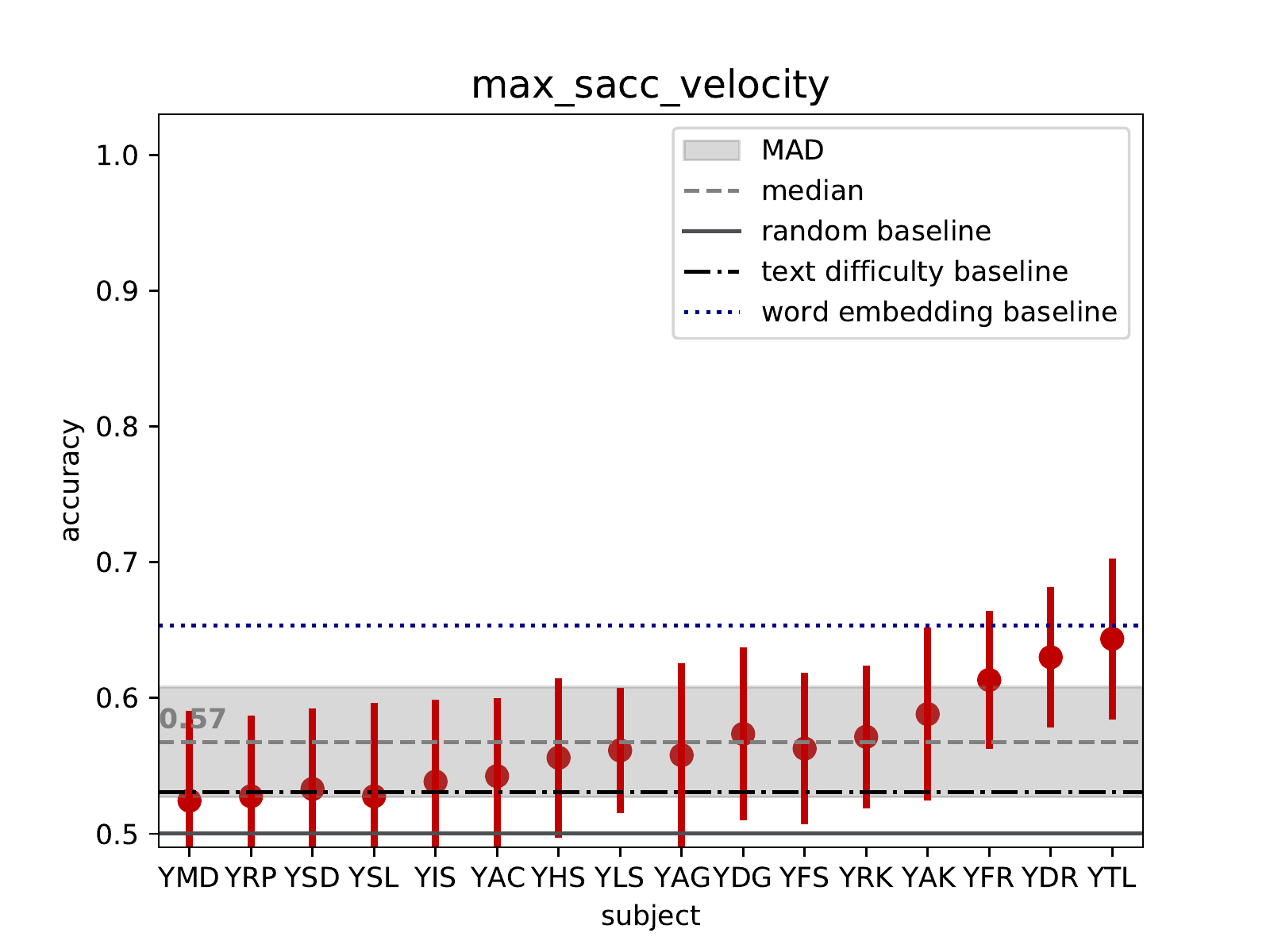} 
    \includegraphics[width=0.32\textwidth]{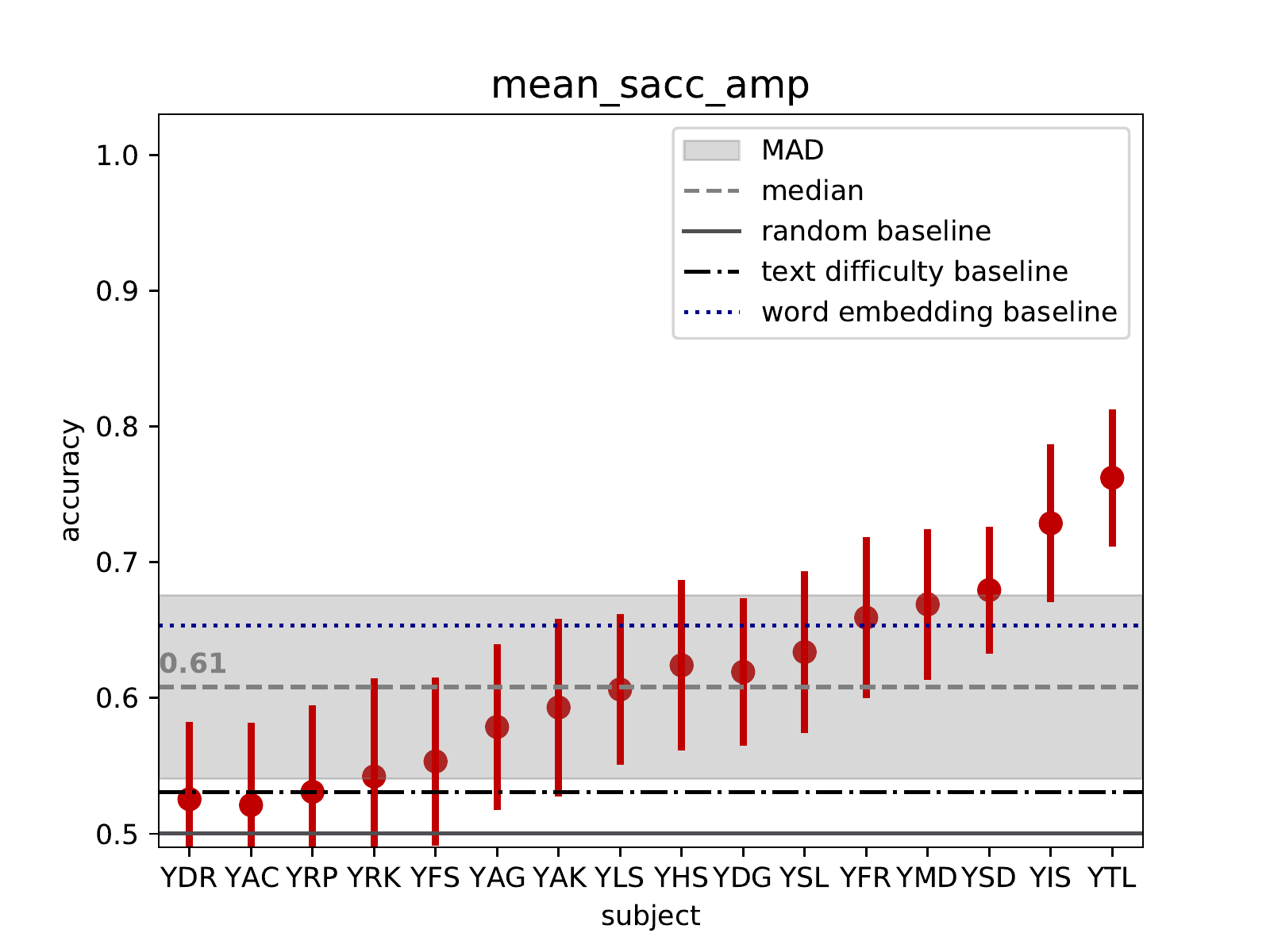} 
    \includegraphics[width=0.32\textwidth]{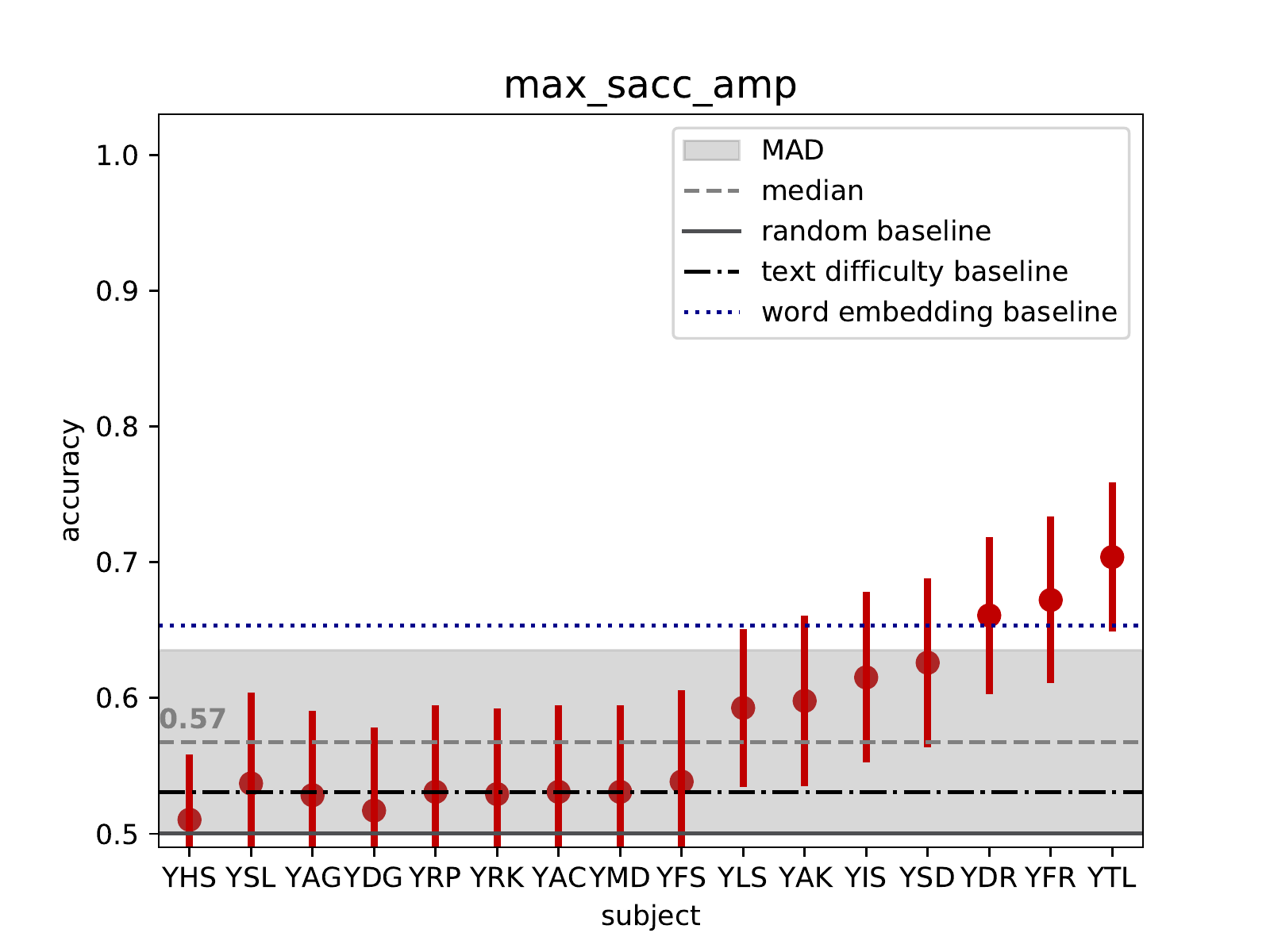} 
    \includegraphics[width=0.32\textwidth]{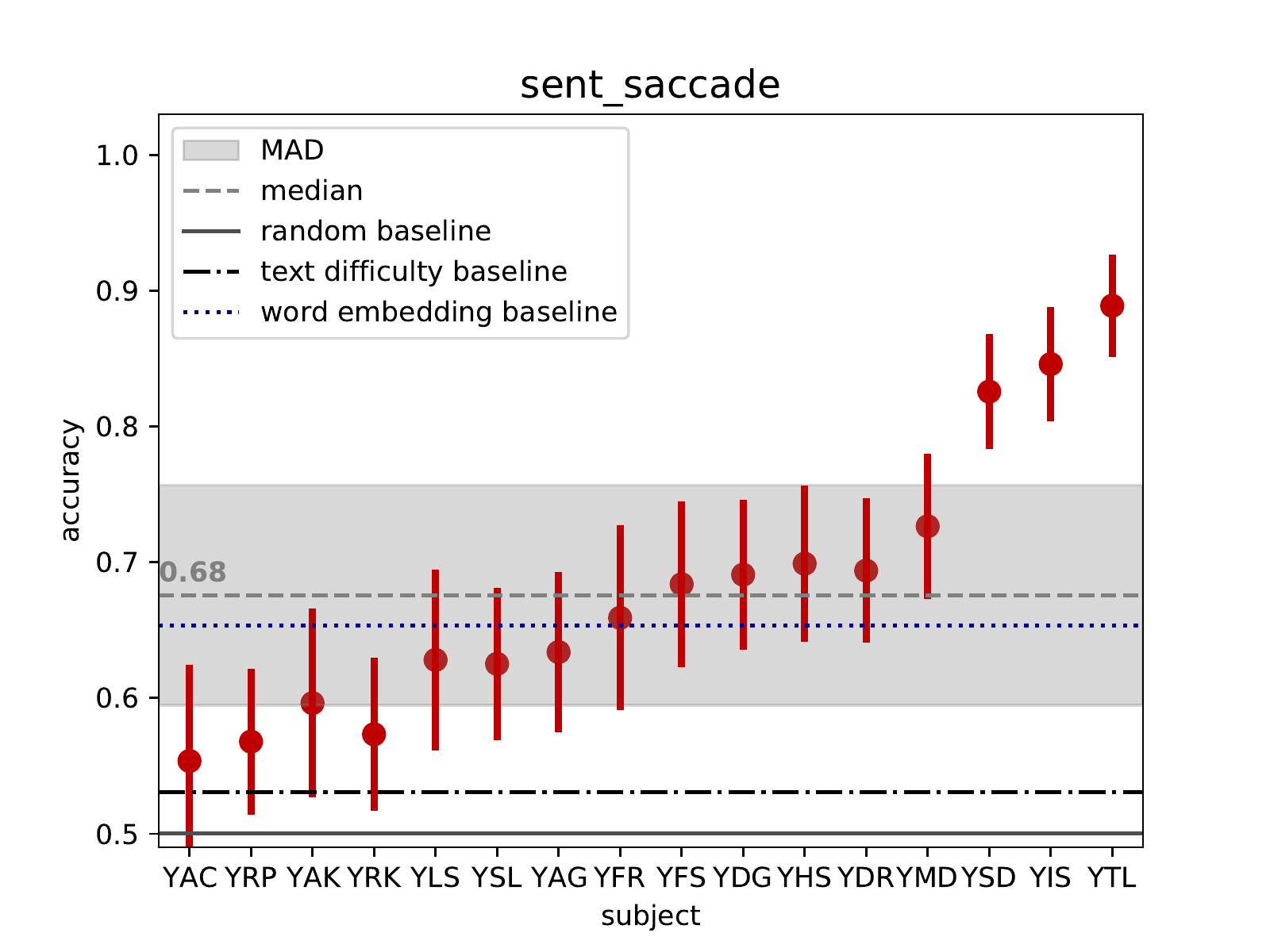} 
    \includegraphics[width=0.32\textwidth]{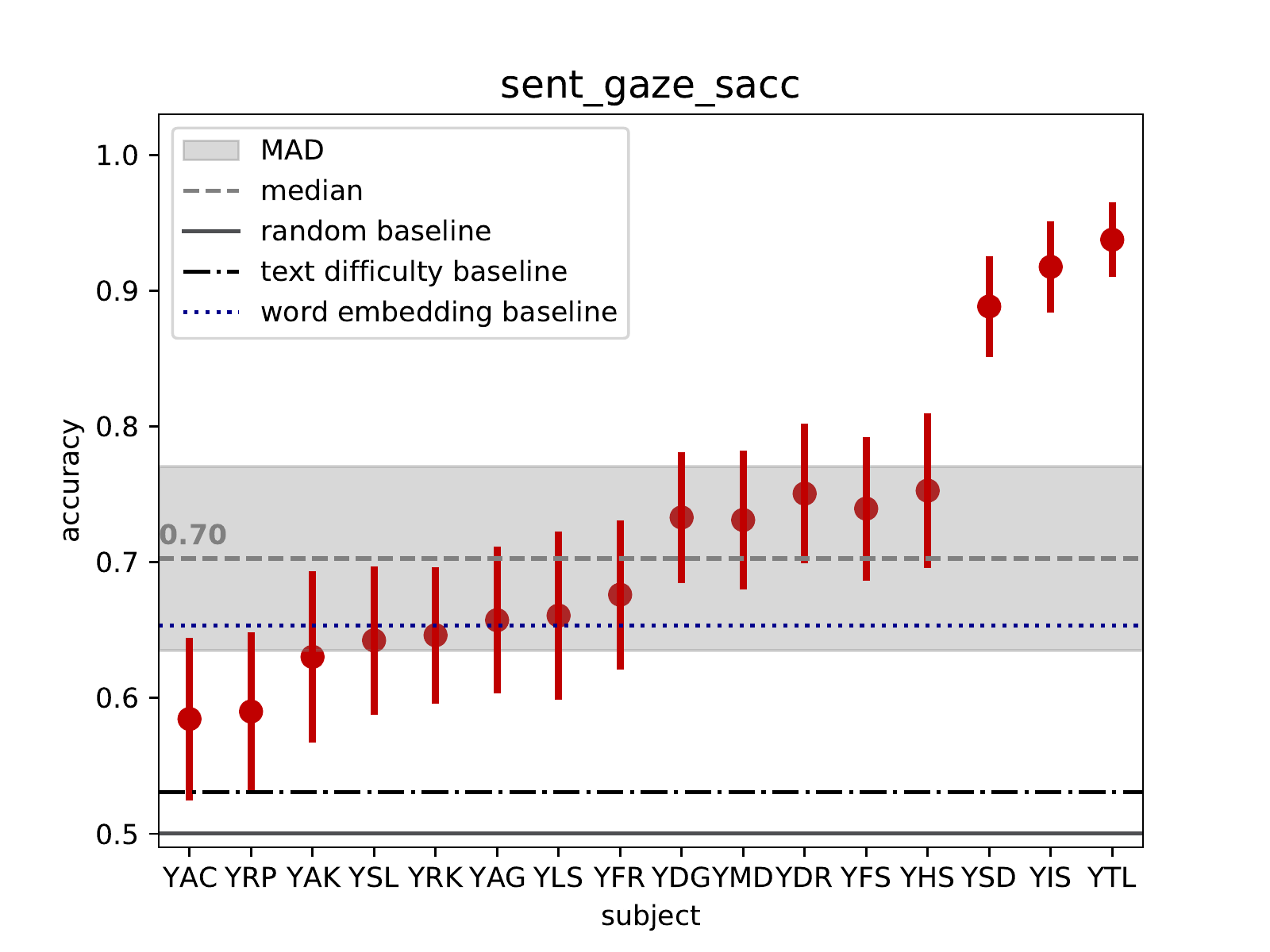} 
    \caption{Eye-tracking sentence-level classification accuracy on the ZuCo 2.0 data.}
    \label{fig:sent-res-z2-et}
\end{figure}

\noindent The results for the sentence-level within-subject models are summarized in Table \ref{tab:sent-level-res-summary}.
The results on all eye-tracking feature sets are presented in Figure \ref{fig:sent-res-z1-et} for ZuCo 1.0 and in Figure \ref{fig:sent-res-z2-et} for ZuCo 2.0.
The results show that the models trained on the three fixation-based features perform similarly, but combining them substantially improves the accuracy (\textit{sent\_gaze}: ZuCo 1.0 = 76\% accuracy, ZuCo 2.0 = 68\% accuracy). Moreover, the models trained on the individual saccade features perform worse than the models trained on the fixation features (with only a few subjects considerably higher than random performance), but again the combination of multiple features is helpful (\textit{sent\_saccade}: ZuCo 1.0 = 77\%, ZuCo 2.0 = 68\%). Finally, we observe that combining fixation and saccade features yields the best results using sentence-level eye-tracking features (\textit{sent\_gaze\_sacc}: ZuCo 1.0 = 82\%, ZuCo 2.0 = 70\%). Notably, the sentence-level eye-tracking features achieve markedly higher performance on both datasets compared to models based on word-level features.

\begin{figure}[ht]
    \centering
    \includegraphics[width=0.32\textwidth]{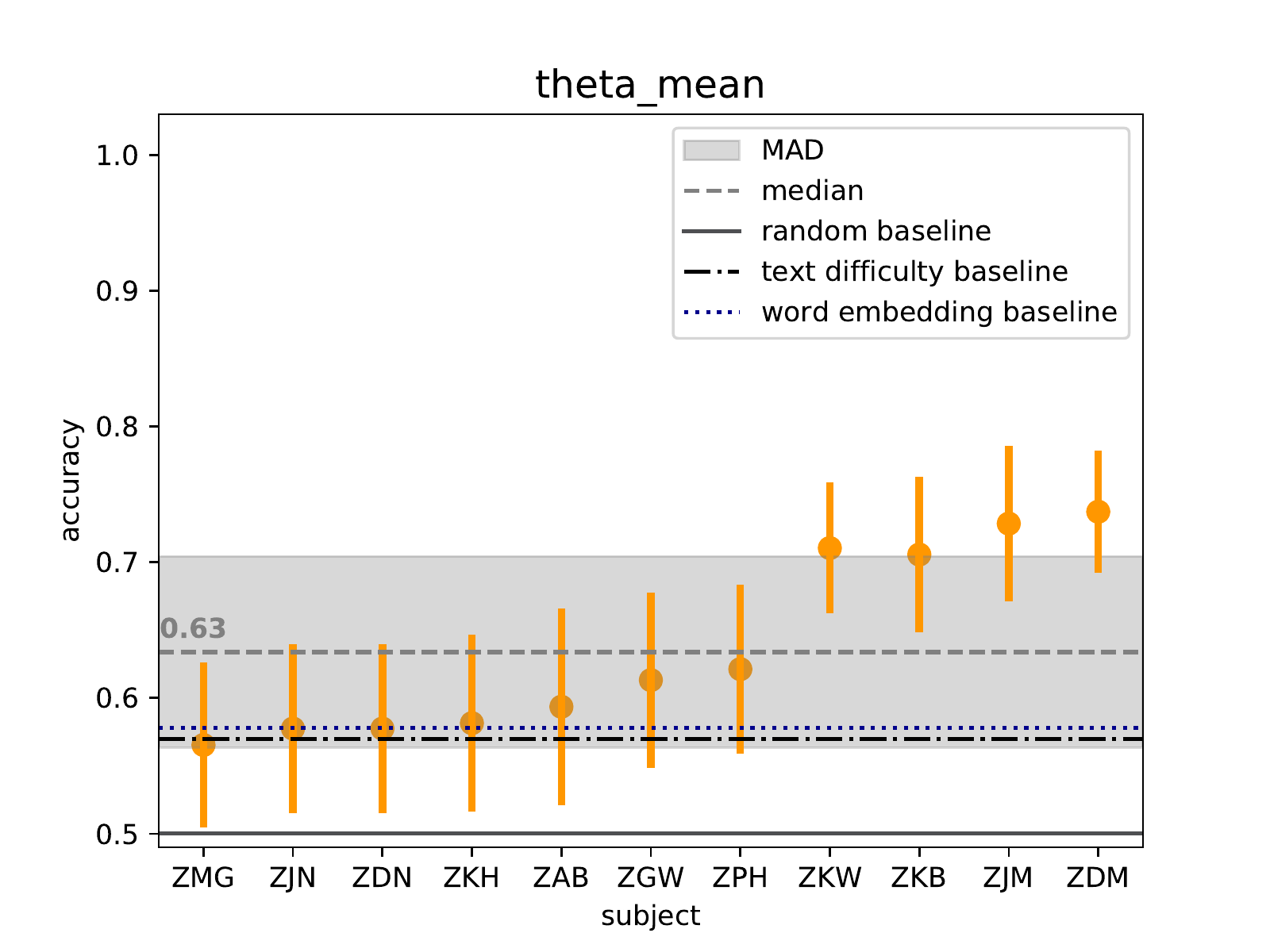} 
    \includegraphics[width=0.32\textwidth]{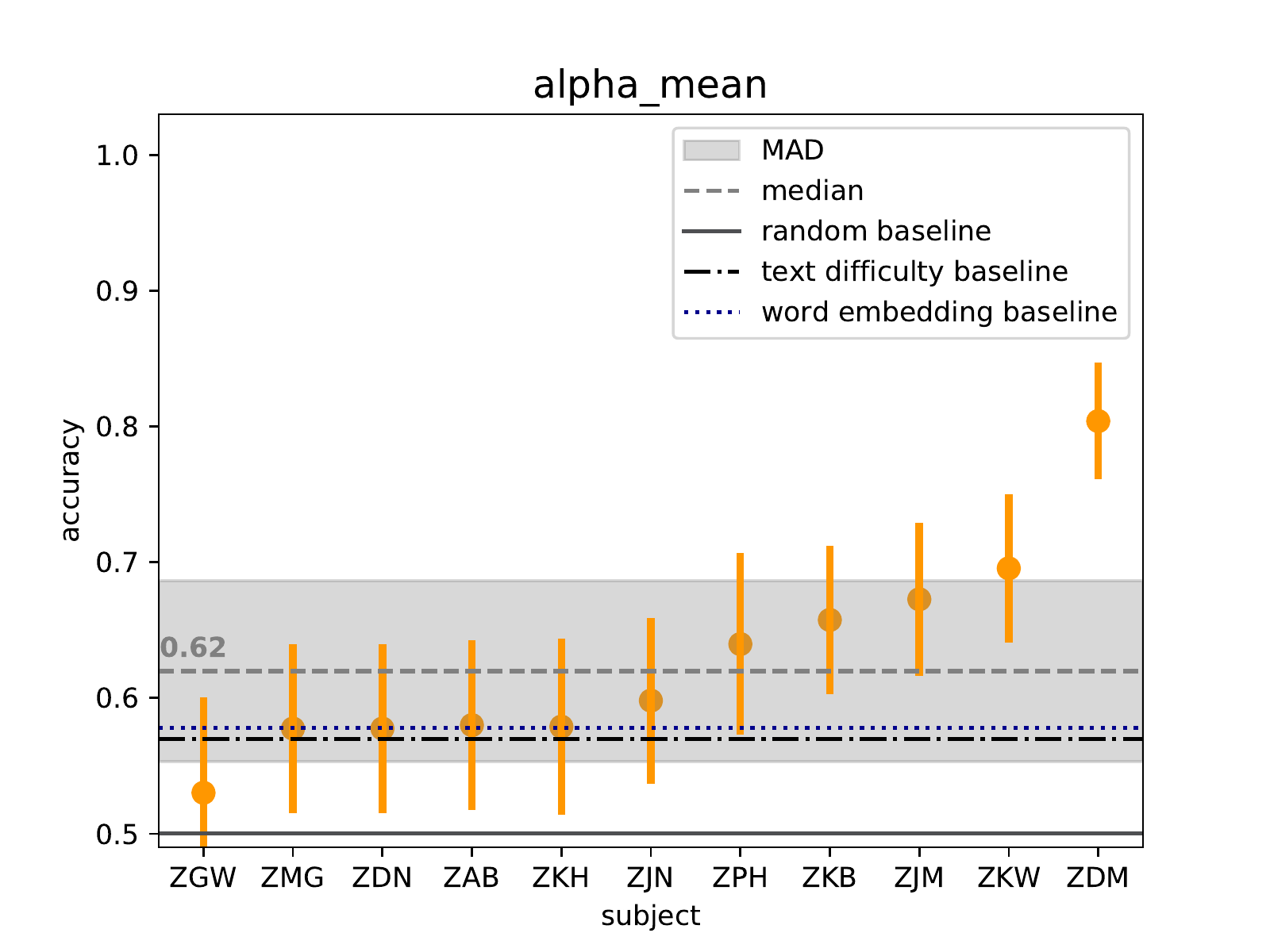} 
    \includegraphics[width=0.32\textwidth]{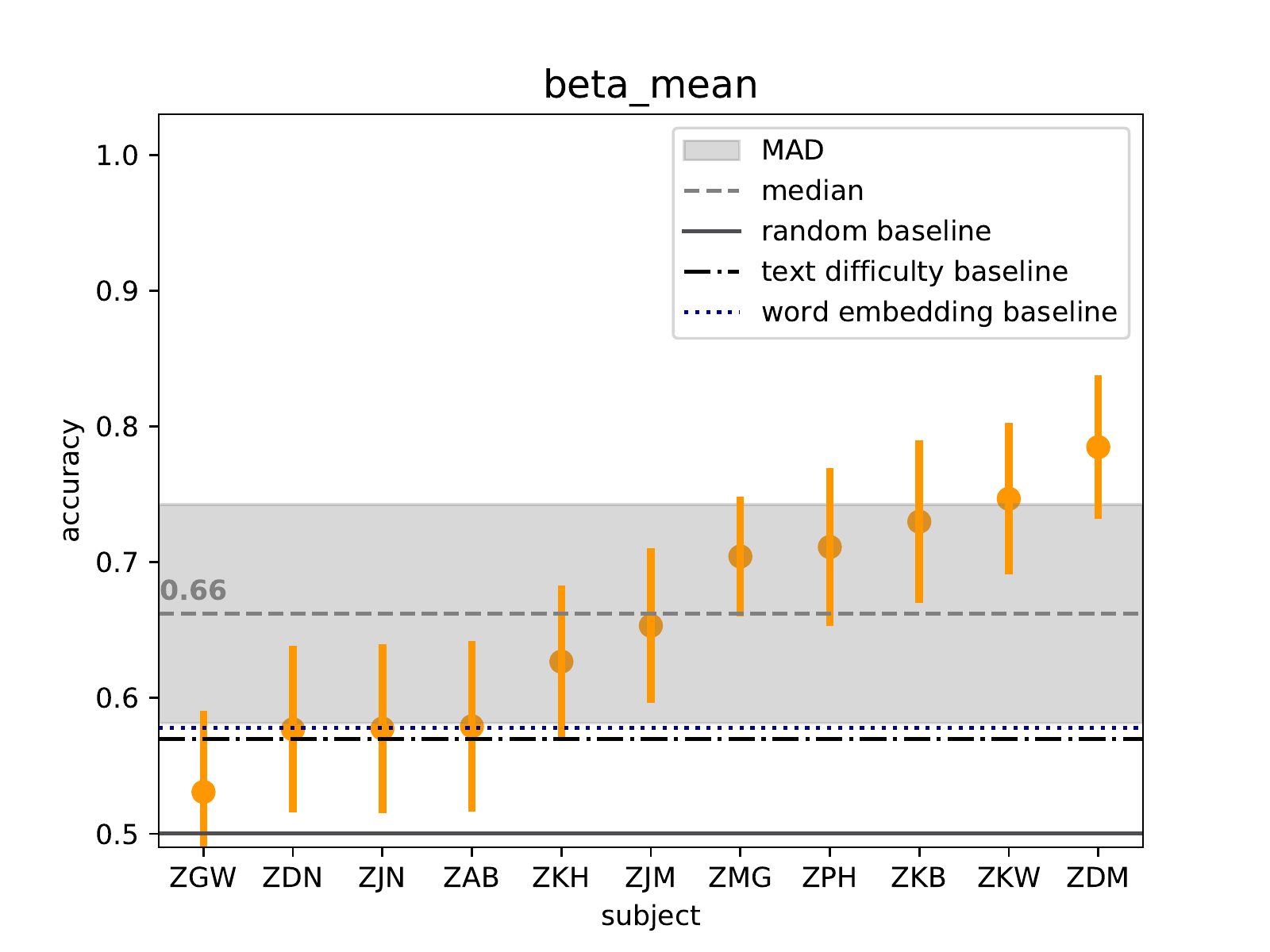} 
    \includegraphics[width=0.32\textwidth]{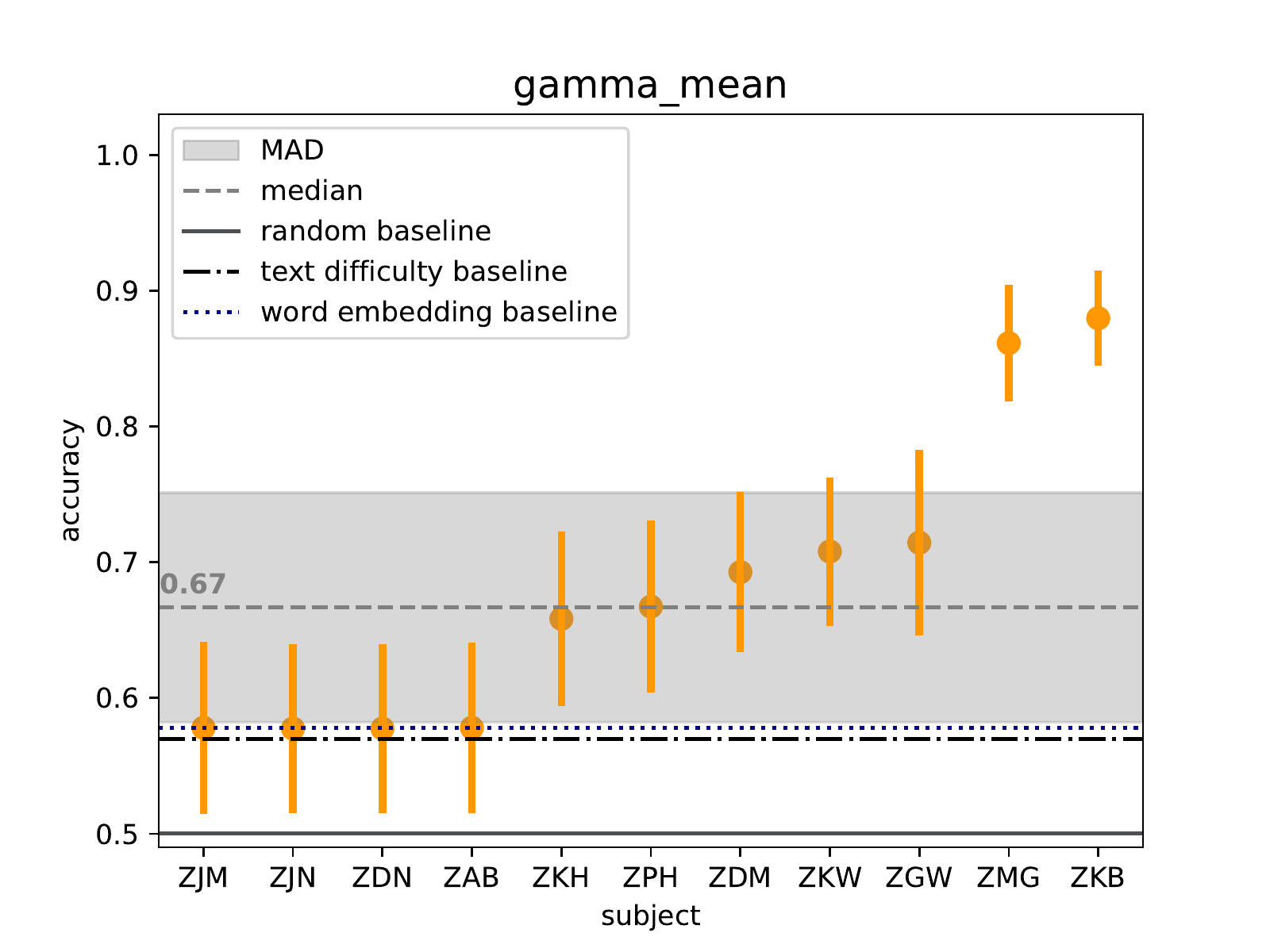} 
    \includegraphics[width=0.32\textwidth]{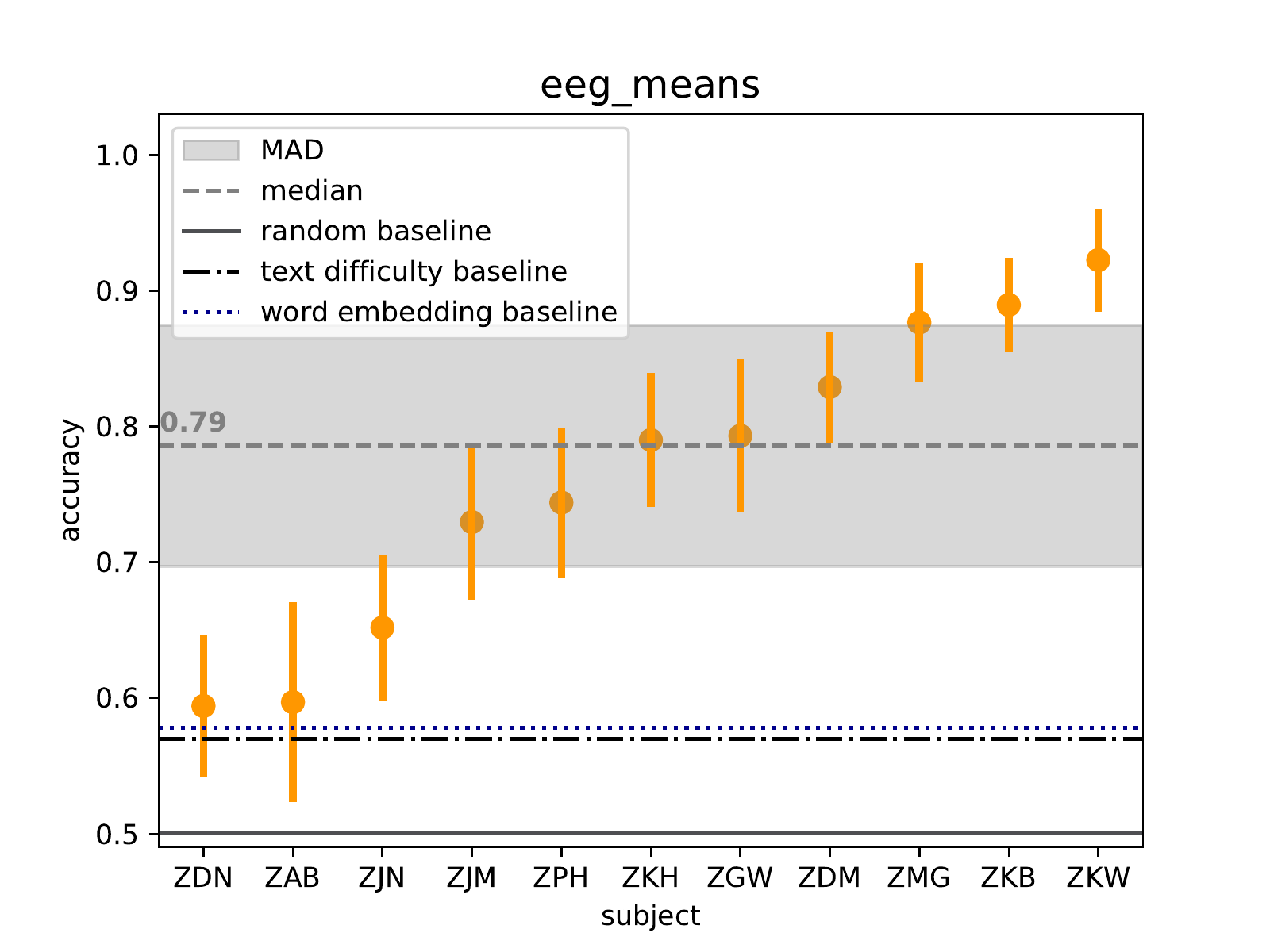} 
    \includegraphics[width=0.32\textwidth]{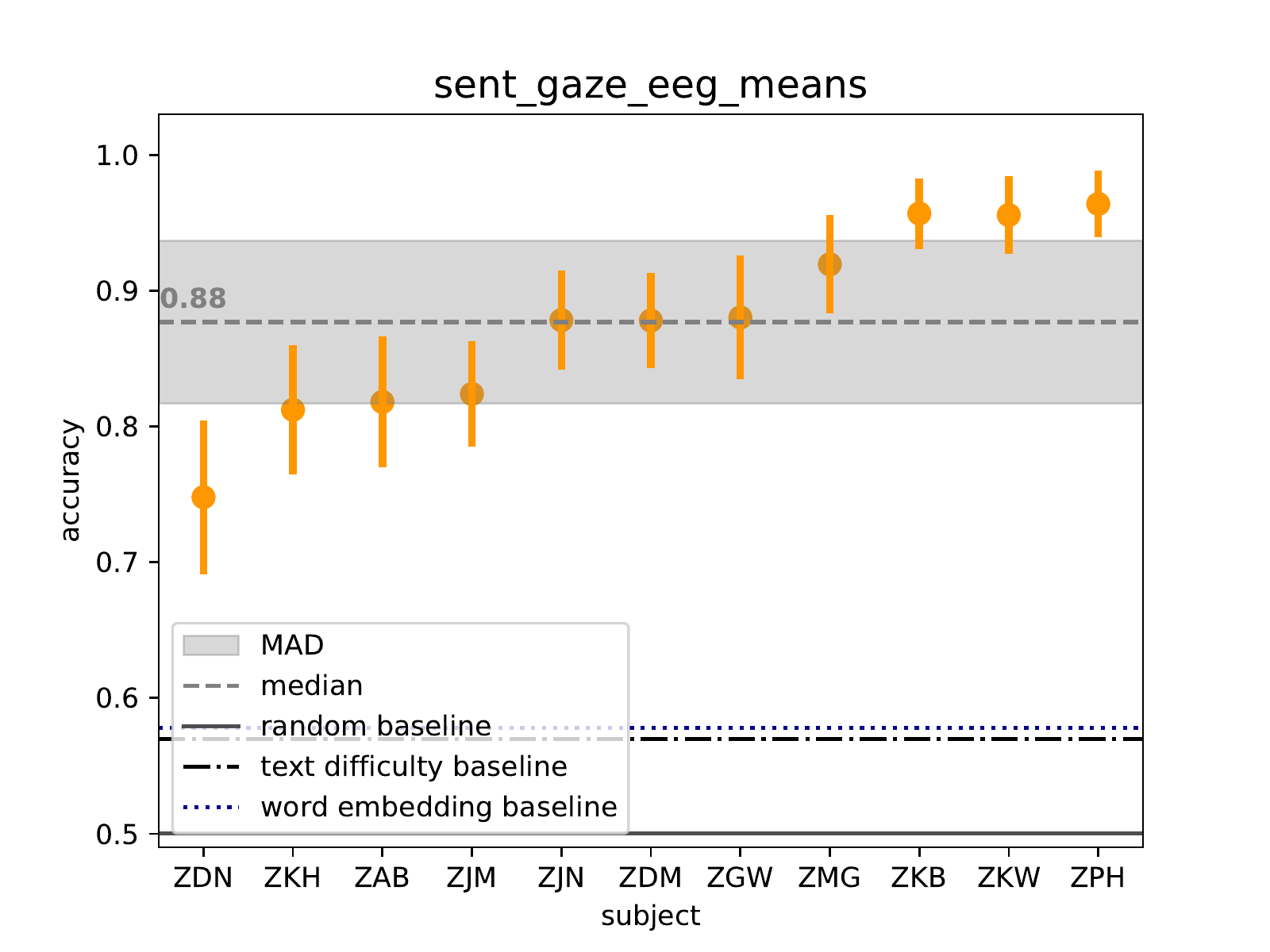} 
\includegraphics[width=0.32\textwidth]{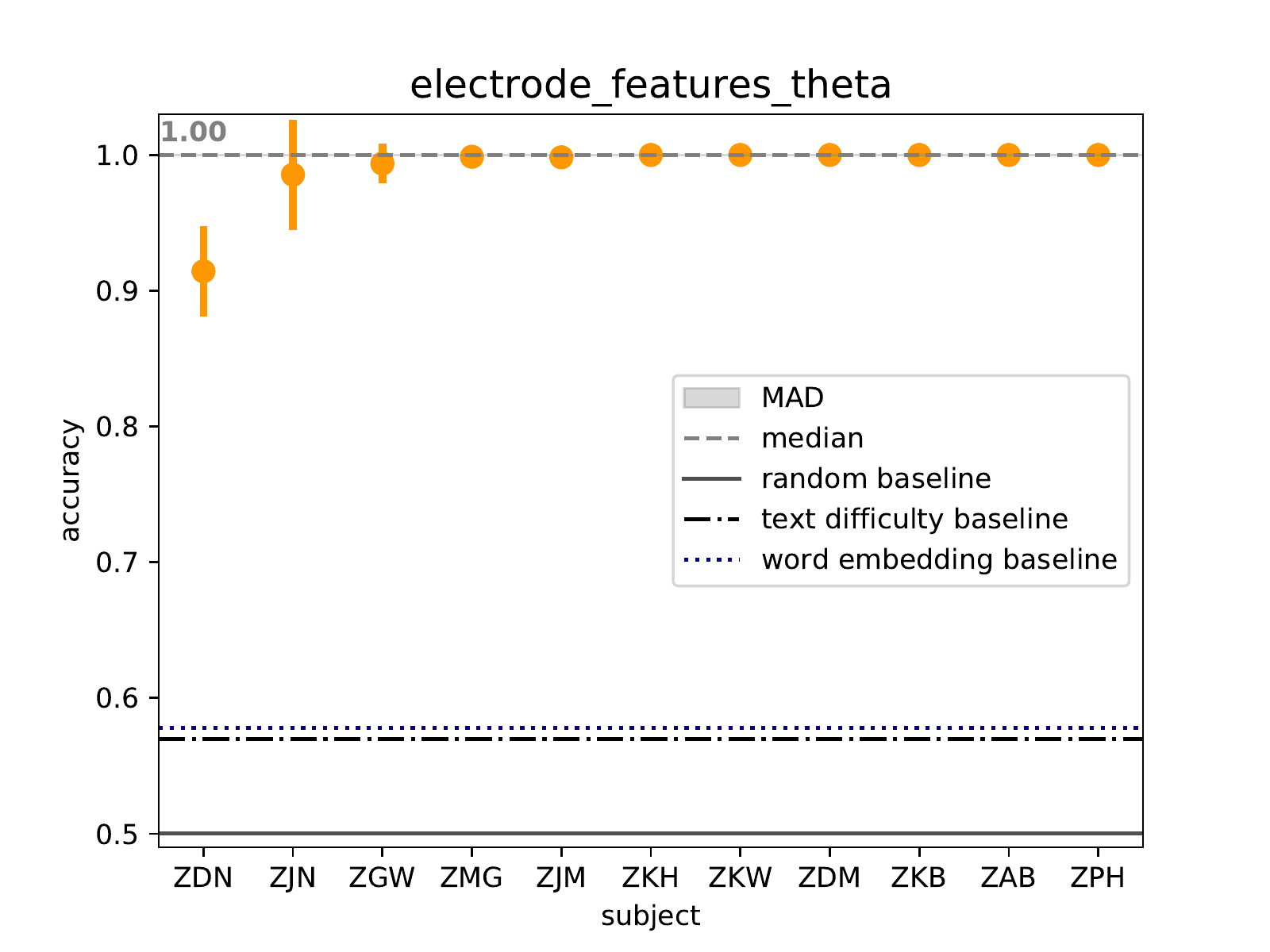} 
    \includegraphics[width=0.32\textwidth]{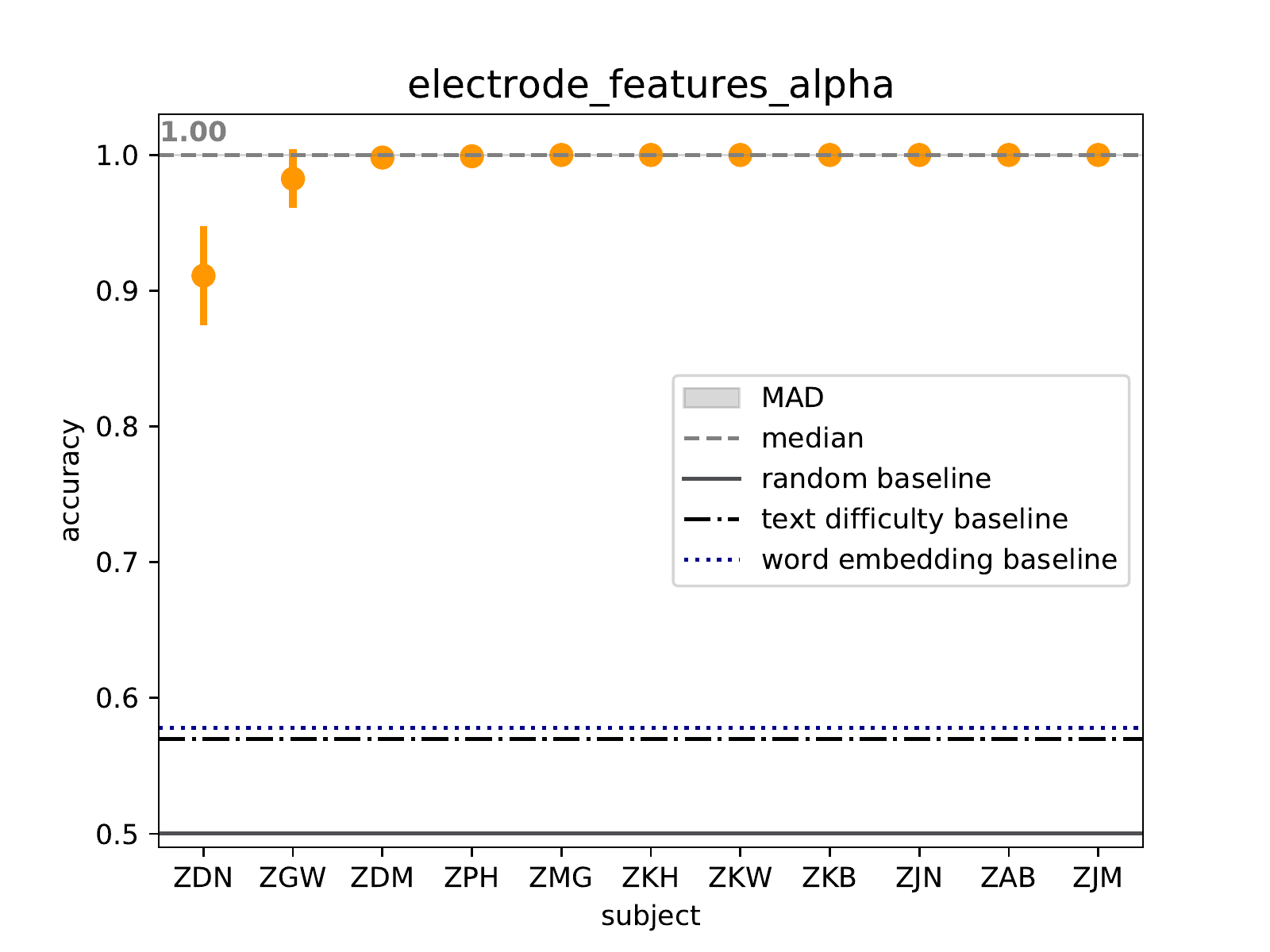} 
        \includegraphics[width=0.32\textwidth]{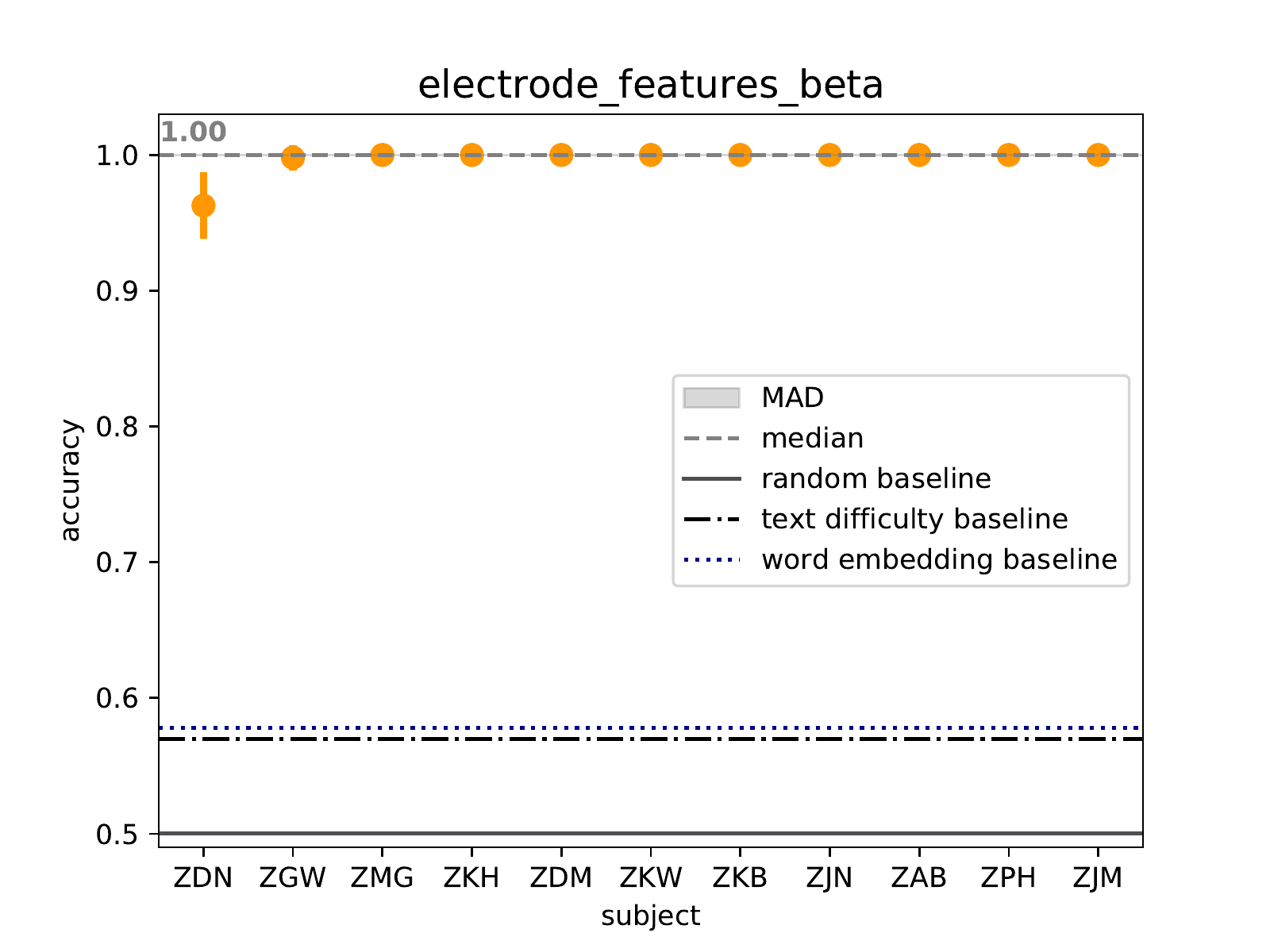}
    \includegraphics[width=0.32\textwidth]{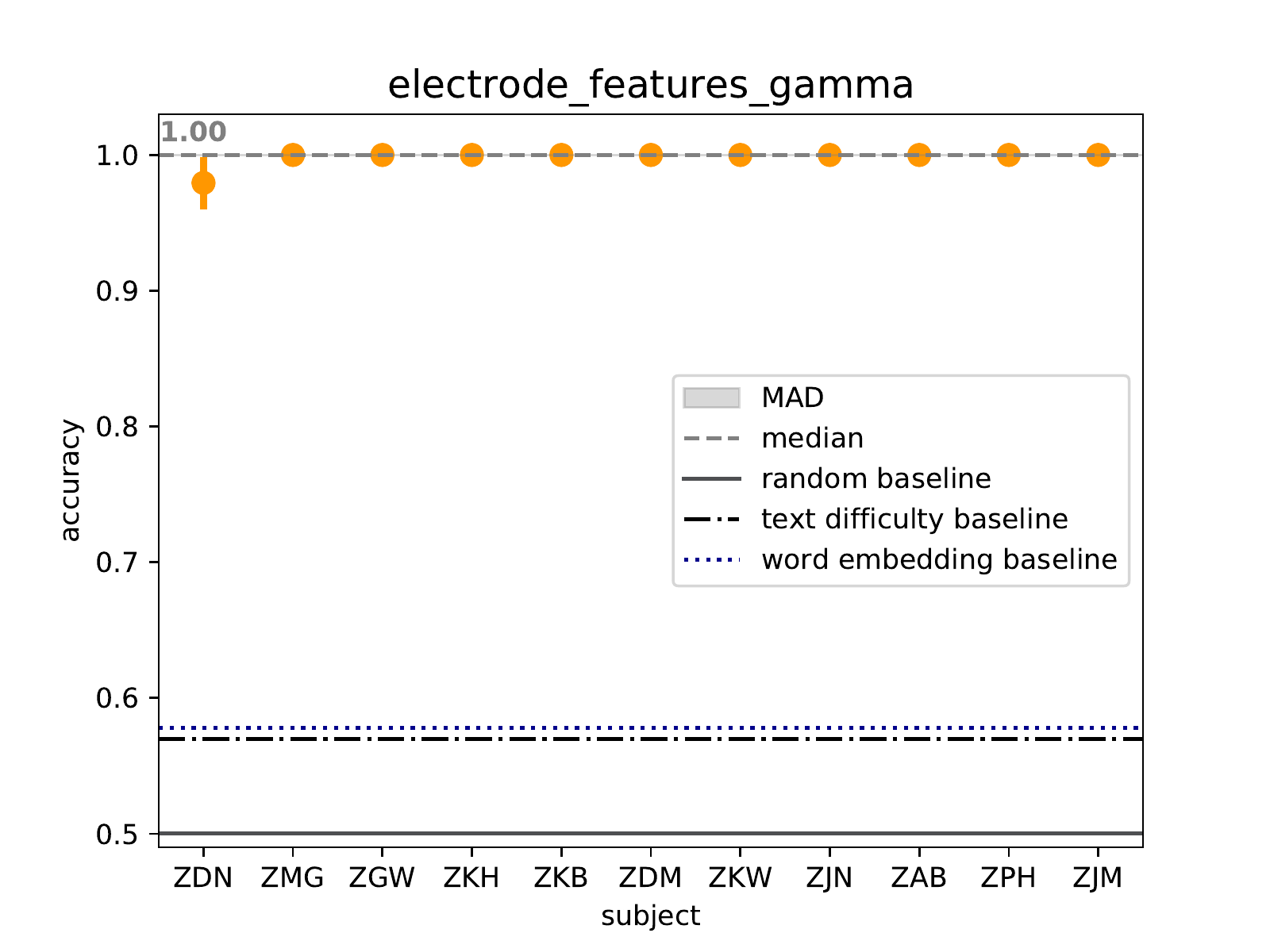} 
     \includegraphics[width=0.32\textwidth]{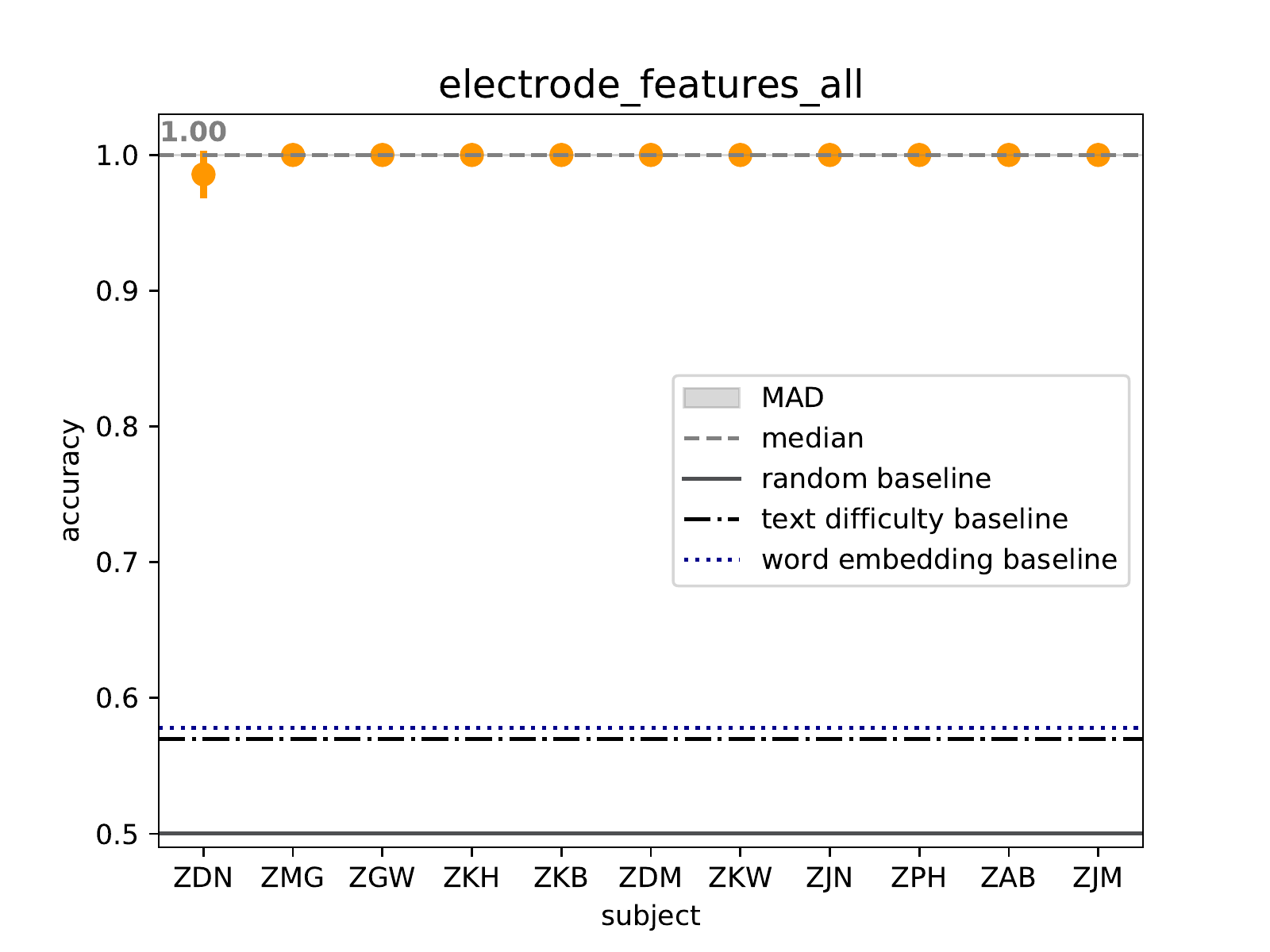} 
    \caption{EEG sentence-level classification accuracy on the ZuCo 1.0 data.}
    \label{fig:sent-res-z1-eeg}
\end{figure}

\begin{figure}[ht!]
    \centering
    \includegraphics[width=0.32\textwidth]{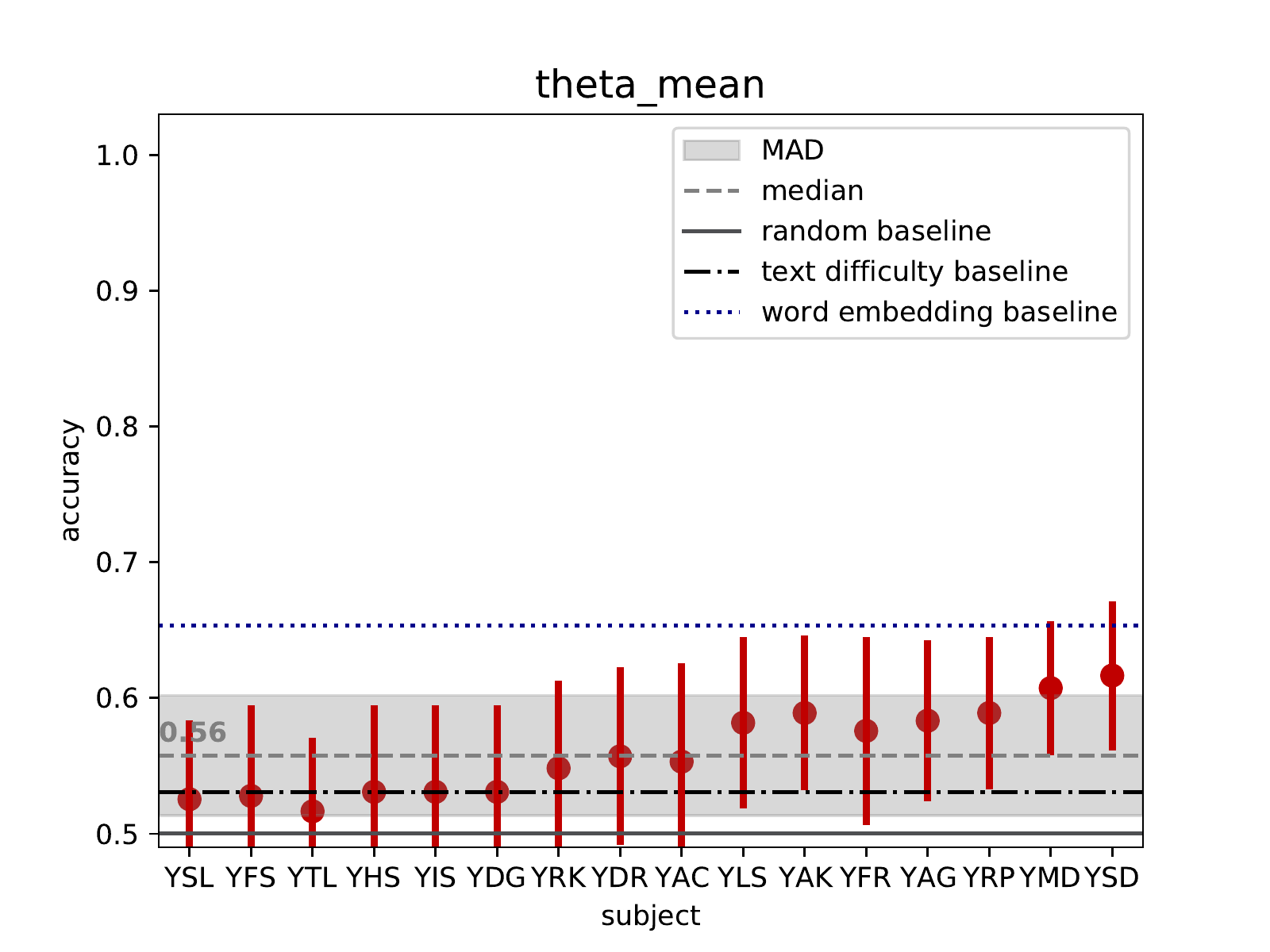} 
    \includegraphics[width=0.32\textwidth]{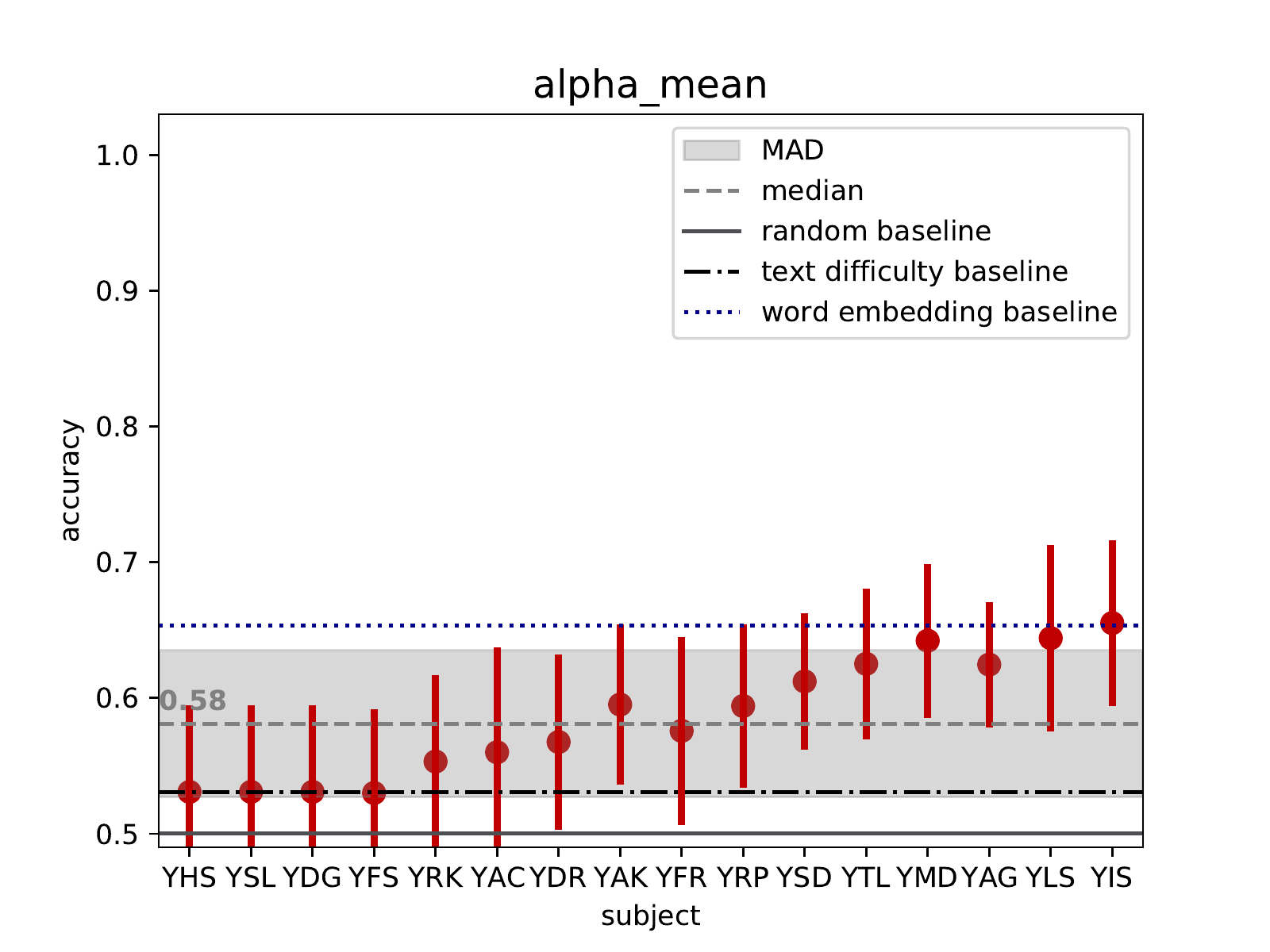} 
    \includegraphics[width=0.32\textwidth]{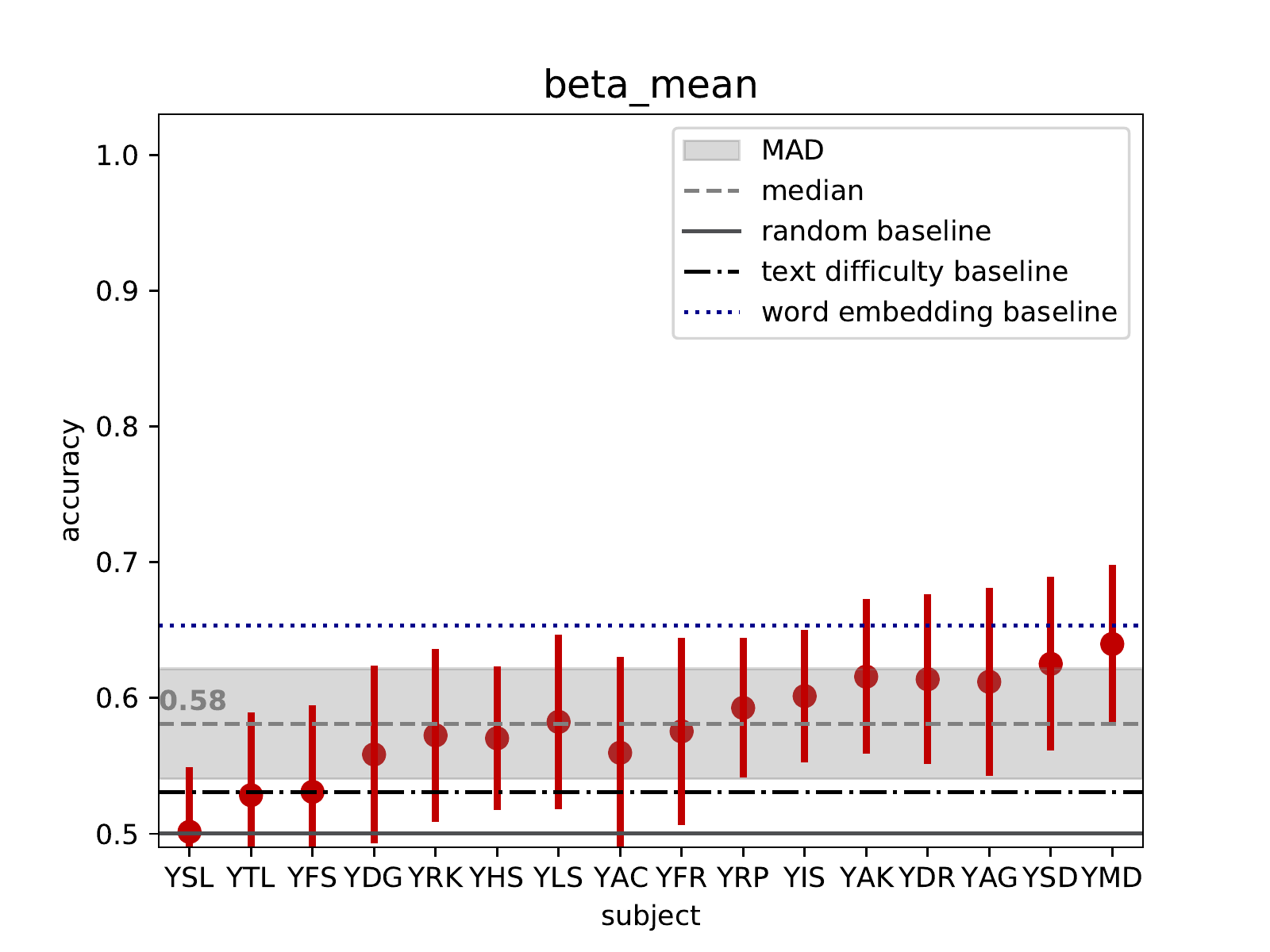} 
    \includegraphics[width=0.32\textwidth]{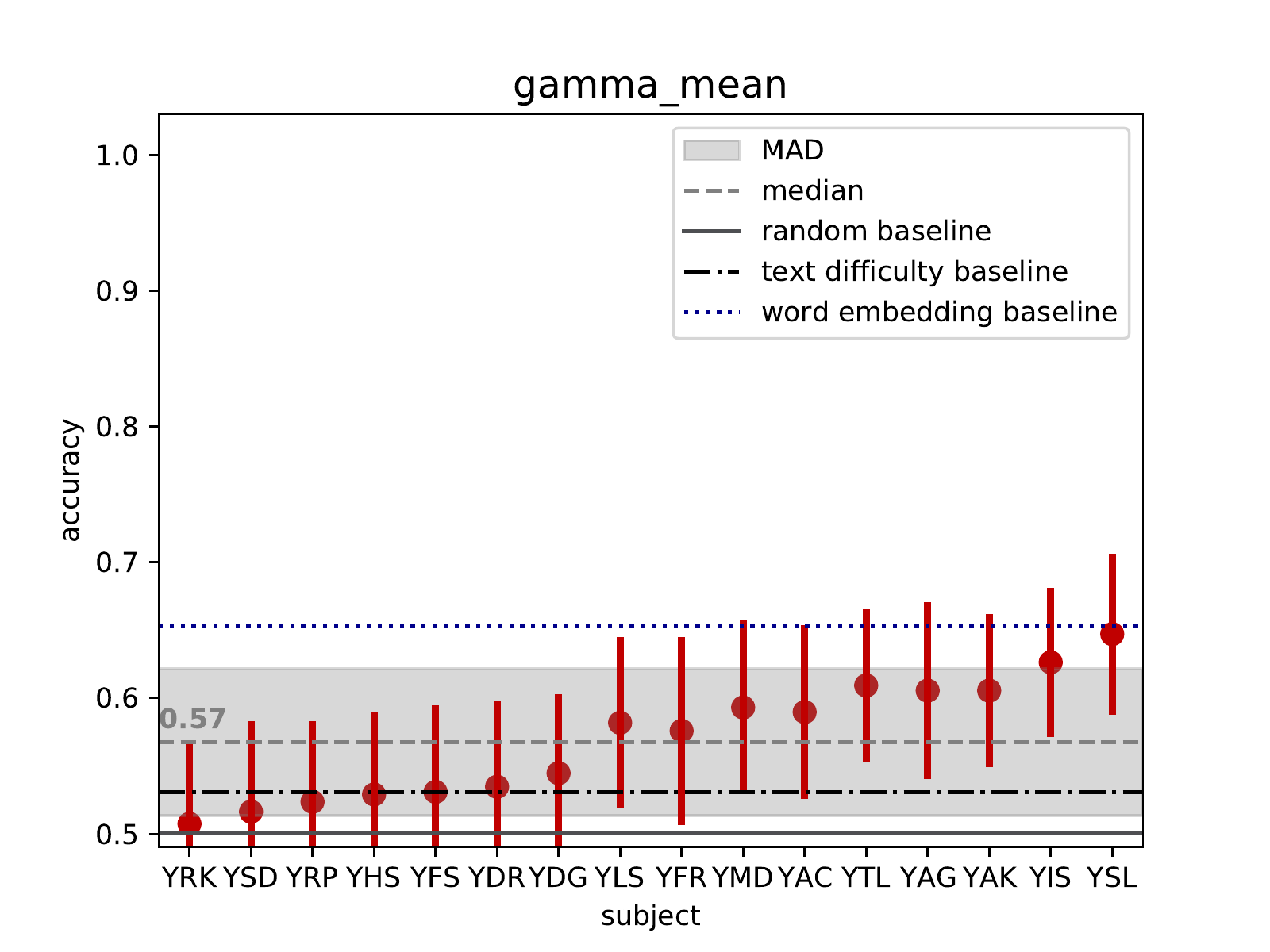} 
    \includegraphics[width=0.32\textwidth]{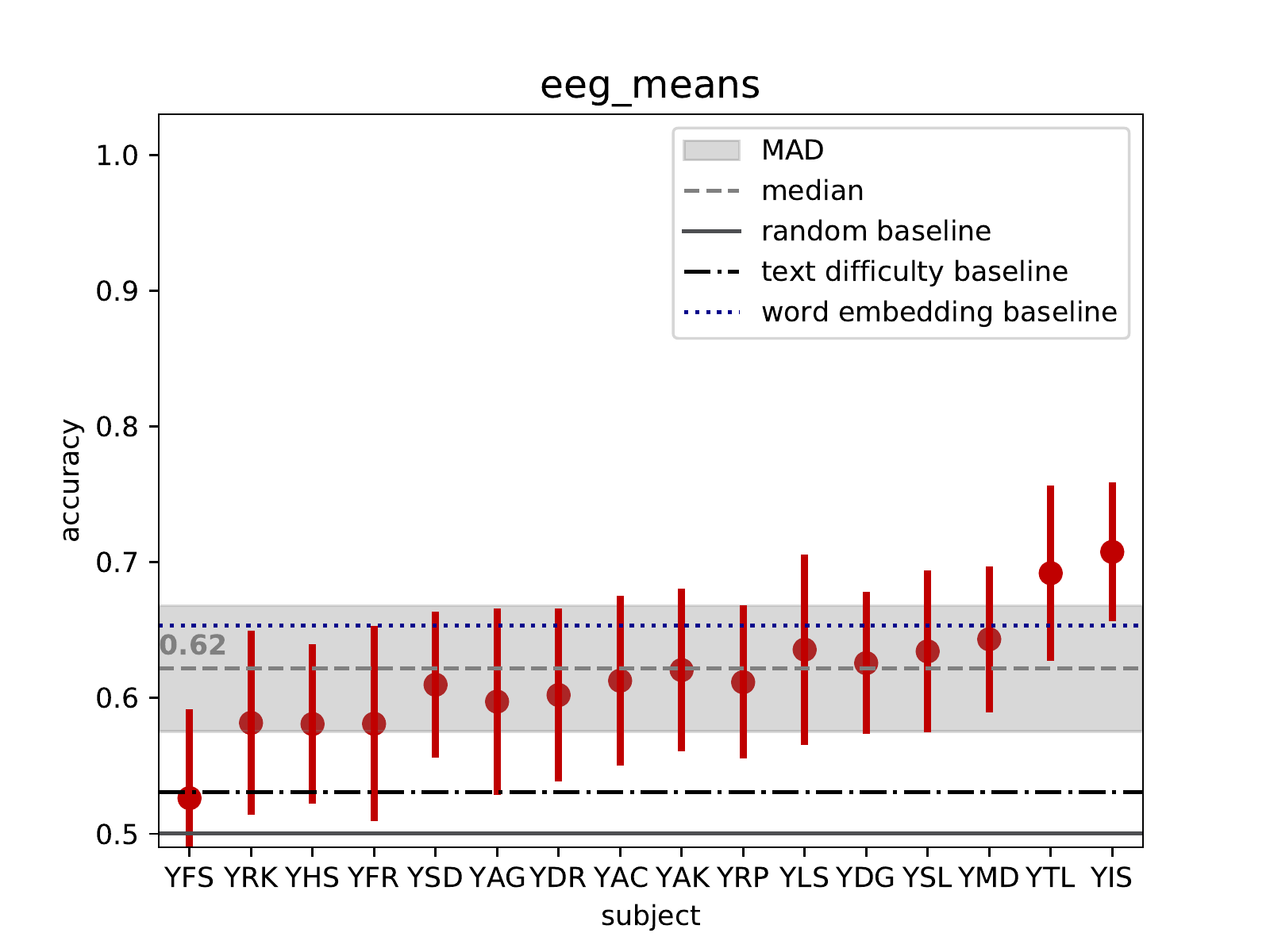} 
    \includegraphics[width=0.32\textwidth]{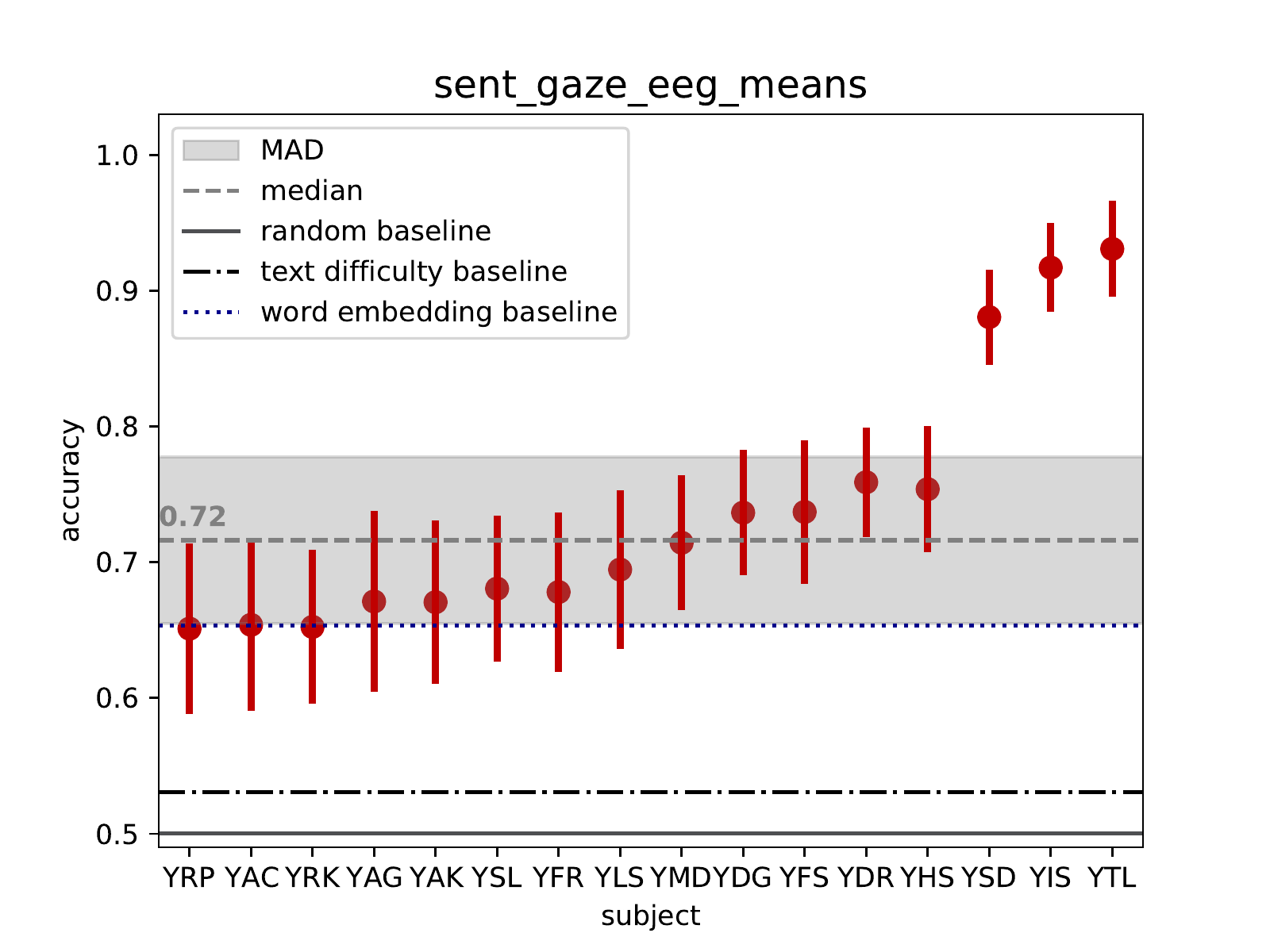} 
    \includegraphics[width=0.32\textwidth]{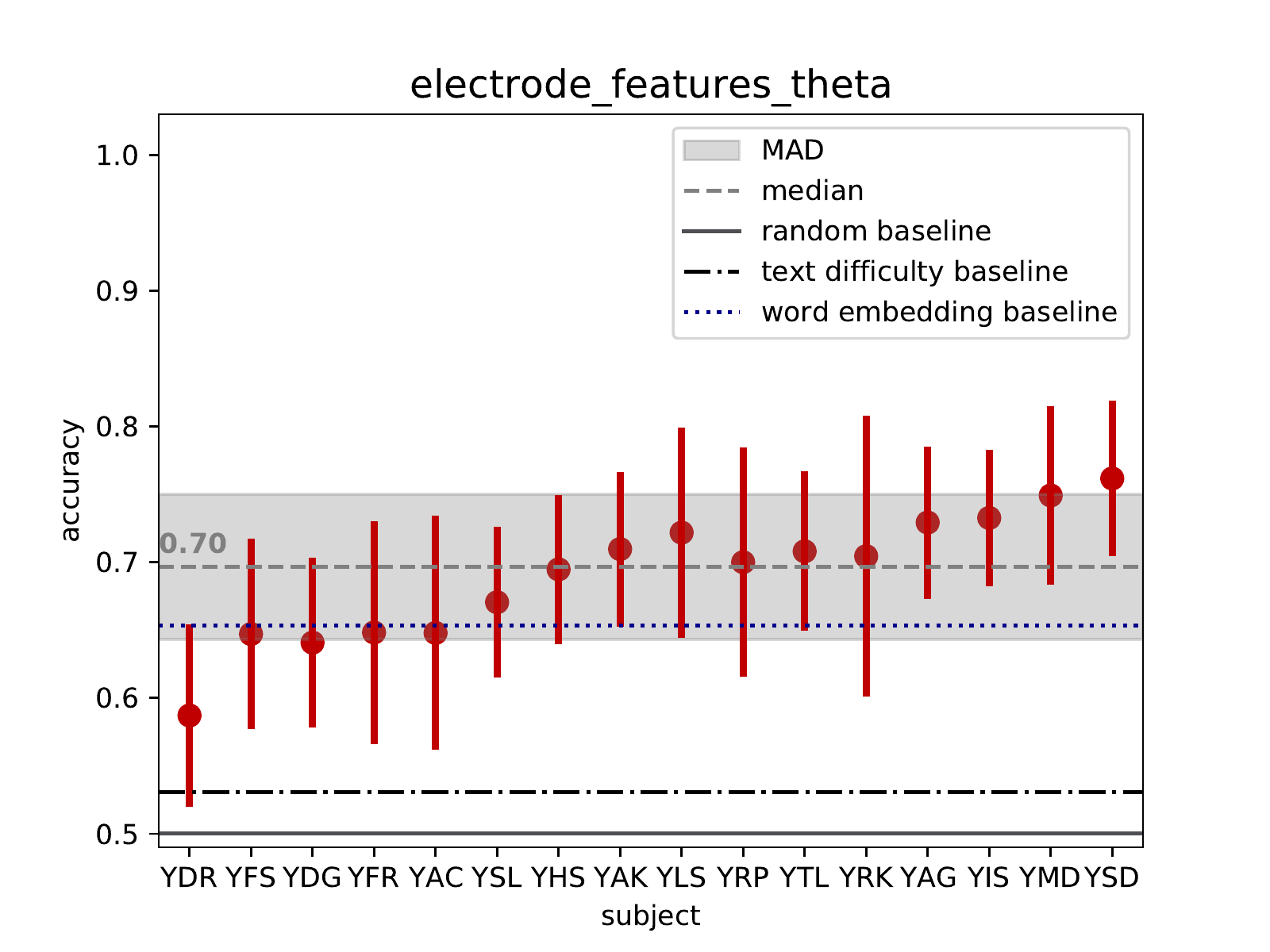} 
    \includegraphics[width=0.32\textwidth]{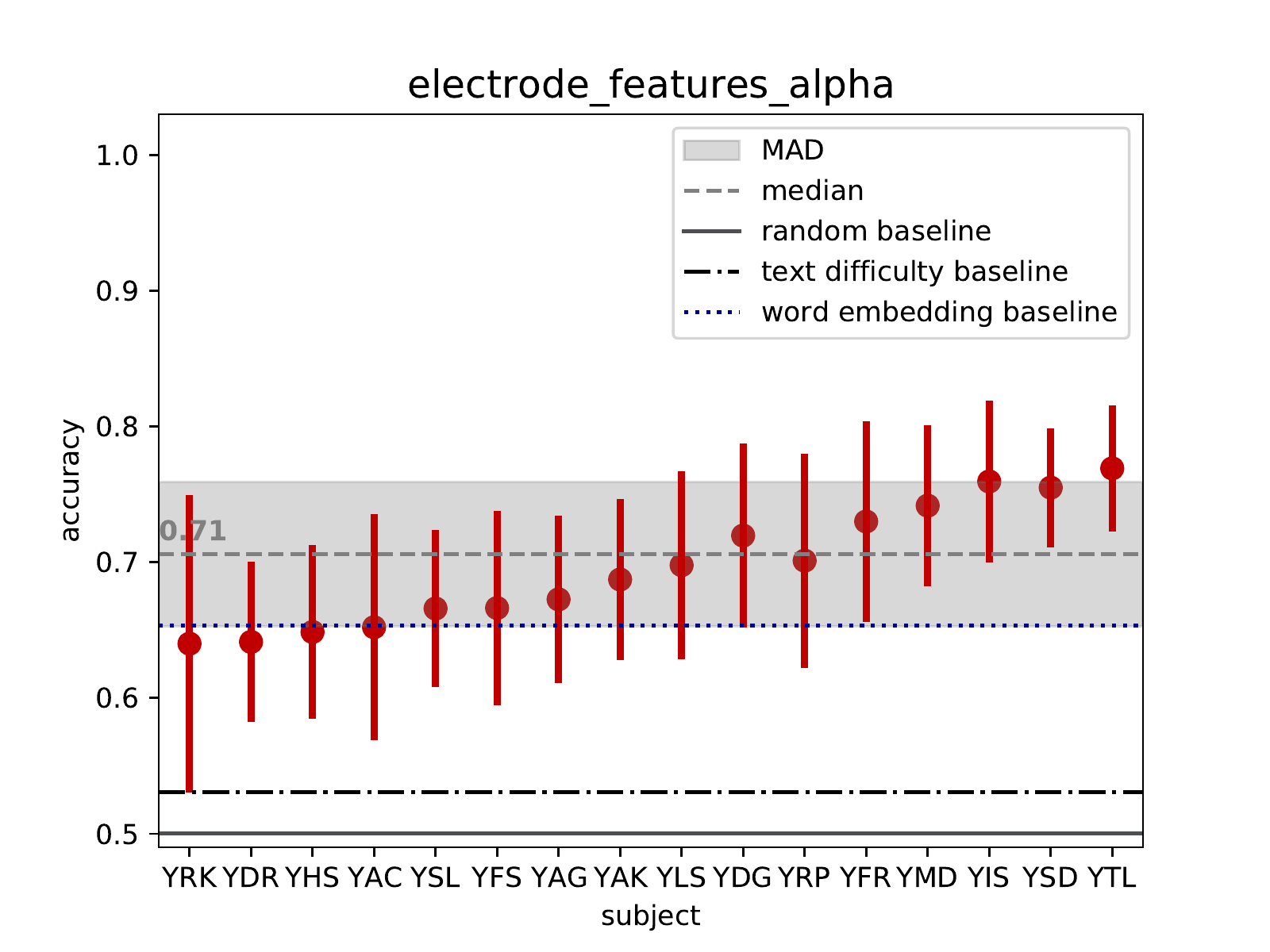} 
    \includegraphics[width=0.32\textwidth]{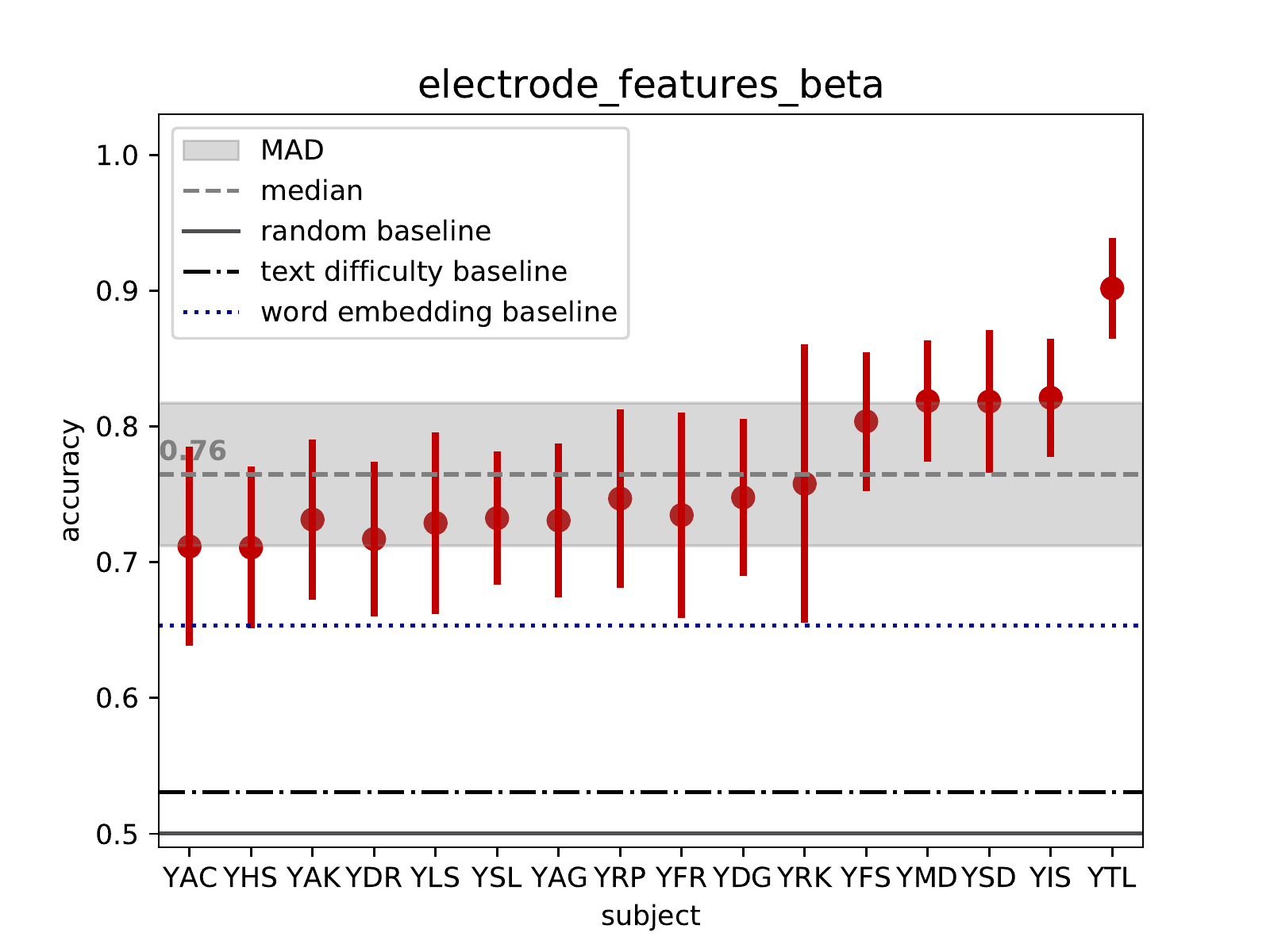}
    \includegraphics[width=0.32\textwidth]{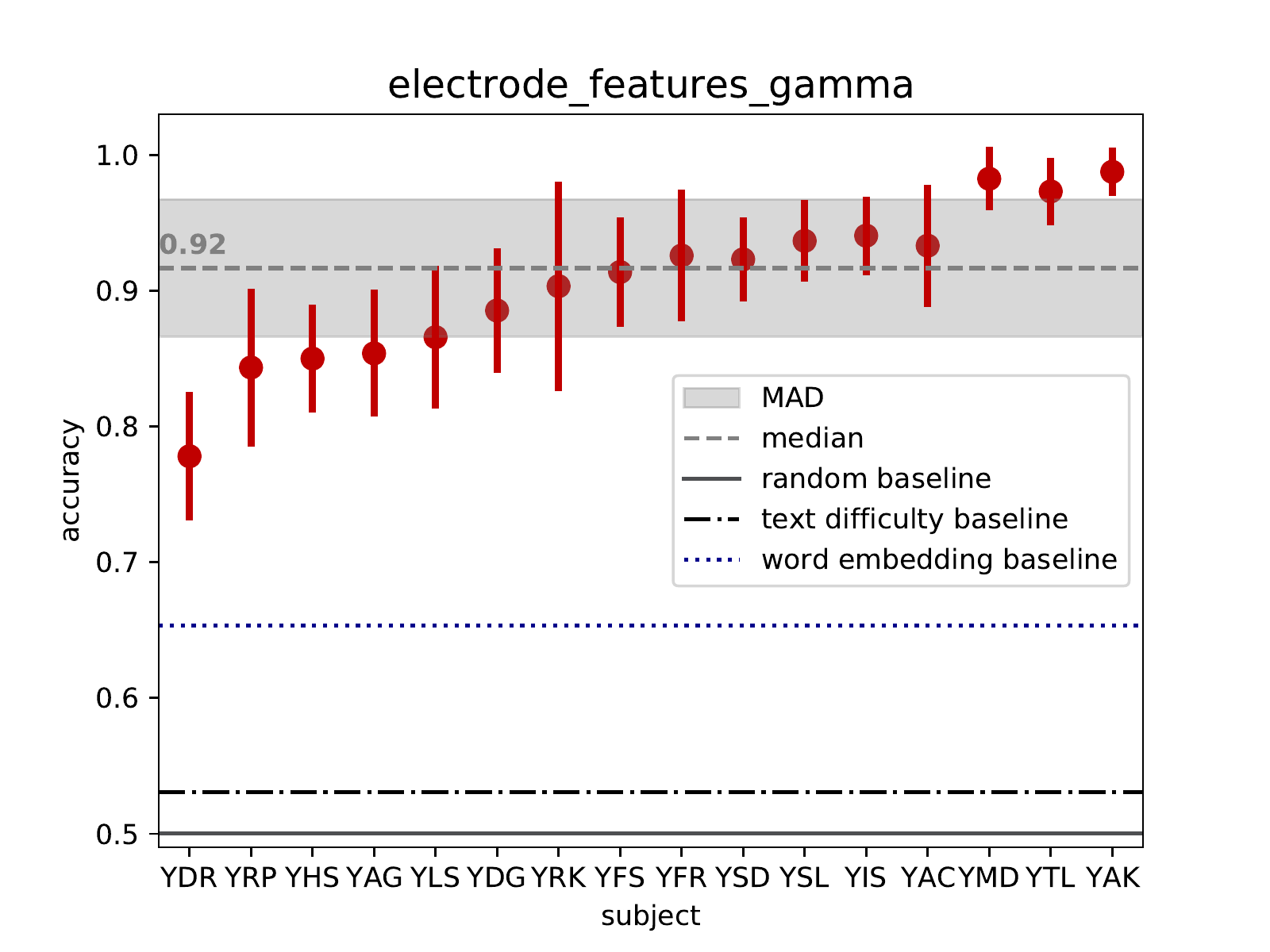} 
    \includegraphics[width=0.32\textwidth]{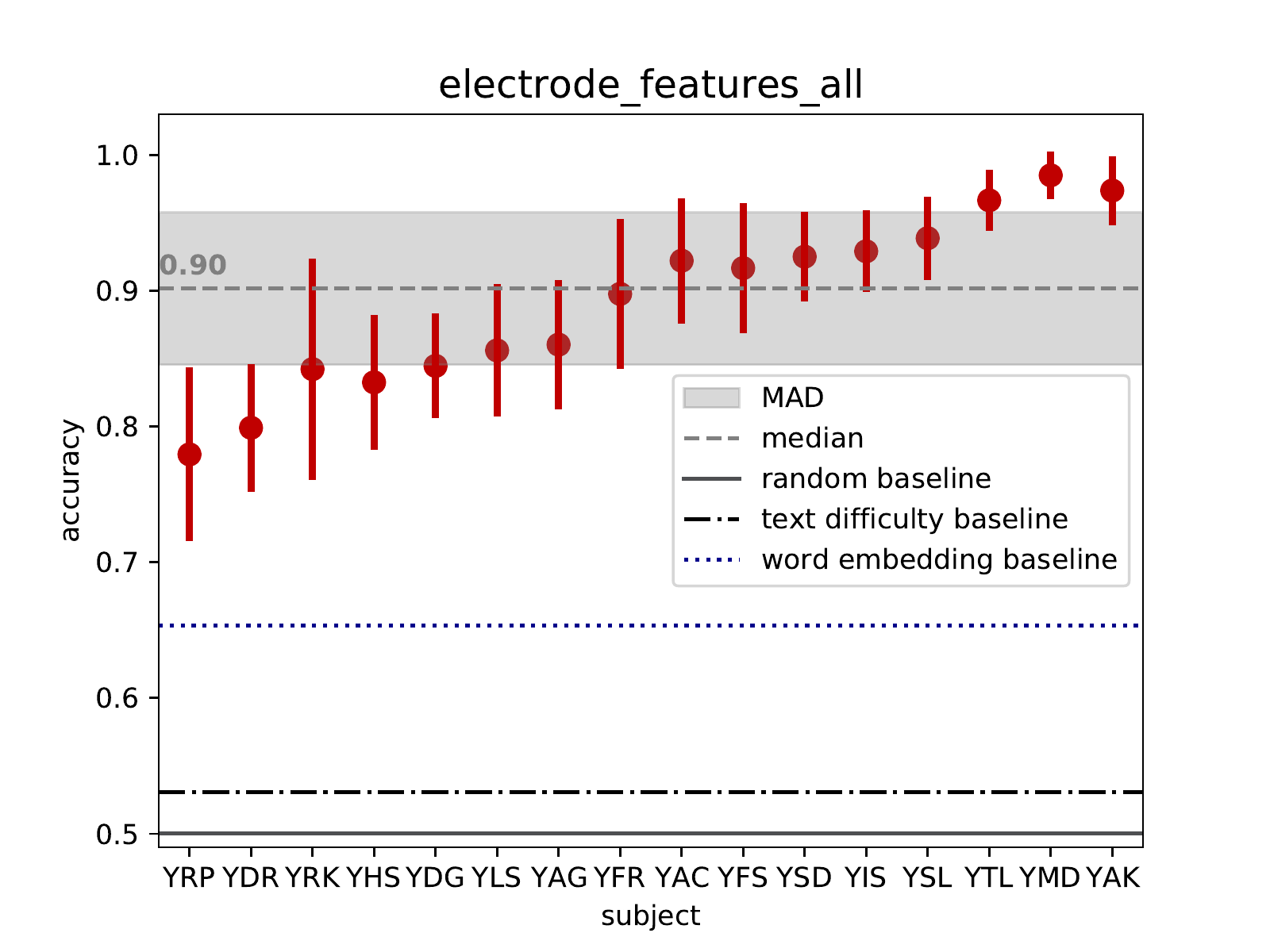} 
    \caption{EEG sentence-level classification accuracy on the ZuCo 2.0 data.}
    \label{fig:sent-res-z2-eeg}
\end{figure}

The results for the EEG feature sets are presented in Figure \ref{fig:sent-res-z1-eeg} for ZuCo 1.0 and in Figure \ref{fig:sent-res-z2-eeg} for ZuCo 2.0. With the EEG mean features, all frequency bands perform similarly (ZuCo 1.0: lowest - alpha = 62\%, highest - gamma = 67\%; ZuCo 2.0: lowest - theta = 56\%, highest - beta = 58\%). Combining all four mean EEG features yields improved results (\textit{eeg\_means}: ZuCo 1.0 = 79\%, ZuCo 2.0 = 62\%). Additionally, the combination of EEG and ET features shows further improvements (\textit{sent\_gaze\_eeg\_means}: ZuCo 1.0 = 88\%, ZuCo 2.0 = 72\%).
However, the most accurate results are achieved using the electrode features. For ZuCo 1.0, most subjects reach 99\% accuracy. The gamma frequency band features yield the best performing model. For ZuCo 2, the results of using the theta, alpha, and beta frequency bands range from 70-76\% accuracy, whereas using gamma frequency band results in 92\% accuracy. This is in line with the quantitative data analysis by \citet{mathur2021dynamic}, who find that the EEG frequency ranges are relatively
passive during natural reading, although sudden spikes can be observed in task-specific reading.

In general, we observe substantially higher performance on models trained on the ZuCo 1.0 data, i.e., 10\% higher accuracy than ZuCo 2.0 for both eye tracking and EEG. Moreover, the gamma frequency band information greatly improves the performance. This is in line with previous research: Adding high frequency band features improves mental task classification \citep{zhang2010improving}.
Finally, the sentence-level EEG features yield significantly higher accuracy for ZuCo 2.0 than word-level models. This might be due to the fact that word-level features result in a much high number of parameters than the sentence-level features, which can be detrimental in relation to the low number of training samples available.

\clearpage

\subsubsection{Leave-one-out Cross-subject Evaluation}
    
\noindent To analyze the generalization capabilities of the models, we additionally performed the previously described cross-subject experiments. 

\begin{figure}[ht]
    \centering
    \includegraphics[width=0.49\textwidth]{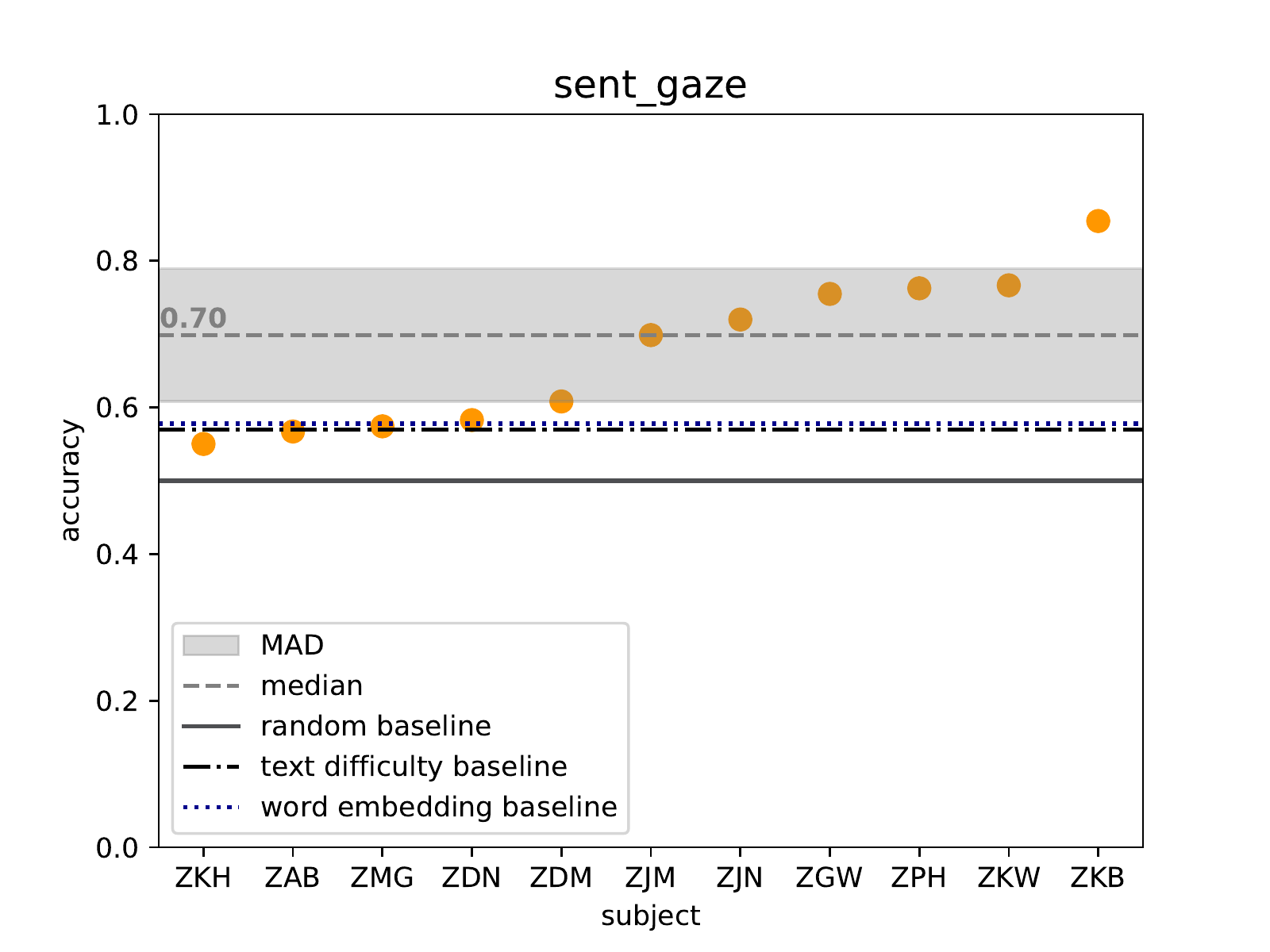} 
    \includegraphics[width=0.49\textwidth]{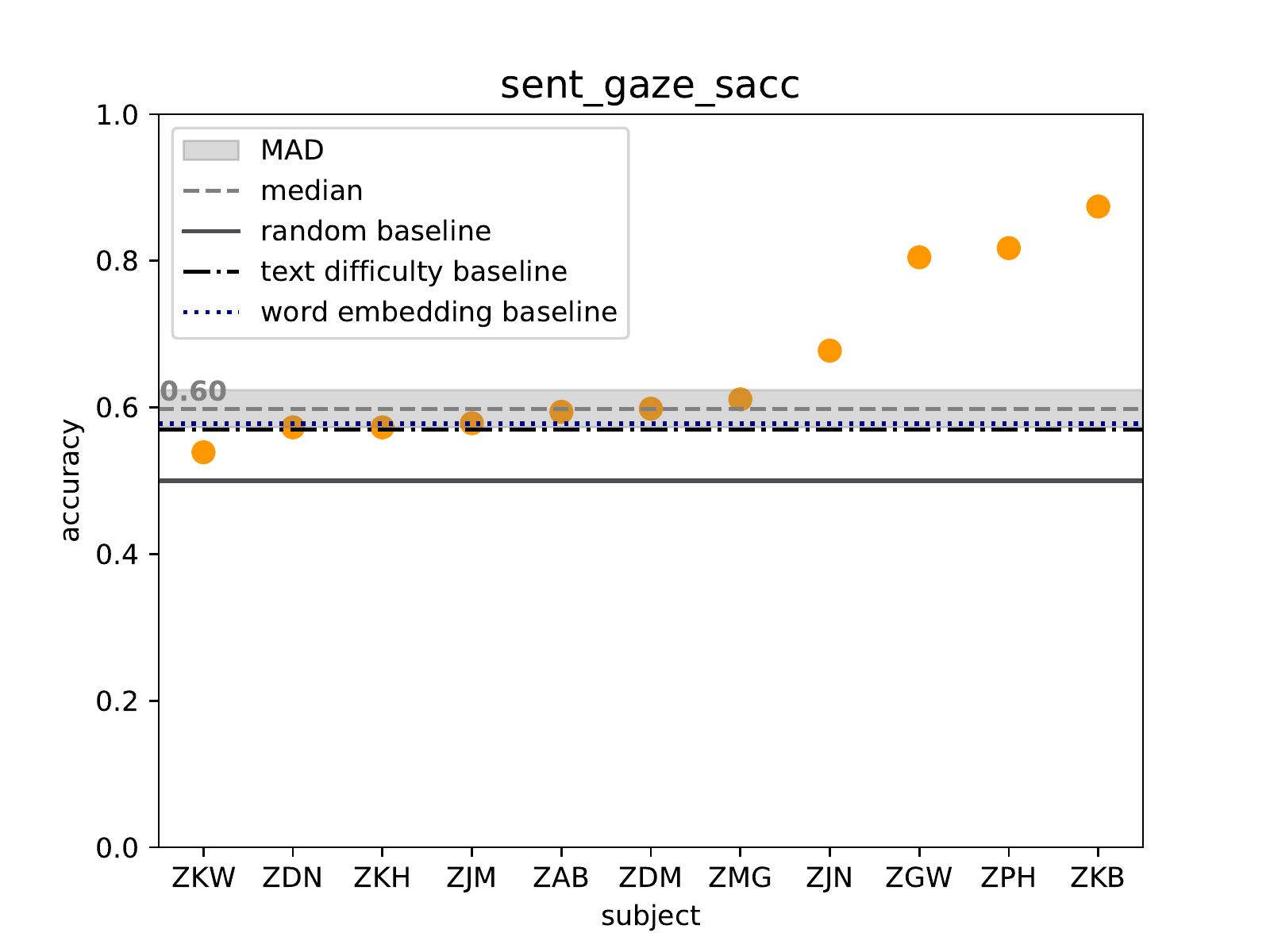} 
    \includegraphics[width=0.49\textwidth]{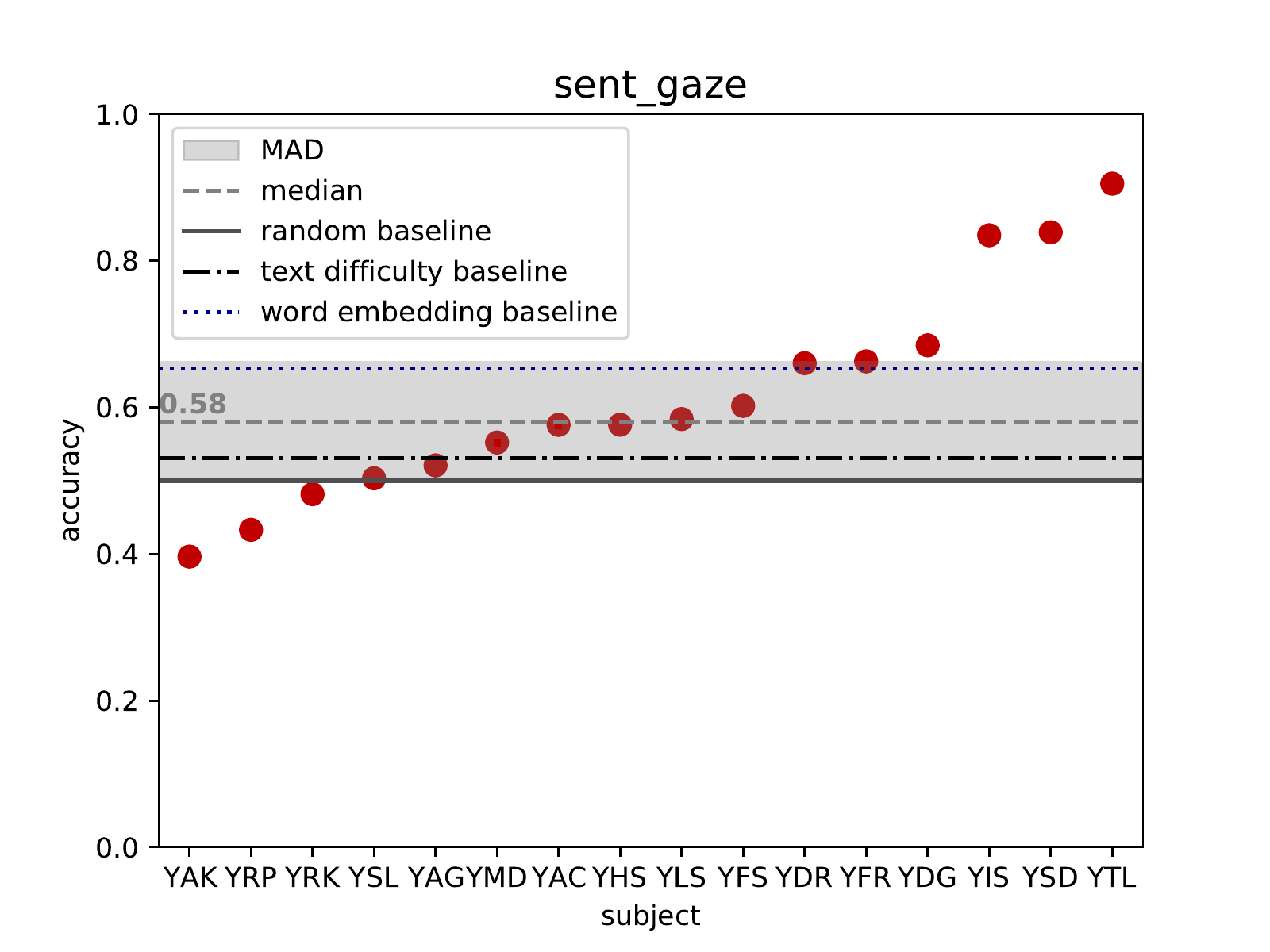} 
    \includegraphics[width=0.49\textwidth]{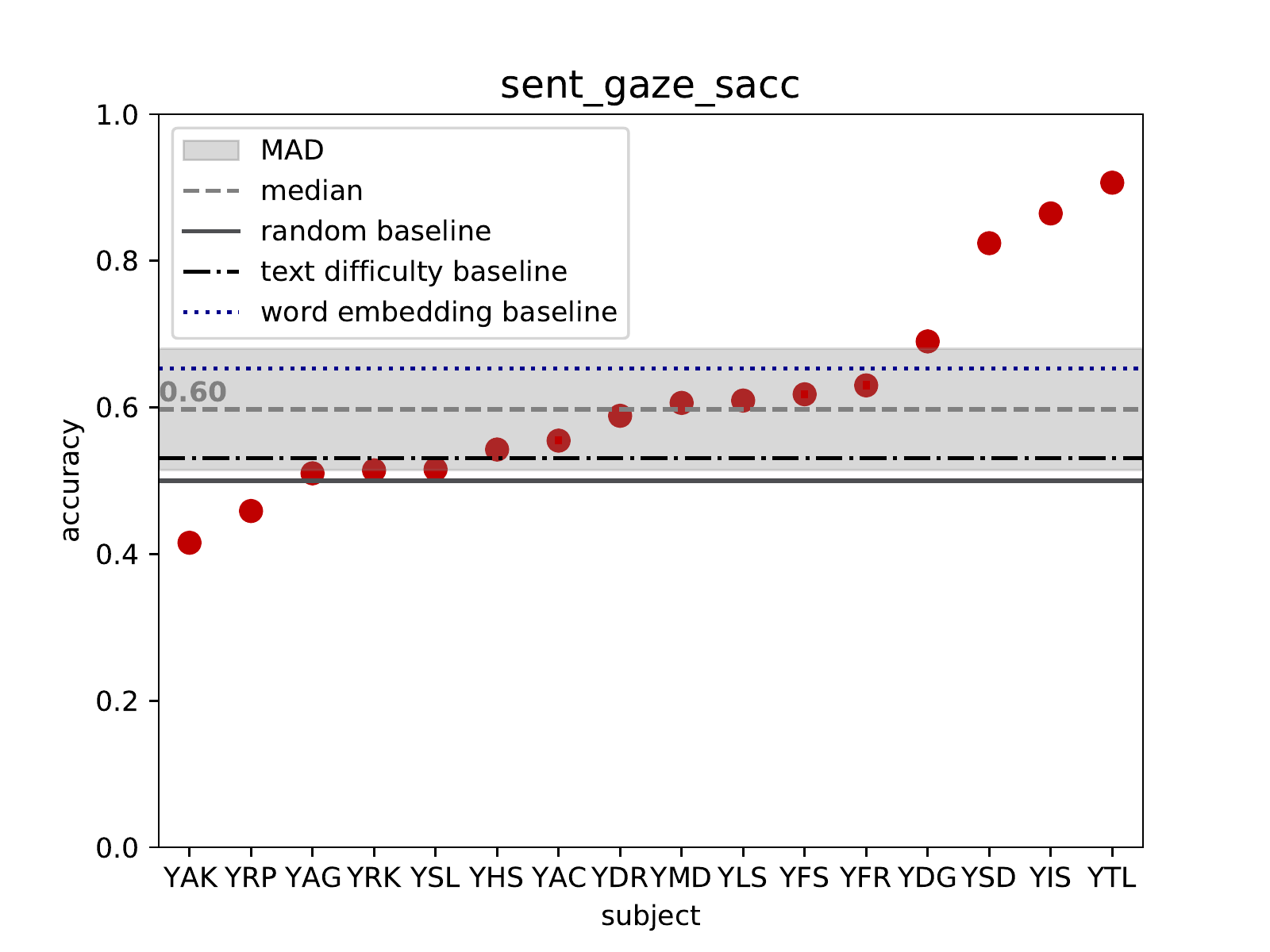} 
    \caption{Cross-subject eye-tracking sentence-level classification accuracy on ZuCo 1.0 and ZuCo 2.0, without saccade feature (left) and with saccade features (right).}
    \label{fig:sent-res-et-cross}
\end{figure}

\begin{figure}[ht]
    \centering
    \includegraphics[width=0.49\textwidth]{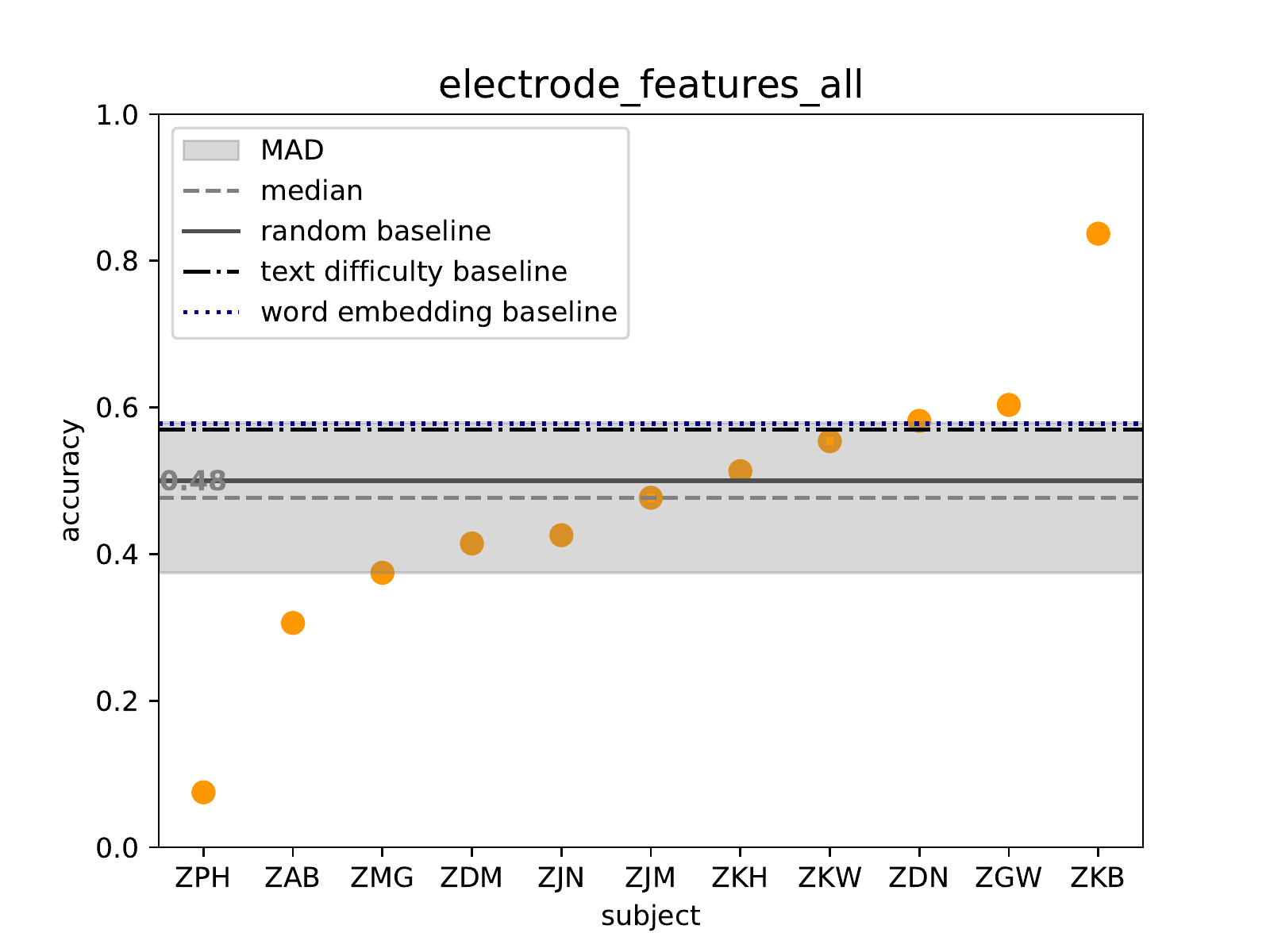} 
    \includegraphics[width=0.49\textwidth]{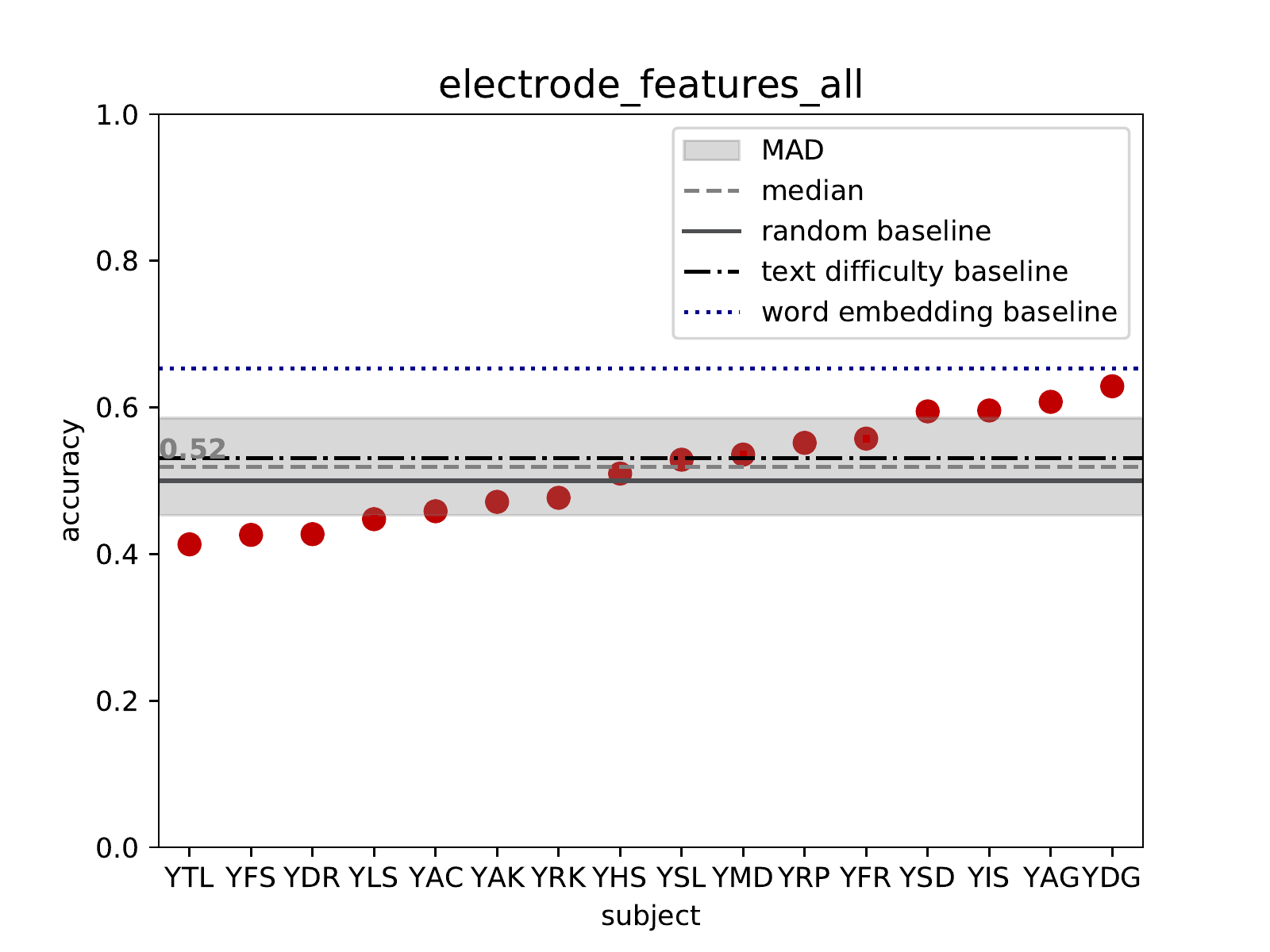} 
    \caption{Cross-subject EEG sentence-level classification accuracy on ZuCo 1.0 and ZuCo 2.0 with all electrode features.}
    \label{fig:sent-res-eeg-all-cross}
\end{figure}

\begin{figure}[ht]
    \centering
    \includegraphics[width=0.49\textwidth]{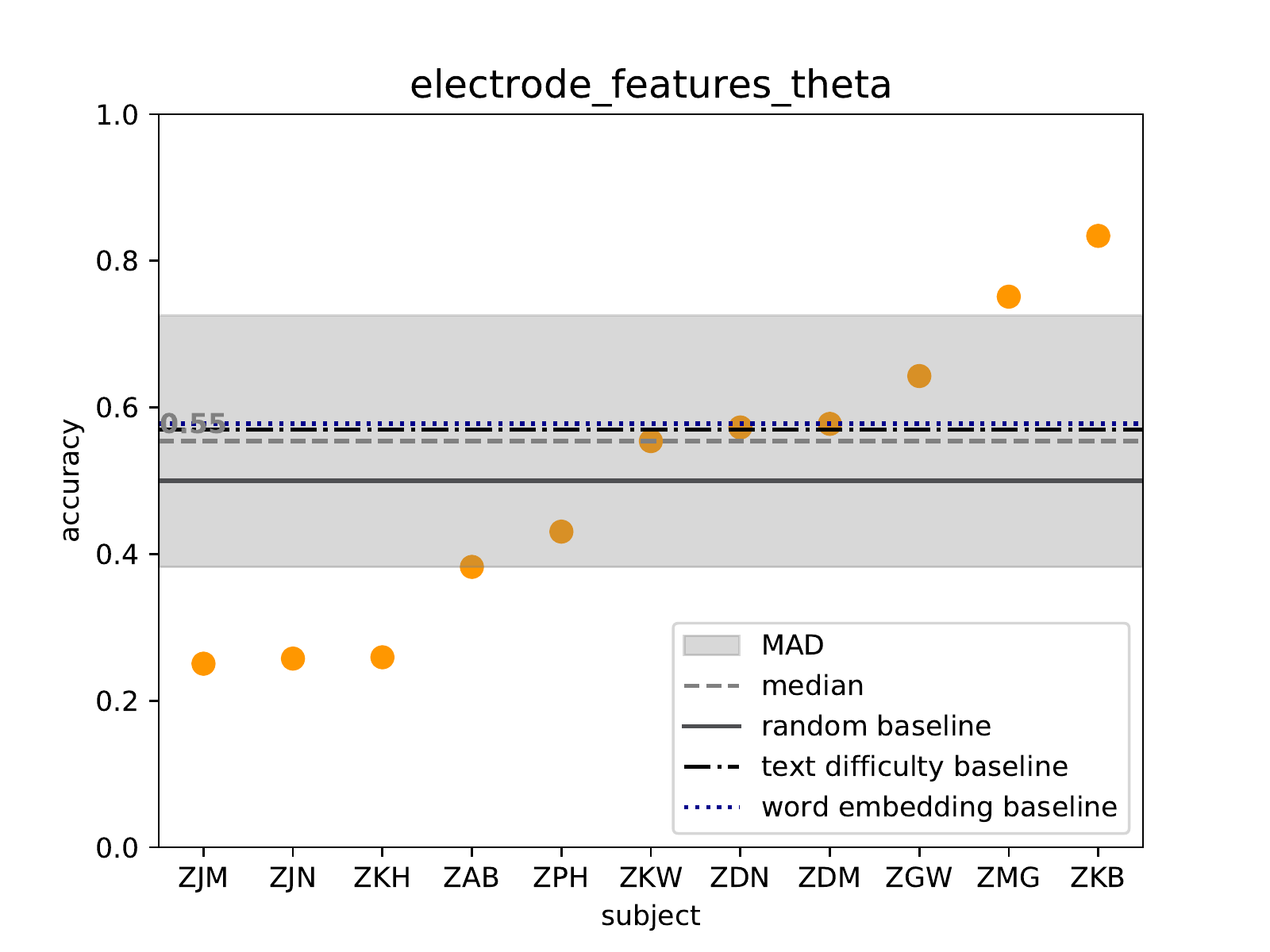} 
    \includegraphics[width=0.49\textwidth]{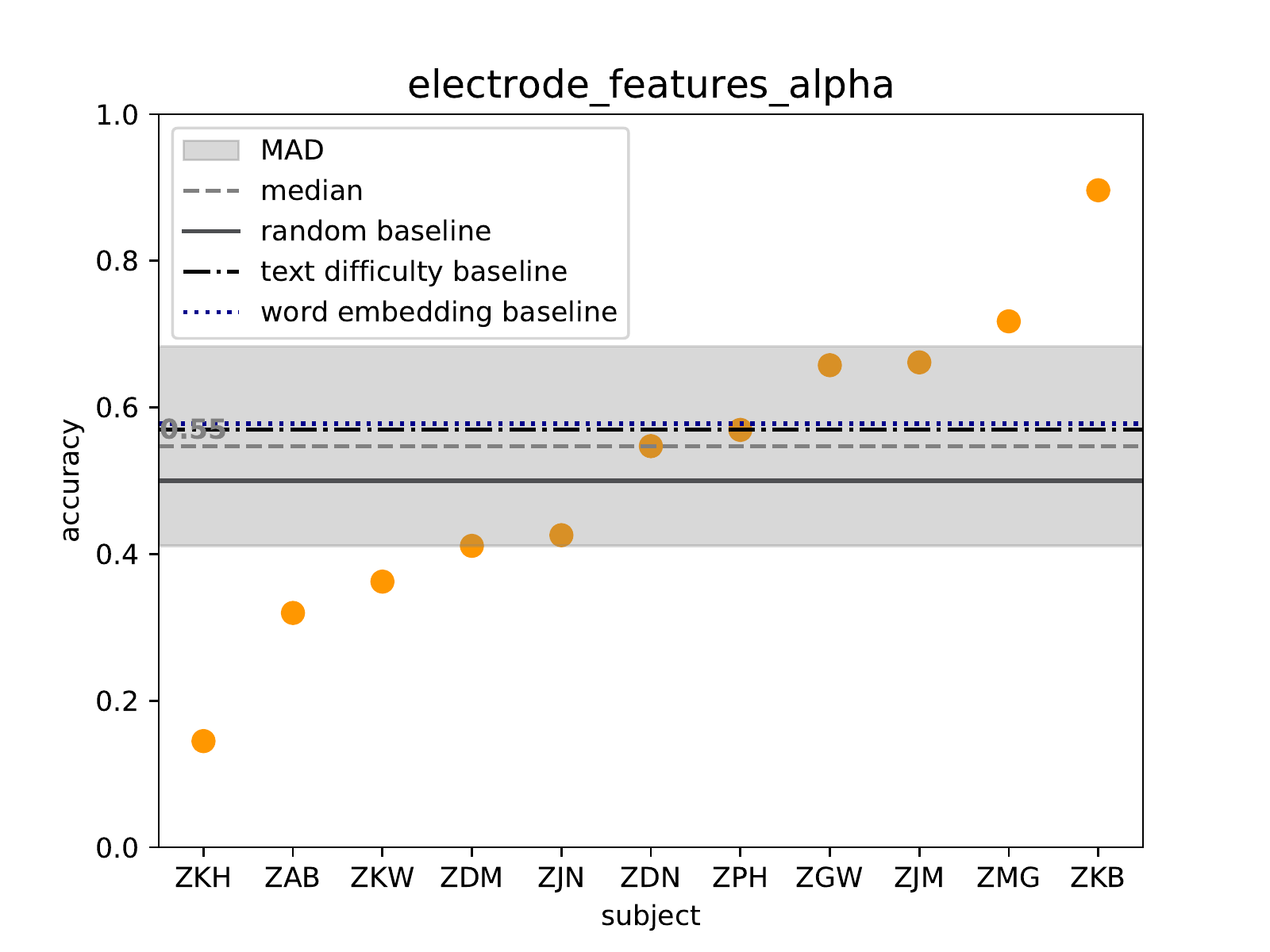} 
    \includegraphics[width=0.49\textwidth]{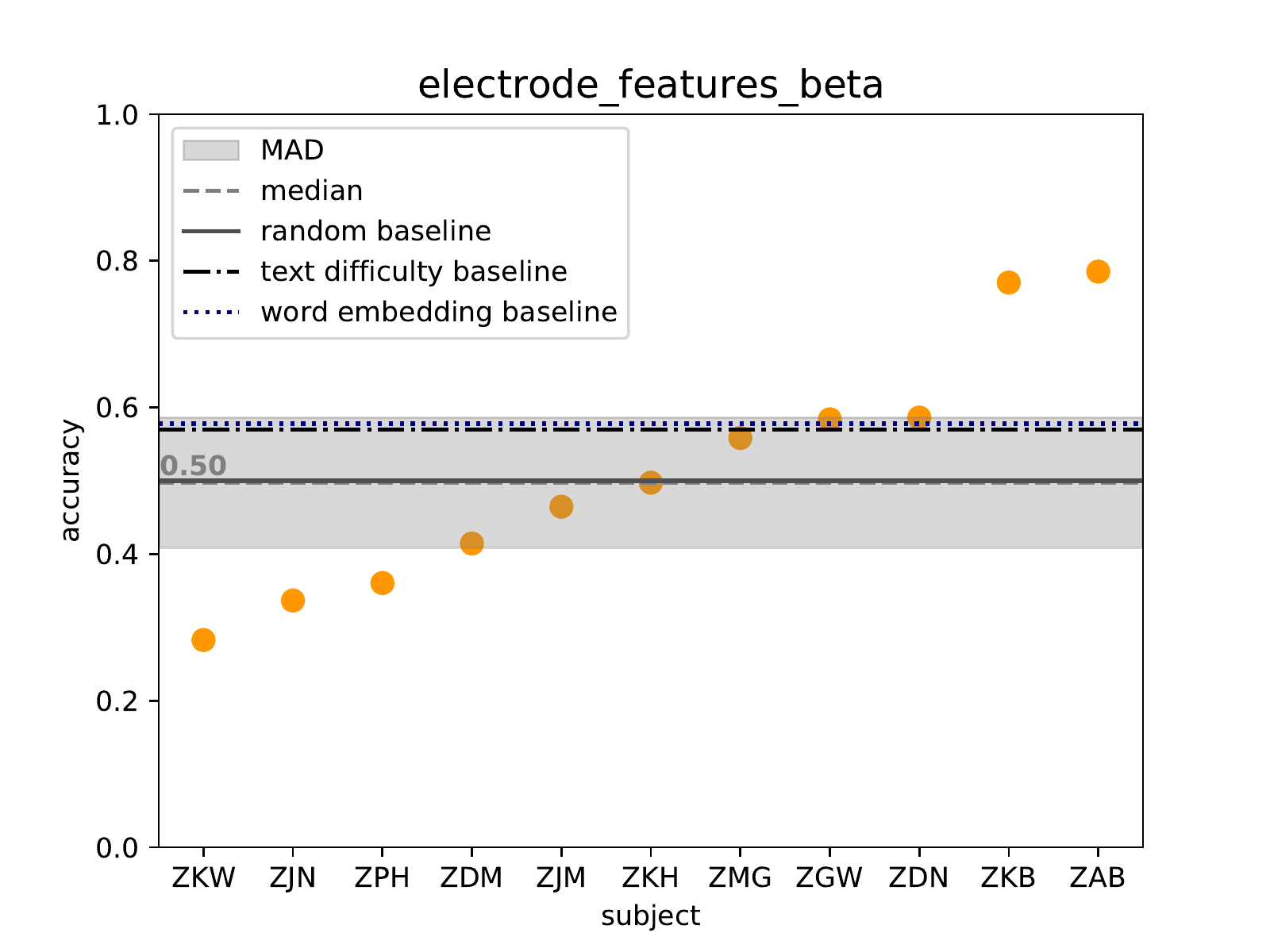} 
    \includegraphics[width=0.49\textwidth]{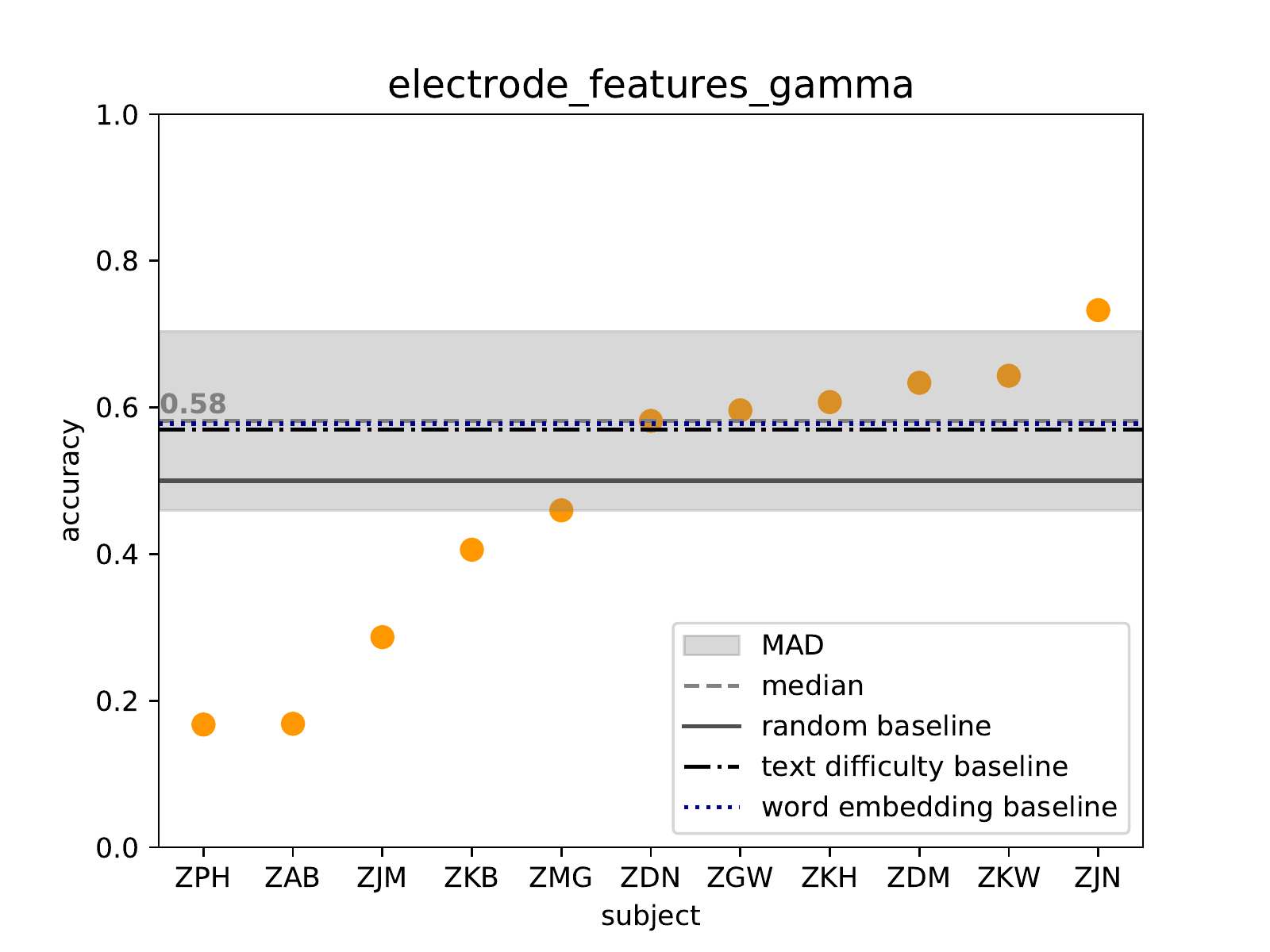} 
    \caption{Cross-subject EEG sentence-level classification accuracy on ZuCo 1.0.}
    \label{fig:sent-res-eeg-bands-cross-z1}
\end{figure}

\begin{figure}[ht]
    \centering
    \includegraphics[width=0.49\textwidth]{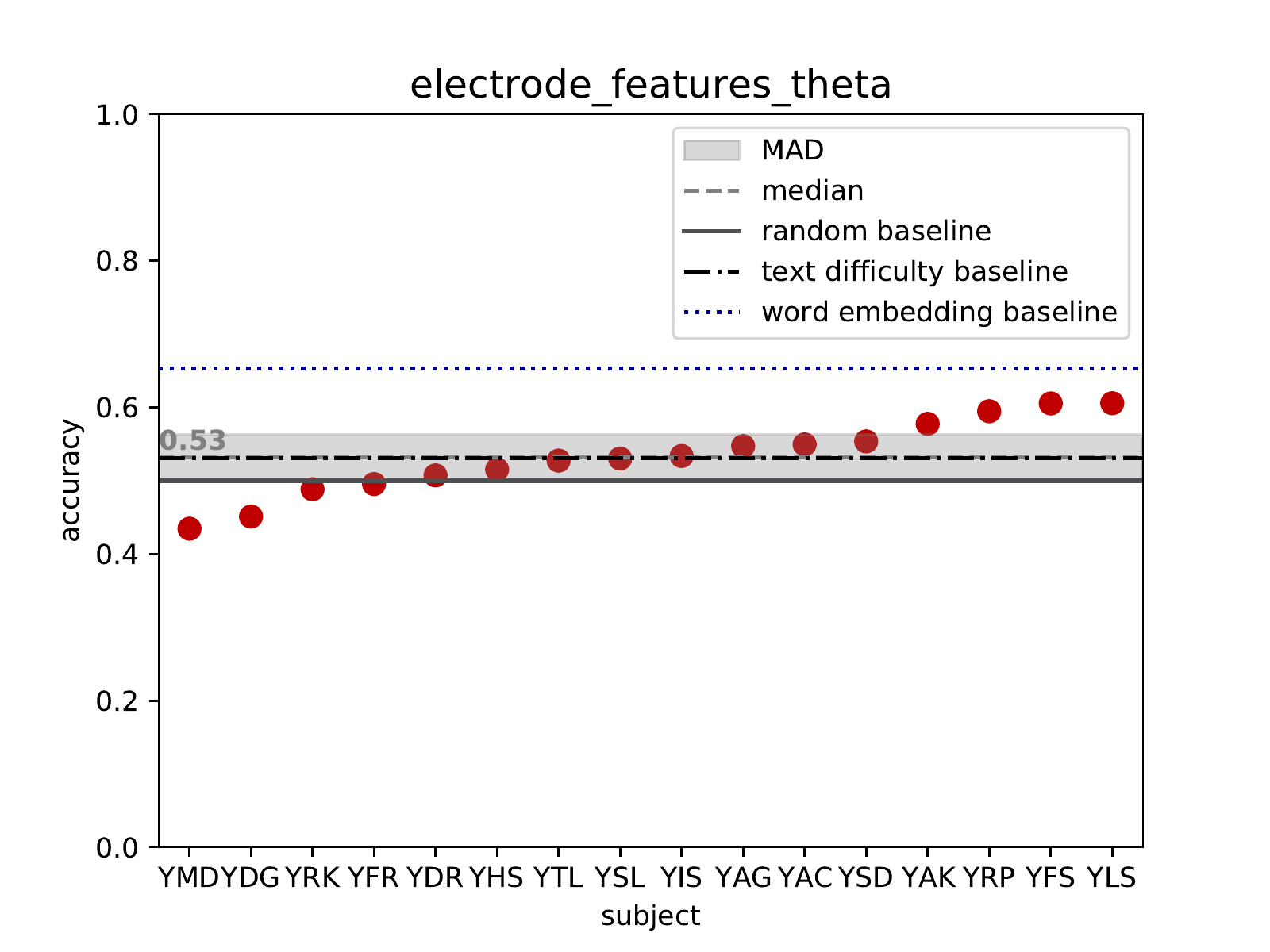} 
    \includegraphics[width=0.49\textwidth]{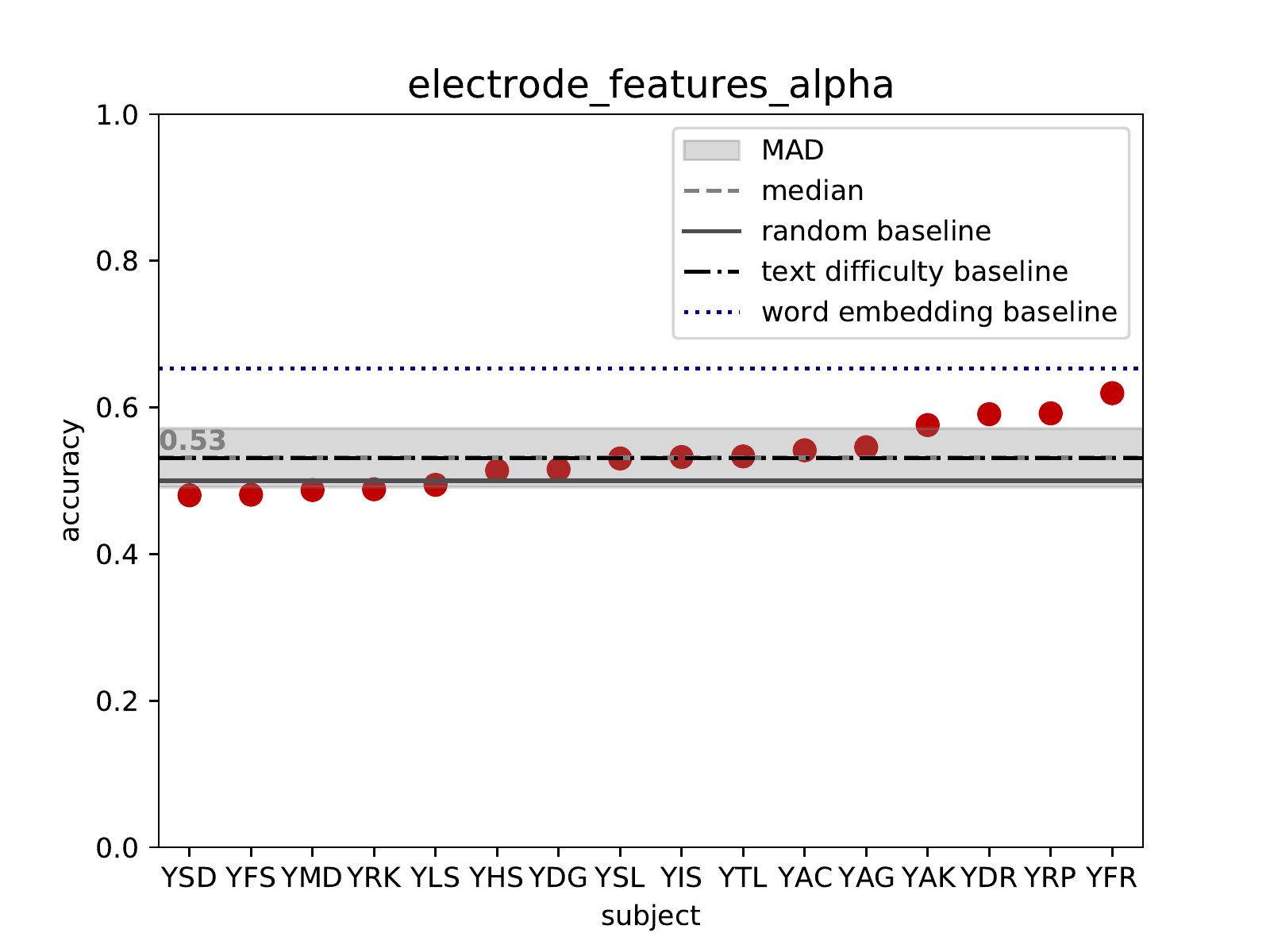} 
    \includegraphics[width=0.49\textwidth]{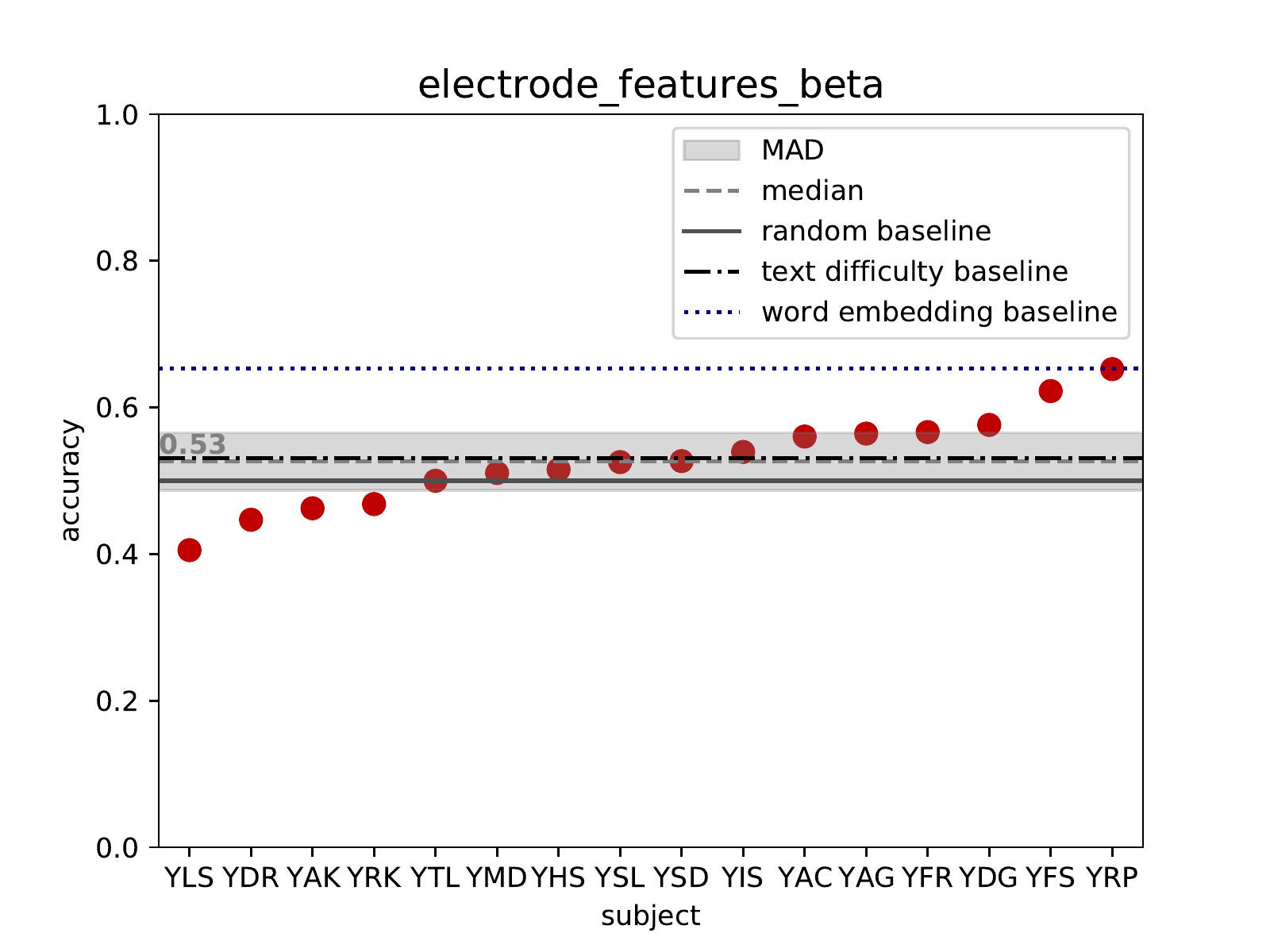} 
    \includegraphics[width=0.49\textwidth]{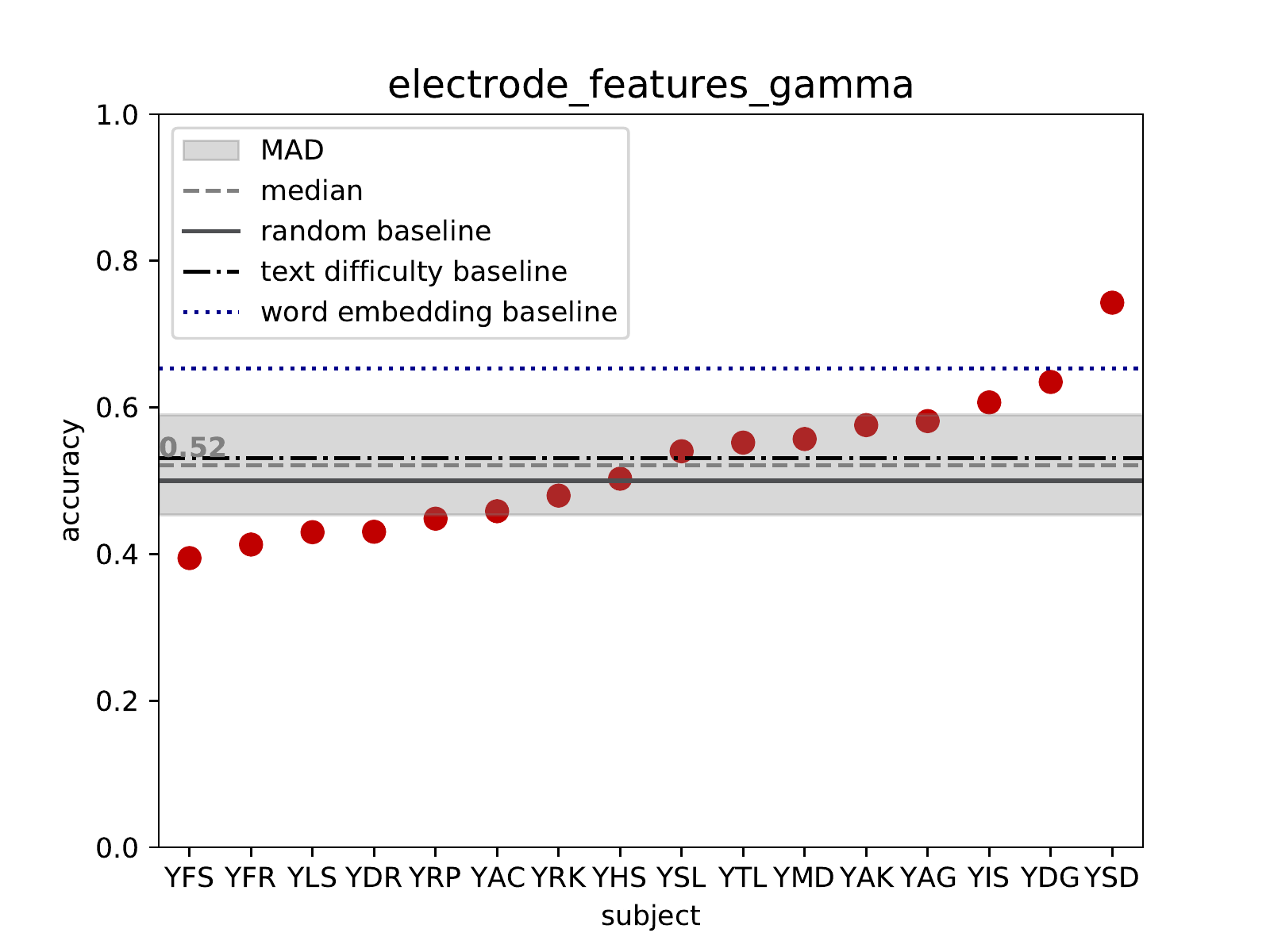} 
    \caption{Cross-subject EEG sentence-level classification accuracy on ZuCo 2.0.}
    \label{fig:sent-res-eeg-bands-cross-z2}
\end{figure}

The results are presented in Figures \ref{fig:sent-res-et-cross} - \ref{fig:sent-res-eeg-bands-cross-z2}. For both ZuCo 1.0 and ZuCo 2.0, when using the fixation-based eye-tracking features (\textit{sent\_gaze}), for some subjects we observe accuracies significantly above the text baselines (ZuCo 1.0 median 70\%; ZuCo 2.0 median 58\%). When including saccade features (\textit{sent\_gaze\_sacc}), the variability between subjects is reduced, but the accuracy does not improve. Moreover, including saccade features results in similar performance for both ZuCo 1.0 and ZuCo 2.0.

\begin{table}[t]
\centering
\begin{tabular}{lcccc}
\toprule
 & \multicolumn{2}{c}{ZuCo 1.0} & \multicolumn{2}{c}{ZuCo 2.0} \\
\textbf{feature set} & \textbf{median} & \textbf{MAD} & \textbf{median} & \textbf{MAD} \\\midrule
sent\_gaze & 0.70 & 0.09 & 0.58 & 0.08 \\
sent\_gaze\_sacc & 0.60 & 0.03 & 0.60 & 0.08 \\\midrule
electrode\_features\_all & 0.48 & 0.10 & 0.52 & 0.07 \\
electrode\_features\_gamma & 0.58 & 0.12 & 0.52 & 0.07 \\
electrode\_features\_beta & 0.50 & 0.09 & 0.53 & 0.04 \\
electrode\_features\_theta & 0.55 & 0.17 & 0.53 & 0.03 \\
electrode\_features\_alpha & 0.55 & 0.14 & 0.53 & 0.04\\\bottomrule
\end{tabular}
\caption{Sentence-level cross-subject result summary.}
\label{tab:cross-subj-results}
\end{table}

However, using the EEG electrode features, the opposite is the case. ZuCo 1.0 shows high variance between the subjects, the mean accuracy across all models is below random (48\%) and only very few subjects achieve performance above the text baseline. The mean performance of the ZuCo 2.0 models is slightly above random (52\%). Additionally, we note that in the cross-subject settings, the sentence-level models again perform better than the word-level models.

%We perform a final cross-subject experiment where we train on subjects from both ZuCo 1.0 and ZuCo 2.0.

%\begin{figure}[ht]
 %   \centering
  %  \includegraphics[width=0.89\textwidth]{figures/sent-results/cross-subj-zucoAll.png} 
  %  \caption{Cross-subject eye-tracking sentence-level classification accuracy on the combined subbjects of ZuCo 1.0 \textit{and} ZuCo 2.0, ... .}
 %   \label{fig:sent-res-et-cross}
%\end{figure}

\clearpage
\section{Control Analyses}\label{sec:control}

\noindent In this section, we perform various secondary analyses to validate our results. Since sentence-level models yield higher performance with fewer parameters and lower computational costs, we focus on these results.

\subsection{Outlier Analysis}
\noindent First, we detect outlier subjects based on the standard deviation of the feature values and investigate whether the worst- and best-performing subjects are the same as these outliers. We define an outlier as a subject whose feature values deviate from the sample mean by two standard deviations (sample mean$\pm$2 std). Table \ref{tab:outliers-z1-z2} shows the mean feature values before scaling, standard deviation and the resulting outlier subjects for ZuCo 1.0 and ZuCo 2.0. For the EEG mean features, only ZKH from ZuCo 1 and YDG, YSD, YFS from ZuCo 2 are marked as outliers for certain features.
For the eye-tracking features, only one subject (ZGW) from ZuCo 1.0 and three subjects (YAG, YRK, YRP) from Zuco 2.0 were marked as outliers. There is no apparent correlation between the model performance and outlier subjects on the within-subject results: On the main EEG feature sets (\textit{electrode\_features\_all} and \textit{electrode\_features\_gamma}), the results of all outlier subjects fall within the median absolute deviation (MAD) across all subjects. On the main eye-tracking feature set (\textit{sent\_gaze\_sacc} including all fixation and saccade features), only the subject YRP from ZuCo 2.0 falls slightly below the MAD.

\begin{table}[ht]
\centering
\small
\begin{tabular}{l|lll|lll}
\toprule
& \multicolumn{3}{c}{\textbf{ZuCo 1.0}} & \multicolumn{3}{c}{\textbf{ZuCo 2.0}} \\
\textbf{Feature}              & Mean    & StD    & Outliers  & Mean    & StD    & Outliers\\\midrule
\textsc{Eye-tracking}                  &         &        &          \\\midrule
fixation\_number     & 1,038   & 0,23   & -   & 0,995    & 0,29   & YAG      \\
omission\_rate       & 0,409   & 0,10   & -    & 0,393    & 0,11   & -     \\
reading\_speed       & 2,341   & 0,73   & -    & 2,240    & 0,63   & -     \\
max\_sacc\_dur       & 57,740  & 18,74  & ZGW    & 76,834   & 32,84  & YAG,YRK  \\
max\_sacc\_velocity  & 576,884 & 104,64 & -  & 1023,973 & 446,90 & YRP      \\
mean\_sacc\_dur      & 18,512  & 3,03   & ZGW    & 22,608   & 5,50   & YRK      \\
mean\_sacc\_velocity & 250,012 & 26,28  & -  & 264,913  & 70,07  & YRP      \\\midrule
\textsc{EEG}                  &         &        &          \\\midrule
theta\_mean          & 2,786   & 0,63   & -    & 2,775    & 0,59   & YDG      \\
alpha\_mean          & 2,585   & 1,16   & ZKH   & 2,475    & 0,98   & YSD      \\
beta\_mean           & 2,707   & 1,19   & ZKH   & 2,325    & 0,64   & -     \\
gamma\_mean          & 1,785   & 0,39   & -    & 1,735    & 0,50   & YFS     \\\bottomrule
\end{tabular}
\caption{Mean feature values before scaling, standard deviation, and the resulting outlier subjects for ZuCo 1.0 (left) and ZuCo 2.0 (right) (sample mean $\pm$2 std).}
\label{tab:outliers-z1-z2}
\end{table}

\subsection{Correlation Analysis} 
\noindent Second, we explore if the model performance depends on a subject's task performance, level of English language proficiency, or reading speed.
Therefore, we investigate whether the performance of the models trained on data from individual subjects correlates with their answer scores in the comprehension questions, their performance in the LexTALE vocabulary test, or their reading speed (see Tables \ref{tab:subjects-stats-z1} and \ref{tab:subjects-stats-z2}). 

Table \ref{tab:correlations-zuco1-zuco2} shows the Spearman correlation coefficients for both datasets. We find no significant correlations for the models trained on EEG electrode features. We do see significant negative correlations between the reading speed of the task-specific sentences and the performance of the models trained on eye-tracking features. This is in line with the preliminary data analysis (Section \ref{sec:prel-data-analysis}), showing that participants who fixate on fewer words during task-specific reading have made a clearer mental distinction between the two reading tasks. These participants exhibit a different reading strategy and therefore have a higher skipping rate when specifically searching for information in a sentence.

\begin{table}[ht]
\centering
\small
\begin{tabular}{l|cc|cc}
\toprule   
 & \multicolumn{2}{c}{\textbf{within-subject}} & \multicolumn{2}{c}{\textbf{cross-subject}} \\
\textbf{ZuCo 1.0} & \textit{sent\_gaze\_sacc} &  \textit{EEG elec.} & \textit{sent\_gaze\_sacc} & \textit{EEG elec.} \\\midrule
Score TSR & 0.04  & 0.20 & 0.15 & -0.29 \\
Score NR & 0.33 & 0.05  & -0.03 & -0.13 \\
Speed TSR & -0.18  & 0.30 & \textbf{-0.42} & 0.02 \\
Speed NR & \textbf{0.58} & \textbf{0.50} & 0.11 & 0.27 \\
LexTALE & 0.22  & 0.00 & 0.40 & \textbf{-0.32} \\\midrule
\textbf{ZuCo 2.0} & \textit{sent\_gaze\_sacc} & \textit{EEG elec.} & \textit{sent\_gaze\_sacc} & \textit{EEG elec.} \\\midrule
Score TSR & 0.54 &  -0.12 & 0.26 & 0.01 \\
Score NR & 0.14 & -0.05  & 0.34 & 0.33 \\
Speed TSR & \textbf{-0.75}* & 0.07  & \textbf{-0.61}* & 0.23 \\
Speed NR & -0.16  & 0.19  & -0.04 & 0.23 \\
LexTALE & 0.27 & \textbf{0.22} &  0.15 & \textbf{0.43} \\\bottomrule
\end{tabular}
\caption{Spearman correlation between the within-subject and cross-subject classification accuracies of all subjects in ZuCo 1.0 (top) and ZuCo 2.0 (bottom) for a given feature set and the control scores or reading speed. The feature sets are ET = sent\_gaze\_sacc, EEG mean = eeg\_means, and EEG elec. = electrode\_features\_all. * marks significant correlations $p<0.05$.}
\label{tab:correlations-zuco1-zuco2}
\end{table}

\subsection{Superiority of Gamma Frequency Band Features}

\begin{figure}[t]
    \centering
    \includegraphics[width=0.9\textwidth]{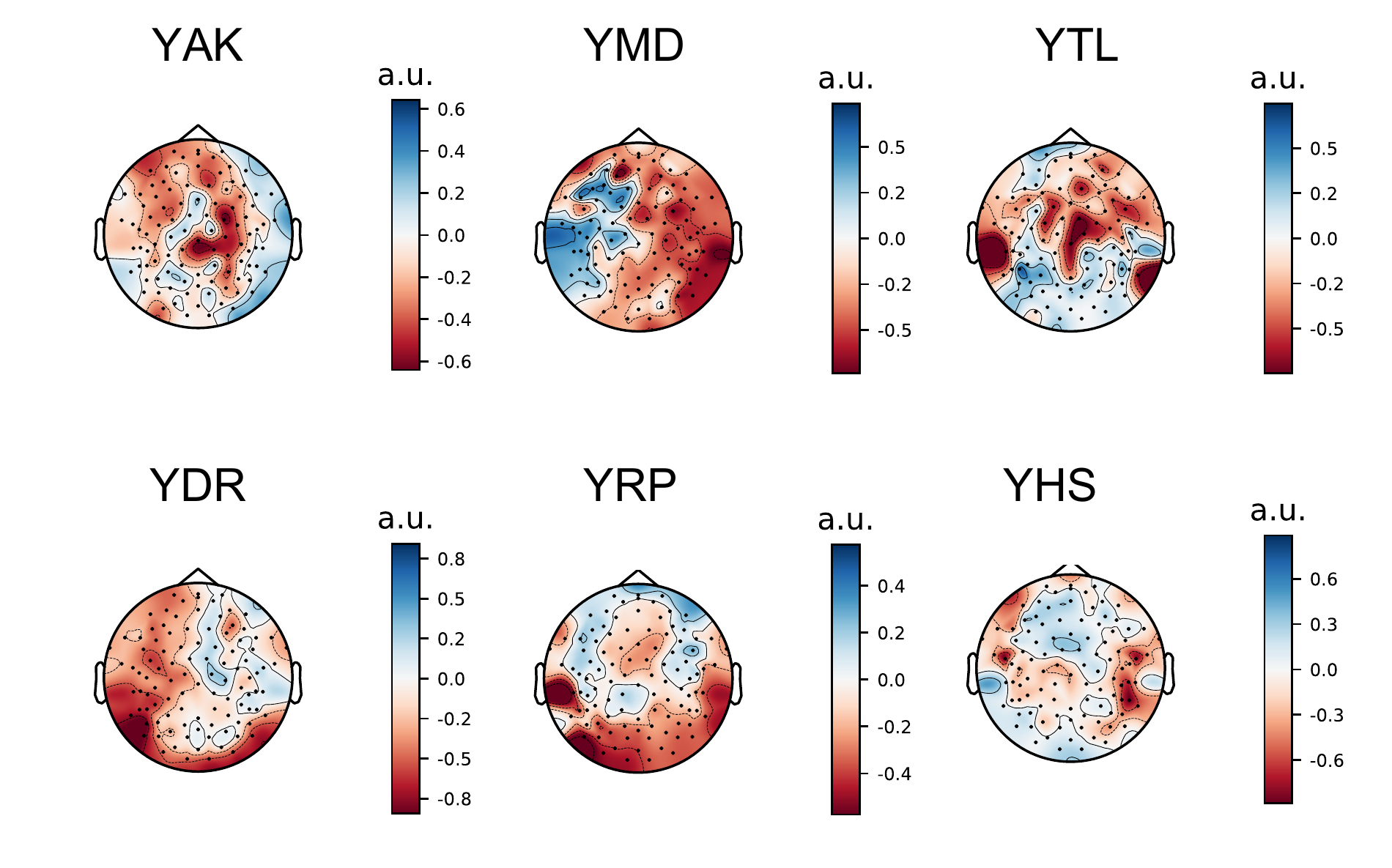}
    \caption{Topography plots of the SVM coefficients for the feature electrode\_features\_gamma for the three best (top) and three worst (bottom) performing subjects from ZuCo 2.0.}
    \label{fig:sent-res-z2-eeg-svm-coefficients-gamma-best-worst}
\end{figure}

\noindent Lastly, to analyze the surprisingly high performance of the gamma frequency band features in all settings, Figure \ref{fig:sent-res-z2-eeg-svm-coefficients-gamma-best-worst} shows topography plots of the SVM coefficients in the within-subject  sentence-level reading task classification for the feature electrode\_features\_gamma for the three best and worst performing subjects of the within-subject models from ZuCo 2.0. To obtain an interpretable topography analysis, the plots were generated by transforming the model into a linear forward model according to the implementation by \citet{haufe2014interpretation}.\footnote{\url{https://mne.tools/stable/auto_examples/decoding/linear_model_patterns.html}}
The topography plots show the gamma distribution for the three best performing subjects exhibiting positive SVM coefficients in a mid frontal electrode cluster, whereas for the worst performing subjects the positive SVM coefficients are pronounced in the occipital electrodes.

\subsection{Subject Classification} 
\begin{figure}[ht]
    \centering
    \includegraphics[width=0.45\textwidth]{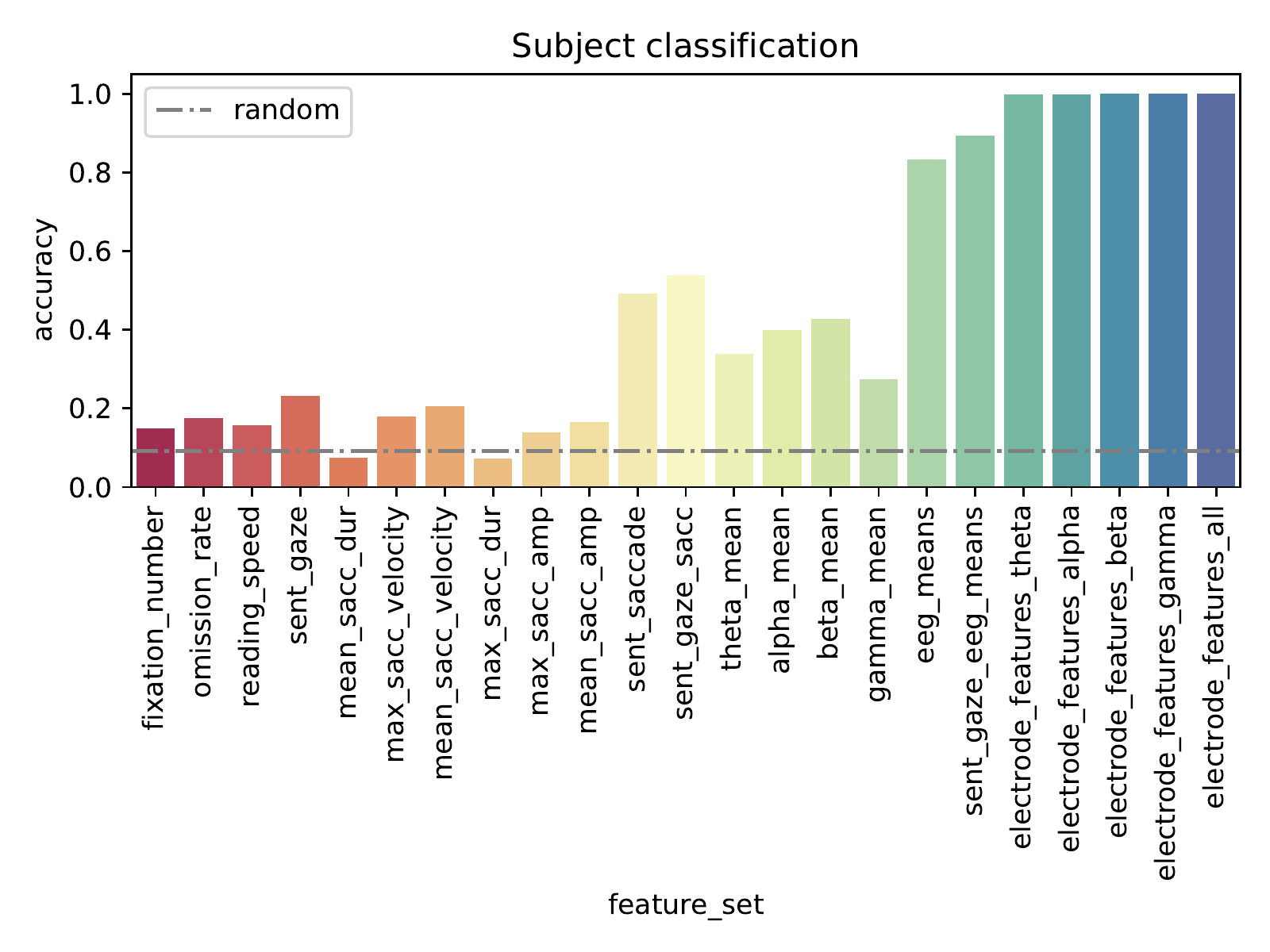} 
    \includegraphics[width=0.45\textwidth]{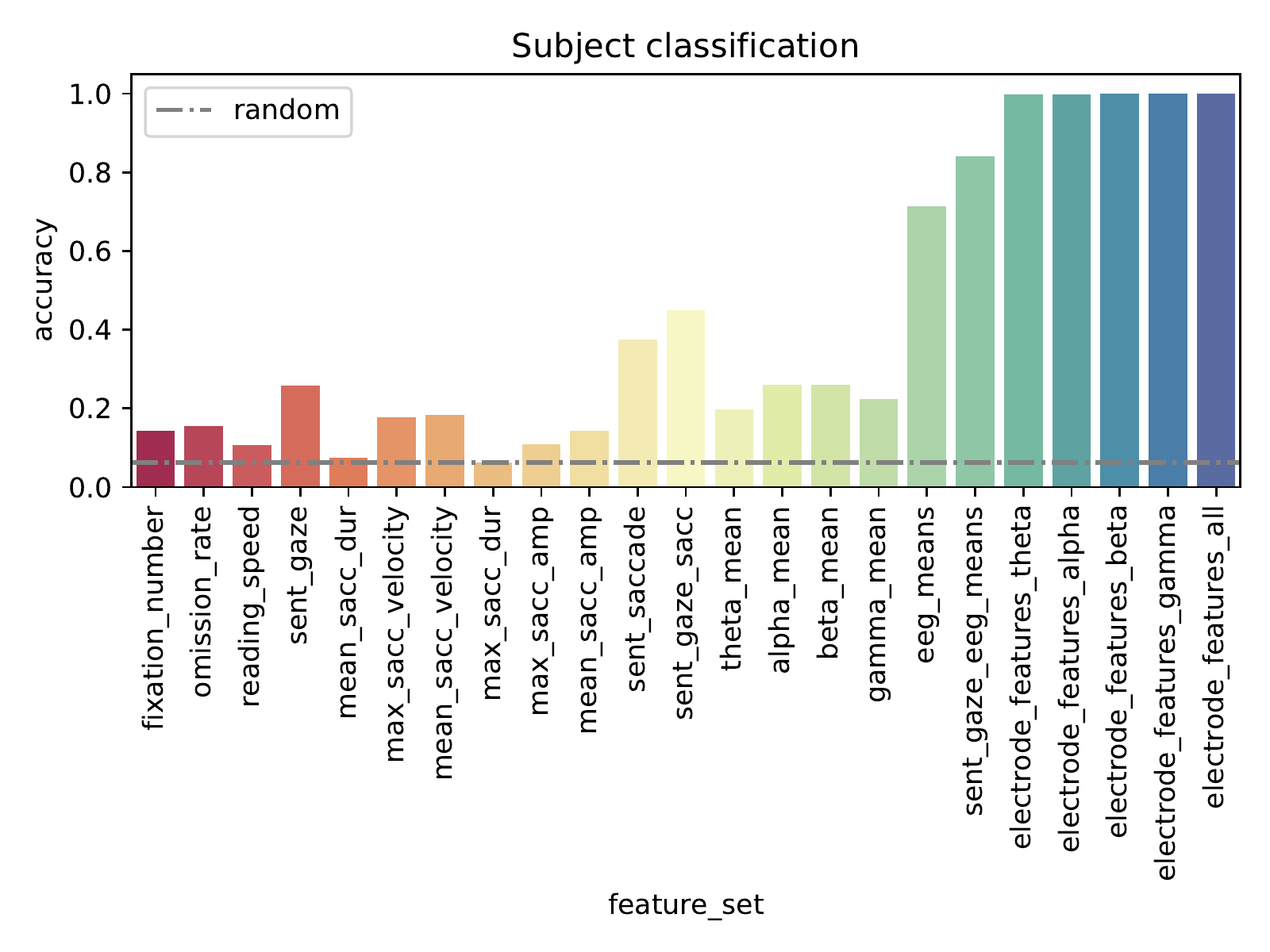} 
    \caption{Subject classification for ZuCo 1.0 (left) and ZuCo 2.0 (right). The results are averaged across all subjects.}
    \label{fig:subj-classification}
\end{figure}

\noindent As a control experiment, we train models to recognize the individual subjects instead of the reading tasks. As shown in Figure \ref{fig:subj-classification}, the EEG electrode features again yield 
the highest accuracy for this subject classification for both ZuCo 1.0 and ZuCo 2.0, while the eye-tracking and EEG mean feature sets do not yield such good performance. However, most of the feature sets do reach an accuracy higher than random, defined as the probability of randomly classifying the correct subject in a balanced dataset (i.e, $1/11$ for ZuCo 1.0 and $1/16$ for ZuCo 2.0. This shows how the EEG electrode features are much more subject-specific than the eye tracking features. This is possibly be due to the loss of information in the eye-tracking features as well as the EEG mean features as they are aggregated on a higher level. This subject classification task opens opportunities for further research exploring privacy preservation in task-specific EEG and eye-tracking signals.

\subsection{Fixation Ablation}

%- Hypothesis: Only a percentage of EEG signal is necessary to accurately predict the reading task (not true...).

\noindent In an additional experiment, we extract the EEG features in their chronological fixation order, instead of the word order within the sentences as in all other experiments in this paper. The input dimension of the feature vectors remains the same. However, instead of averaging the electrode value across all fixations, we average only on the first 10\%, 20\%, 50\% or 75\% of fixations. For this control experiment, we use the gamma frequency band features, since these yielded the best results.

We test how the performance of the reading task classification is affected when taking only a proportion of the fixated words. We hypothesize that only a small percentage of the EEG signals is necessary to accurately predict the reading task, speculating that the mental state is different enough between the reading tasks from the beginning.

\begin{figure}[ht!]
    \centering
    \includegraphics[width=0.48\textwidth]{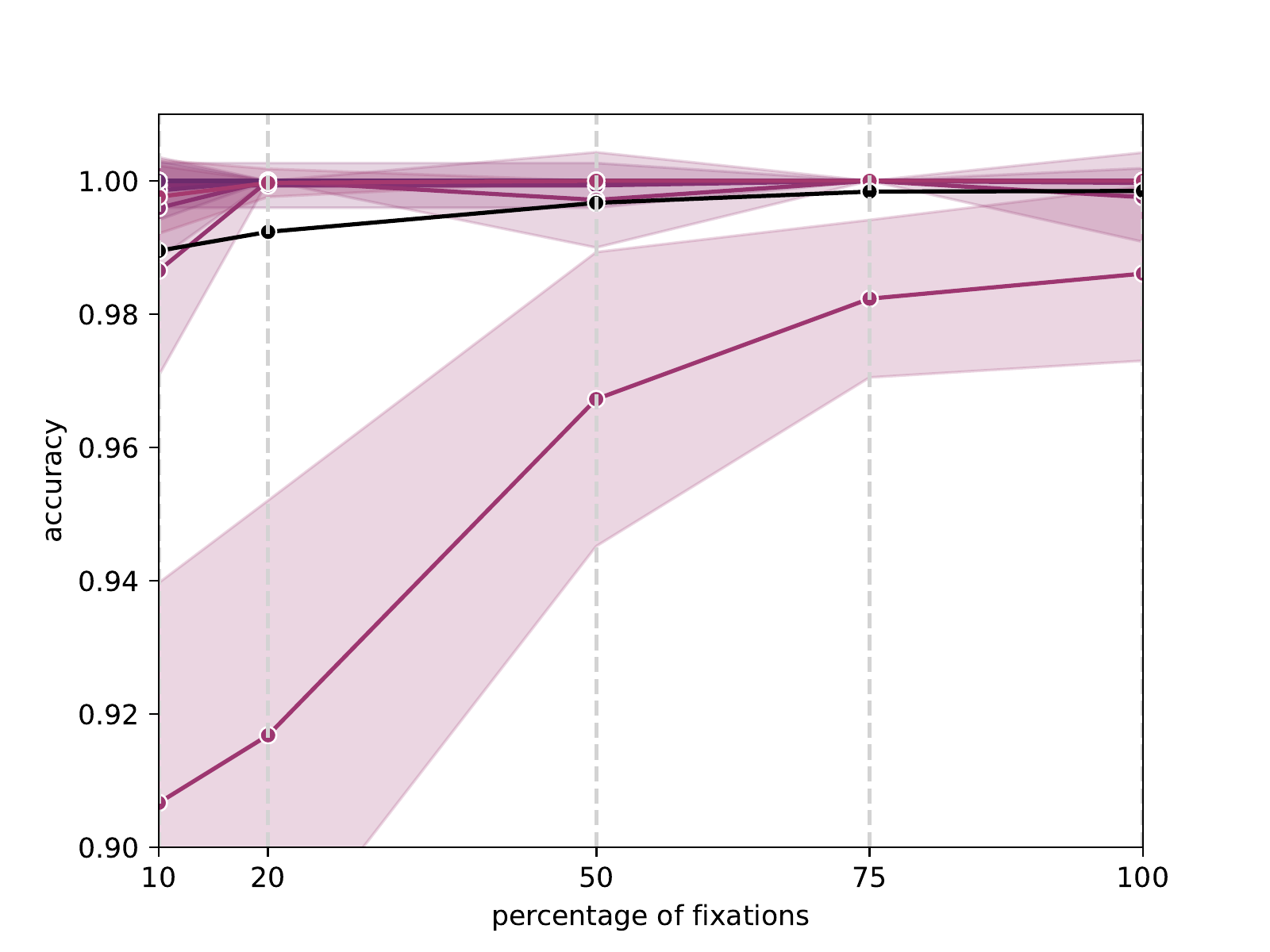} 
    \includegraphics[width=0.48\textwidth]{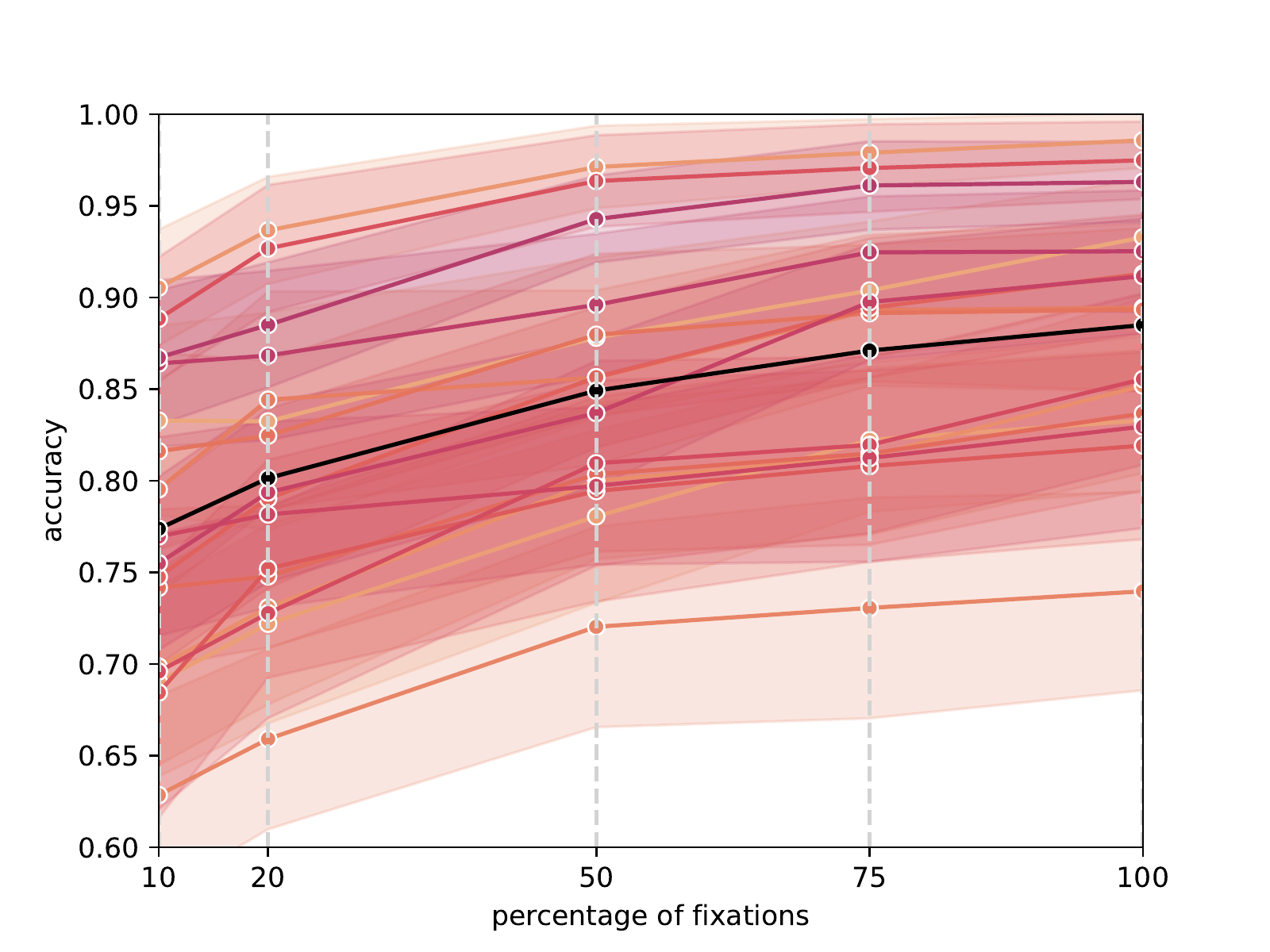} 
    \caption{Classification accuracy for each subject for the EEG electrode features of the gamma frequency band in fixation order on sentence level, with varying fixation percentage. ZuCo 1.0 on the left, ZuCo 2.0 on the right.}
    \label{fig:fix-sent-res-z1-z2-subjs}
\end{figure}

Figure \ref{fig:fix-sent-res-z1-z2-subjs} shows that this hypothesis holds true for the participants of ZuCo 1.0, where only one subject shows a substantial decrease in classification accuracy with a lower percentage of fixations. However, for most subjects even 10\% of the EEG signal is enough for an accurate classification (\textgreater 99\% accuracy). This might be due to the fact that the NR and TSR reading tasks were recorded in separate sessions in ZuCo 1.0. For ZuCo 2.0, the results show that models trained only on the first 10-50\% of EEG signals do yield lower accuracy. The highest results are achieved when training on the EEG signals of all fixations in a sentence. These results show the impact of the session bias in ZuCo 1.0: The fact that the reading tasks were recorded in different sessions presents as a systematic differentiation in the brain activity signals. Therefore, the EEG features only from the beginning fixations of every sentence are sufficient for accurate classification. In contrast, for ZuCo 2.0, recorded in a single session, the accuracy decreases with a lower proportion of the signals.

\subsection{Session \& Block Classification}

\subsubsection*{ZuCo 1.0: Session Classification}
\noindent  Auto-correlations between the trials (or blocks) of an experimental session due to a range of different factors such as  familiarization with the task or mental fatigue is a well-known challenge in experimental psychology  \citep{Baayen2017}. Moreover, it has been demonstrated that a subject's eye movements show session-specific properties \citep{Bargary2017}.  % TODO add sentence about trial and session effects in ET and EEG.
%Session effects in ET:  Bagary2017
%Trial effects in ET: 
%Session effects in EEG: 
%Trial effects in EEG
%Baayen2010,Masson2013: trial effects in lexical decision RTs
%---> I suggest to only cite Baayen2017 for trial effects in general. 
We therefore perform further analyses in an attempt to explore and quantify the session and block effects that emerge from the experiment design.
For ZuCo 1.0, we can test the session effect by adding the sentences recorded in a third task, the sentiment reading data (see description in Appendix \ref{app:sentiment-reading}). As described above, for ZuCo 1.0, the normal reading and task-specific reading paradigms were recorded in separate sessions. In each of the two sessions, half of the sentiment reading (SR) data was also recorded. Hence, by adding these sentences and training a model to classify in which session each sentence was recorded, we can get an estimate of the session-specific information contained in the EEG signals. By investigating this session bias, we explore whether the model exploits some uncontrolled systematic differences between the sessions to perform the task classification. By adding sentences of the same task recorded in both sessions, we hypothesize that the performance of this session classification task (including NR, TSR, and SR sentences) should be lower than for the reading task classification with only NR and TSR sentences, because the model cannot rely on differences between the recording sessions. This would show that session-specific information is partly a cause of the high accuracy for ZuCo 1.0. 

\begin{figure}[t!]
    \centering
    \begin{subfigure}[A]{0.49\textwidth}
    \includegraphics[width=\textwidth, trim = 0 0 0 300]{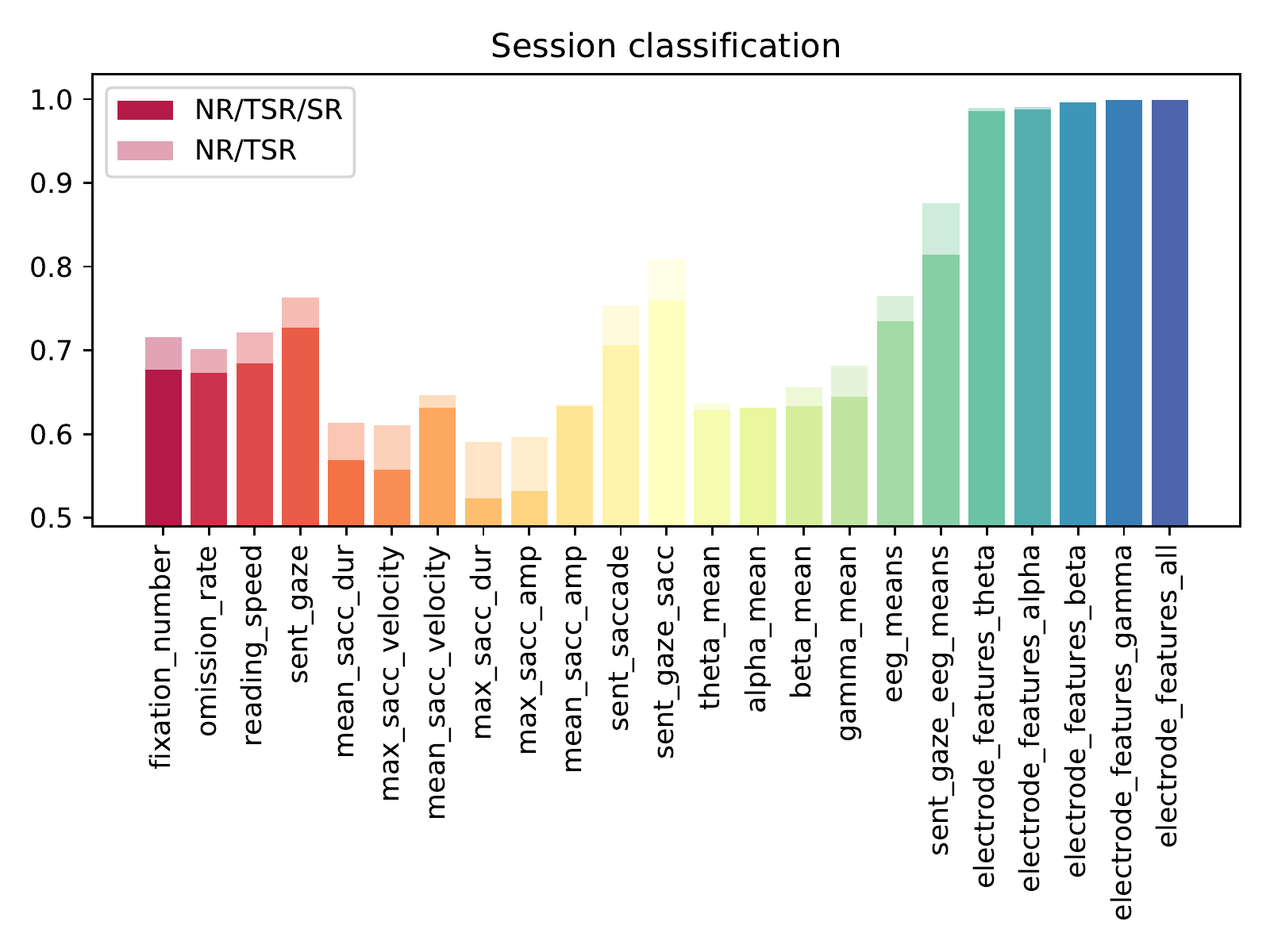} \caption{}
    \end{subfigure}
    \begin{subfigure}[B]{0.45\textwidth}
    \includegraphics[width=\textwidth]{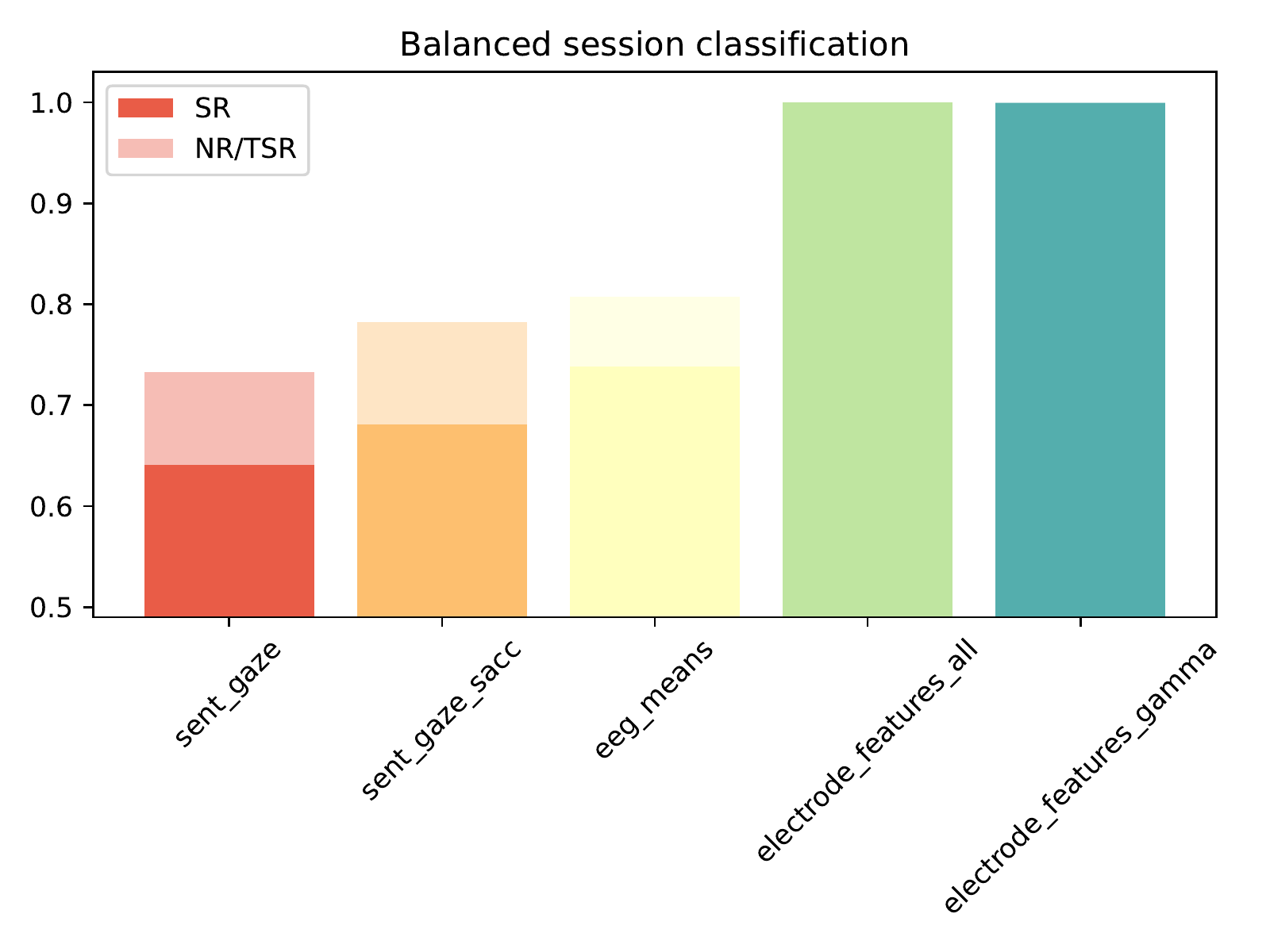} \caption{}
    \end{subfigure}
    \caption{Session classification. \textbf{(a)} Classifying the ZuCo 1.0 sentences into the two recording sessions, with and without the additional sentences. In the latter case, the labels are identical to all reading task classification experiments presented previously on ZuCo 1.0. \textbf{(b)} Session classification with balanced datasets. Classifying the ZuCo 1.0 sentences into the two recording sessions to quantify the session effect.}
    \label{fig:session-class-z1}
\end{figure}

Figure \ref{fig:session-class-z1} shows how adding the sentence from the additional paradigm recorded in both sessions of ZuCo 1.0, decreases the performance for most features, albeit not by much. Notably, the differences are larger for eye-tracking features and the accuracy is almost not affected for EEG electrode features. 

\begin{figure}[t!]
    \centering
    \includegraphics[width=0.7\textwidth]{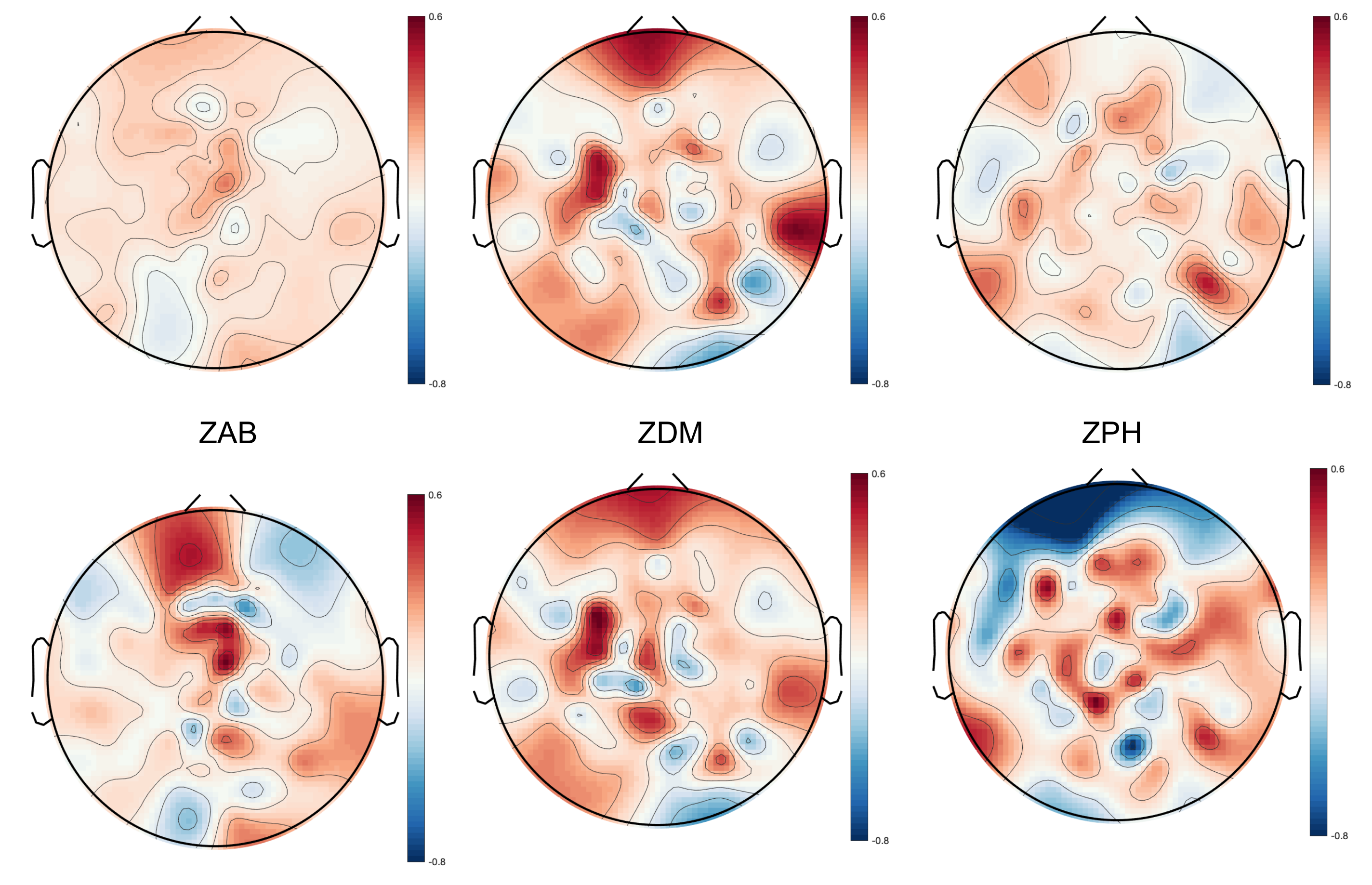}
    \caption{Topography plots of the SVM coefficients for all electrodes (gamma) - top for reading task classification and bottom for session classification including the additional sentences.}
    \label{fig:topo-session-class}
\end{figure}

Figure \ref{fig:topo-session-class} shows the topography plots of the SVM coefficients for the gamma activity of all electrodes for three randomly chosen subjects, the top row for reading task classification and the bottom for this experiment of session classification including the additional sentences. Again, the plots were generated according to the implementation by \citet{haufe2014interpretation}. Similar to the analysis based only on the ZuCo 2.0 data set, the highest SVM coefficients also seem to be in a mid frontal electrode cluster, although there is not a clear pattern shared across different subjects. 

Additionally, for ZuCo 1.0 we perform the session classification on completely balanced datasets, i.e., the same number of samples for the first and second session during training and testing the model. Figure \ref{fig:session-class} (b) shows the results for the most relevant eye tracking and EEG feature sets. We compare the performance of classifying the additional sentences of the sentiment reading task (SR) recorded in two sessions, to classifying the same number of sentences from NR and TSR. We see that when classifying these SR sentences, the performance drops between 5\% - 10\% in accuracy for eye-tracking (with and without saccadic features) as well as for the mean EEG features. However, for the EEG electrode features, the performance does not drop.

These additional session classification experiments show that the session effect does impact the reading task classification for the eye-tracking features and the mean EEG features, but it cannot entirely explain the high performance in the ZuCo 1.0 models, especially for the models with EEG electrode features, where the performance does not decrease at all. From these results, we can conclude that the session bias is more noticeable in aggregated features than in individual electrode features.

\begin{figure}[t!]
    \centering
    \includegraphics[width=0.7\textwidth]{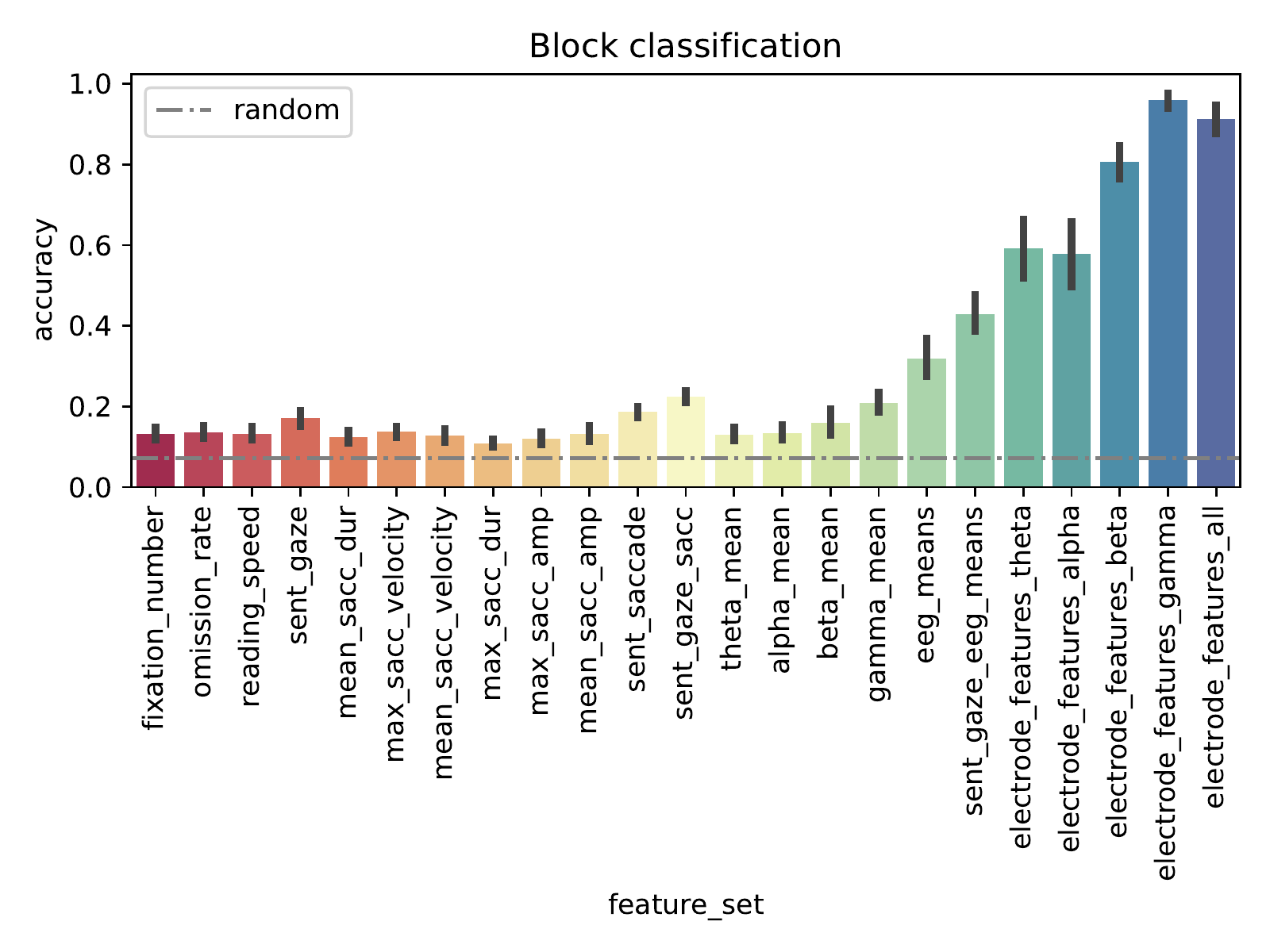} 
    \caption{Classifying the ZuCo 2.0 sentences into the 14 recording blocks.}
    \label{fig:block-class}
\end{figure}

\subsubsection*{ZuCo 2.0: Block Classification}
\noindent For ZuCo 2.0, all sentences were recorded in the same session. Therefore, there are no session-specific biases. However, we can analyze the effect of the order in which the sentences were recorded. As described in Section \ref{sec:data}, the sentence blocks of normal reading and task-specific sentences were alternated. Each of the 14 blocks contains approx. 50 sentences of either normal reading or task-specific reading. Hence, we also test the performance of classifying the sentences into these 14 blocks with within-subject sentence-level models. 

\begin{figure}[t!]
    \centering
    \includegraphics[width=0.32\textwidth, trim=100 0 20 0 ]{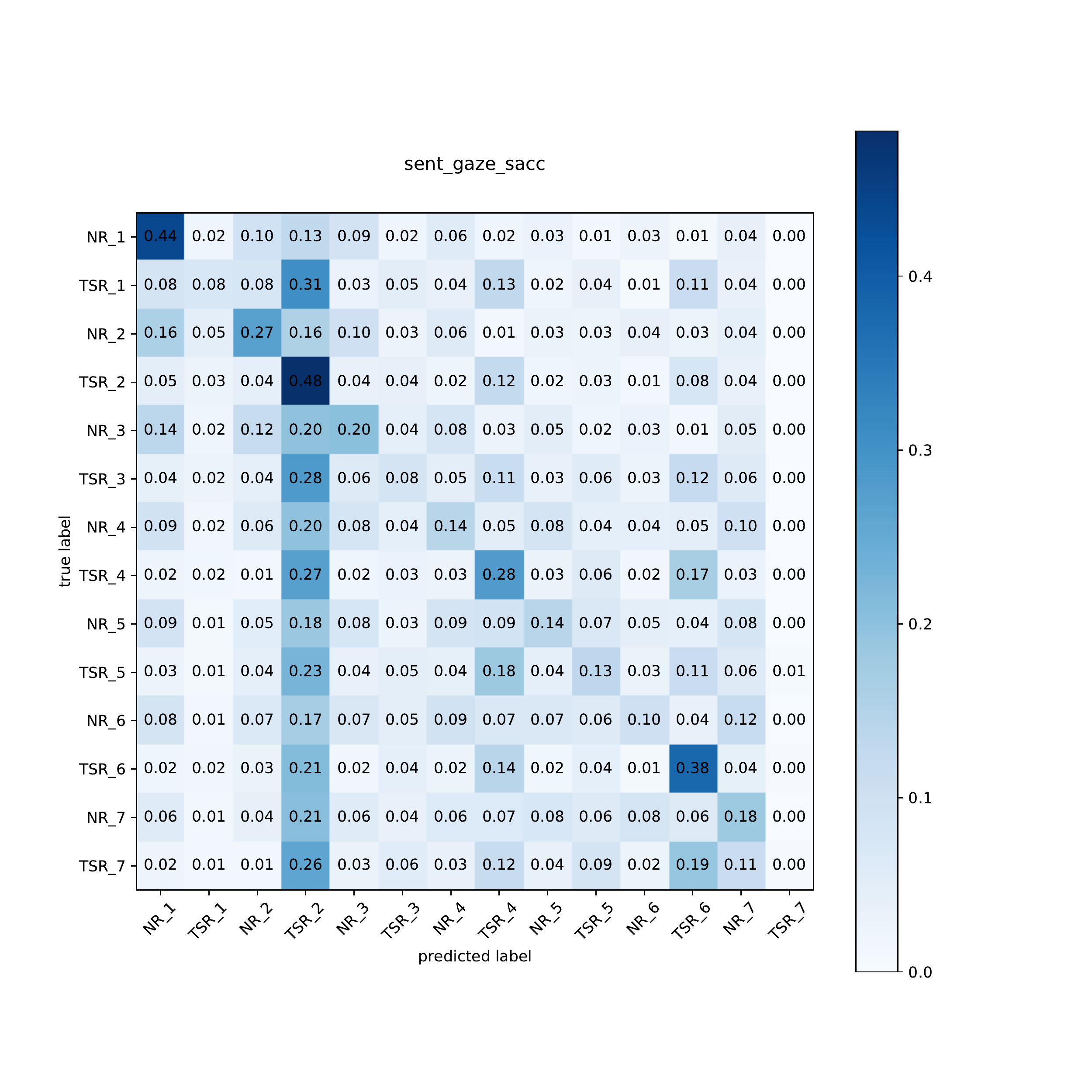}
    \includegraphics[width=0.32\textwidth,trim=100 0 20 0 ]{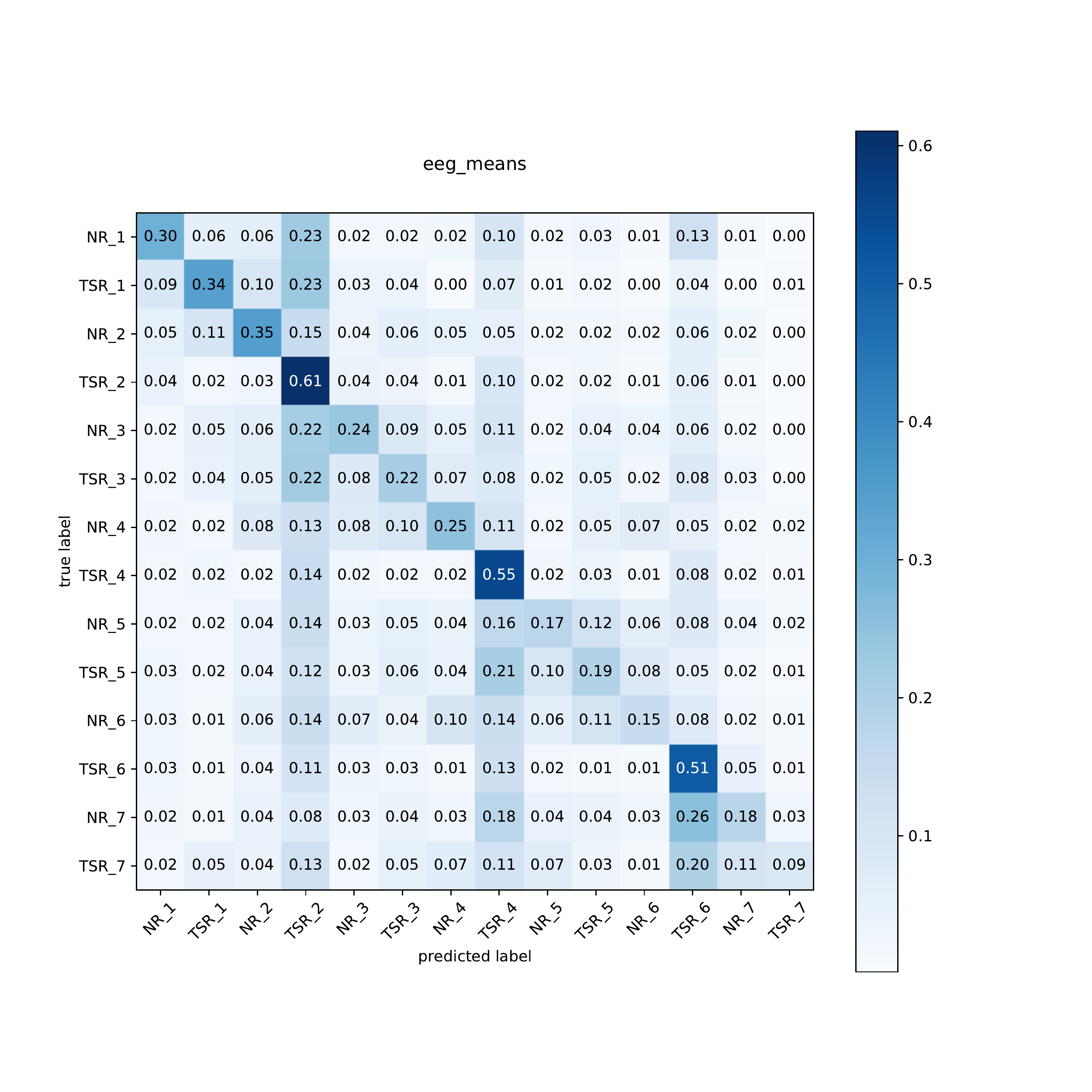}
    \includegraphics[width=0.32\textwidth,trim=100 0 20 0 ]{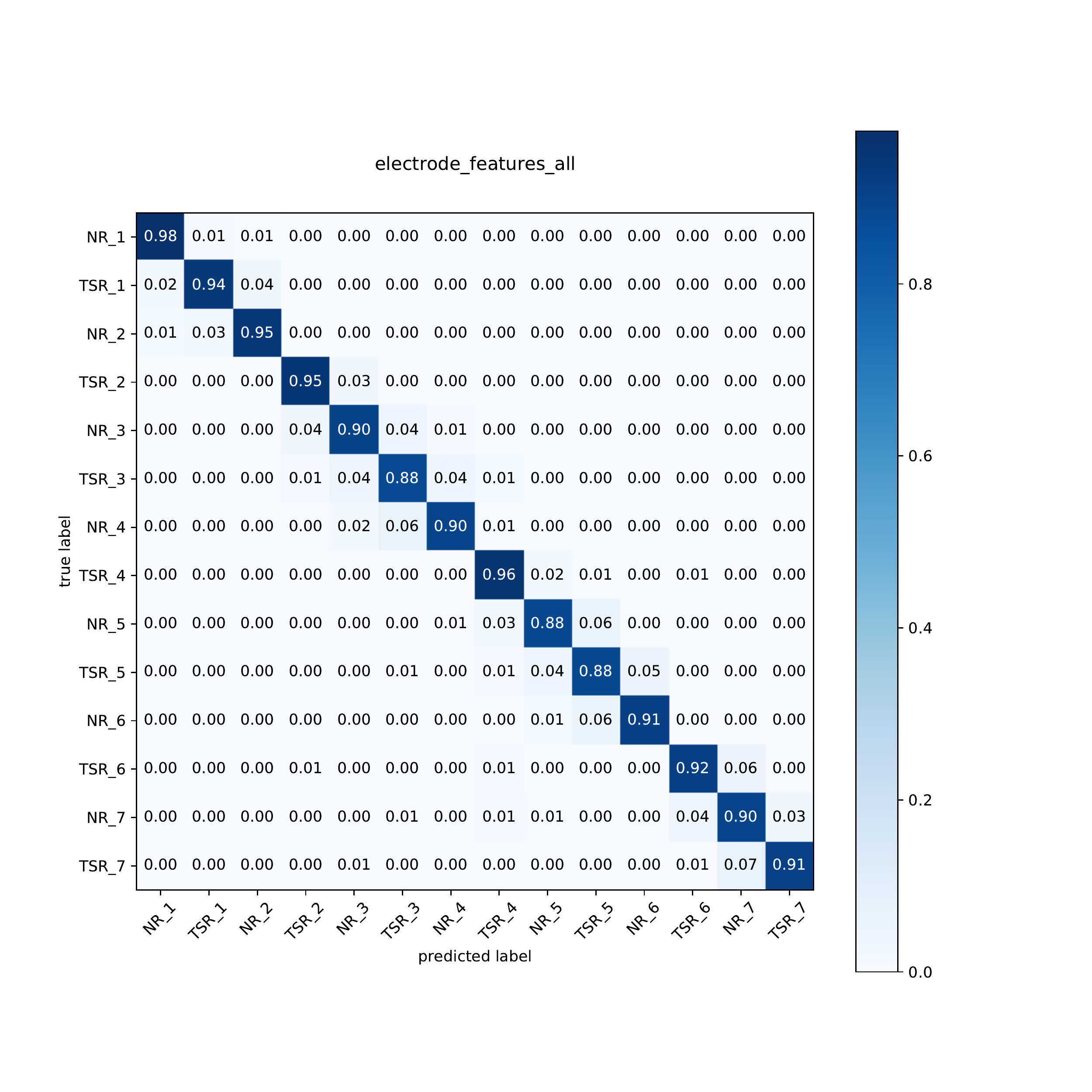} 
    \caption{Confusion matrices for block classification of ZuCo 2.0 for the three feature sets \textit{sent\_gaze\_sacc}, \textit{eeg\_means} and \textit{electrode\_features\_all}.}
    \label{fig:block-cms}
\end{figure}

The results are averaged across all subjects. Figure \ref{fig:block-class} shows how all eye-tracking features as well as the mean EEG features perform only slightly above random for this block classification task, while the electrode features -- especially from the gamma frequency band -- still achieve high accuracy. A closer look at the confusion matrices presented in Figure \ref{fig:block-cms}, shows that with eye-tracking features the blocks are often confounded within the same reading task (left), which shows that effectively the reading patterns between the two tasks are substantially different. When using EEG mean features to classify the sentence into recording blocks, the blocks are more often confounded with neighboring blocks in terms of recording order (middle). Finally, when using the EEG electrode features, the few classification mistakes also occur mostly within neighboring blocks (right). This pattern can be explained by the nature of the EEG experiments, where the electrode impedance was tested and corrected after every 3-4 blocks. 

One concern with the recording blocks is that they were recorded in the same order for all participants. Hence, in an attempt to quantify the effect of the variability in the data between the different sequential recording blocks of ZuCo 2.0, we perform an ablation study and train the binary reading task classification using sentence-level models with a decreasing number of blocks. This means that the models are trained on data ranging from one randomly selected block per reading task (i.e., one NR block and one block) up to six blocks per reading task, and tested on sentences of the remaining blocks. The hypothesis is that if systematic information about the reading tasks is encoded in the order of the blocks, the classification performance will increase when adding more block to the training data.

\begin{figure}[t!]
    \centering
    \includegraphics[width=0.62\textwidth]{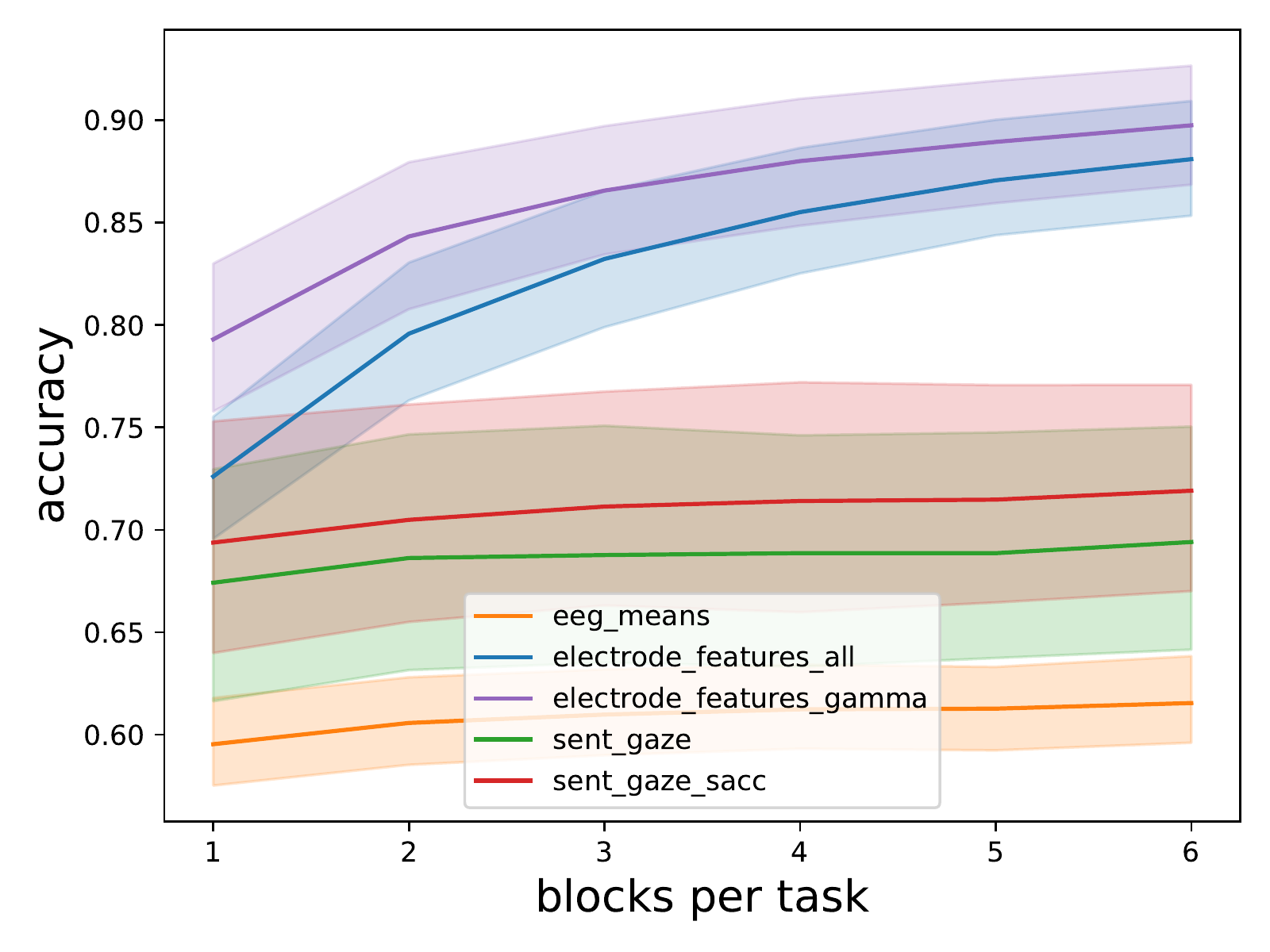} 
    \caption{Reading task classification on the ZuCo 2.0 data with decreasing number of recording blocks in the training data.}
    \label{fig:blocks-ablation}
\end{figure}

The results are shown in Figure \ref{fig:blocks-ablation}. We observe that the performance of both the eye-tracking features as well as the EEG mean features only increases slightly with additional blocks, while accuracy of the models trained on the EEG electrode features (both all frequency bands and gamma) increase substantially with data from more blocks. This shows that the EEG electrode features are more sensitive to the information about the recording blocks encoded in the brain activity signals and that the reading task classification benefits from this information.

\section{Conclusion}

\noindent Reading is a complex cognitive process that requires the simultaneous processing of visual and higher-level linguistic information including  syntactic, semantic and discourse integration. Identifying task-specific reading patterns can improve models of human reading and provide insights into human language understanding and how we perform linguistic tasks. This knowledge can then be applied to machine learning algorithms for natural language processing.

Accurate reading task classification can improve the manual labelling process for a variety of NLP tasks, as these processes are closely related to identifying reading intents. Recognizing reading patterns for estimating reading effort has additional applications such as as the diagnosis of reading impairments such as dyslexia \citep{rello2015detecting} and attention deficit disorder \citep{tor2021automated}.

The rise of more accessible behavioral and physiological datasets tailored towards machine learning allows their use to advance natural language processing methods. The data quality validation of both datasets, including a detailed analysis of eye tracking and EEG features and a comparison to previous studies, can be found in the original publications \citep{hollenstein2018zuco,hollenstein2020zuco}.

In this work, we presented various machine learning methods for reading task classification. The ML models learn to distinguish between a normal reading task and task-specific information-searching reading. We develop models that learn from eye movements or EEG signals. While the within-subject evaluation yields high classification accuracy, further research is required to improve the cross-subject generalization capabilities of the models. This is crucial for any potential applications of this classification task. Furthermore, we extensively analyze the differences between ZuCo 1.0 and ZuCo 2.0, and how the session bias of ZuCo 1.0 impacts the classification results. We address some of the open challenges in building robust ML models for task classification based on eye movement and brain activity data, including signal-to-noise ratio, high inter-subject variability, and generalization across sessions and datasets.

\bibliographystyle{unsrtnat} 
\bibliography{cog-sci-nlp-2021}

\clearpage
\appendix

\section{Data Collection}\label{app:data-collection}

\noindent Please refer to \citet{hollenstein2018zuco} and \citet{hollenstein2020zuco} for detailed descriptions of the data acquisition, preprocessing and feature extraction methods for ZuCo 1.0 and ZuCo 2.0, respectively.

\subsection{Participants}

\noindent For ZuCo 1.0, data were recorded from 12 healthy adults (between 22 and 54 years old; all right-handed; 5 female subjects).
For ZuCo 2.0, data were recorded from 18 healthy adults (between 23 and 52 years old; 2 left-handed; 10 female subjects). 
The native language of all participants is English, originating from Australia, Canada, UK, USA or South Africa. In addition, all subjects completed the standardized LexTALE test to assess their vocabulary and language proficiency (Lexical Test for Advanced Learners of English; \citealp{lemhofer2012introducing}). All participants gave written consent for their participation and the re-use of the data prior to the start of the experiments. The study was approved by the Ethics Commission of the University of Zurich.

\begin{table}[h]
\centering
\small
\begin{tabular}{lccccc}
\toprule
\textbf{ID} & \textbf{LexTALE} & \textbf{Score NR} & \textbf{Score TSR} & \textbf{Speed NR} & \textbf{Speed TSR} \\\midrule
ZKW & 96.25 & 91.67 & 94.84 & 11.73 & 6.14 \\
ZDN & 97.5 & 86.11 & 92.87 & 4.10 & 2.93 \\
ZPH & 97.5 & 94.44 & 97.05 & 7.55 & 2.71 \\
ZMG & 100 & 88.89 & 95.82 & 5.33 & 3.73 \\
ZAB & 100 & 86.11 & 90.42 & 5.14 & 3.32 \\
ZJN & 97.5 & 83.33 & 79.12 & 11.30 & 7.10 \\
ZKH & 81.25 & 83.33 & 93.12 & 6.43 & 5.57 \\
ZGW & 91.25 & 86.11 & 92.14 & 8.06 & 4.17 \\
ZJS & 97.5 & 91.67 & 93.86 & 4.18 & 2.88 \\
ZKB & 100 & 86.11 & 95.33 & 8.43 & 2.48 \\
ZDM & 100 & 80.56 & 96.81 & 5.13 & 3.32 \\
ZJM & 77.5 & 97.22 & 96.56 & 8.73 & 6.30 \\\midrule
\textbf{mean} & \textbf{94.69} & \textbf{87.96} & \textbf{93.16} & \textbf{7.18} & \textbf{4.22} \\\bottomrule
\end{tabular}
\caption{ZuCo 1.0 Subject demographics, LexTALE scores, and control scores and reading speed (i.e. seconds per sentence) for each task. The * next to the subject ID marks a bilingual subject.}
\label{tab:subjects-stats-z1}
\end{table}

\begin{table}[h]
\centering
\small
\begin{tabular}{lccccc}
\toprule
\textbf{ID} & \textbf{LexTALE} & \textbf{Score NR} & \textbf{Score TSR} & \textbf{Speed NR} & \textbf{Speed TSR} \\\midrule
YAC & 76.25\% & 82.61\% & 83.85\% & 5.27 & 4.96 \\
YAG & 93.75\% & 91.30\% & 56.92\% & 7.64 & 8.73 \\
YAK & 100.00\% & 74.07\% & 96.41\% & 3.83 & 5.89 \\
YDG & 100.00\% & 91.30\% & 96.67\% & 4.97 & 3.93 \\
YDR & 85.00\% & 78.26\% & 96.92\% & 4.32 & 2.32 \\
YFR & 85.00\% & 89.13\% & 94.36\% & 6.48 & 4.79 \\
YFS & 90.00\% & 91.30\% & 96.15\% & 3.96 & 2.85 \\
YHS & 90.00\% & 78.26\% & 97.69\% & 3.30 & 2.40 \\
YIS & 97.50\% & 89.13\% & 98.46\% & 5.82 & 2.58 \\
YLS & 93.75\% & 91.30\% & 92.31\% & 5.57 & 5.85 \\
YMD & 100.00\% & 86.96\% & 95.64\% & 7.50 & 6.24 \\
YMS & 86.25\% & 89.13\% & 95.38\% & 7.68 & 3.35 \\
YRH & 81.25\% & 86.96\% & 95.64\% & 5.14 & 4.32 \\
YRK & 85.00\% & 97.83\% & 96.15\% & 7.35 & 7.70 \\
YRP & 82.50\% & 78.26\% & 90.00\% & 7.14 & 8.37 \\
YSD & 95.00\% & 93.48\% & 94.36\% & 5.01 & 2.87 \\
YSL & 71.25\% & 84.78\% & 83.85\% & 6.73 & 6.14 \\
YTL* & 81.25\% & 80.43\% & 94.10\% & 7.48 & 3.23 \\\midrule
\textbf{mean}  & \textbf{88.54\%} & \textbf{86.36\%} & \textbf{91.94\%} & \textbf{5.84} & \textbf{4.81} \\\bottomrule
\end{tabular}
\caption{ZuCo 2.0 Subject demographics, LexTALE scores, and control scores and reading speed (i.e. seconds per sentence) for each task. The * next to the subject ID marks a bilingual subject.}
\label{tab:subjects-stats-z2}
\end{table}

\subsection{EEG Data}\label{app:eeg-data}

\noindent In this section, we present the EEG data extracted from the ZuCo corpus for this work. We describe the acquisition and preprocessing procedures as well as the feature extraction.

\subsection{Sentiment Reading Task}\label{app:sentiment-reading}
\noindent ZuCo 1.0 includes a third reading task. We only use this data for the control analyses. Therefore, we describe it here.

For this task, the subjects were presented with positive, negative, or neutral sentences from the Stanford Sentiment Treebank \citep{socher2013recursive}. The Stanford Sentiment Treebank contains single sentences extracted from movie reviews with manually annotated sentiment labels. We randomly selected 400 very positive, very negative, or neutral sentences (4\% of the full treebank). The 400 selected sentences are comprised of 123 neutral, 137 negative and 140 positive sentences. The sentences were split into two blocks, one for each recording session.

The participants were asked to read the sentences normally. The objective was to analyze the elicitation of emotions and opinions during reading. As a control condition, the subjects had to rate the quality of the described movies in 47 of the 400 sentences. The average response accuracy compared to the original labels of the Stanford Sentiment Treebank is 79.53\%.

\section{Model Parameters}

\noindent Table \ref{tab:params-lstm} shows the hyper-parameters used in the word-level models presented in Section \ref{sec:word-level}.

\begin{table}[h]
\centering
\small
\begin{tabular}{lc}
\toprule
\textbf{Parameter} & \textbf{Value} \\\midrule
Learning rate & 0.001 \\
LSTM dimension & 64 \\
Dense dimension & 64 \\
Batch size & 40 \\
Epochs & 200 \\
Patience & 104 \\
Min. delta & 0.0000001 \\\bottomrule
\end{tabular}
\caption{Hyper-parameters of word-level LSTM model.}
\label{tab:params-lstm}
\end{table}

\end{document}